\documentclass[10pt,twocolumn,letterpaper]{article}

\usepackage{cvpr}
\usepackage{times}
\usepackage{epsfig}
\usepackage{graphicx}
\usepackage{amsmath}
\usepackage{amssymb}
\usepackage{placeins}
\usepackage{subcaption}
\usepackage{caption}
\usepackage{multirow}
\usepackage[export]{adjustbox}
\usepackage{balance}

\newcommand{\fig}[1]{Figure~\ref{fig:#1}}

\newcommand{\p}{\mathcal{P}}

\newcommand{\x}{\mathbf{x}}
\newcommand{\n}{\mathbf{n}}
\newcommand{\D}{\mathcal{D}}

\usepackage[pagebackref=true,breaklinks=true,letterpaper=true,colorlinks,bookmarks=false]{hyperref}

\cvprfinalcopy % *** Uncomment this line for the final submission

\ifcvprfinal\fi
\begin{document}

%%%%%%%%% TITLE
\title{Relightable 3D Head Portraits from a Smartphone Video}

\author{Artem Sevastopolsky$^{1,2}$ \qquad Savva Ignatiev$^2$ \qquad Gonzalo Ferrer$^2$ \\
Evgeny Burnaev$^2$ \qquad Victor Lempitsky$^{1,2}$\\
\ \\
% For a paper whose authors are all at the same institution,
% omit the following lines up until the closing ``}''.
% Additional authors and addresses can be added with ``\and'',
% just like the second author.
% To save space, use either the email address or home page, not both
$^1$ Samsung AI Center, Moscow, Russia\\
$^2$ Skolkovo Institute of Science and Technology (Skoltech), Moscow, Russia\\
\ \\
\texttt{\{a.sevastopol, v.lempitsky\}@samsung.com}
}

% \maketitle
\twocolumn[{
\renewcommand\twocolumn[1][]{#1}
    \maketitle
    \begin{center}
        % \begin{figure*}[t!]
            \centering
            \adjincludegraphics[trim={0 {.23\height} 0 {.27\height}},clip,width=\textwidth]{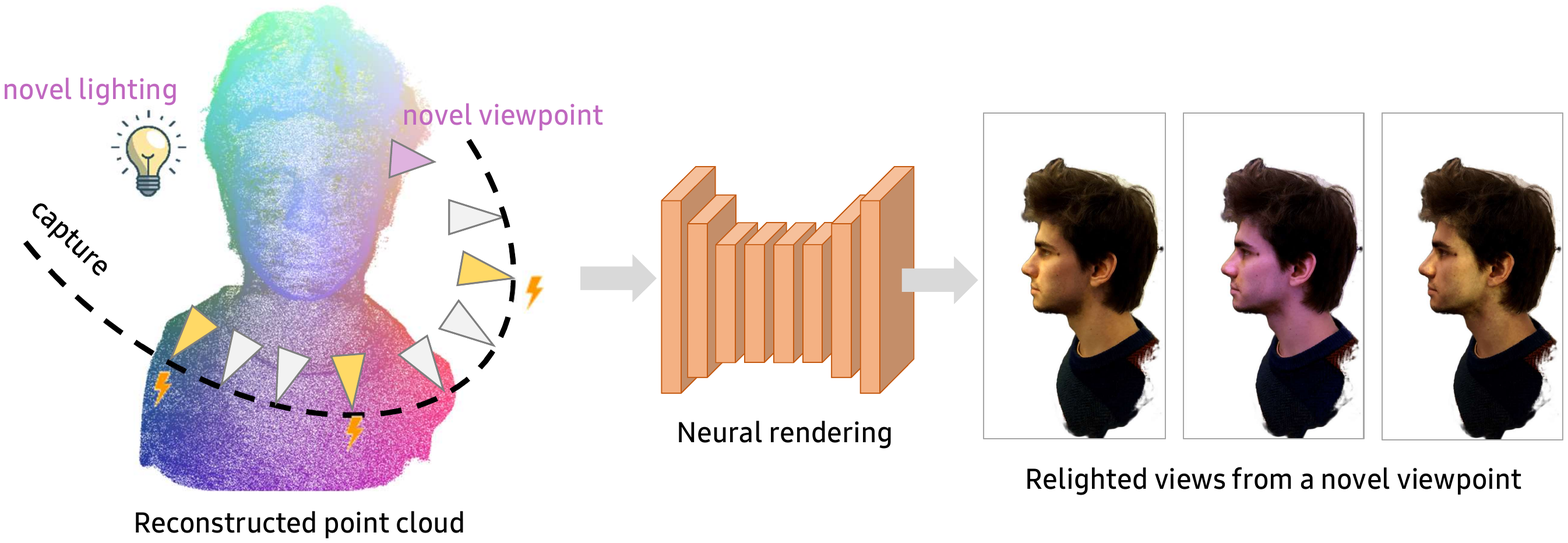}
            \captionof{figure}{%Main application of the method. The method requires a video of a person made by a smartphone camera with the flash blinking at a certain frequency. By the Structure-from-Motion (SfM) software, a camera pose is estimated for each of the frames extracted from the video, and a dense point cloud is reconstructed. The noise in the point cloud is then filtered by a deep segmentation network. After training, our neural pipeline receives the point cloud, rasterized onto a novel camera view, and predicts several light-invariant feature maps, which, given novel lighting conditions, are fused into the relighted image.
            Our method generates relightable 3D portraits. It takes a smartphone video of a person taken with blinking flash. Using Structure-from-Motion (SfM) software, a camera pose is estimated for each of the frames, and a dense point cloud is reconstructed.  After fitting, our neural rendering pipeline receives the point cloud rasterized onto a novel camera view. Several maps of lighting-related properties are then predicted. Based on these maps, images relighted with novel lighting conditions can be rendered.}
            \label{fig:teaser}
        % \end{figure*}
    \end{center}
}]

\begin{abstract}
In this work, a system for creating a relightable 3D portrait of a human head is presented. Our neural pipeline operates on a sequence of frames captured by a smartphone camera with the flash blinking (``flash-no flash'' sequence). A coarse point cloud reconstructed via structure-from-motion software and multi-view denoising is then used as a geometric proxy. Afterwards, a deep rendering network is trained to regress dense albedo, normals, and environmental lighting maps for arbitrary new viewpoints. Effectively, the proxy geometry and the rendering network constitute a relightable 3D portrait model, that can be synthesized from an arbitrary viewpoint and under arbitrary lighting, e.g. directional light, point light, or an environment map. The model is fitted to the sequence of frames with human face-specific priors that enforce the plausability of albedo-lighting decomposition and operates at the interactive frame rate. We evaluate the performance of the method under varying lighting conditions and at the extrapolated viewpoints and compare with existing relighting methods.
\end{abstract}

\section{Introduction}

\begin{figure*}[t!]
    \centering
    \includegraphics[width=\textwidth]{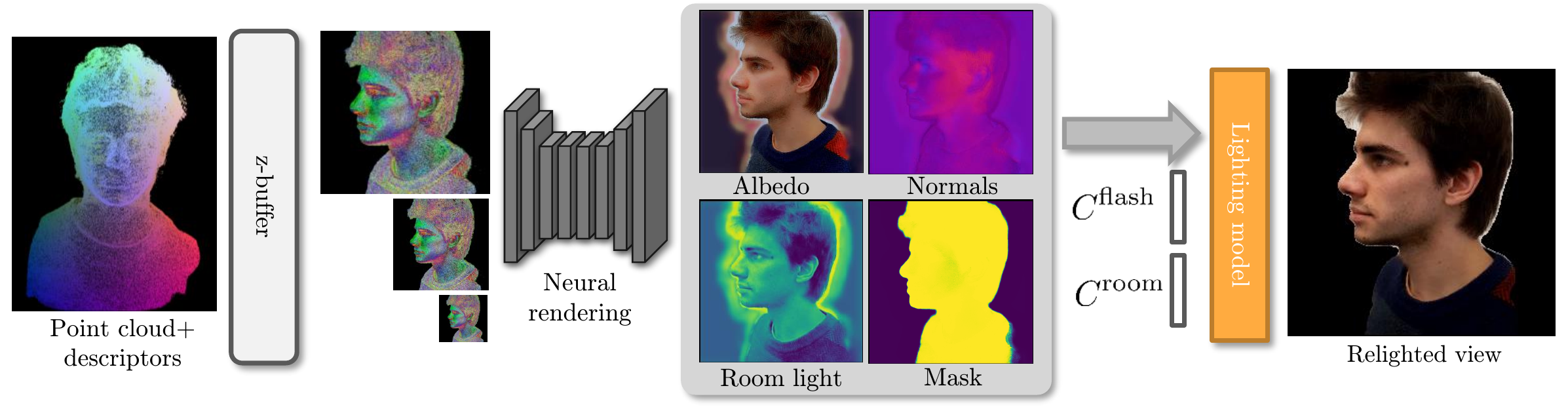}
    \caption{The overview of our rendering process. The point cloud with the neural descriptors is rasterized according to the desired camera position at several resolutions using a simple z-buffer algorithm. The neural rendering process then translates the rasterizations into a set of maps, including the per-pixel albedo, the per-pixel normals, the per-pixel lighting intensity w.r.t.~the original room, and the foreground mask. The lighting model can then reproduce the training image, while taking the dominant colors of the flash and the room lighting into account. At test time, new lighting parameters can be plugged into the lighting model instead.}
    \label{fig:scheme}
\end{figure*}

The rise of mobile photography comes hand-in-hand with the pervasiveness of two-dimensional displays. As three-dimensional display devices such as VR headsets, AR glasses, 3D monitors are becoming wide-spread, expanding mobile photography to 3D content acquisition becomes an interesting research direction. In this work, we describe a system for obtaining photorealistic 3D models of human heads or upperbodies (which we refer to as \textit{3D human portraits}) from handheld video sequences, e.g.\ shot by a smartphone camera.

Generally, acquiring photorealistic 3D models from videos is a heavily-researched direction~\cite{Mildenhall20,Aliev20,Agrawal20}. However, most 3D displays project the 3D models into either the user surrounding (AR glasses) or the virtual environment (VR headsets, 3D monitors). Thus, to enhance the realism of such models, they should be made \textit{relightable} in a realistic way according to their environment (real or synthetic). 

Building realistic \textit{relightable} 3D models is a far less investigated area of research. Existing works either focus on single-view 3D relightable reconstruction~\cite{Zhou19, Zhang20a} (limiting the complexity of the model that can be obtained), or on the acquisition of relightable models using specialized equipment (\textit{light stages})~\cite{Guo19,Zhang20b}. In contrast, our system expects only a handheld video of a person made by a smartphone with interleaved flash and no-flash frames. We assume that the person remains static during the video acquisition (which takes few dozens of seconds). 

Given such a video, our system estimates a relightable model. We follow the neural point-based graphics approach~\cite{Aliev20} and use a point cloud estimated by structure-from-motion as a geometric proxy, whereas the photometric information is encoded in the vectorial neural descriptors of individual points in the cloud. The points and their descriptors can then be rasterized for novel camera viewpoints, and the rasterizations are processed by a rendering network~\cite{Aliev20}. In our case, the rendering network outputs per-pixel albedo values, the normal direction, and the shadow map.

During the scene reconstruction process, we fit this model to the video frames estimating the parameters of the scene lighting and the camera flash along the way. We use human face-specific priors to disentangle lighting and albedo factors. While the priors are only measured in the facial part of the scene, they facilitate disentanglement across the whole 3D portrait. Once reconstructed, the scene (portrait) can be rendered from new viewpoints and with new lighting at interactive speeds. % Example novel view/novel lighting renderings are shown in \fig{teaser}.

% To sum up, our contributions are as follows:
Our contributions are as follows:
\begin{itemize}
    \item We propose a new neural rendering model that supports relighting and is based on point cloud geometry.
    \item We suggest a new albedo-lighting disentangling approach that uses face priors.
    \item Combining the two contributions, we build a system for relightable 3D model creation that is suitable for human heads and upperbodies, uses handheld videos, and supports realtime rendering.
\end{itemize}

We evaluate the resulting system on a number of videos, and compare it to the alternative approaches~\cite{Zhou19, Chen20}.

\section{Related work}

Creating 3D models from registered video sequences is a problem with a long history~\cite{Kutulakos00,Vogiatzis07,Furukawa09}. Estimating the photometric properties for such models is known to be a challenging problem in itself~\cite{Kong14,Taheri12,Zhi20}. Recently, several works have suggested to model objects using coarse geometry and \textit{neural} photometric descriptors. The rasterization of such descriptors is processed into realistic RGB views using rendering networks. The geometry can take the form of a mesh~\cite{Thies19} or a point cloud~\cite{Aliev20,Lassner20,Wiles20}. Our approach builds on the neural point-based graphics framework~\cite{Aliev20} and expands it significantly to add the relighting capability.

Estimating surface properties that allow arbitrary relighting is also a research area with long history~\cite{Dong14,Palma13,Li20}. There is an emerging direction to estimate surface properties from flash and no-flash images of the same scene~\cite{Boss20,Cao20}. Several approaches focus on relighting 2D portraits without 3D head reconstruction~\cite{Zhou19,Zhang20a}. The recent work~\cite{Lattas20} estimates the relightable 3D portrait from a single image but restricts the reconstruction to the face (3DMM area) with hair and garment. Several approaches come up with an impressive full body relightable model, yet rely on dedicated light stage equipment~\cite{Meka19,Guo19,Zhang20b,Chen20,Meka20}. Our work thus achieves a different balance between completeness of the model and the ease of acquisition. It is capable of reconstructing a 3D model of the entire head (and shoulders), and does so from a handheld video sequence that can be acquired from a mobile phone. 

Even though our reconstruction is not restricted to the face region, the success of our method relies on the use of face priors. The face region thus serves as a ``probe'' facilitating the reconstruction. Similar idea has been proposed for the reconstruction of non-relightable head portraits in~\cite{Xu20}.

\section{Method}

We start by discussing the lighting model used in our approach. After that, we detail our approach to geometric modeling based on neural point-based graphics. In the end of this section, we discuss the model construction (fitting process), including the loss terms and the prior terms.

\subsection{Lighting model}

For a given point $\x$ in space which belongs to the surface of a volumetric object, the level of radiance of the light emitted at the location $\x$ in direction $\omega_o$ is commonly described by the rendering equation~\cite{Kajiya86}:

\begin{equation}
    \begin{aligned}
        &L_o(\x, \omega_o) \\
        &= \int\limits_{S} f_r(\x, \omega_i, \omega_o) L_i(\x, \omega_i) v(\x, \omega_i) \cdot \langle -\omega_i, \n(\x) \rangle d\omega_i,
    \label{eq:rendering_eq}
    \end{aligned}
\end{equation}

\noindent where $L_i(\x, \omega_i)$ defines the incoming radiance to $\x$ in direction $\omega_i$, $S$ stands for the upper hemisphere w.r.t.\ the surface tangent plane at $\x$ with the unit normal $\n(\x)$. Also, $f_r(\x, \omega_i, \omega_o)$ is the ratio of scattered light intensity in the direction $\omega_o$ and the incoming at direction $\omega_i$, which is usually referred to as the bidirectional reflectance distribution function (BRDF). Furthermore, $v(\x, \omega_i)$ is a visibility term (equals to 1 if a point $\x$ is reachable by the light from direction $\omega_i$, or 0 if there is an occlusion in that direction), and $L_o(\x, \omega_o)$ defines the total radiance coming from $\x$ in direction $\omega_0$. In terms of this model, BRDF is a material property that describes its spatially-varying light scattering properties. Since all images in our experiments are RGB, values $L_o(\x, \omega_o)$, $L_i(\x, \omega_i)$, $f_r(\x, \omega_i, \omega_o)$ are all in $\mathbb{R}^3$, with each component (channel) calculated independently.

In our setup, we capture the subject using a set of frames taken with an environment lighting, and another set frames in the same environment additionally lighted by a camera flash. To model these two sets of frames, we decompose the incoming radiance with visibility into two terms:
\begin{gather}
L_i(\x, \omega_i)\,v(\x, \omega_i) = L_i^\mathrm{room}(\x, \omega_i) v^\mathrm{room}(\x, \omega_i) + \nonumber\\ \mathrm{F} \cdot L_i^\mathrm{flash}(\x, \omega_i) v^\mathrm{flash}(\x, \omega_i)\,,
\end{gather}
where $L_i^\mathrm{room}(\x, \omega_i)$ stands for the environmental (room) light, $L_i^\mathrm{flash}(\x, \omega_i)$ stands for the flash light, and $\mathrm{F}$ indicated if the photo is taken with the flash turned on ($\mathrm{F} = 1$) or turned off ($\mathrm{F} = 0$). Accordingly, $v^\mathrm{room}$ and $v^\mathrm{flash}$ model the light occlusion for the environmental and the flash lighting respectively. 

We further assume that the BRDF is Lambertian (constant at each point) $f_r(\x, \omega_i, \omega_o) = \rho(\x)$ (with the normalizing constant omitted), i.e.\ corresponds to a diffuse surface, and $\rho(\x)$ stands for the surface albedo at $\x$. We note that through the use of neural rendering, our system can model some amount of non-Lambertian effects since the albedo in our model can be effectively made view-dependent. Applying the modifications to~(\ref{eq:rendering_eq}), we get:

\begin{equation}
    \begin{aligned}
        & L_o(\x, \omega_o) =\\
        & \rho(\x) \int\limits_{S} L_i^\mathrm{room}(\x, \omega_i) v^\mathrm{room}(\x, \omega_i) \cdot \langle -\omega_i, \n(\x)\rangle d\omega_i
        \\
        & + \mathrm{F} \cdot \rho(\x) \int\limits_{S} L_i^\mathrm{flash}(\x, \omega_i) v^\mathrm{flash}(\x, \omega_i) \cdot \langle -\omega_i, \n(\x)\rangle d\omega_i
        \label{eq:rendering_eq_2}
    \end{aligned}
\end{equation}

Suppose now that the integral in the first part is equal to $\overline{s}(x) \in \mathbb{R}^3$ and defines the shadowing caused both by the room lamps and occlusions (e.g.\ a nose casting shadow on a cheek, in case of a human head). We are going to model it as a product of color temperature and grayscale shadowing $\overline{s}(x) = C^\mathrm{room} \cdot s(x),\, C^\mathrm{room} \in \mathbb{R}^3,\, s(x) \in \mathbb{R}$. As for the second (flash) part, we explicitly model the incoming light radiance $L_i^\mathrm{flash}(\x, \omega_i) = \frac{C^\mathrm{flash}}{d(\x)^2}$, where $d(\x)$ is the distance from the flash to $\x$, and $C^\mathrm{flash} \in \mathbb{R}^3$ is a constant vector proportional to the color temperature and intensity of the flash. In our experiments we assume that the flash is far enough, and hence, $d(\x) \approx d$, with $d$ being the distance from the camera to its closest point in the point cloud, and the light rays from the flash are approximately parallel. Since on a smartphone, flashlight and the camera lens are usually co-located at close proximity (at $\omega_o$), we assume that $v^\mathrm{flash}(\x, \omega_i) = 1$ for all $\x$ observed in a flashlighted image. This brings us to the transformation of~(\ref{eq:rendering_eq_2}) into (\ref{eq:rendering_eq_3}): % This assumption is practical in our case, as we will be working with the point cloud not a triangular mesh.

\begin{equation}
    \begin{aligned}
        & L_o(\x, \omega_o) = \rho(\x) C^\mathrm{room} s(x) + \mathrm{F} \cdot \rho(\x) \frac{C^\mathrm{flash}}{d^2} \langle\n(\x), -\omega_o\rangle
        \label{eq:rendering_eq_3}
    \end{aligned}
\end{equation}

Note that in~(\ref{eq:rendering_eq_3}), the decomposition of the light intensity $L_o(\x, \omega_o)$ into separate components is ambiguous, as is common to most lighting-albedo estimation problems. In particular, there is an inverse proportional relationship between $\rho(\x)$ and both $C^\mathrm{room}$ and $C^\mathrm{flash}$. This ambiguity will be resolved in Section~\ref{sub:loss_functions} through the use of appropriate priors.

\subsection{Geometric modeling}

After discussing the lighting model, we proceed to the geometric modeling. We assume that a set of RGB photographs of a human head $\mathcal{I} = \{I_1, \dots, I_P\}$, $I_k \in \mathbb{R}^{\mathrm{H} \times \mathrm{W} \times 3}$, all are taken by a mobile camera in the same environment and featuring the head from various angles. A subset  $\mathcal{I}' = \{I_{s_1}, \dots, I_{s_M}\}$ of photos features the head additionally lighted by a camera flash. As a first step, we use structure-from-motion (SfM) methods to reconstruct camera poses $C_1, \dots, C_P$ of each image: $C_k = (K_k, [R_k, t_k])$. Then, similarly, we reconstruct a dense point cloud $\p = \{p_1, \dots, p_N\},\, p_i=\{x_i, y_i, z_i\}$ from all images. In our experiments, we used Agisoft Metashape~\cite{Agisoft}, while any other software such as COLMAP~\cite{Schoenberger2016sfm,Schoenberger2016mvs}, etc. could be employed instead.

\paragraph{Segmentation and filtering.} As we aim to model the subject without background, we estimate the foreground segmentation of individual photographs, and filter out the estimated point cloud along the way. We use the newly proposed U$^2$-Net~\cite{Qin20} segmentation network designed for a salient object segmentation task. We fine-tune the pre-trained U$^2$-Net model with a warm-up learning schedule on the Supervisely~\cite{Supervisely} human segmentation dataset to make it more suitable for our task. After that, we pass photographs $\mathcal{I}$ through the fine-tuned model and obtain the sequence of initial `soft' masks $\mathcal{M}^0 = \{M_1^0, \dots, M_P^0\}$, $M_k^0 \in \mathbb{R}^{\mathrm{H} \times \mathrm{W} \times 1}$. Finally, we add multi-view consistency of the silhouettes via visual hull estimation~\cite{Laurentini94} using point cloud as a geometric proxy, and using the weights of the segmentation network as a parameterization. This is done by fine-tuning the weights of the segmentation network to minimize the inconsistency of segmentations across the views, which results in obtaining the refined masks $\mathcal{M} = \{M_1, \dots, M_P\}$.

\subsubsection{Neural rendering.} Our rendering approach (\fig{scheme}) is based on a deep neural network predicting albedo, normals, and room shadows from a point cloud rasterized onto each camera view, and follows the Neural Point-Based Graphics (NPBG) pipeline~\cite{Aliev20} to accomplish that. Following their approach, points in the point cloud are augmented with neural descriptors, i.e.\ multi-dimensional latent vectors characterizing point properties: $\D = \{d_1, \dots, d_N\},\, d_k \in \mathbb{R}^L$ ($L=8$ in all our experiments).

The neural rendering stage then starts with the rasterization of the descriptors onto the canvas associated with the camera $C_k$. This is performed the way suggested in~\cite{Aliev20}: the \textit{raw image} $S[0] = S[0](\p, \D, C_k) \in \mathbb{R}^{H \times W \times L}$ is formed by Z-Buffering the point cloud (for each pixel, finding the closest to the camera point which projects to this pixel). The descriptor of each of the closest points is assigned to the respective pixel. In case there are no point which project to some pixel, a null descriptor is assigned to that pixel instead. Similarly, we construct a set of auxiliary raw images $S[1], \dots, S[T]$ of spatial sizes $\frac H{2^t} \times \frac W{2^t}$ and perform the rasterization of descriptors onto these images by the same algorithm. A pyramid of the raw images is introduced to cover the point cloud at several scales. The highest resolution raw image $S[0]$ features the richest spatial detail, while the lower resolution ones feature less surface bleeding (see~\cite{Aliev20} for the detailed explanation).  % The lowest resolution image S[T] has coarse geometric detailization, but has the least surface bleeding, while the intermediate raw images S[0], . . . , S[T−1] achieve different detailization-bleeding tradeoffs

The second step involves transforming the set of raw images $S[0], \dots, S[T]$ by a deep neural network $f_\phi(S[0] \dots, S[T])$ closely following the U-Net~\cite{Ronneberger15} structure with gated convolutions~\cite{Yu19}. As suggested in~\cite{Aliev20}, each of the raw images is passed as an input (or concatenated) to the first layer of the network encoder of the respective resolution. The output of the rendering network is the set of dense maps (in the case of~\cite{Aliev20}, the maps contain output RGB values).

In our case, the last layer of the network outputs an eight-channel tensor with several groups:
\begin{itemize}
    \item The first group contains three channels and uses sigmoid non-linearity. These channels contain the albedo values $A \in \mathbb{R}^{H \times W \times 3}$. Each pixel of $A$ describes the spatially-varying reflectance properties (albedo $\rho(\x)$) of the head surface (skin, hair, or other parts). 
    
    \item The second group also has three channels, and uses groupwise $L_2$ normalization in the end. This group contains the rasterized normals $N \in \mathbb{R}^{H \times W \times 3}$, with each pixel containing a normal vector $\n(\x)$ of the point at the head surface in the world space. 
    
    \item Next, one-channel group that uses the sigmoid non-linearity, corresponds to the grayscale shadowing $S \in \mathbb{R}^{H \times W}$, caused by the variations in the room lighting and consequently occurring occlusions. 
    
    \item Finally, the last one-channel group that also uses the sigmoid non-linearity, defines the segmentation mask $M \in \mathbb{R}^{H \times W}$, with each pixel containing the predicted probability of the pixel belonging to the head and not the background.
\end{itemize}

Given the output of the rendering network, the final rendered image is defined by fusing the albedo, normals, and room shadows, as prescribed by the lighting model~(\ref{eq:rendering_eq_3}):
\begin{equation}
    \begin{aligned}
        \mathcal{I} = A \cdot C^\mathrm{room} \cdot S + \mathrm{F} \cdot A \frac{C^\mathrm{flash}}{d^2} \langle N, -\omega_o\rangle,
    \end{aligned}
    \label{eq:final_image}
\end{equation}
\noindent where the scalar product $\langle N, -\omega_o \rangle$ is applied to each pixel. The color temperature vectors $C^\mathrm{room}$ and $C^\mathrm{flash}$  both in $\mathbb{R}^3$ are considered part of the model and are shared between all pixels. These vectors are estimated from training data as discussed below. At inference time, feature maps $A$ and $N$, as well as $C^\mathrm{room}, C^\mathrm{flash}$ parameters, can be used to render the human head under a different modified lighting, such as directional lighting $\left( \mathcal{I}' = A \cdot \lfloor \langle N, -\omega \rangle \rfloor \right)$ or other models such as spherical harmonics (SH, see~\cite{Ramamoorthi01, Gkitsas20}). In the latter case, the albedo $A$ is multiplied by a non-linear function of the pixel normals $N$ that incorporates the SH coefficients. When a $360^\circ$ panorama of a new environment is available, one can obtain the values of the SH coefficients (up to the predefined order) by the panorama integration. As described in~\cite{Ramamoorthi01}, choosing the order of 3 or higher, which results in at least 9 coefficients per color channel, can often yield the expressive lighting effects. In case of the third-order SH, the relighted image is defined as a quadratic form: %TODO: THIS IS WORTH DISCUSSING

\begin{equation}
    \begin{aligned}
        \mathcal{I}'[i, j] = A[i, j] \cdot [N[i, j]\,\, 1]^T \cdot \mathbf{M}(\mathrm{SH}_\mathrm{coef}) \cdot [N[i, j]\,\, 1],
    \end{aligned}
    \label{eq:SH}
\end{equation} %TODO double-check

\noindent where the $4 \times 4$ matrix $\mathbf{M}(\mathrm{SH}_\mathrm{coef})$ linearly depends on the 27 SH coefficients $\mathrm{SH}_\mathrm{coef}$ (the expression for $\mathbf{M}$ can be found in~\cite{Ramamoorthi01}).

% (?) Our lighting model does not account for specularity and other properties and only assumes the diffuse material. Instead, we rely on the deep rendering network $f_\phi$ that can compensate for missing view-dependent effects.

\subsection{Model fitting}
\label{sub:loss_functions}

\paragraph{Generic scene losses.} Our model contains a large number of parameters that are fitted to the data. During fitting, the obtained images $\mathcal{I}$~(\ref{eq:final_image}) are compared with the ground truth image $\mathcal{I}^\mathrm{gt}$ by a loss constructed from several components that are described below. 

The \textbf{main loss} equals to the evaluated mismatch between the estimated lighted image $\mathcal{I}$ and the ground truth image $\mathcal{I}^\mathrm{gt}$:

\begin{equation}
    \begin{aligned}
    L_\mathrm{final}&(\phi, \D, C^\mathrm{room}, C^\mathrm{flash}) \\
    & = \Delta\big(\mathcal{I}(\phi, \D, C^\mathrm{room}, C^\mathrm{flash}), \mathcal{I}^\mathrm{gt}\big),
    \end{aligned}
\end{equation}

\noindent whereas the mismatch function $\Delta$ is employed to compare pair of images:
\begin{equation}
    \begin{aligned}
        \Delta(I_1, I_2) = \mathrm{VGG}(I_1, I_2) + \beta \cdot L_1(\mathrm{pool}(I_1), \mathrm{pool}(I_2))
    \end{aligned}
\end{equation}
\noindent Here, $\mathrm{VGG}(\cdot, \cdot)$ is a perceptual mismatch based on the layers of VGG-16 network~\cite{Simonyan14}, $L_1(\cdot, \cdot)$ refers to the mean absolute deviation, $\mathrm{pool}(I)$ is the average pooling of an image $I$ with $K \times K$ kernel ($K=4$ in our experiments) and $\beta$ is a balancing coefficient introduced to equalize the range of values of two terms ($\beta = 2500$ was selected). While VGG encourages matching of high-frequency details, $L_1$ rewards matching colors. A naive optimization of $L_1$ can lead to blurring and loss of details, therefore, we evaluate the $L_1$-term over the downsampled images. %TODO DISCUSS THE MISMATCH LOSS HERE, INTRODUCE FORMULA.

%DISCUSS AND INTRODUCE THE MASK LOSS HERE.

Since the system must be able to segment the rendered head from an arbitrary viewpoint, we introduce the \textbf{segmentation loss} constraining the predicted mask $M$:
\begin{equation}
    \begin{aligned}
    L_\mathrm{mask}(\phi, \D) = -\log \mathrm{Dice}(M(\phi, \D), M_i),
    \end{aligned}
\end{equation}
where the Dice function is a common choice of the mismatch for segmentation~\cite{Sudre17} (evaluated as the pixel-wise F1 score). Effectively, with this loss, the network learns to extrapolate the precalculated masks $\mathcal{M}$ to new viewpoints. 

% DISCUSS AND INTRODUCE THE ROOM SHADING LOSS HERE.
For non-flashlighted images, the rendered image is equal to $\mathcal{I} = A \cdot C^\mathrm{room} \cdot S$ according to (\ref{eq:final_image}). In practice, this creates a certain ambiguity between the learned maps $A$ and $S$. Since $A$ participates in both terms of~(\ref{eq:final_image}), the high-frequency component of the renderings tends to be saved in $S$ by default. The following \textbf{room shading loss} implicitly requires $A$ to be fine-grained instead:

\begin{equation}
    \begin{aligned}
        L_\mathrm{TV}(\phi, \D) = \mathrm{TV}(S(\phi, \D))
    \end{aligned}
\end{equation}

\noindent where $\mathrm{TV}$ is the Total Variation loss based on $L_1$.

\paragraph{Face prior losses.} To further regularize the learning process, we use the particular property of the scene that we handle the presence of face regions. For that, for each training image we perform face alignment using a pretrained PRNet system~\cite{Feng18}. Given an arbitrary image $I$ containing a face, PRNet estimates a \textit{face alignment map}, i.e.~a tensor $\mathrm{Posmap}$ of size 256 x 256 that maps the UV coordinates (in a predefined, fixed texture space associated with the human face) to the screen-space coordinates of the image $I$. Let $\mathrm{Posmap}_1, \dots, \mathrm{Posmap}_P$ define the position maps calculated for each image in the training sequence. By the operation $I \odot \mathrm{Posmap}$, we denote the bilinear sampling (backward warping) of an image $I$ onto $\mathrm{Posmap}$, which results into mapping the visible part of $I$ into the UV texture space. The mapping thus constructs a colored (partial) face texture. 

We employ the gathered face geometry data in two ways. Firstly, throughout the fitting we estimate an albedo half-texture $\mathcal{T}_A$ of size $256 \times 128$ (only for the left part of the albedo). $\mathcal{T}_A$ is initialized by taking a pixel-wise median of the projected textures for all flashlighted images $\mathcal{T}_F = \mathrm{median}(I_{s_1} \odot \mathrm{Posmap}_{s_1}, \dots, I_{s_M} \odot \mathrm{Posmap}_{s_M})$ and averaging left and flipped right halves. The \textbf{symmetry loss} is a facial prior that encourages the symmetry of albedo by comparing it with the learned albedo texture: 

\begin{equation}
    \begin{aligned}
        L_\mathrm{symm}(\phi, \D, \mathcal{T}_A) = \Delta(& A(\phi, \D) \odot \mathrm{Posmap}_i, \\
        &\, [\mathcal{T}_A,\, \mathrm{flip}(\mathcal{T}_A)])
    \end{aligned}
\end{equation}

\noindent (mismatch is evaluated only where $A(\phi, \D) \odot \mathrm{Posmap}_i$ is defined). $[\mathcal{T}_A,\, \mathrm{flip}(\mathcal{T}_A)]$ denotes the concatenated texture and its horizontally flipped version. Note that this loss both makes albedo more symmetric and matches albedo texture $\mathcal{T}_A$ colors with the ones of the learned albedo. The balance between these two factors is controlled by the learning rate for $\mathcal{T}_A$. Incorporation of this loss is mainly inspired by the recent work~\cite{Wu20} and employed here in a simpler setting (without confidence estimation), as the UV texture space of PRNet is symmetric by design. For our task, the symmetry loss helps resolving the decomposition of an image into albedo and shadows, as the opposite points of the face (e.g. left and right cheeks) can have similar albedo, while casted shadows are most often non-symmetric.

Additionally, to select the gamma for albedo, we introduce the \textbf{albedo color matching loss} $L_\mathrm{cm}(\phi, \D) = L_1(A(\phi, \D) \odot \mathrm{Posmap}_i, \mathcal{T}_F)$ (also calculated only for valid texels of $A(\phi, \D) \odot \mathrm{Posmap}_i$).

Another type of data which can be inferred from PRNet outputs is the normals for the face part. Each face alignment map, along with the estimated depth (also by PRNet) and a set of triangles, defines a triangular mesh. We render the meshes estimated for each view, smoothen them, calculate the face normals and render the projected normals onto the respective camera view (technically, all performed in the Blender rendering engine). Then, the rendered normals are rotated by the camera rotation matrix for their conversion to the world space. We will introduce the notation $N_1, \dots, N_P$ for the estimated normal images for the facial part of size $H \times W$. The normals predicted by the network are matched with the PRNet normals at the facial region (defined by the mask $M_{face}$) by the \textbf{normal loss}:

\begin{equation}
    \begin{aligned}
    L_\mathrm{normal}(\phi, \D) = \Delta(N(\phi, \D) \cdot M_{face},\,\, N_i \cdot M_{face})
    \end{aligned}
\end{equation}

The composite loss is expressed as follows:

\begin{equation}
    \begin{aligned}
        L(&\phi, \D, C^\mathrm{room}, C^\mathrm{flash}, \mathcal{T}_A) = L_\mathrm{final}(\phi, \D, C^\mathrm{room}, C^\mathrm{flash})  \\
        &+ \alpha_\mathrm{normal} L_\mathrm{normal}(\phi, \D) + \alpha_\mathrm{symm} L_\mathrm{symm}(\phi, \D, \mathcal{T}_A)  \\
        &+ \alpha_\mathrm{cm} L_\mathrm{cm}(\phi, \D) + \alpha_\mathrm{TV} L_\mathrm{TV}(\phi, \D) \\
        &+ \alpha_\mathrm{mask} L_\mathrm{mask}(\phi, \D)
    \end{aligned}
\end{equation}

In our experiments, the separate losses were balanced as follows: $\alpha_\mathrm{normal} = 0.1,\, \alpha_\mathrm{symm} = 0.02,\, \alpha_\mathrm{cm} = 100,\, \alpha_\mathrm{TV} = 50,\, \alpha_\mathrm{mask} = 10^3$.

\paragraph{Optimization.} The learnable part of the system is trained for one scene by the backpropagation of the loss to the rendering network parameters $\phi$, point descriptors $\D$, and to the auxiliary parameters $C^\mathrm{room}, C^\mathrm{flash}, \mathcal{T}_A$. We use Adam~\cite{Kingma14} with the same learning rate for $\phi$ and $\D$, while the rest of learnable parameters feature different learning rates, empirically selected according to the range of their possible values. At each step, a sample training image is selected, and a forward pass followed by the gradient step is made. Train-time augmentations include random zoom in/out and subsequent cropping of a small patch.
\section{Experiments}

\subsection{Real head portraits}
\label{sec:exp_real}

We now show results for our method. Our experimental dataset consists of five sequences taken by a mobile camera of Samsung Galaxy S8 smartphone. Capturing was made in Open Camera app~\cite{OpenCamera20} with controlled white balance, 30 FPS, 1/200 s shutter speed and ISO of approximately 200 (selected for each sequence according to the lighting conditions). During each recording, the camera flash was blinking regularly once per second, each time turning on for $\sim$ 0.1 s. Each subject was photographed for 15-25 seconds and was asked to remain still during shooting. The individual frames were extracted from the resulting videos at 6 FPS, which included all the flashlighted photos. The camera poses and point clouds were reconstructed by Agisoft Metashape~\cite{Agisoft} with \textit{Highest} precision of cameras alignment and \textit{High} quality of point cloud reconstruction; the latter is a tradeoff between the geometry completeness and the number of points. Normals and meshes for the face part were collected by inferring PRNet~\cite{Feng18} with MTCNN~\cite{Zhang16} face detector for each of the captured frames. The meshes estimated by PRNet were smoothened by five iterations of Laplacian smoothing~\cite{Sorkine04}.

\begin{figure*}[h]
    % trim: left lower right upper
    \adjincludegraphics[trim={{.26\width} {.44\height} {.61\width} {.35\width}},clip,width=.135\linewidth]{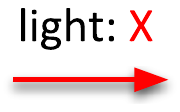}
    \adjincludegraphics[trim={{.26\width} {.44\height} {.61\width} {.35\width}},clip,width=.135\linewidth]{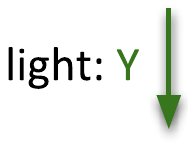}
    \adjincludegraphics[trim={{.26\width} {.44\height} {.61\width} {.35\width}},clip,width=.135\linewidth]{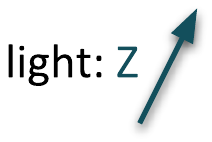}
    \hfill
    \adjincludegraphics[trim={{.26\width} {.44\height} {.61\width} {.35\width}},clip,width=.135\linewidth]{images/schemes/arrowsX.pdf}
    \adjincludegraphics[trim={{.26\width} {.44\height} {.61\width} {.35\width}},clip,width=.135\linewidth]{images/schemes/arrowsY.pdf}
    \hfill
    \adjincludegraphics[trim={{.26\width} {.44\height} {.61\width} {.35\width}},clip,width=.135\linewidth]{images/schemes/arrowsX.pdf}
    \adjincludegraphics[trim={{.26\width} {.44\height} {.61\width} {.35\width}},clip,width=.135\linewidth]{images/schemes/arrowsY.pdf}
    \hfill
    
    \adjincludegraphics[trim={0 0 0 {.3\height}},clip,width=.135\linewidth]{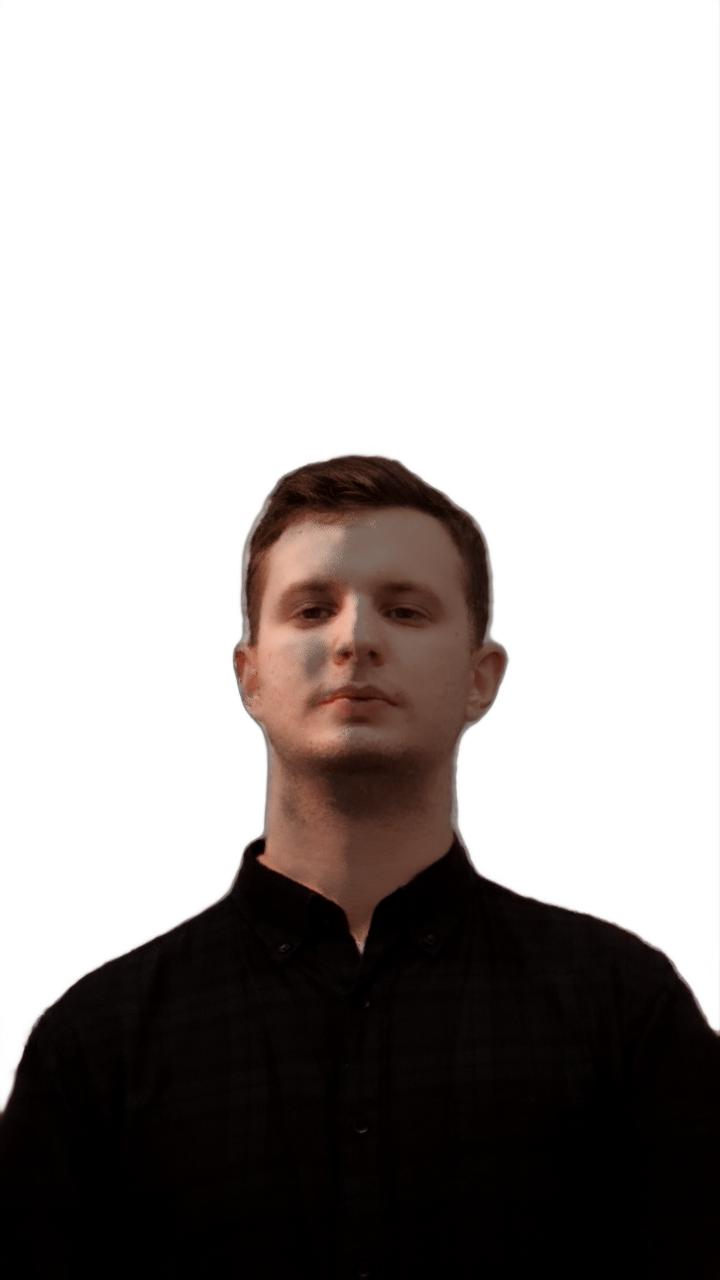}
    \adjincludegraphics[trim={0 0 0 {.3\height}},clip,width=.135\linewidth]{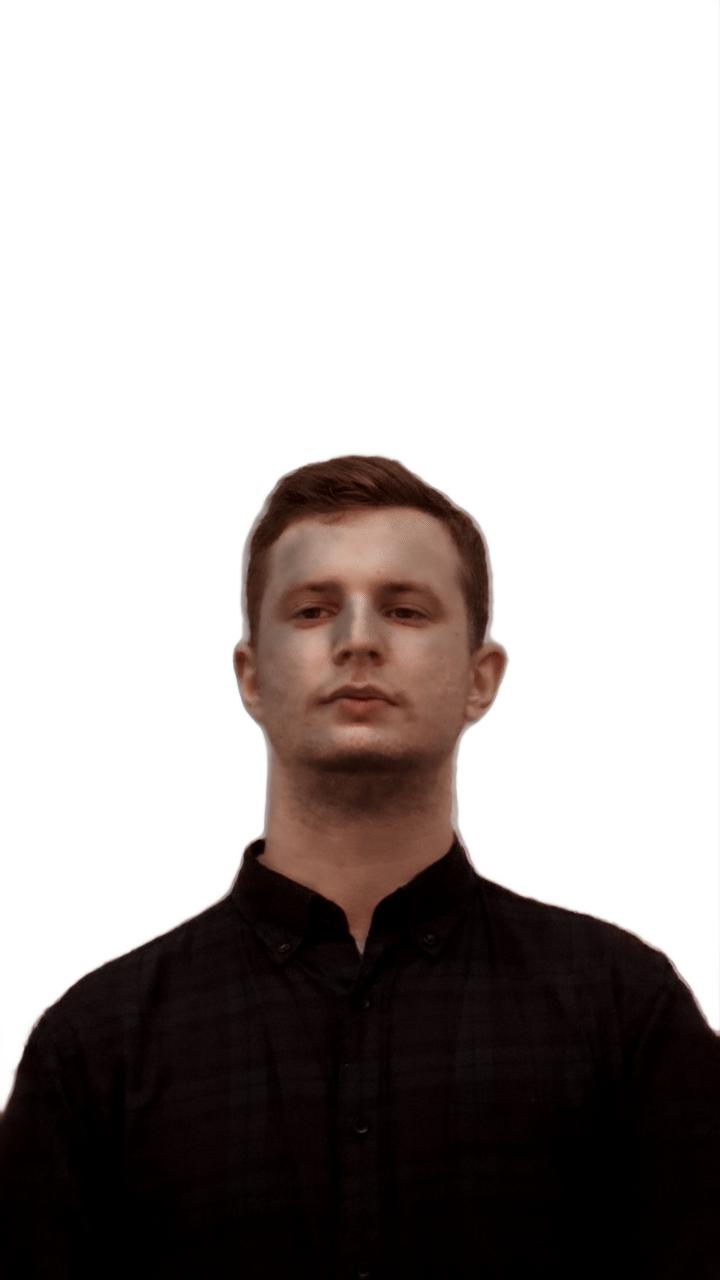}
    \adjincludegraphics[trim={0 0 0 {.3\height}},clip,width=.135\linewidth]{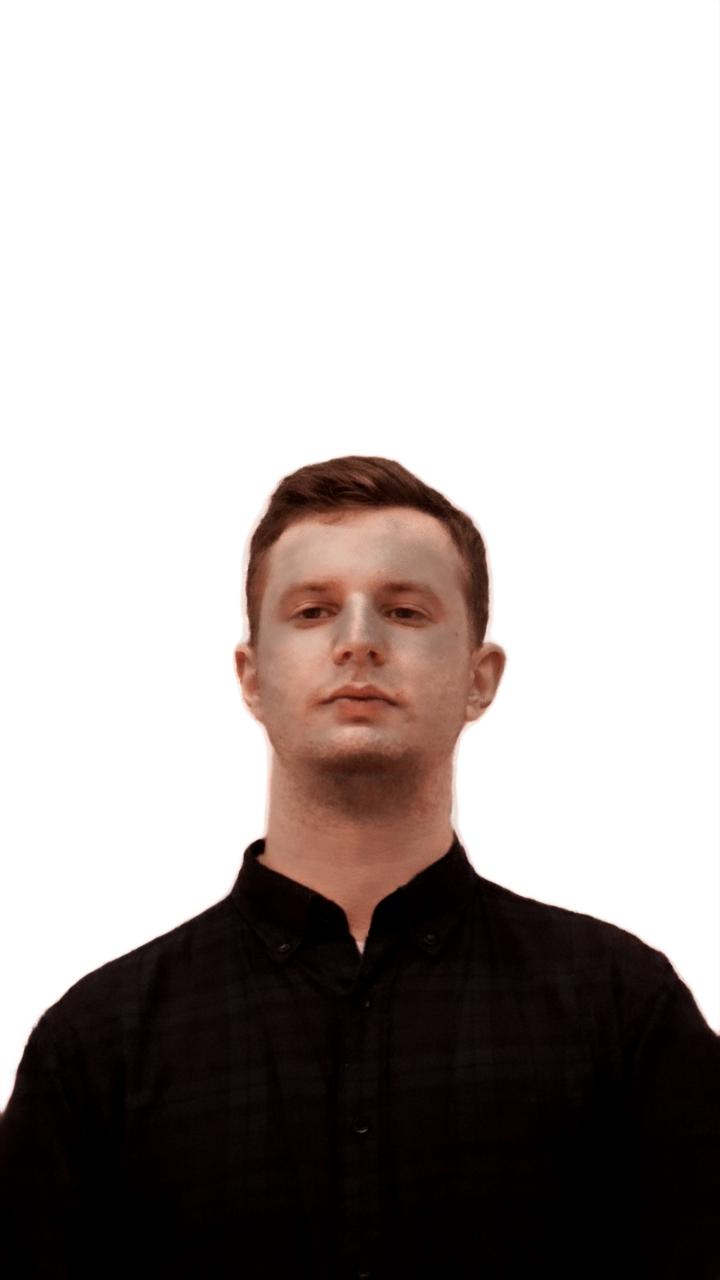}
    \hfill
    \adjincludegraphics[trim={0 0 0 {.3\height}},clip,width=.135\linewidth]{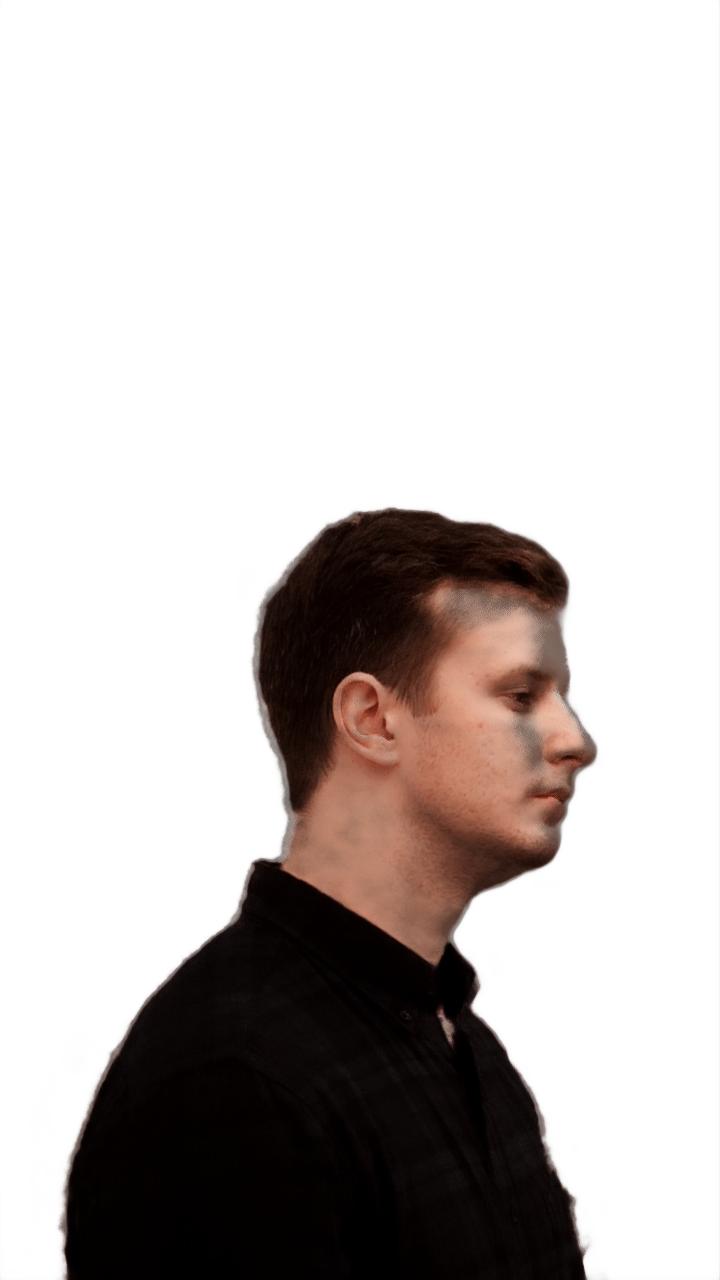}
    \adjincludegraphics[trim={0 0 0 {.3\height}},clip,width=.135\linewidth]{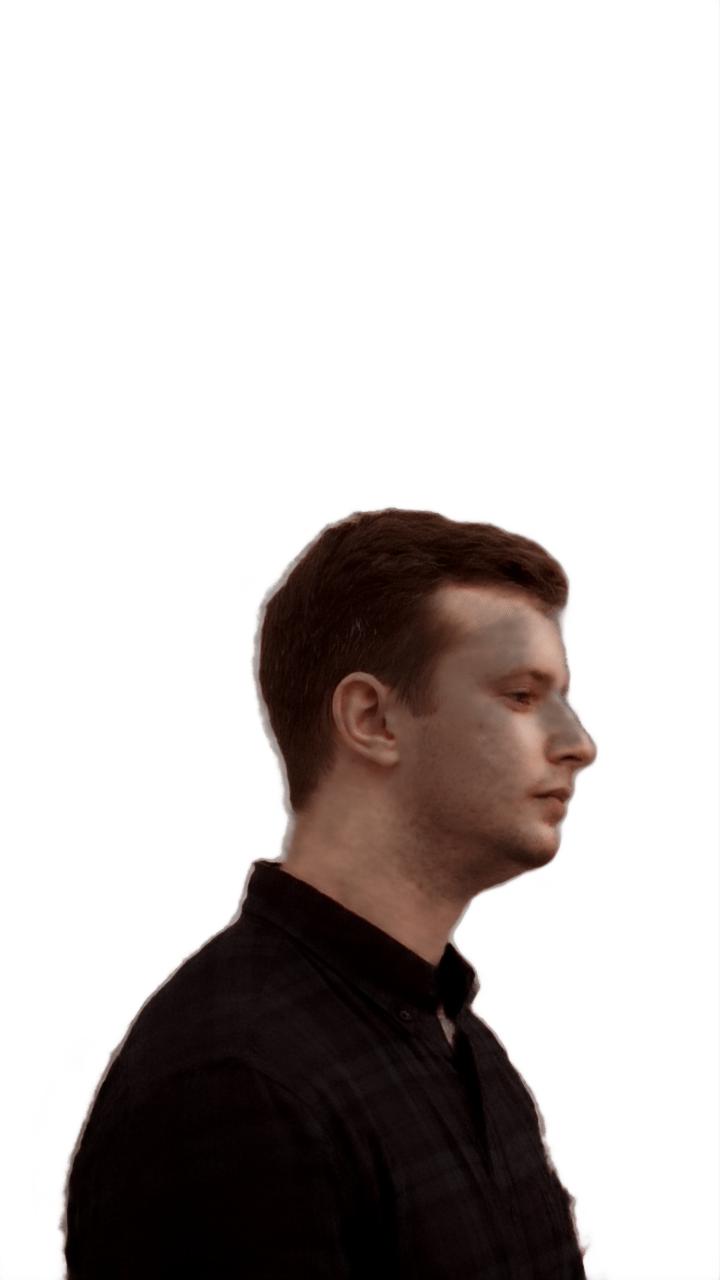}
    \hfill
    \adjincludegraphics[trim={0 0 0 {.3\height}},clip,width=.135\linewidth]{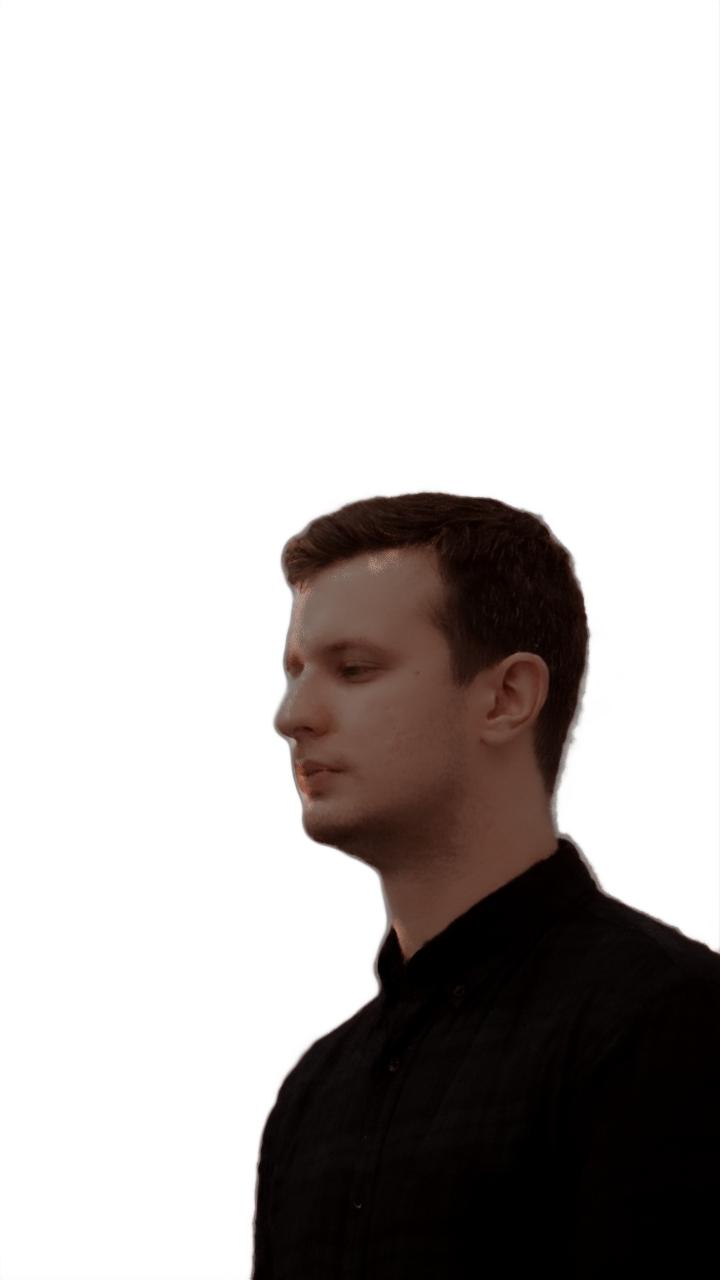}
    \adjincludegraphics[trim={0 0 0 {.3\height}},clip,width=.135\linewidth]{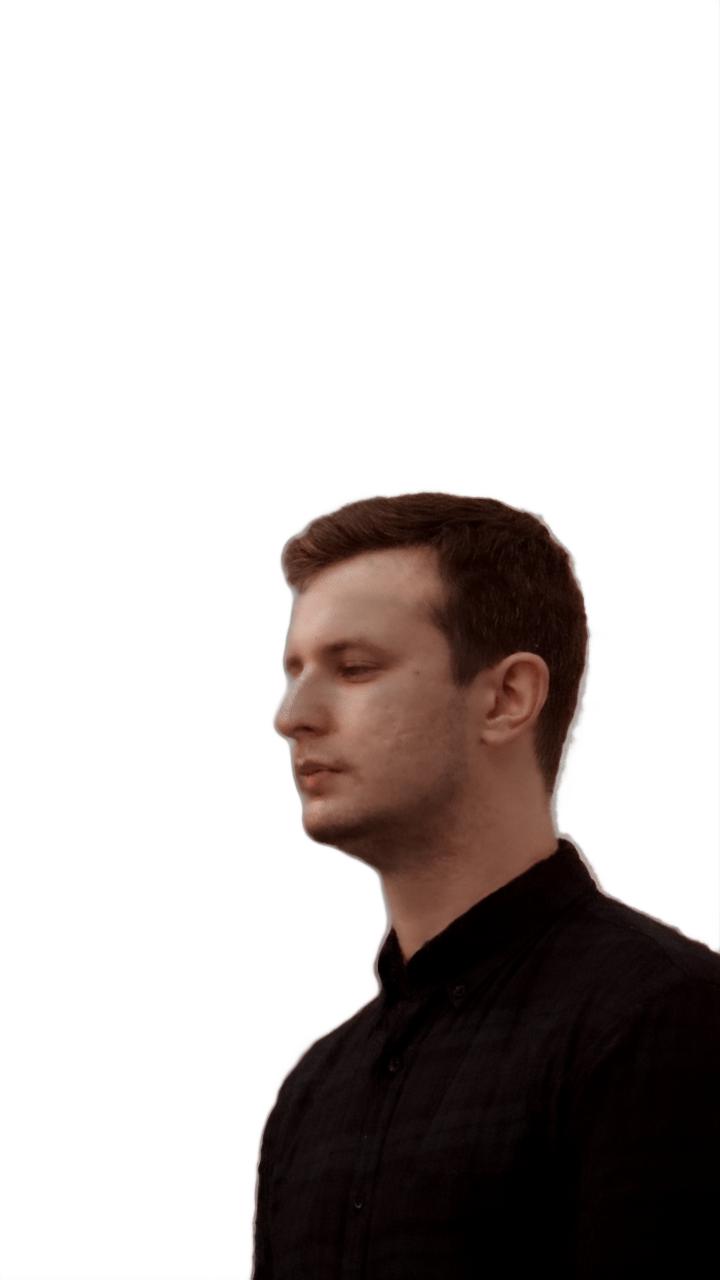}
    
    \caption{Qualitative comparison of the smartphone-captured real head portraits rendering under several novel viewpoints and novel (non-colocated) directional light. Light directions in the top row are defined in the \textbf{world space} coordinate axes associated with the frontal head view. The people are additionally lighted by constant ambient (indirectly emitted) light for better appearance. \textit{Electronic zoom-in recommended.}}
    \label{fig:real_relighting}
\end{figure*}

For each subject, ten frames (five flashlighted and five non-flashlighted) covering a head from all sides were selected for validation, while the rest were used for training. Our method is fitted to each sequence separately. The pipeline is trained for 80'000 steps. At each step, a random frame (and the respective camera viewpoint) is selected, and a random patch of size $512 \times 512$ is extracted by sampling the patch center inside from the foreground region of the ground truth mask. In Fig.~\ref{fig:real_relighting}, the results of simultaneous relighting and view interpolation are depicted. For this evaluation, we show the results at novel viewpoints and directional+ambient lighting (three sample light directions were selected as up, right and forward vectors naturally associated with the frontal head view). The final rendered image under such conditions is given by: $\mathcal{I}' = A \cdot \left( \alpha + (1 - \alpha) \cdot \lfloor \langle N, -\omega \rangle \rfloor \right)$, where $\alpha$ stands for the ambient light, emitted regardless of the normal direction ($\alpha=0.5$ was taken for the evaluation), $\omega$ defines the novel light direction, and the dot product $\langle N, -\omega \rangle$ is applied pixel-wise. Note that the normals are challenging to estimate for regions which remain completely black under the flashlight (e.g. a black jacket).

% \begin{equation}
%     \begin{aligned}
%         \mathcal{I}' = A \cdot \left( \alpha + (1 - \alpha) \cdot \lfloor (N, -\omega) \rfloor \right),
%     \end{aligned}
%     \label{eq:dir_amb_lighting}
% \end{equation} 

\begin{figure*}[h]
    % \adjincludegraphics[trim={{.26\width} {.44\height} {.61\width} {.35\width}},clip,width=.135\linewidth]{images/schemes/arrowsX.pdf}
    % \adjincludegraphics[trim={{.26\width} {.44\height} {.61\width} {.35\width}},clip,width=.135\linewidth]{images/schemes/arrowsY.pdf}
    % \adjincludegraphics[trim={{.26\width} {.44\height} {.61\width} {.35\width}},clip,width=.135\linewidth]{images/schemes/arrowsZ.pdf}
    % \hfill
    % \adjincludegraphics[trim={{.26\width} {.44\height} {.61\width} {.35\width}},clip,width=.135\linewidth]{images/schemes/arrowsX.pdf}
    % \adjincludegraphics[trim={{.26\width} {.44\height} {.61\width} {.35\width}},clip,width=.135\linewidth]{images/schemes/arrowsY.pdf}
    % \hfill
    % \adjincludegraphics[trim={{.26\width} {.44\height} {.61\width} {.35\width}},clip,width=.135\linewidth]{images/schemes/arrowsX.pdf}
    % \adjincludegraphics[trim={{.26\width} {.44\height} {.61\width} {.35\width}},clip,width=.135\linewidth]{images/schemes/arrowsY.pdf}
    % \hfill
    
    % 1
    \adjincludegraphics[trim={0 0 0 {.3\height}},clip,width=.135\linewidth]{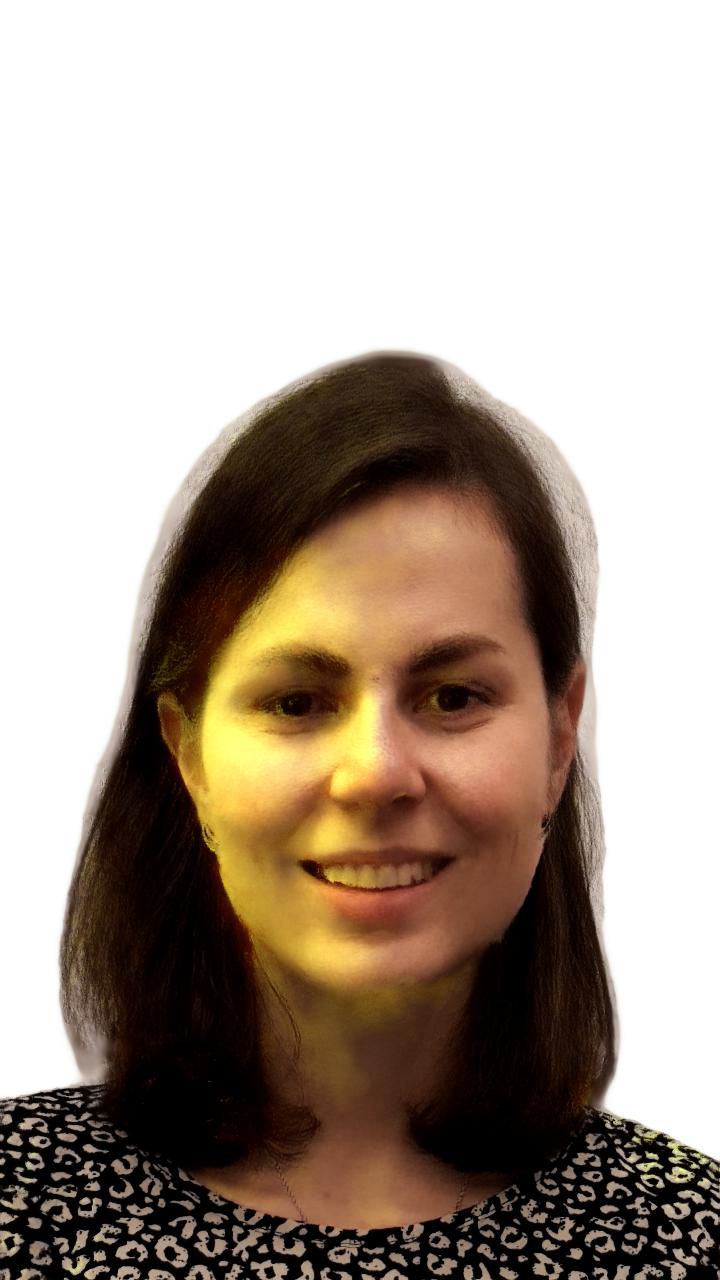}
    \adjincludegraphics[trim={0 0 0 {.3\height}},clip,width=.135\linewidth]{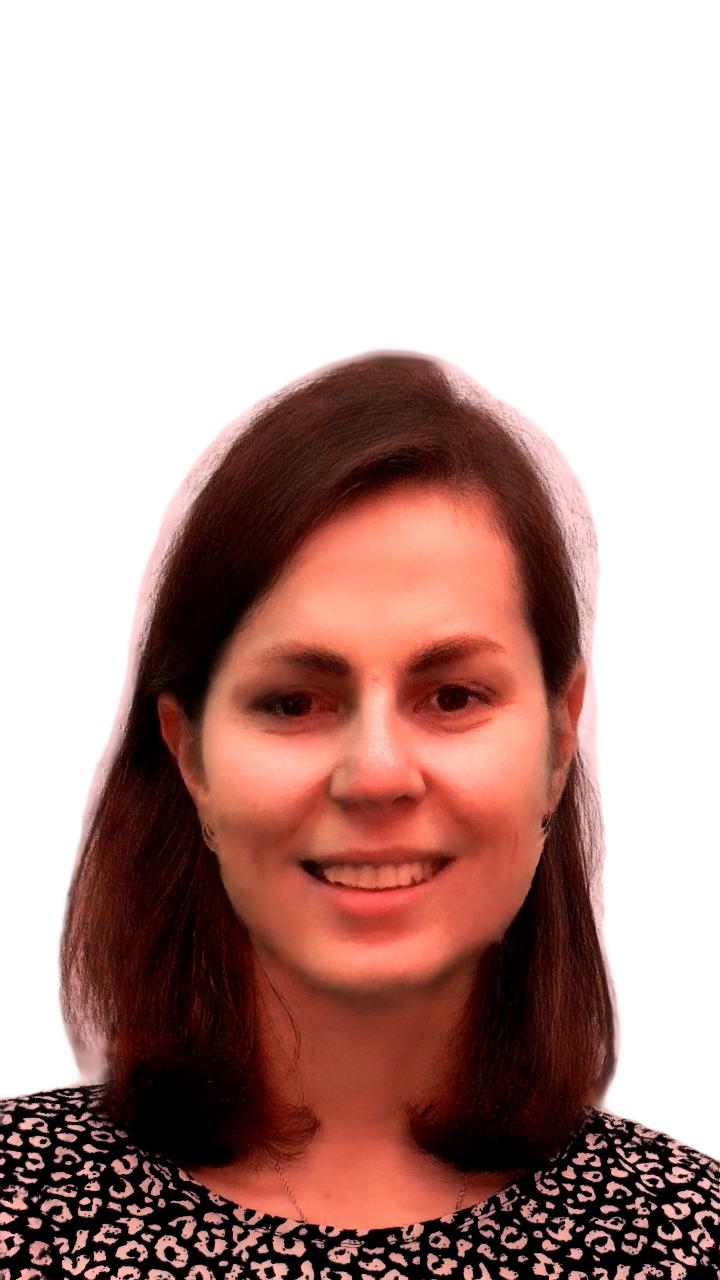}
    \adjincludegraphics[trim={0 0 0 {.3\height}},clip,width=.135\linewidth]{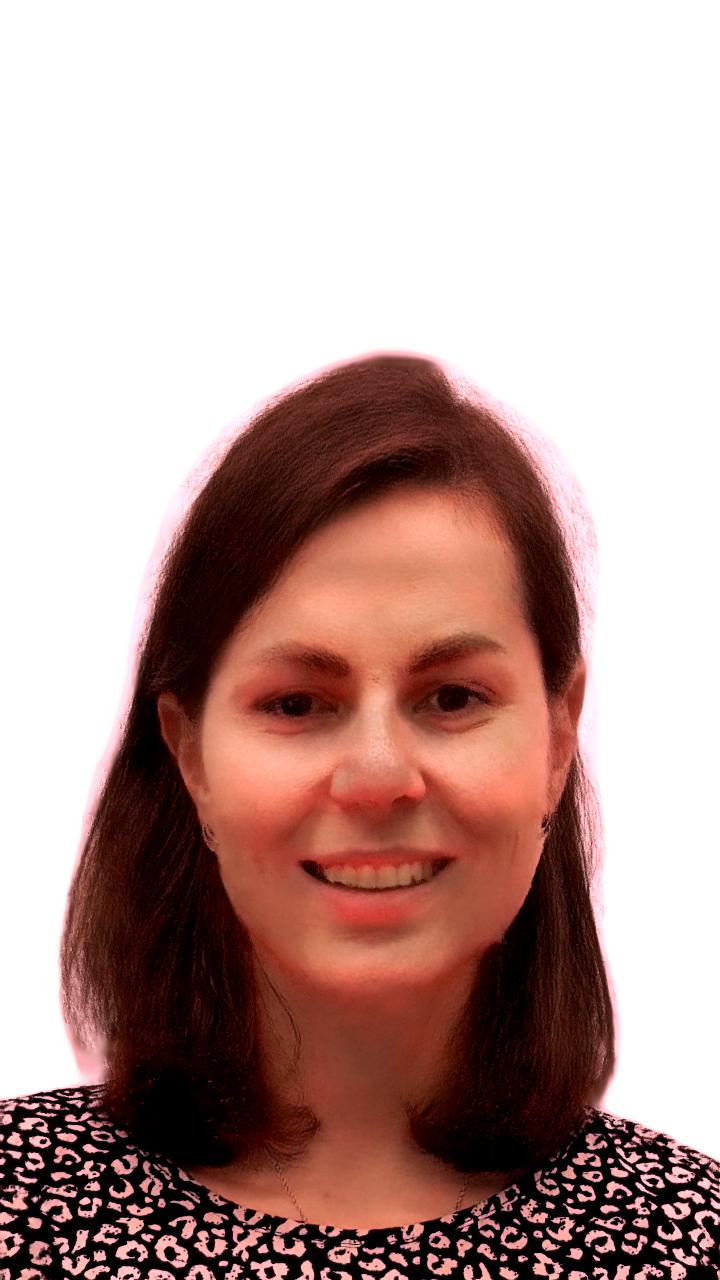}
    \hfill
    \adjincludegraphics[trim={0 0 0 {.3\height}},clip,width=.135\linewidth]{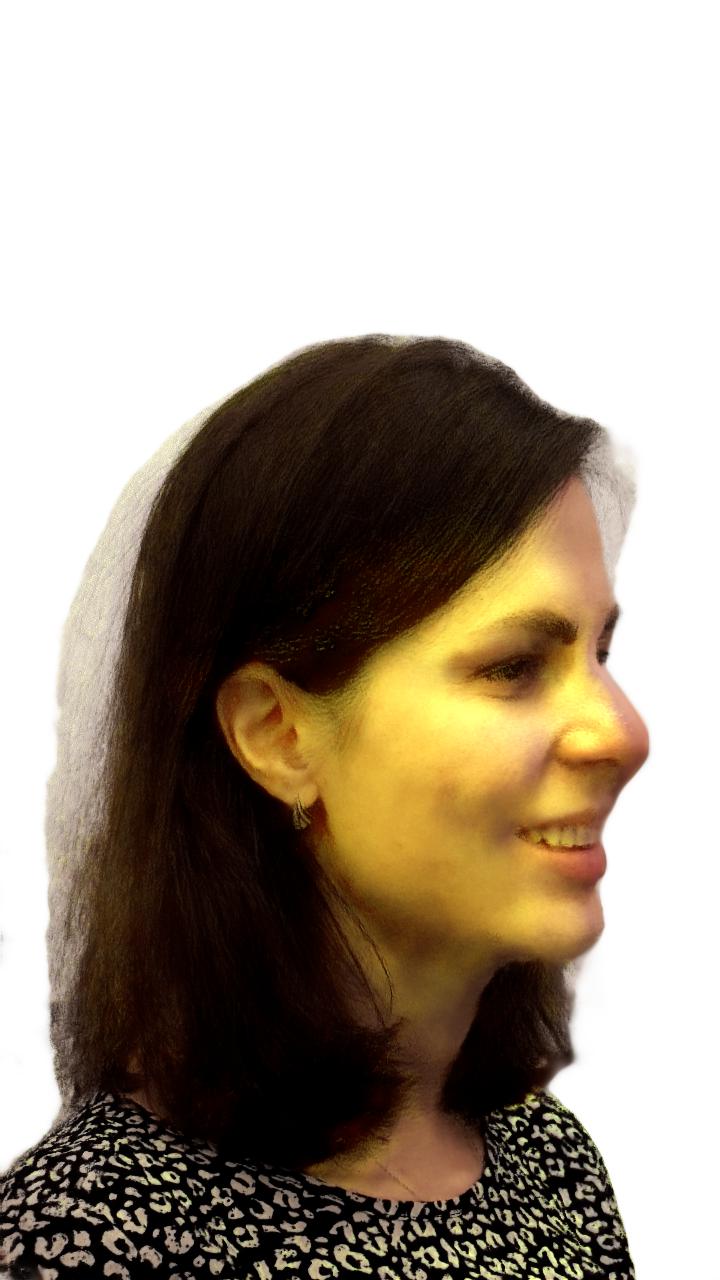}
    \adjincludegraphics[trim={0 0 0 {.3\height}},clip,width=.135\linewidth]{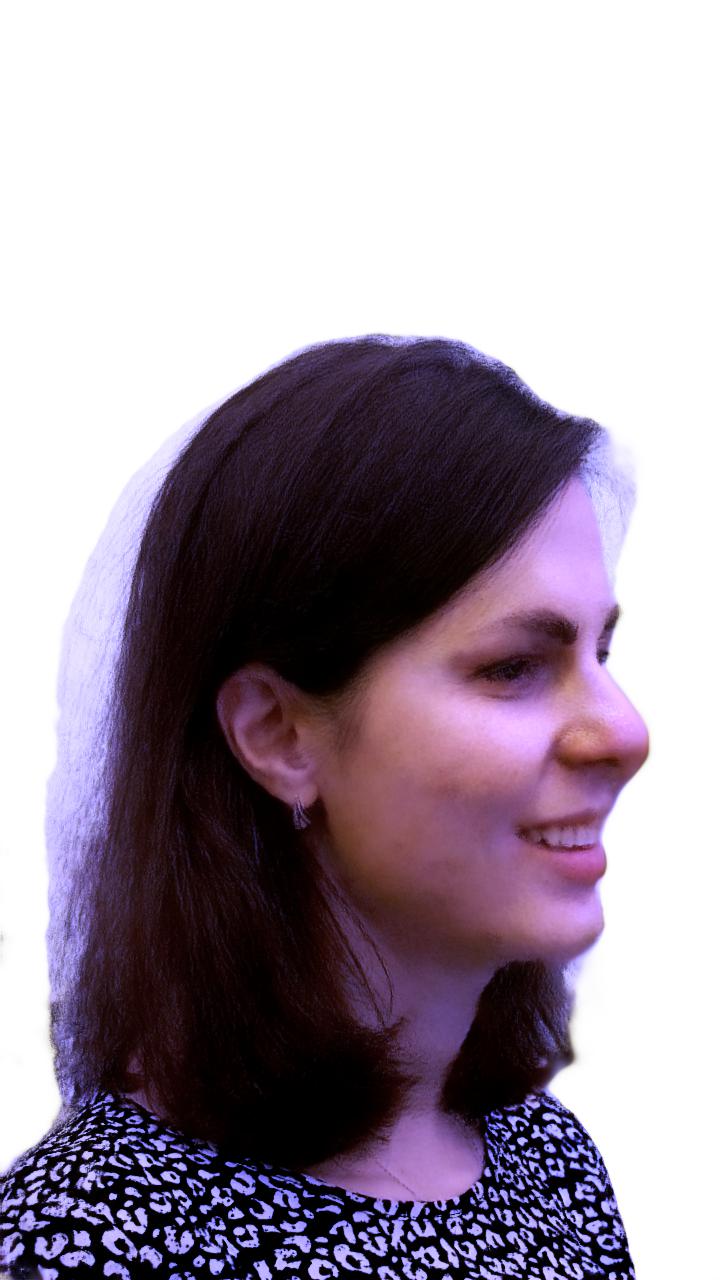}
    % \adjincludegraphics[trim={0 0 0 {.3\height}},clip,width=.135\linewidth]{images/real/Person4_Lprof_light_to_forward_additional_lighting.jpg}
    \hfill
    \adjincludegraphics[trim={0 0 0 {.3\height}},clip,width=.135\linewidth]{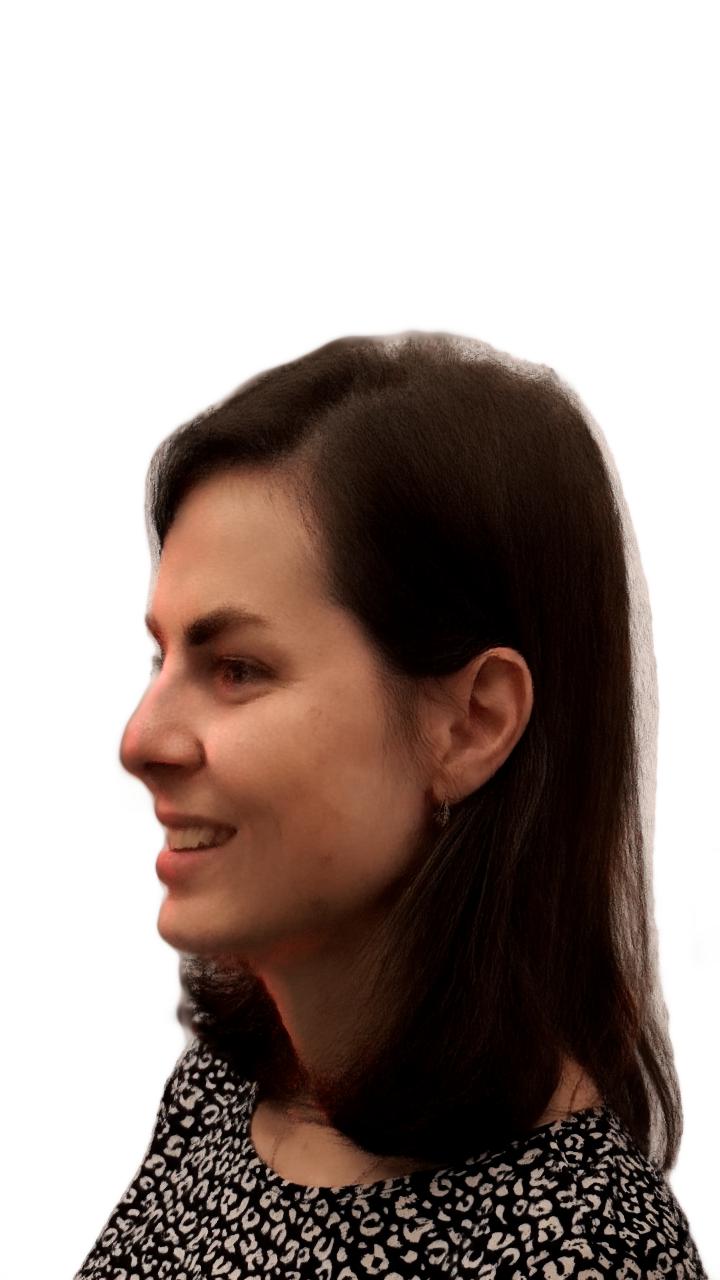}
    \adjincludegraphics[trim={0 0 0 {.3\height}},clip,width=.135\linewidth]{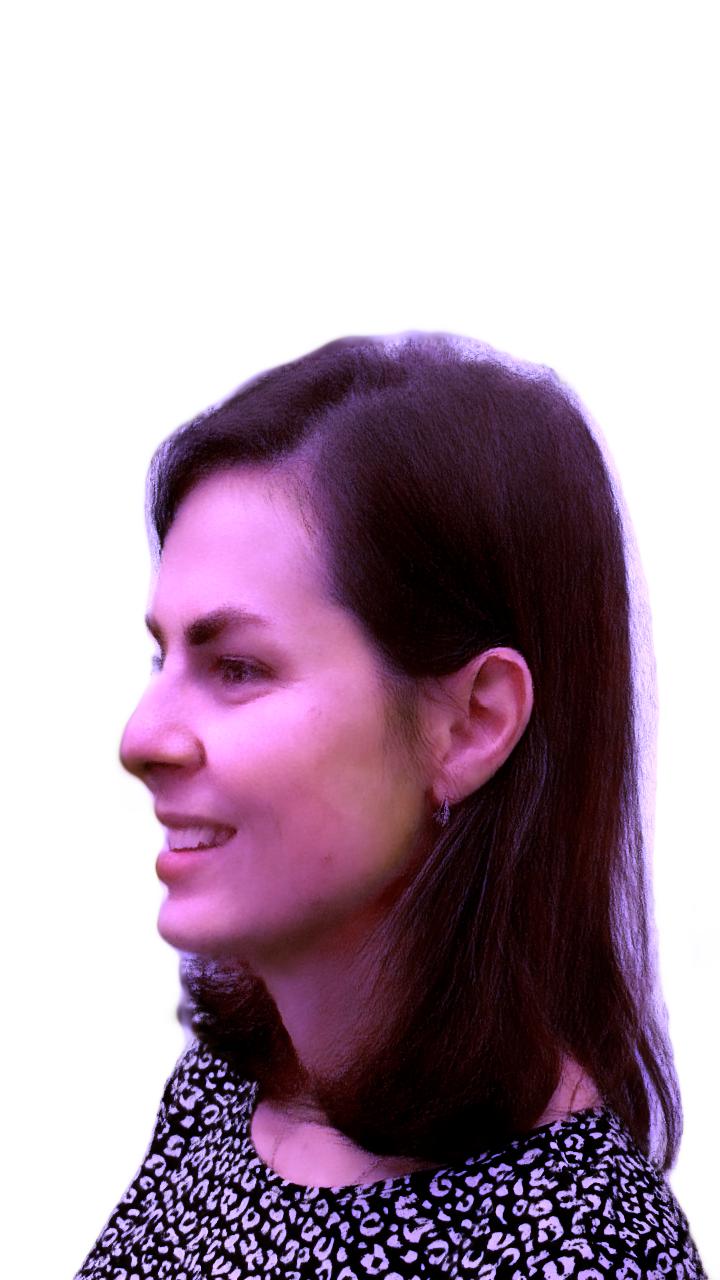}
    % \adjincludegraphics[trim={0 0 0 {.3\height}},clip,width=.135\linewidth]{images/real/Person4_Rprof_light_to_forward_additional_lighting.jpg}
    
    % 2
    \adjincludegraphics[trim={0 0 0 {.3\height}},clip,width=.135\linewidth]{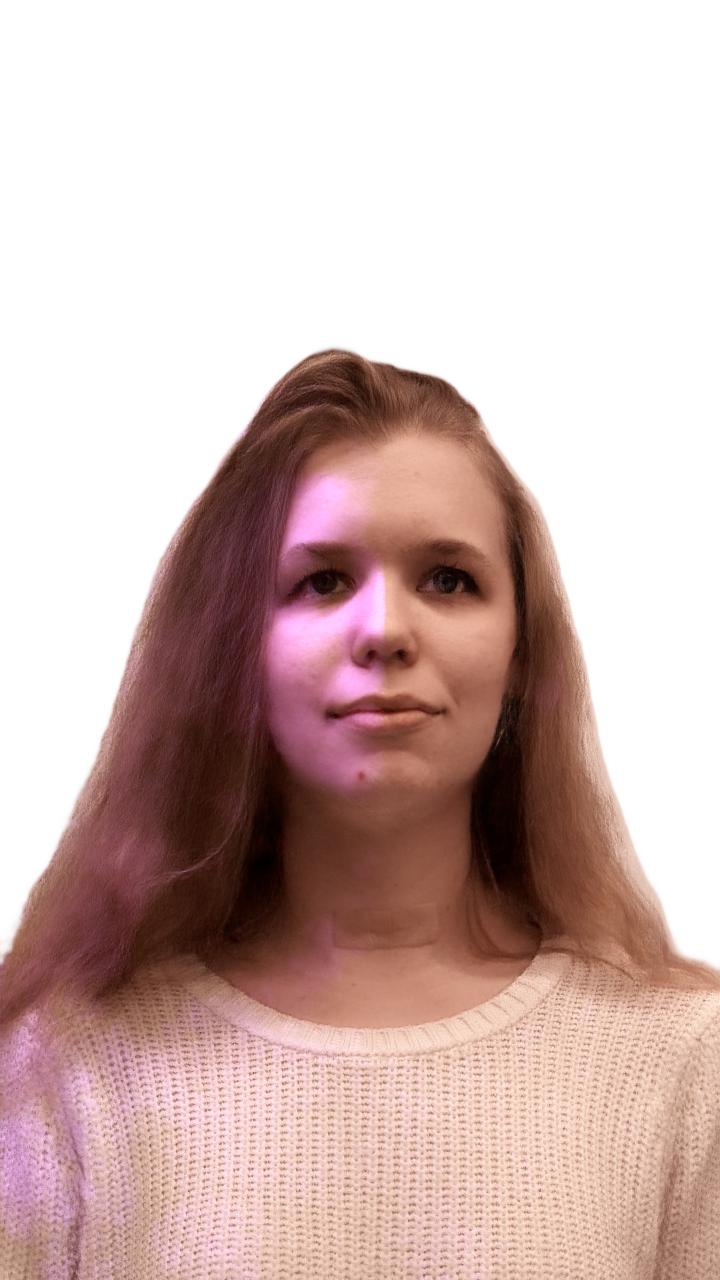}
    \adjincludegraphics[trim={0 0 0 {.3\height}},clip,width=.135\linewidth]{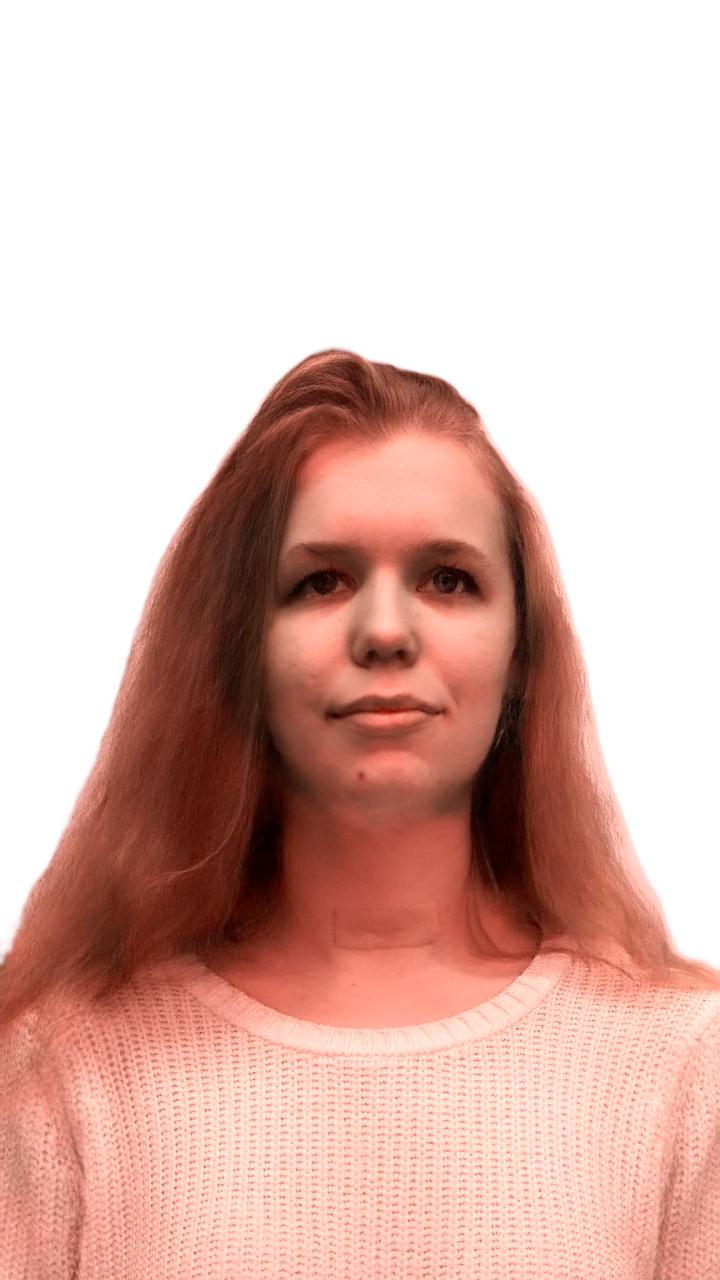}
    \adjincludegraphics[trim={0 0 0 {.3\height}},clip,width=.135\linewidth]{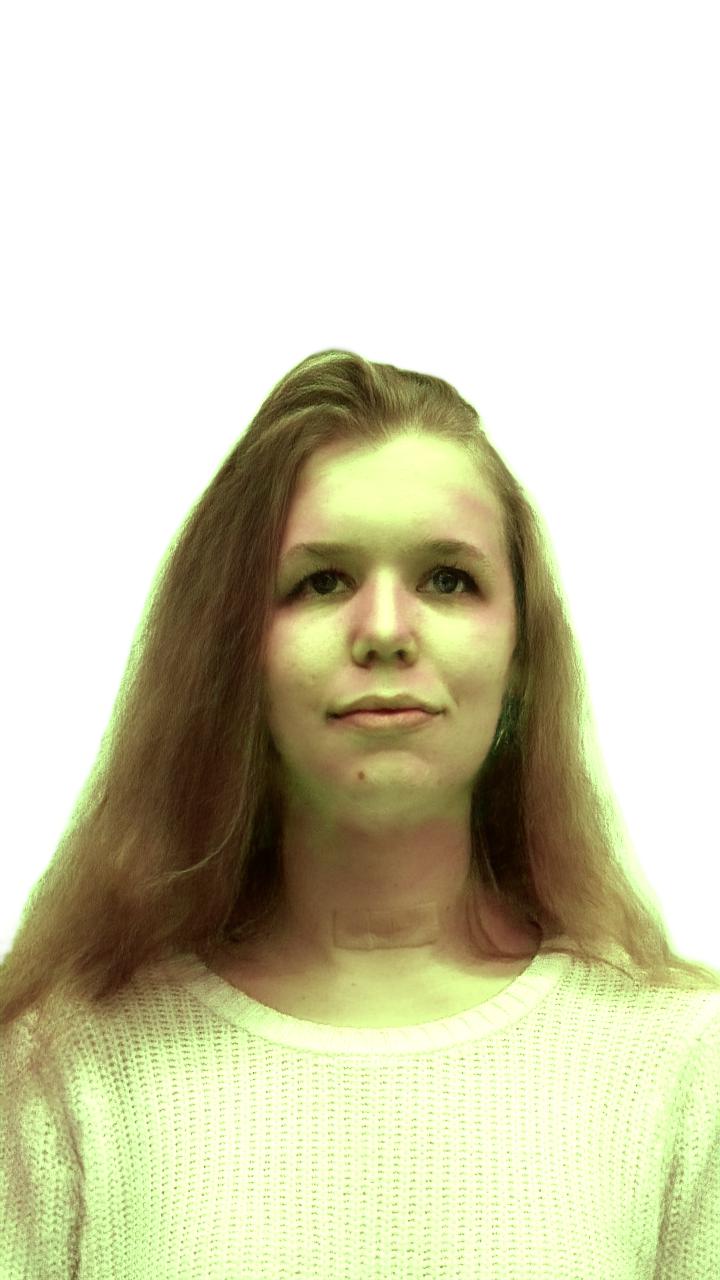}
    \hfill
    \adjincludegraphics[trim={0 0 0 {.3\height}},clip,width=.135\linewidth]{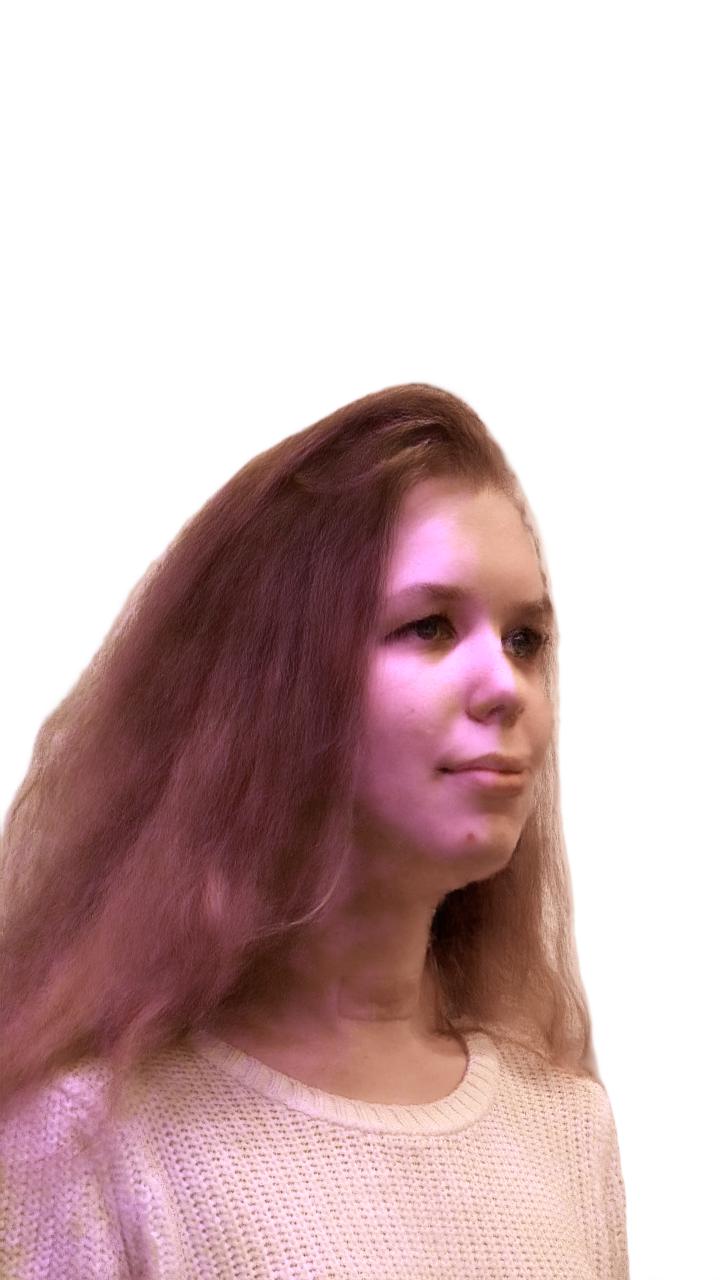}
    \adjincludegraphics[trim={0 0 0 {.3\height}},clip,width=.135\linewidth]{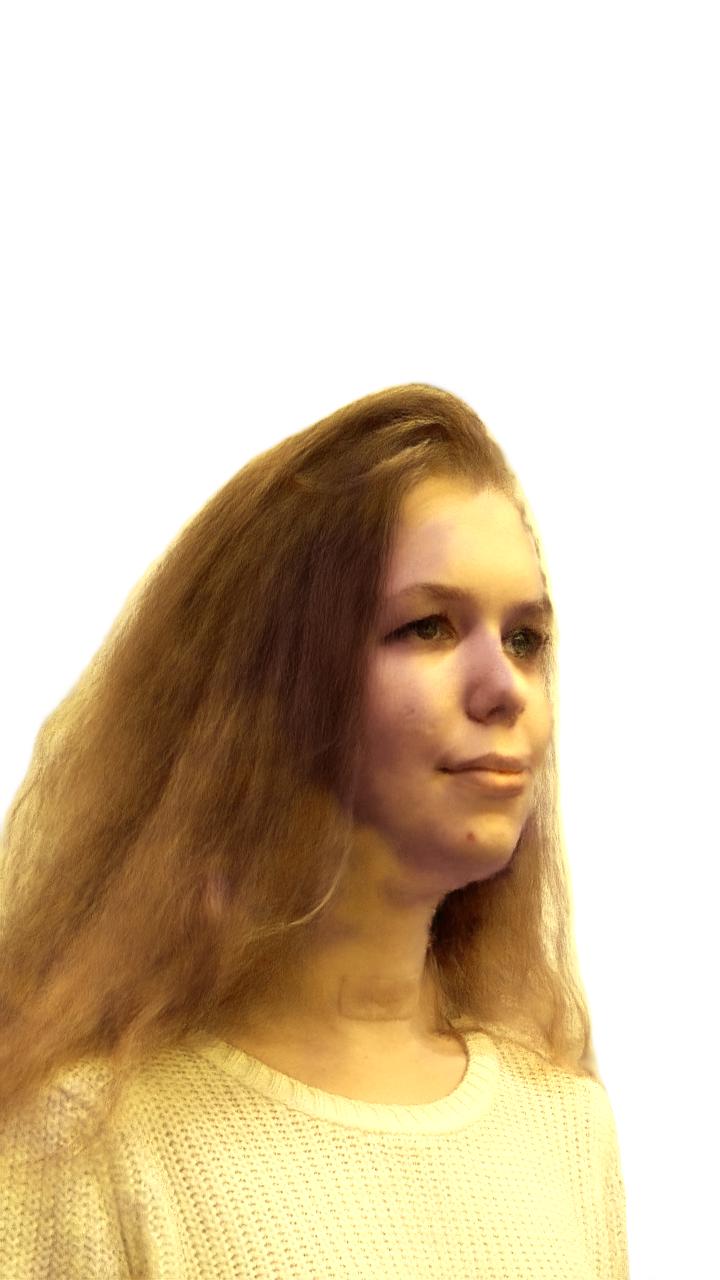}
    % \adjincludegraphics[trim={0 0 0 {.3\height}},clip,width=.135\linewidth]{images/real/Person5_Lprof_light_to_forward_additional_lighting.jpg}
    \hfill
    \adjincludegraphics[trim={0 0 0 {.3\height}},clip,width=.135\linewidth]{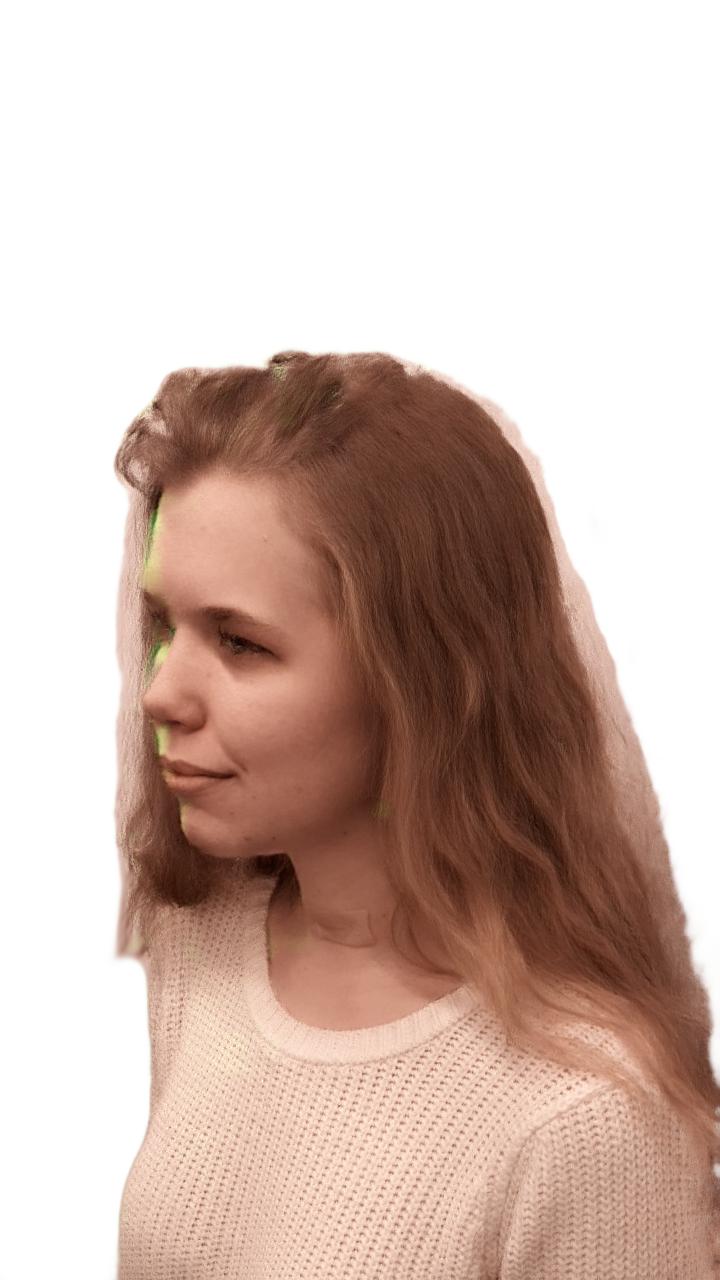}
    \adjincludegraphics[trim={0 0 0 {.3\height}},clip,width=.135\linewidth]{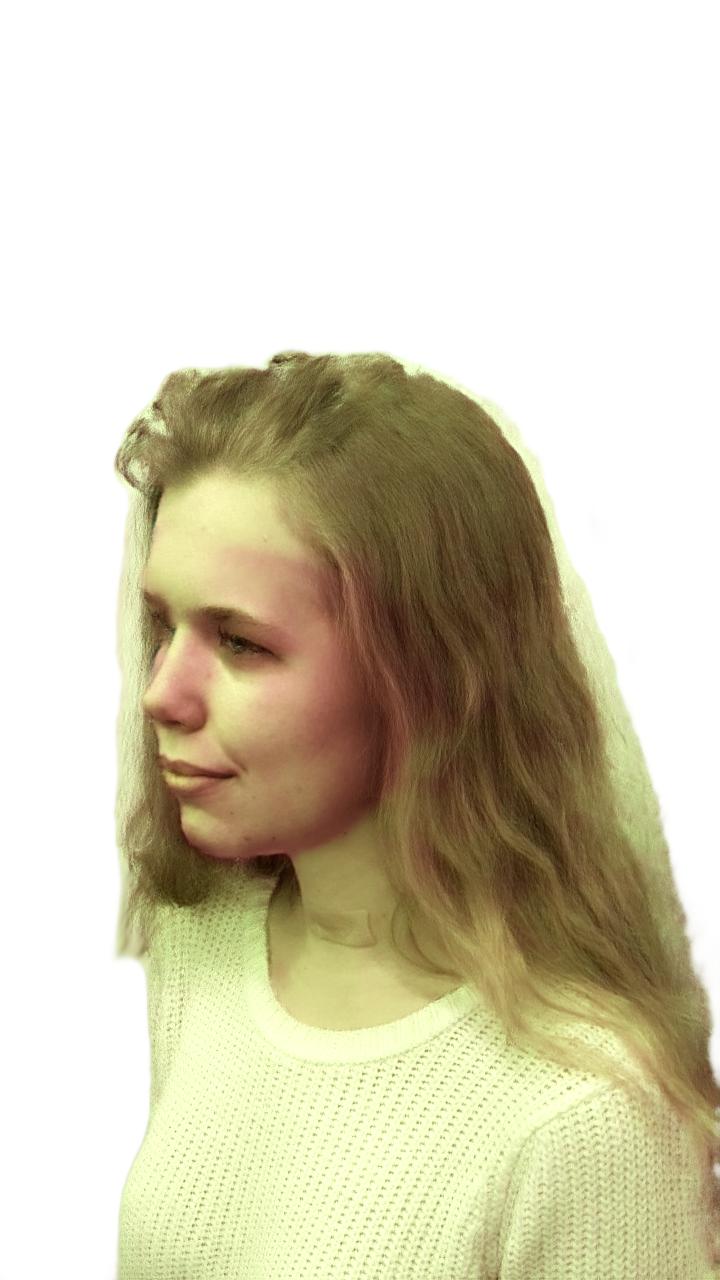}
    % \adjincludegraphics[trim={0 0 0 {.3\height}},clip,width=.135\linewidth]{images/real/Person5_Rprof_light_to_forward_additional_lighting.jpg}
    
    % 3
    \adjincludegraphics[trim={0 0 0 {.3\height}},clip,width=.135\linewidth]{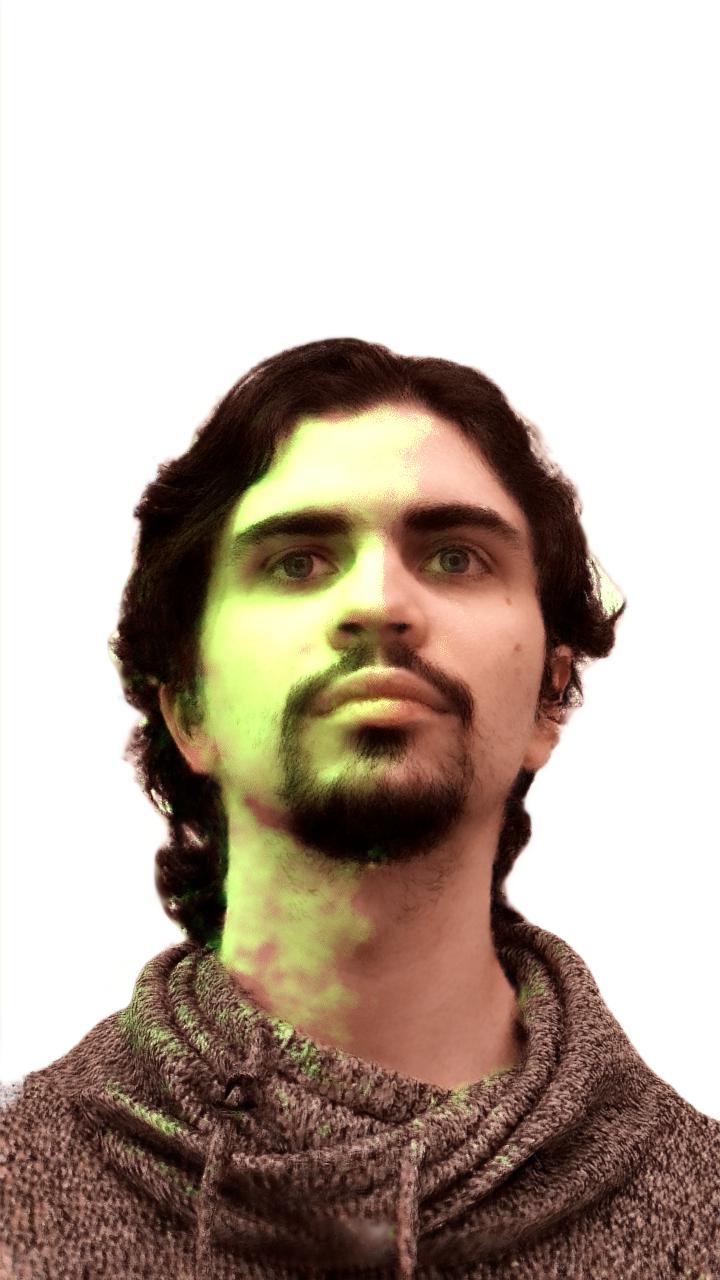}
    \adjincludegraphics[trim={0 0 0 {.3\height}},clip,width=.135\linewidth]{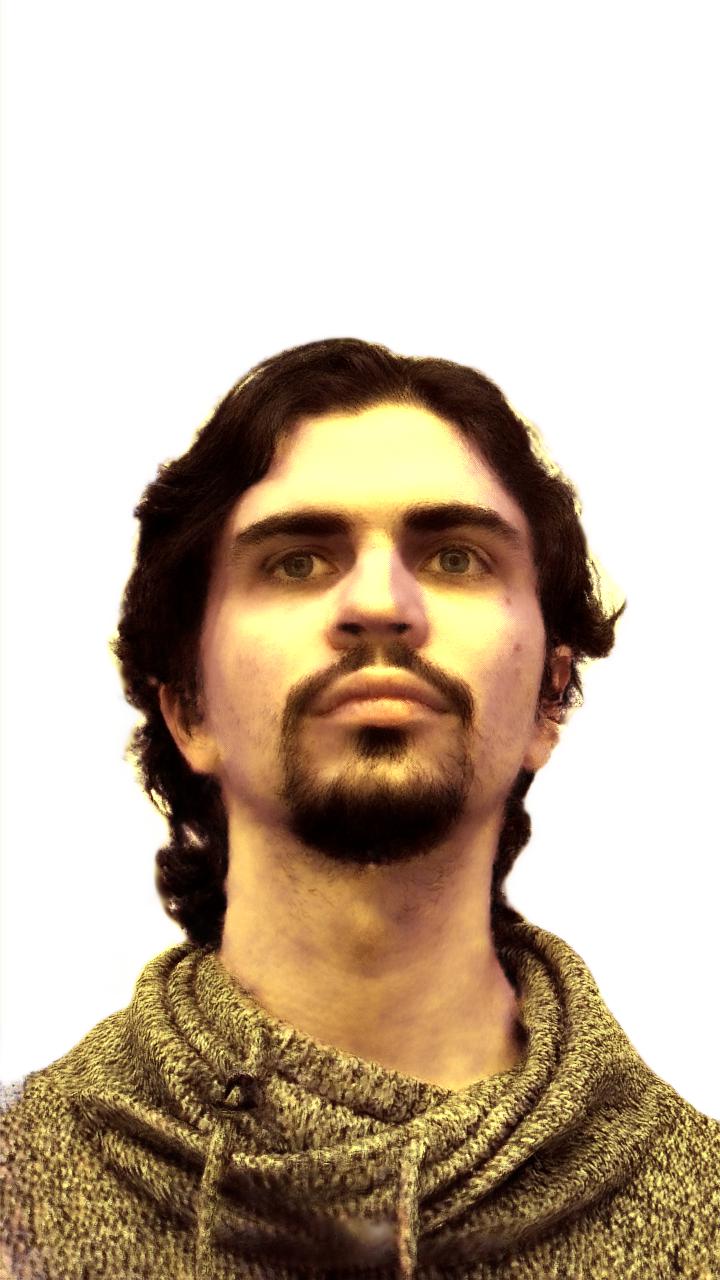}
    \adjincludegraphics[trim={0 0 0 {.3\height}},clip,width=.135\linewidth]{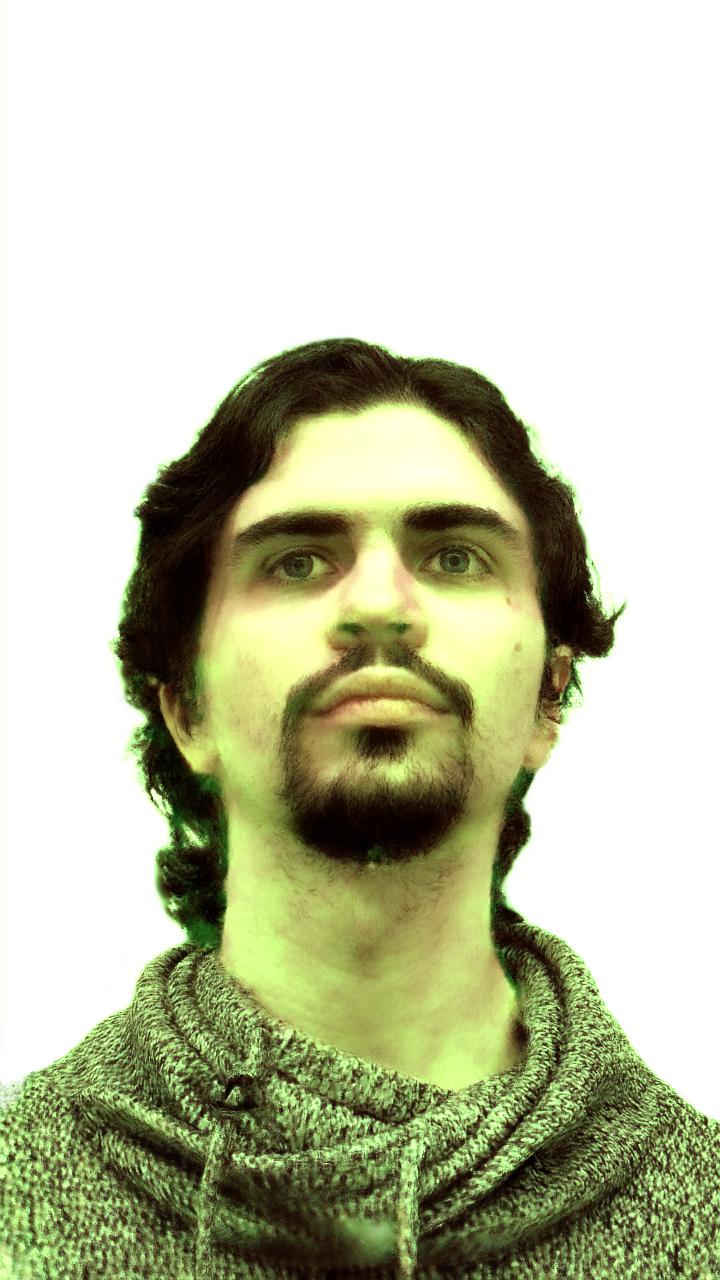}
    \hfill
    \adjincludegraphics[trim={0 0 0 {.3\height}},clip,width=.135\linewidth]{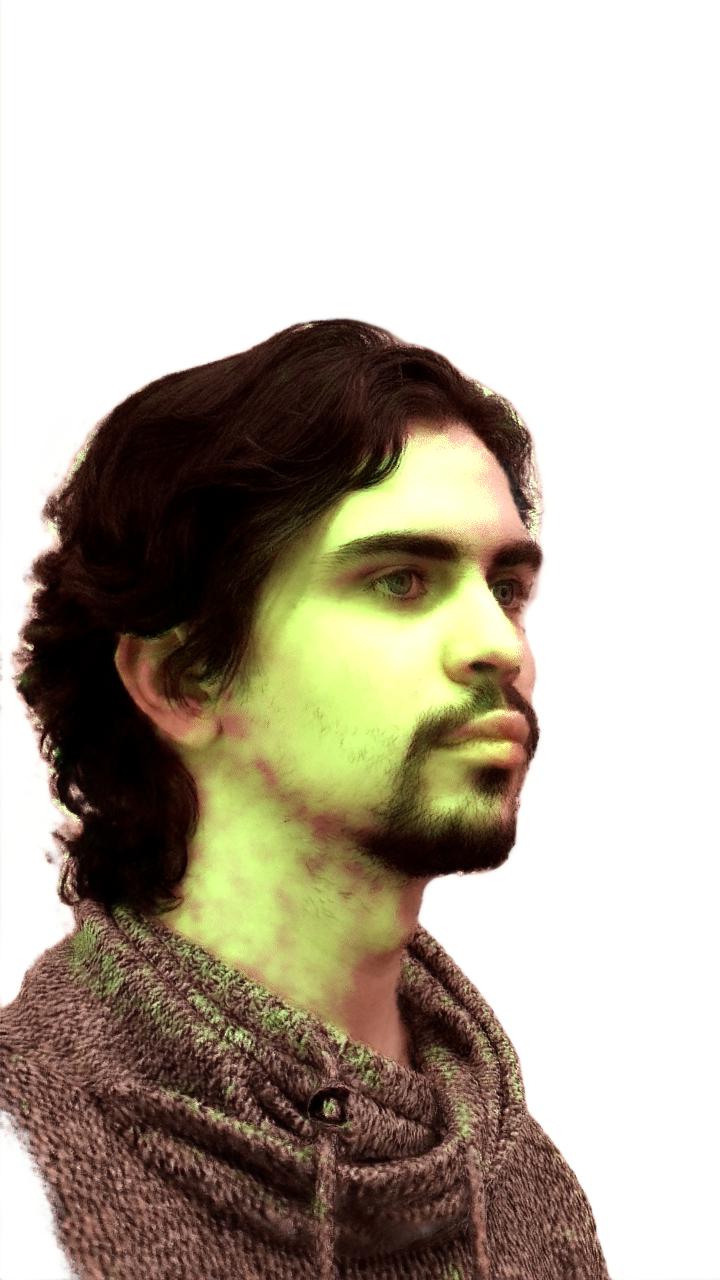}
    \adjincludegraphics[trim={0 0 0 {.3\height}},clip,width=.135\linewidth]{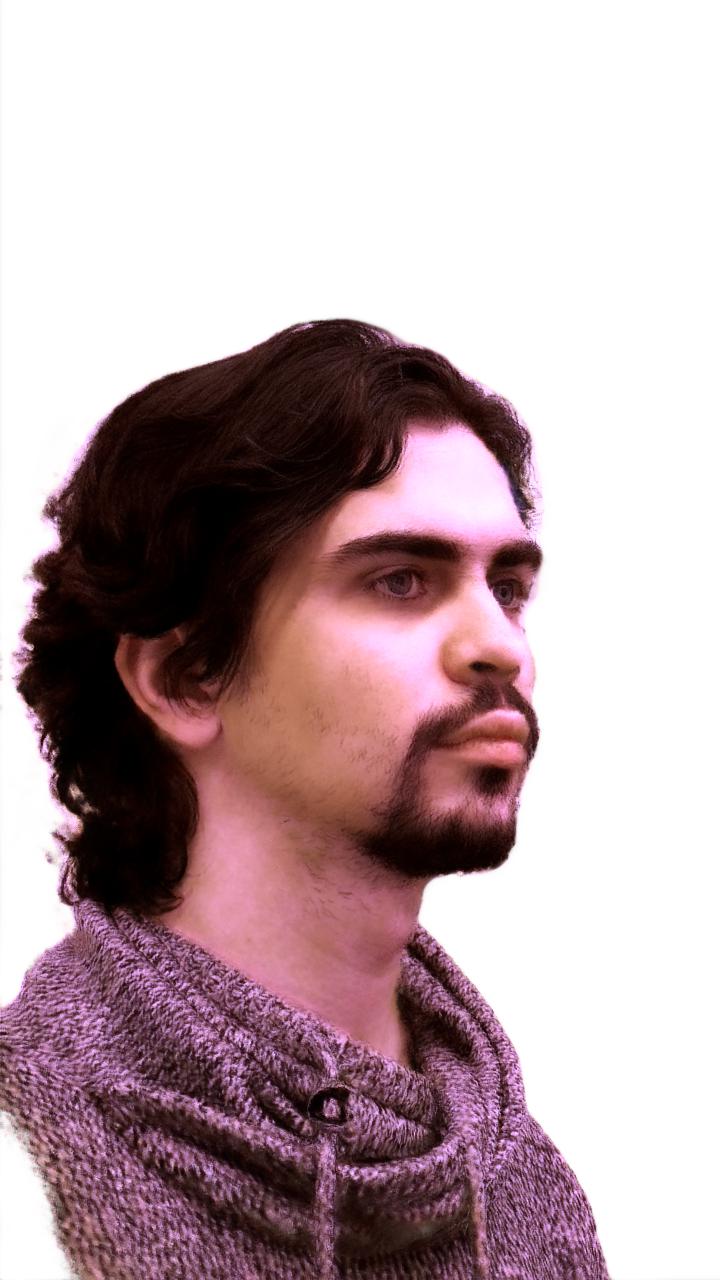}
    % \adjincludegraphics[trim={0 0 0 {.3\height}},clip,width=.135\linewidth]{images/real/Person2_Lprof_light_to_forward_additional_lighting.jpg}
    \hfill
    \adjincludegraphics[trim={0 0 0 {.3\height}},clip,width=.135\linewidth]{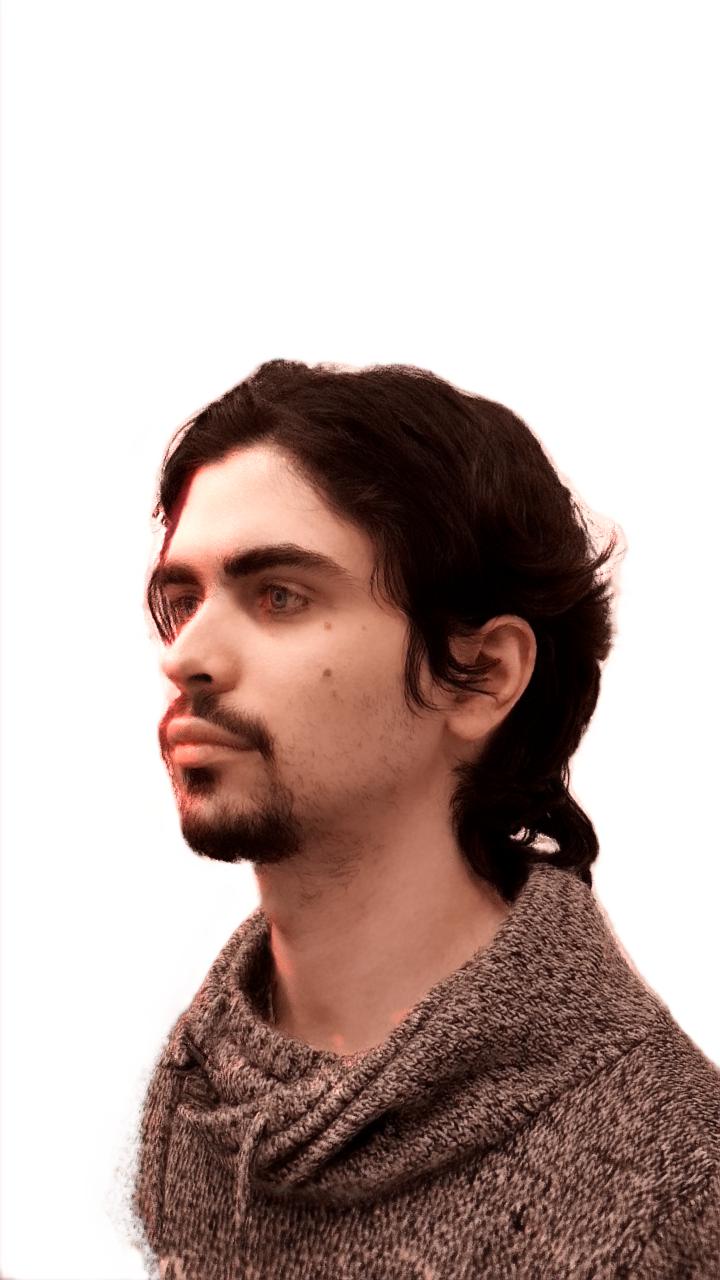}
    \adjincludegraphics[trim={0 0 0 {.3\height}},clip,width=.135\linewidth]{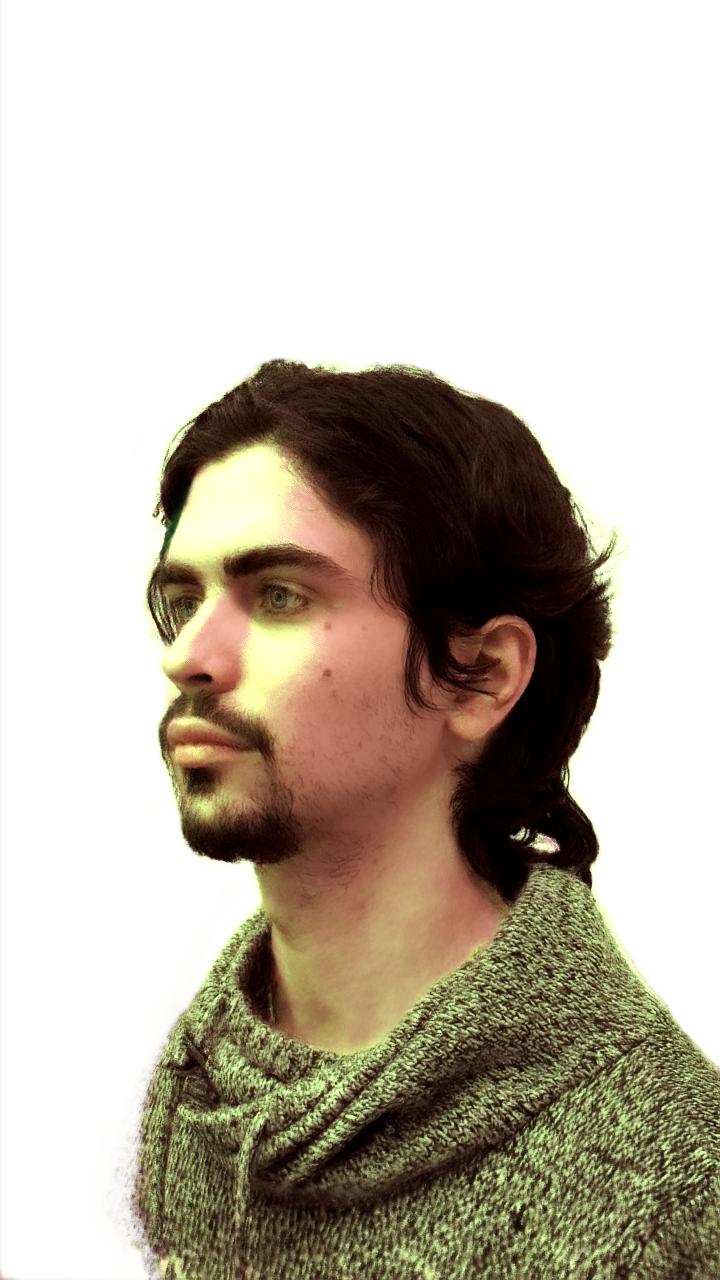}
    % \adjincludegraphics[trim={0 0 0 {.3\height}},clip,width=.135\linewidth]{images/real/Person3_Rprof_light_to_forward_additional_lighting.jpg}

    \caption{Adding the light to the original sequences. Compared to the previous figure (\textit{relighting} case), here we investigate \textit{additional lighting} -- the case when a point light is introduced while the room lighting is left in place. The color of a point light is chosen randomly for each rendering among red, green, blue, purple, yellow, and brown. The light directions are the same as in Fig.~\ref{fig:real_relighting}. The results are for the viewpoints from validation. \textit{Electronic zoom-in recommended.}}
    \label{fig:real_additional_lighting}
\end{figure*}

A similar comparison is shown in Fig.~\ref{fig:real_additional_lighting} for the case of \textit{additional lighting}, i.e.\ adding a new point light source, similar to flash light, to the captured environmental lighting of the room. In this case, a head is relighted by plugging novel $d$ (distance from the novel light source) and novel $\omega_o$ (direction of the novel light source) into~(\ref{eq:rendering_eq_3}). The renderings can look more appealing in the additional lighting setting than in the relighting setting in certain cases, as the estimated albedo is only rendered here in conjunction with soft shadows and shading.

\begin{figure*}[h]
    \centering
    \rotatebox{90}{ }
    \hspace{0.2cm}
    \begin{subfigure}{.135\textwidth}
        \centering
        \includegraphics[width=.5\textwidth]{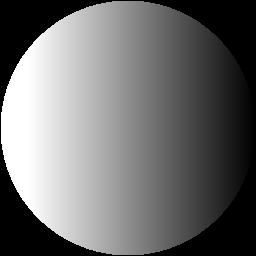}
    \end{subfigure}
    \begin{subfigure}{.135\textwidth}
        \centering
        \includegraphics[width=.5\textwidth]{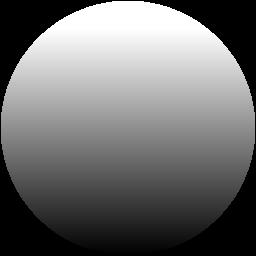}
    \end{subfigure}
    \begin{subfigure}{.135\textwidth}
        \centering
        \includegraphics[width=.5\textwidth]{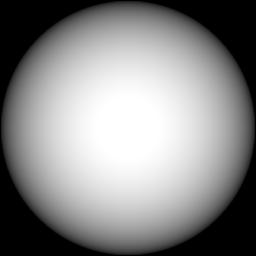}
    \end{subfigure}
    \hfill
    \begin{subfigure}{.135\textwidth}
        \centering
        \includegraphics[width=.5\textwidth]{images/real/DSR_person1/0003_crop_light_00.jpg}
    \end{subfigure}
    \begin{subfigure}{.135\textwidth}
        \centering
        \includegraphics[width=.5\textwidth]{images/real/DSR_person1/0003_crop_light_01.jpg}
    \end{subfigure}
    \hfill
    \begin{subfigure}{.135\textwidth}
        \centering
        \includegraphics[width=.5\textwidth]{images/real/DSR_person1/0003_crop_light_00.jpg}
    \end{subfigure}
    \begin{subfigure}{.135\textwidth}
        \centering
        \includegraphics[width=.5\textwidth]{images/real/DSR_person1/0003_crop_light_01.jpg}
    \end{subfigure}
    
    % Ours relighted
    \rotatebox{90}{Ours (relighting)}
    \adjincludegraphics[clip,trim={0 {.3\height} 0 0},width=.135\textwidth]{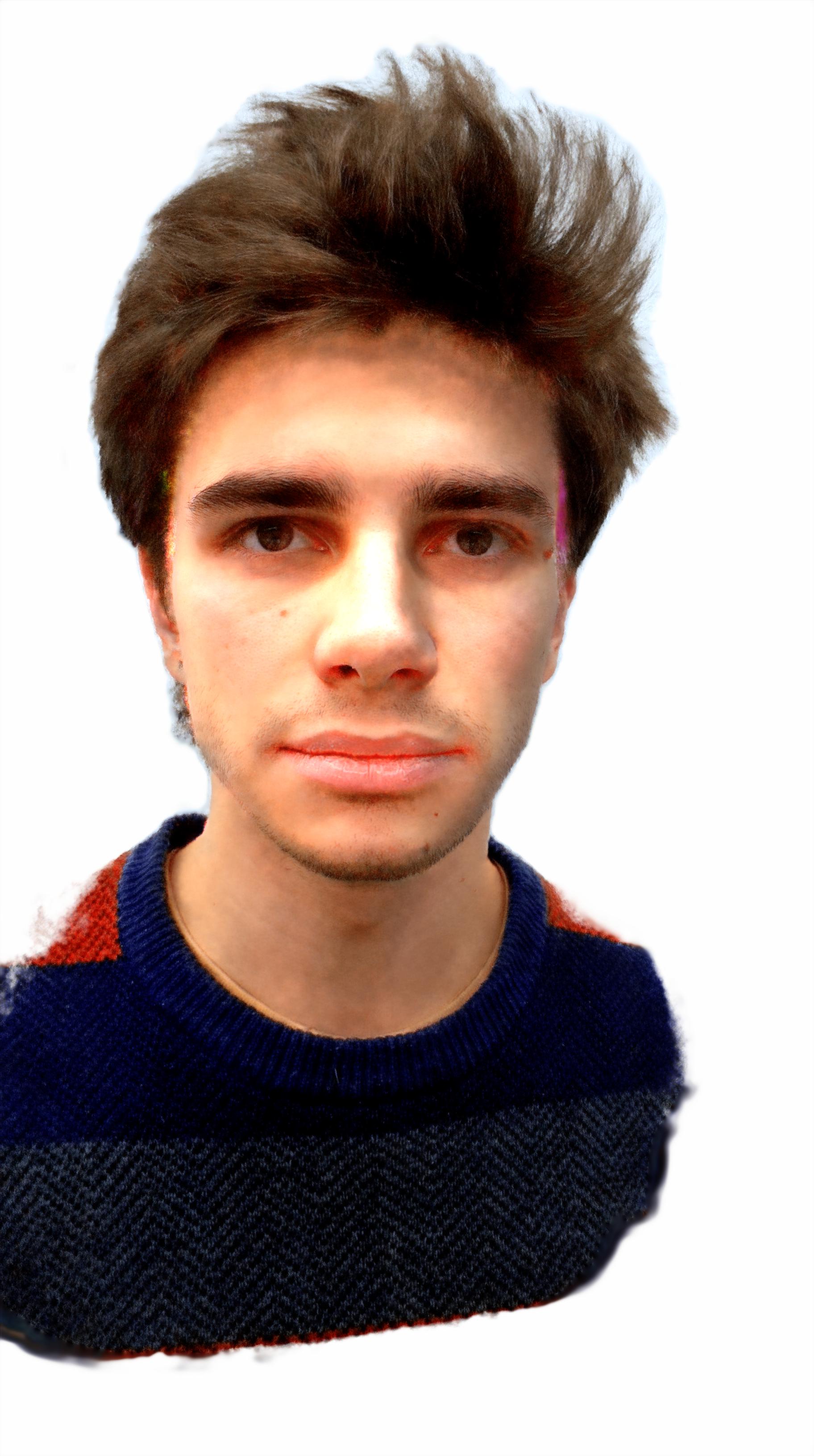}
    \adjincludegraphics[clip,trim={0 {.3\height} 0 0},width=.135\textwidth]{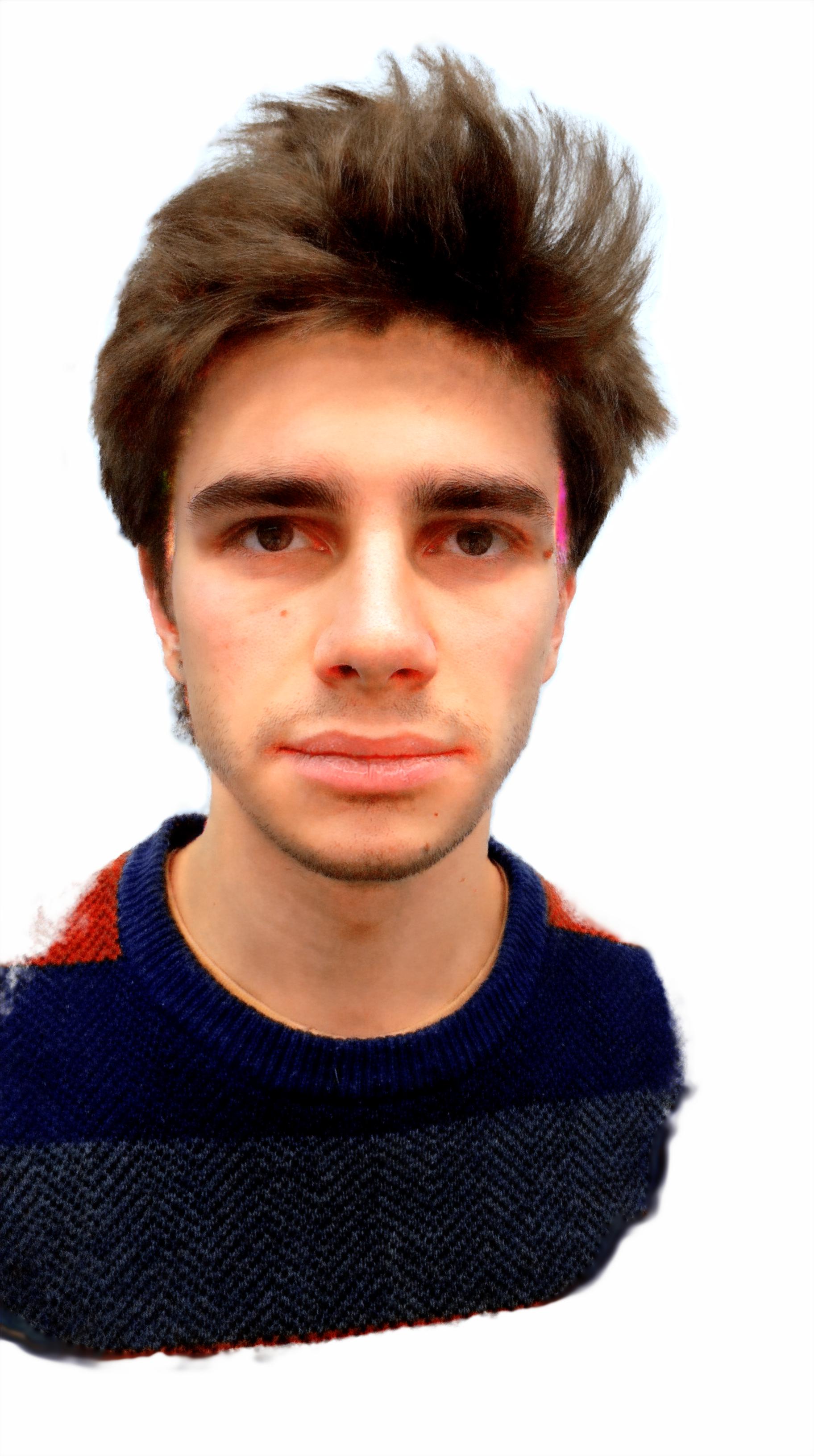}
    \adjincludegraphics[clip,trim={0 {.3\height} 0 0},width=.135\textwidth]{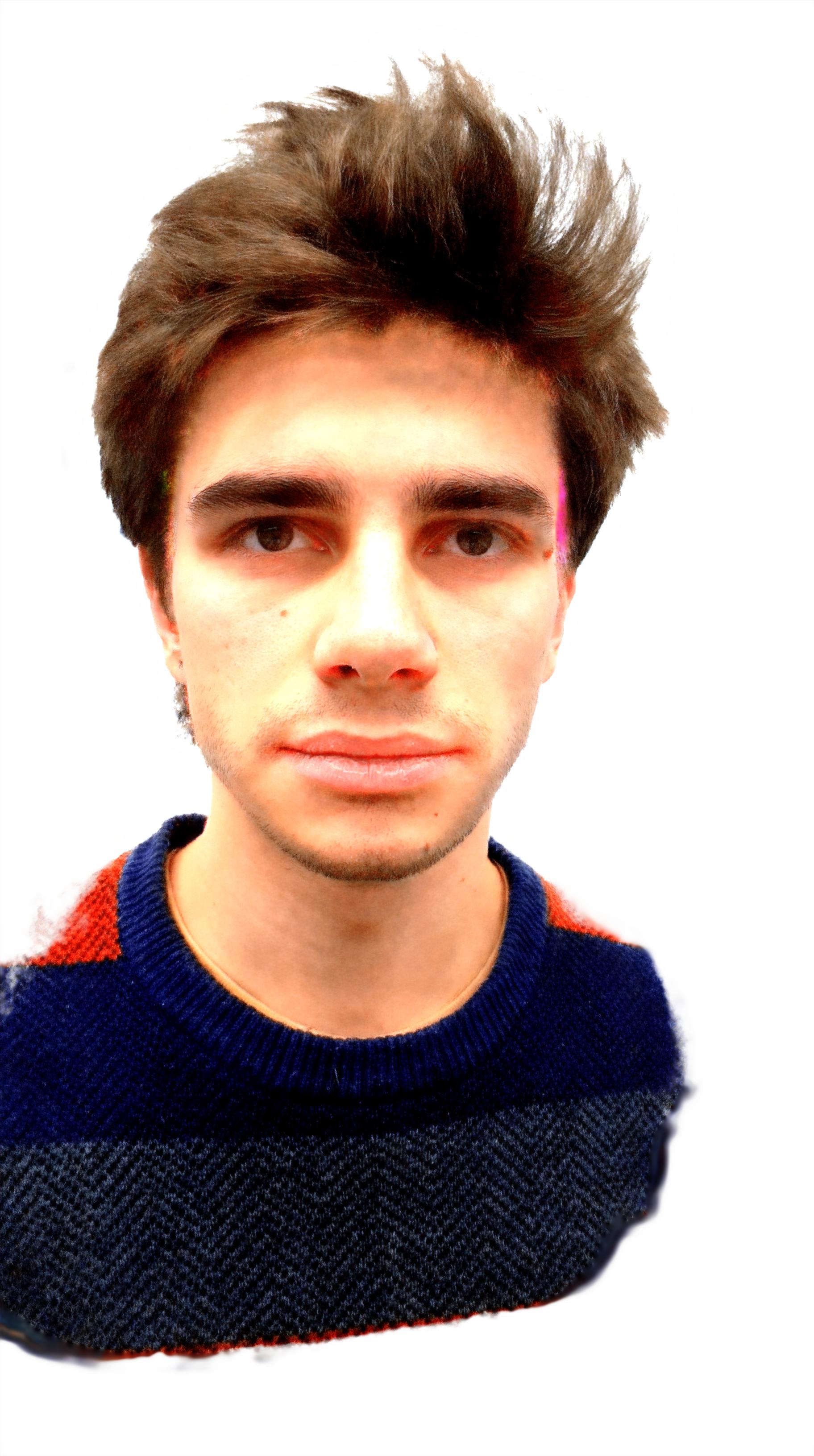}
    \hfill
    \adjincludegraphics[clip,trim={0 {.3\height} 0 0},width=.135\textwidth]{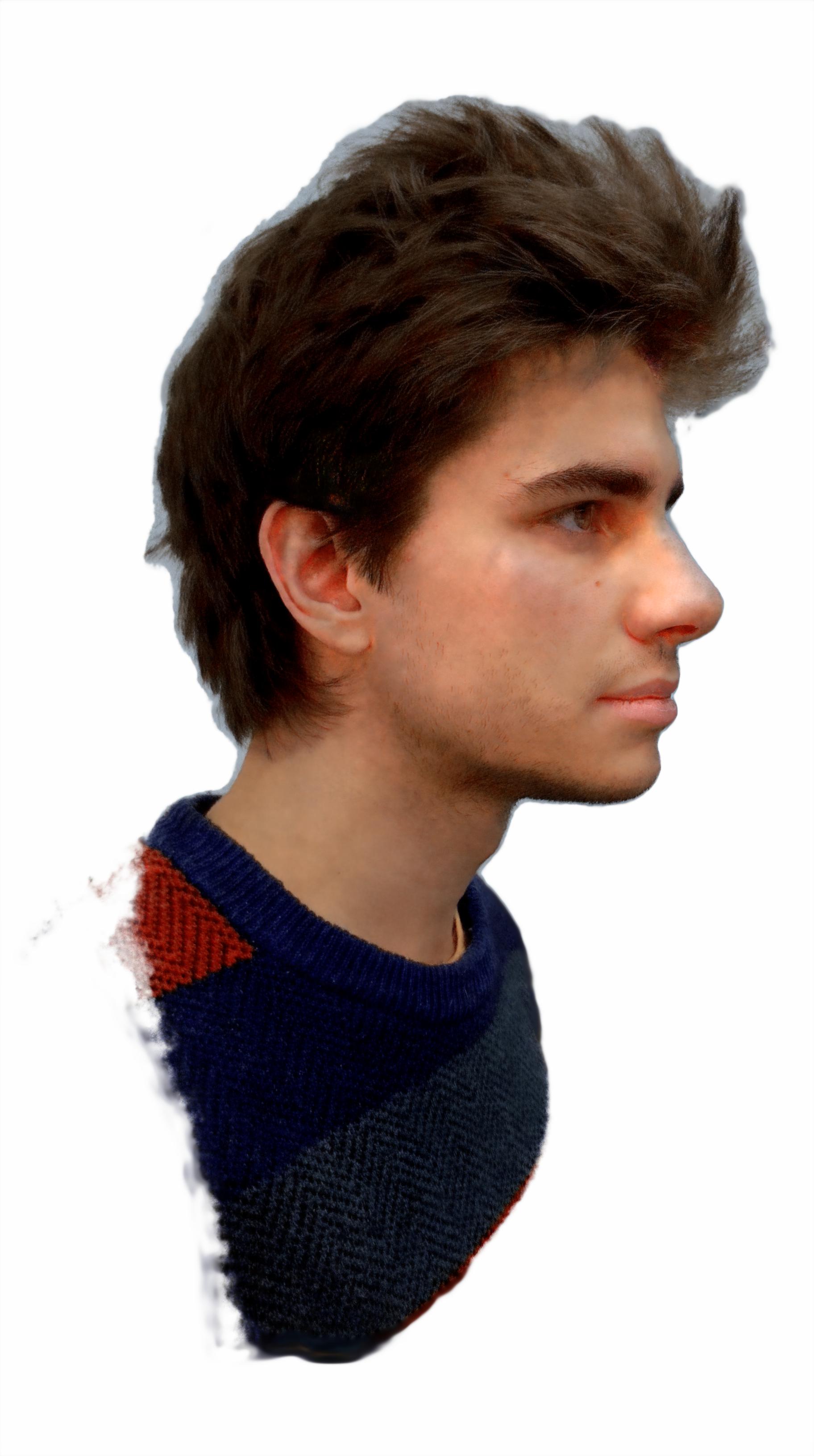}
    \adjincludegraphics[clip,trim={0 {.3\height} 0 0},width=.135\textwidth]{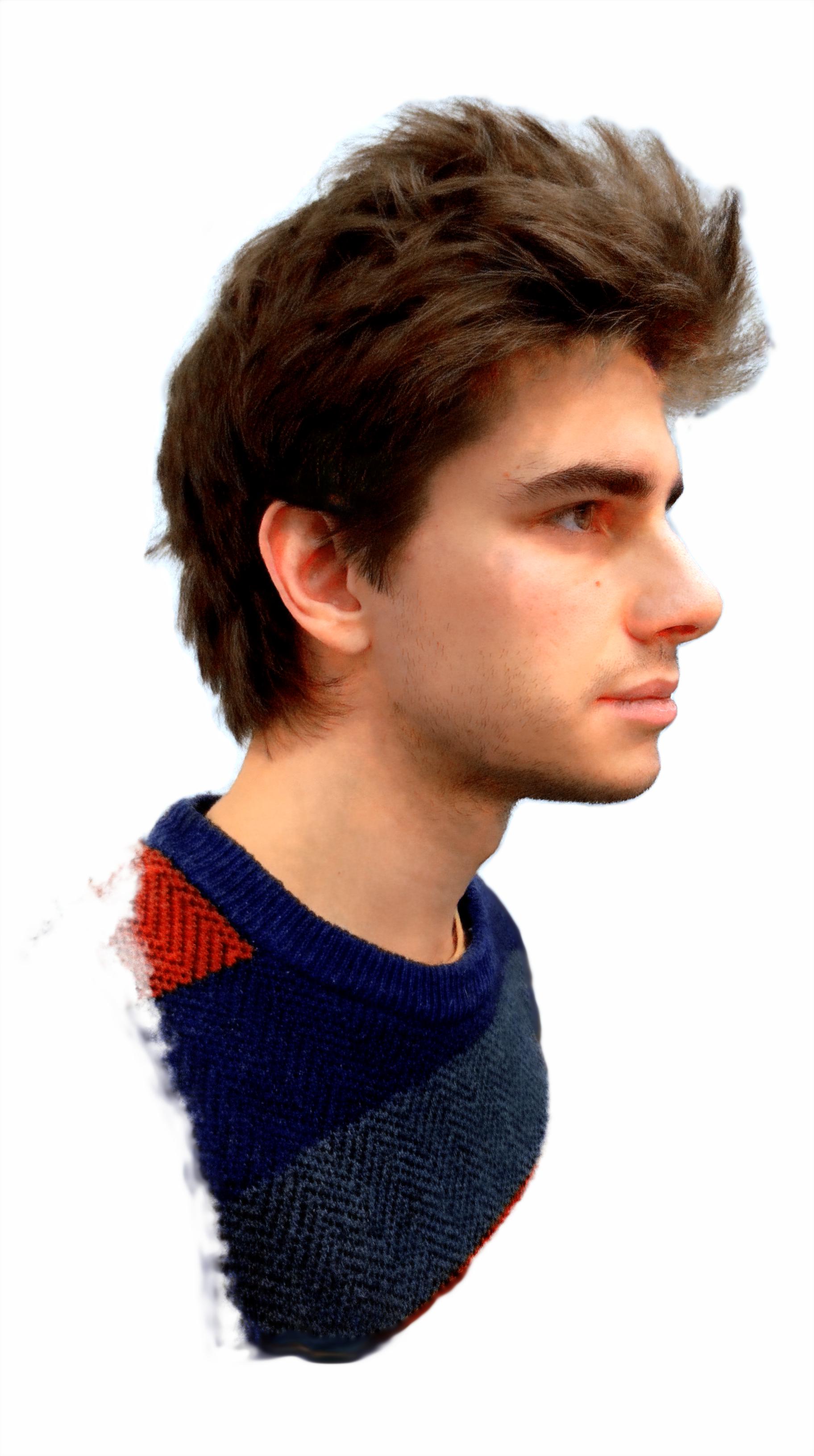}
    \hfill
    \adjincludegraphics[clip,trim={0 {.3\height} 0 0},width=.135\textwidth]{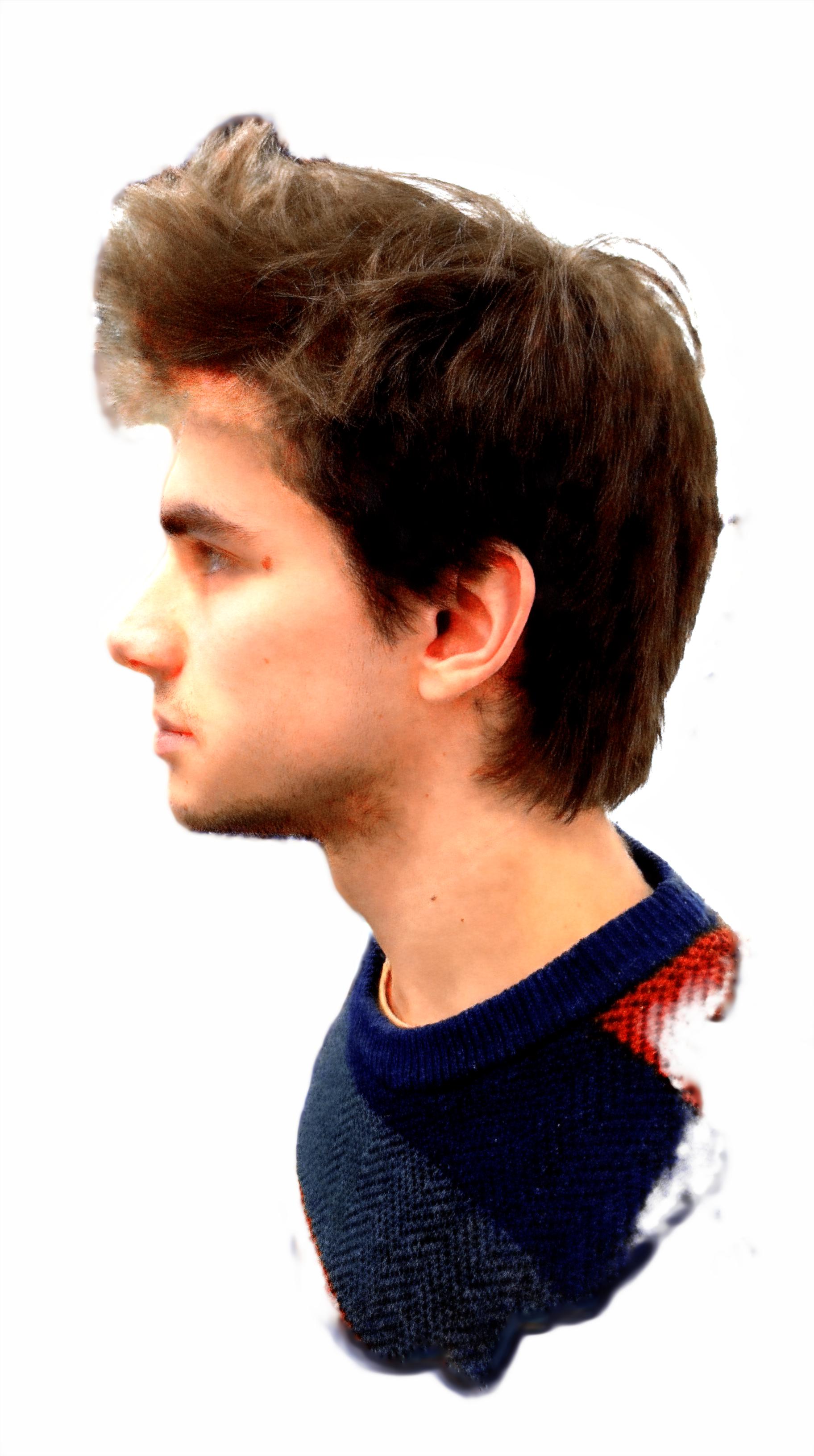}
    \adjincludegraphics[clip,trim={0 {.3\height} 0 0},width=.135\textwidth]{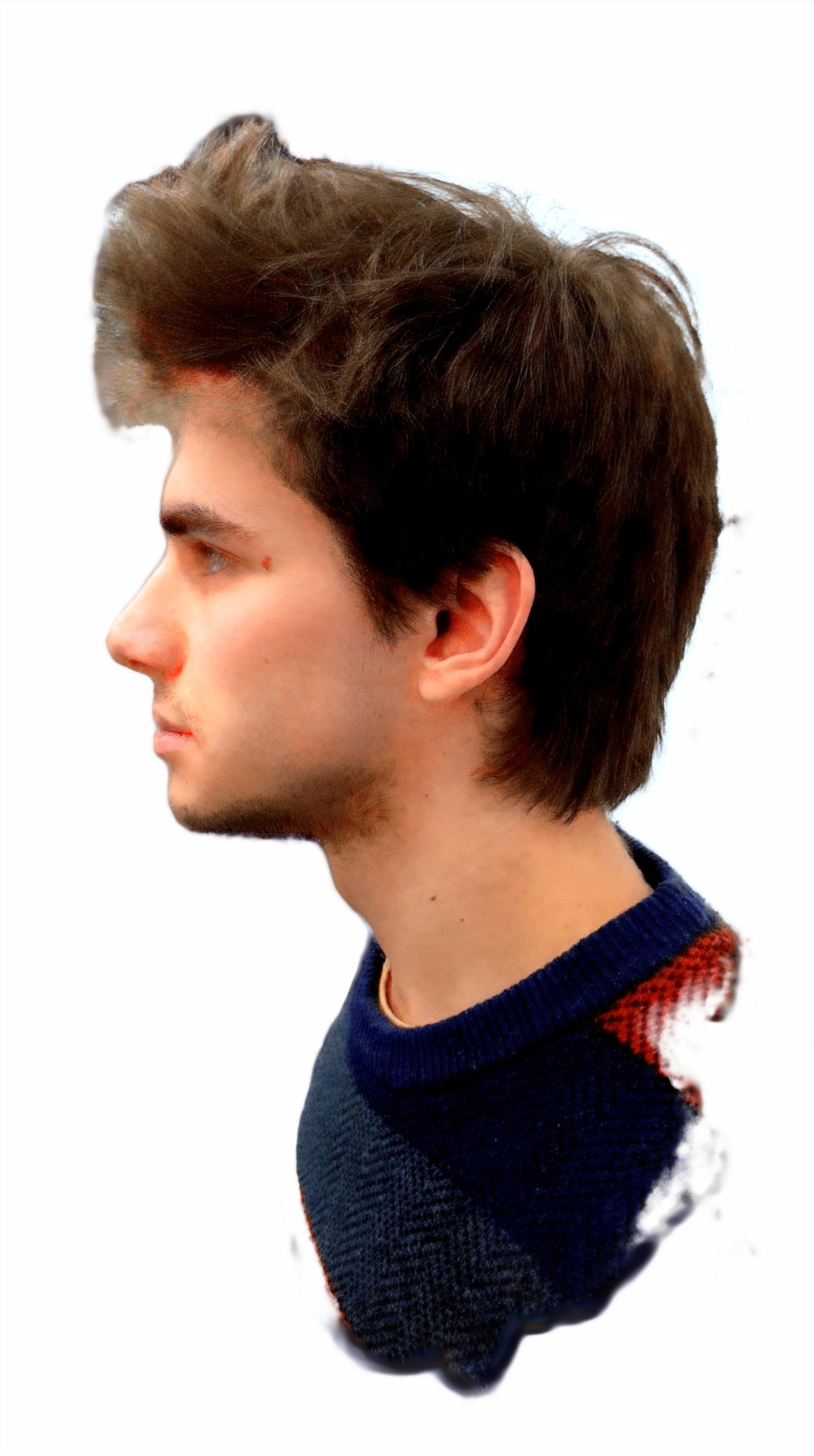}
    
    % Ours additional lighting
    % \rotatebox{90}{Ours (add. lighting)}
    % \includegraphics[width=.135\textwidth]{images/real/SH_add_lighting_person1/to_right_viewer_config_0080_infer.jpg}
    % \includegraphics[width=.135\textwidth]{images/real/SH_add_lighting_person1/to_down_viewer_config_0080_infer.jpg}
    % \includegraphics[width=.135\textwidth]{images/real/SH_add_lighting_person1/to_forward_viewer_config_0080_infer.jpg}
    % \hfill
    % \includegraphics[width=.135\textwidth]{images/real/SH_add_lighting_person1/to_right_viewer_config_0130_infer.jpg}
    % \includegraphics[width=.135\textwidth]{images/real/SH_add_lighting_person1/to_down_viewer_config_0130_infer.jpg}
    % \hfill
    % \includegraphics[width=.135\textwidth]{images/real/SH_add_lighting_person1/to_right_viewer_config_0003_infer.jpg}
    % \includegraphics[width=.135\textwidth]{images/real/SH_add_lighting_person1/to_down_viewer_config_0003_infer.jpg}
    
    \rotatebox{90}{DPR~\cite{Zhou19}}
    \includegraphics[width=.135\textwidth]{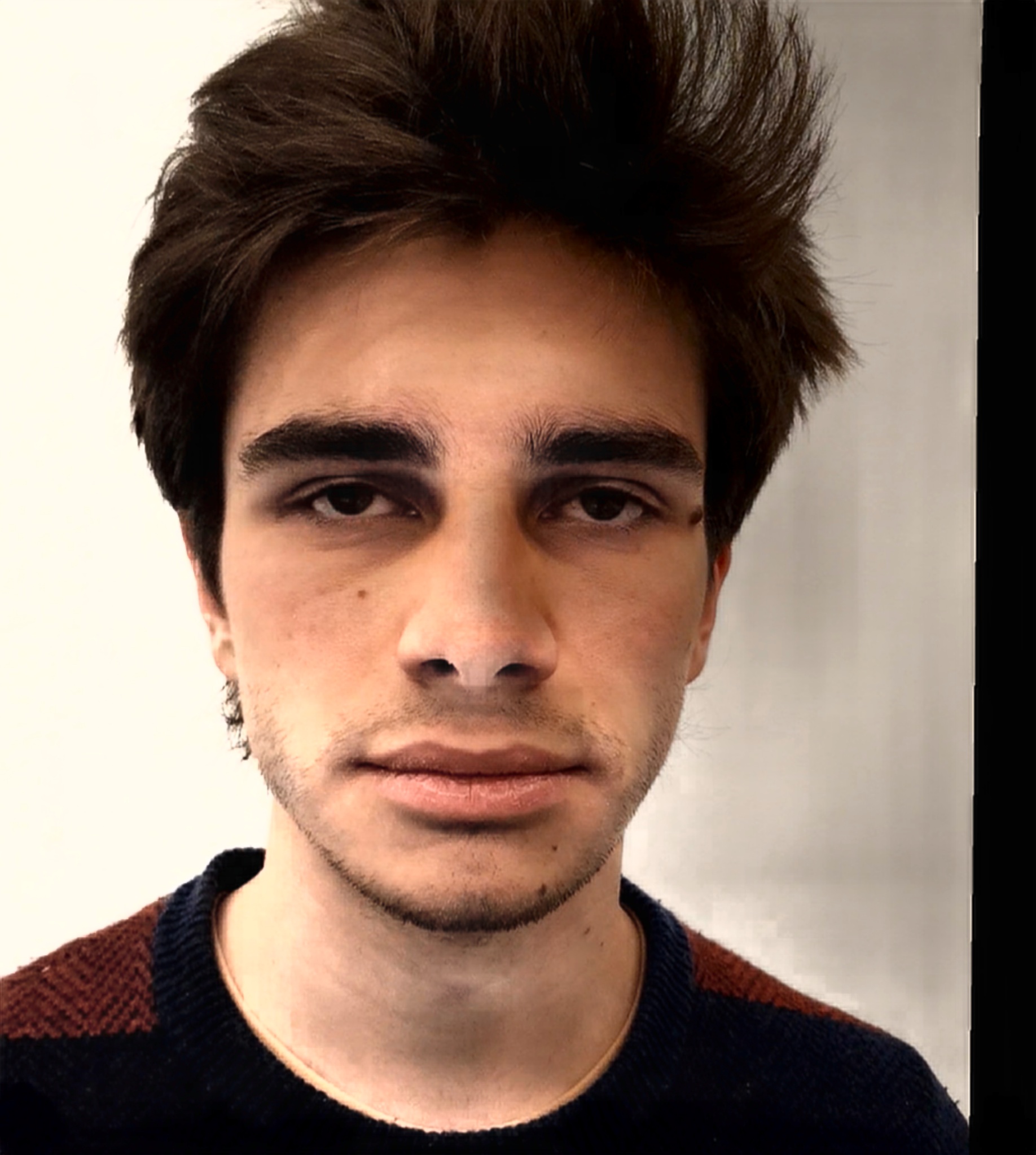}
    \includegraphics[width=.135\textwidth]{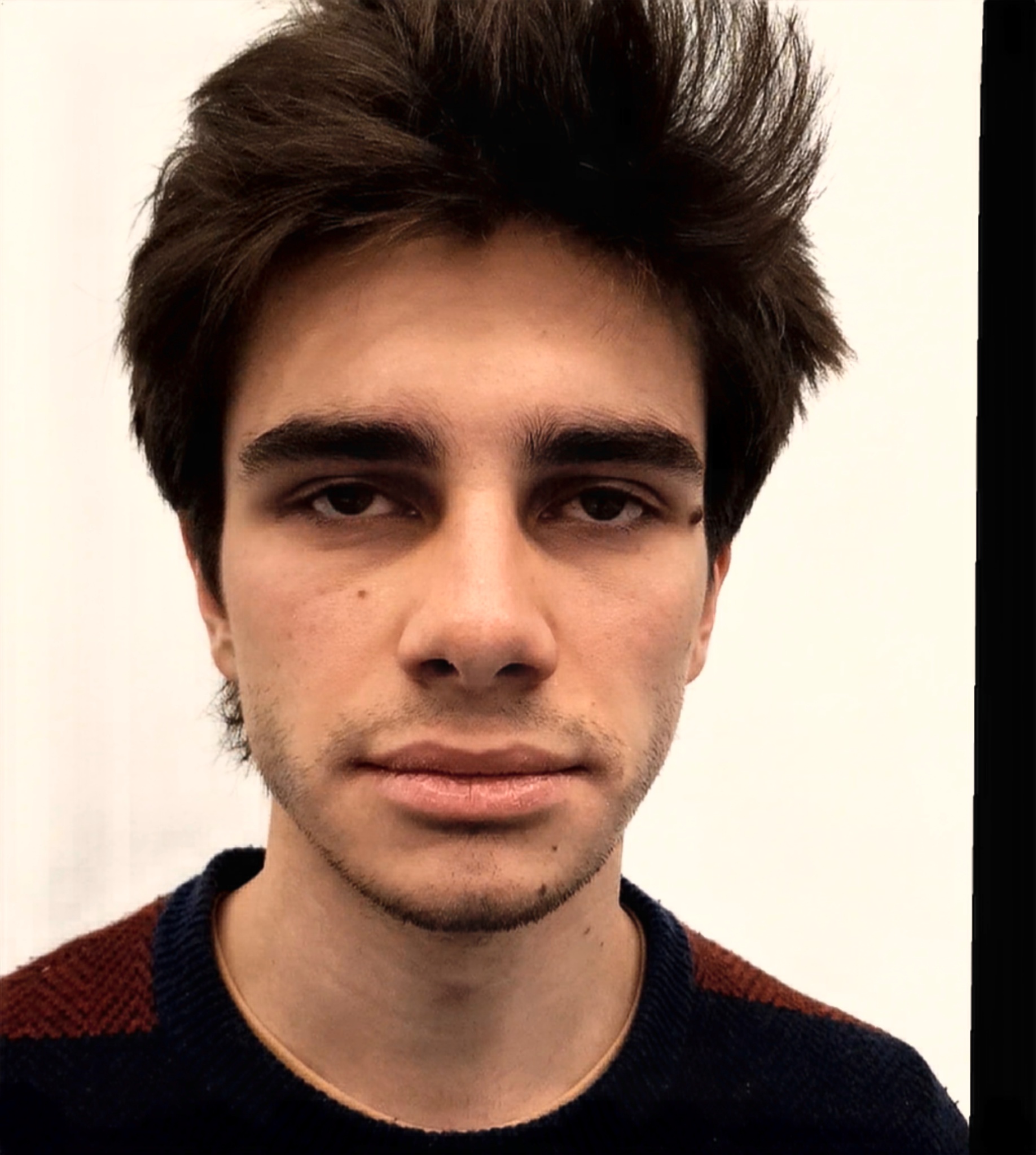}
    \includegraphics[width=.135\textwidth]{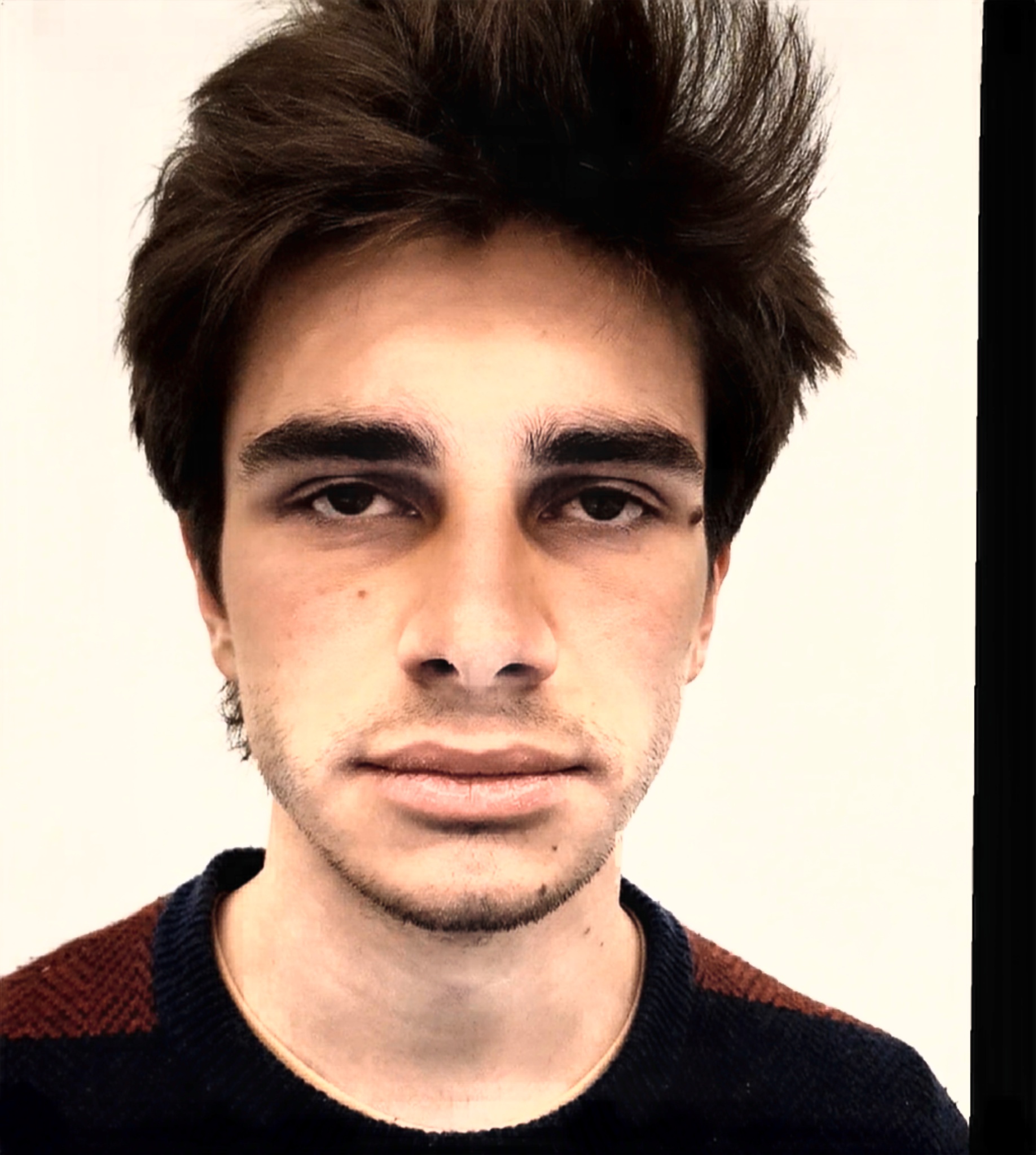}
    \hfill
    \includegraphics[width=.135\textwidth]{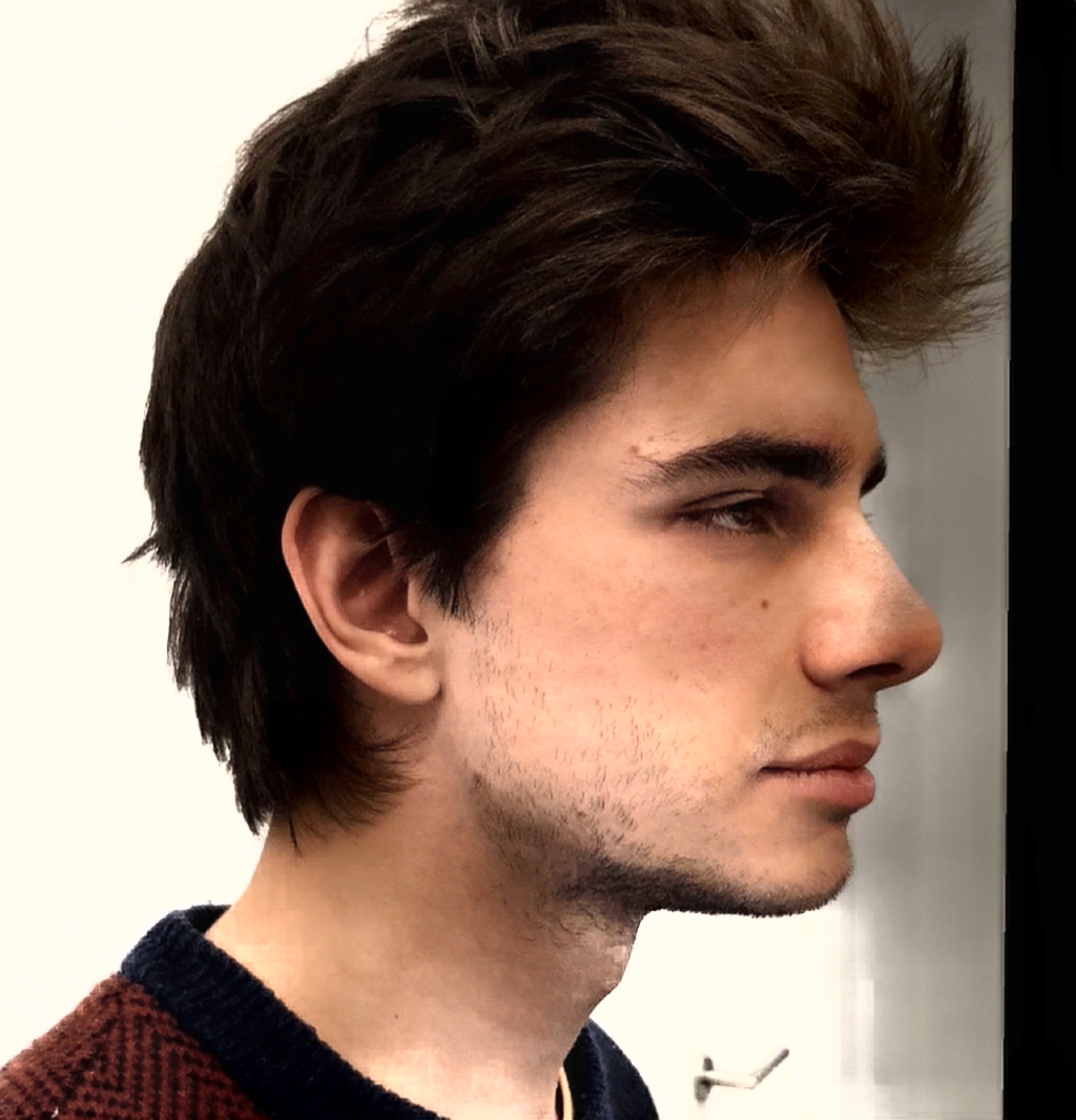}
    \includegraphics[width=.135\textwidth]{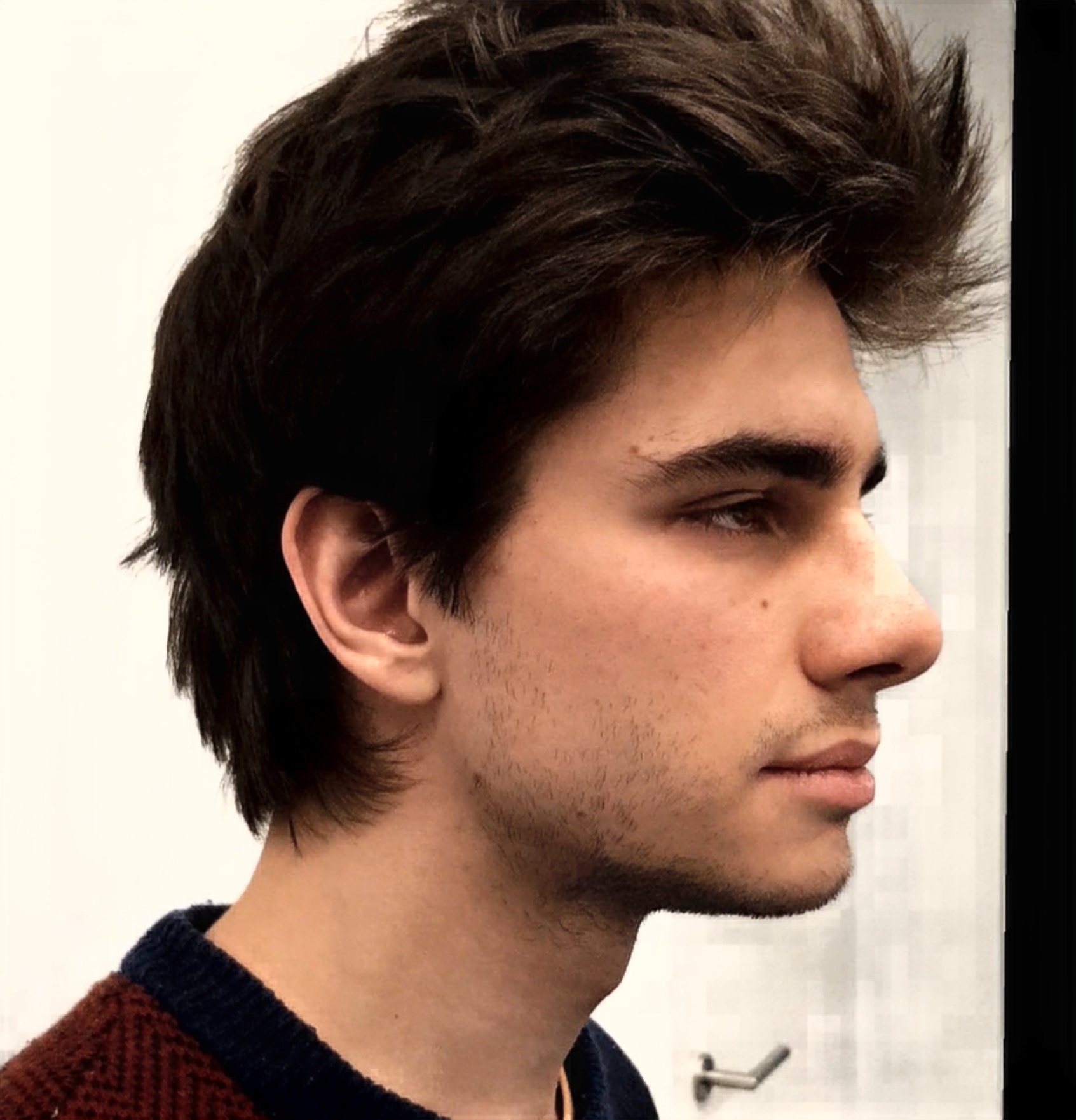}
    \hfill
    \includegraphics[width=.135\textwidth]{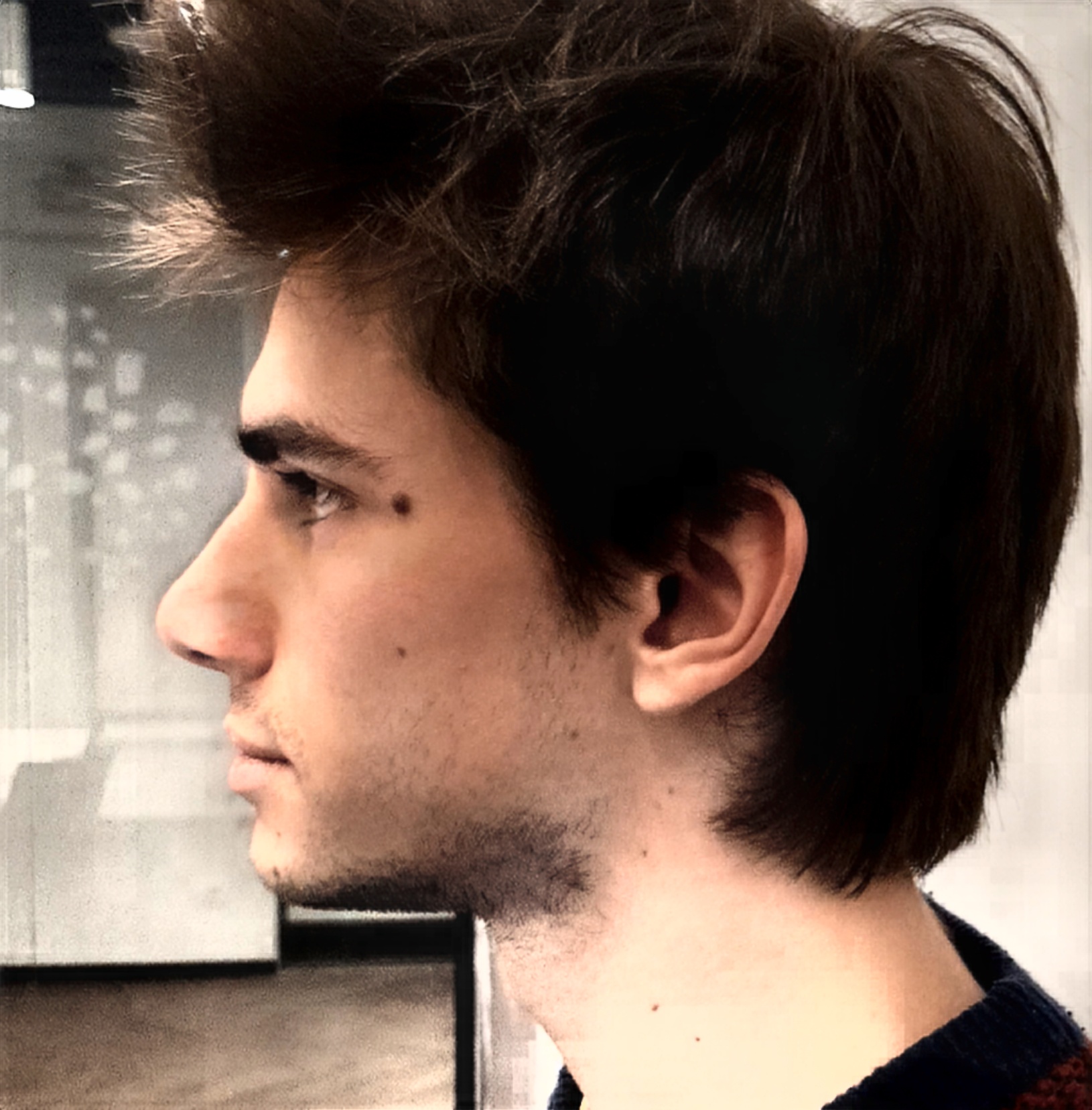}
    \includegraphics[width=.135\textwidth]{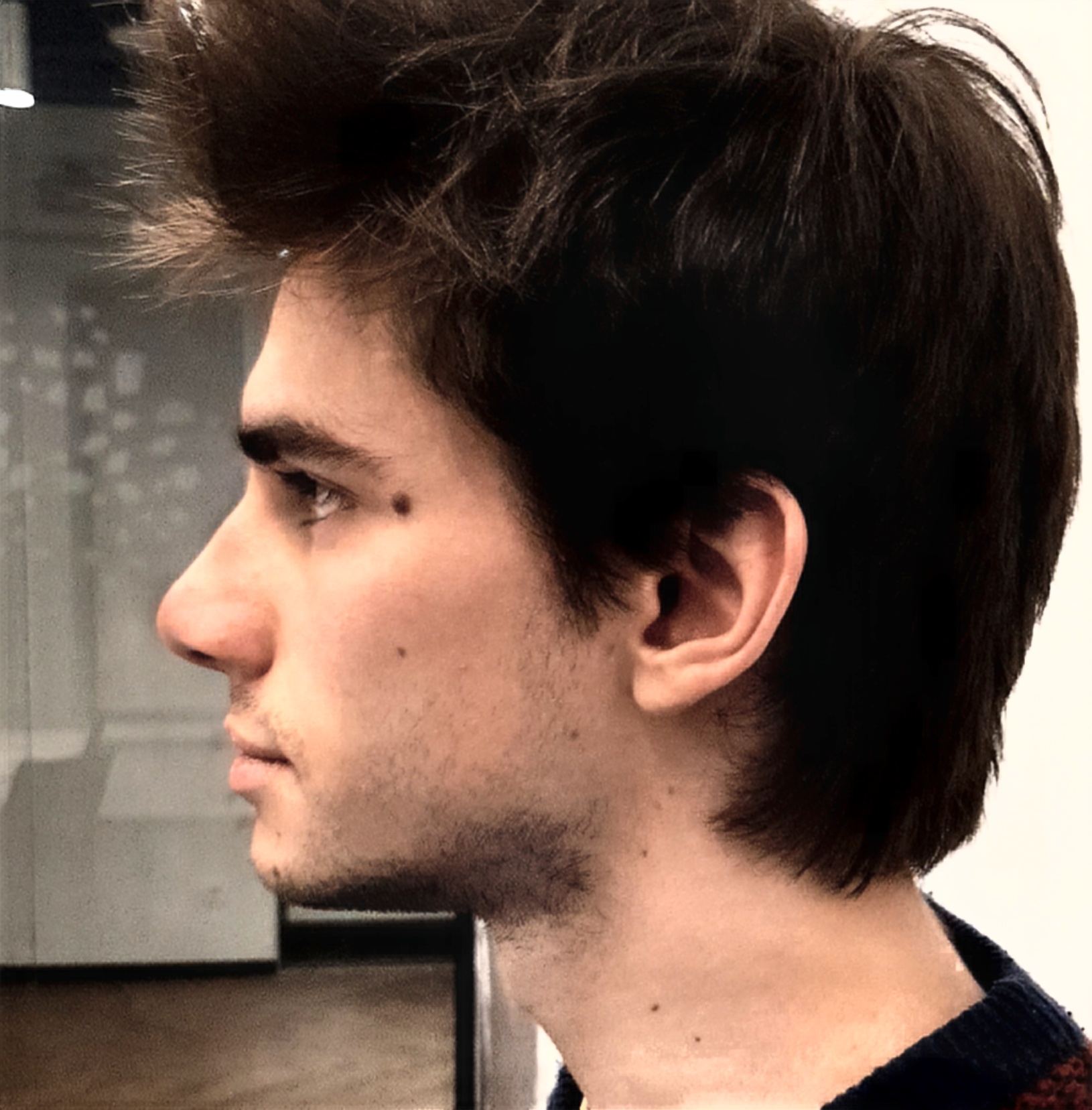}
    
    \caption{Qualitative comparison of the sample smartphone-captured real head portrait under several viewpoints from validation (corresponding to non-flashlighted images) and Spherical Harmonics (SH) lighting, corresponding to ambient and $1^\mathrm{st}$-order SH light. Under such lighting, our system can be compared with others, such as the DPR~\cite{Zhou19} method. The top row depicts a sphere, lighted by the selected SH lighting and rendered from the respective viewpoint. While DPR is able to realistically reshade an image just from a single photo, we observe it is sensitive to cropping and can create inconsistent lighting at the side views.}%In the \textit{Ours (add. lighting)} row, we replace the ambient component of SH light by the environmental lighting of the captured room. In general, we found that it leads to better appearance of subjects, and is closer to the behavior of DPR. } %Our method requires a set of photos instead of one, but is more consistent across views and can render the head from novel viewpoints.}
    \label{fig:real_SH}
\end{figure*}

\begin{figure*}[h]
    \centering
    \begin{subfigure}{.11\linewidth}
        \centering 
        Predicted
    \end{subfigure}
    \begin{subfigure}{.11\linewidth}
        \centering 
        Nearest video frame
    \end{subfigure}
    \hfill
    \begin{subfigure}{.11\linewidth}
        \centering 
        Predicted
    \end{subfigure}
    \begin{subfigure}{.11\linewidth}
        \centering 
        Nearest video frame
    \end{subfigure}
    \hfill
    \begin{subfigure}{.11\linewidth}
        \centering 
        Predicted
    \end{subfigure}
    \begin{subfigure}{.11\linewidth}
        \centering 
        Nearest video frame
    \end{subfigure}
    \hfill
    \begin{subfigure}{.11\linewidth}
        \centering 
        Predicted
    \end{subfigure}
    \begin{subfigure}{.11\linewidth}
        \centering 
        Nearest video frame
    \end{subfigure}
    \vspace{0.1cm}
    
    \begin{subfigure}{.11\textwidth}
        \centering
        \includegraphics[width=\textwidth]{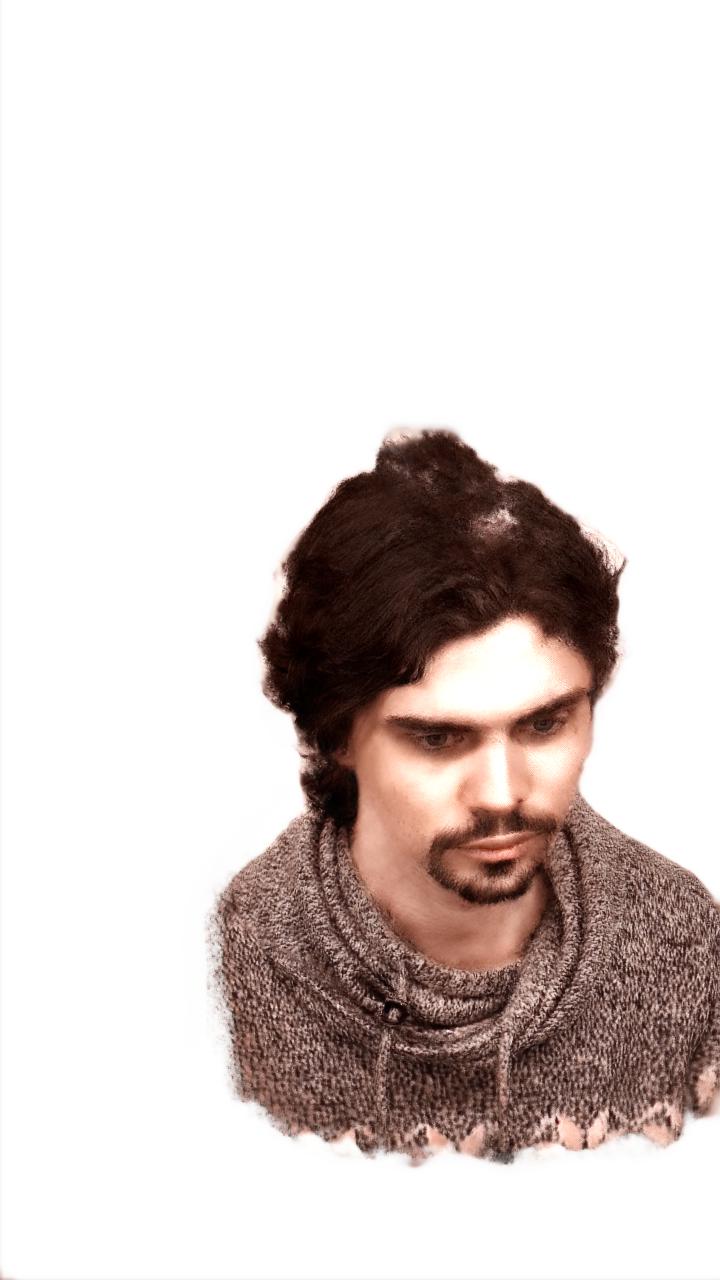}
    \end{subfigure}
    \begin{subfigure}{.11\textwidth}
        \centering
        \includegraphics[width=\textwidth]{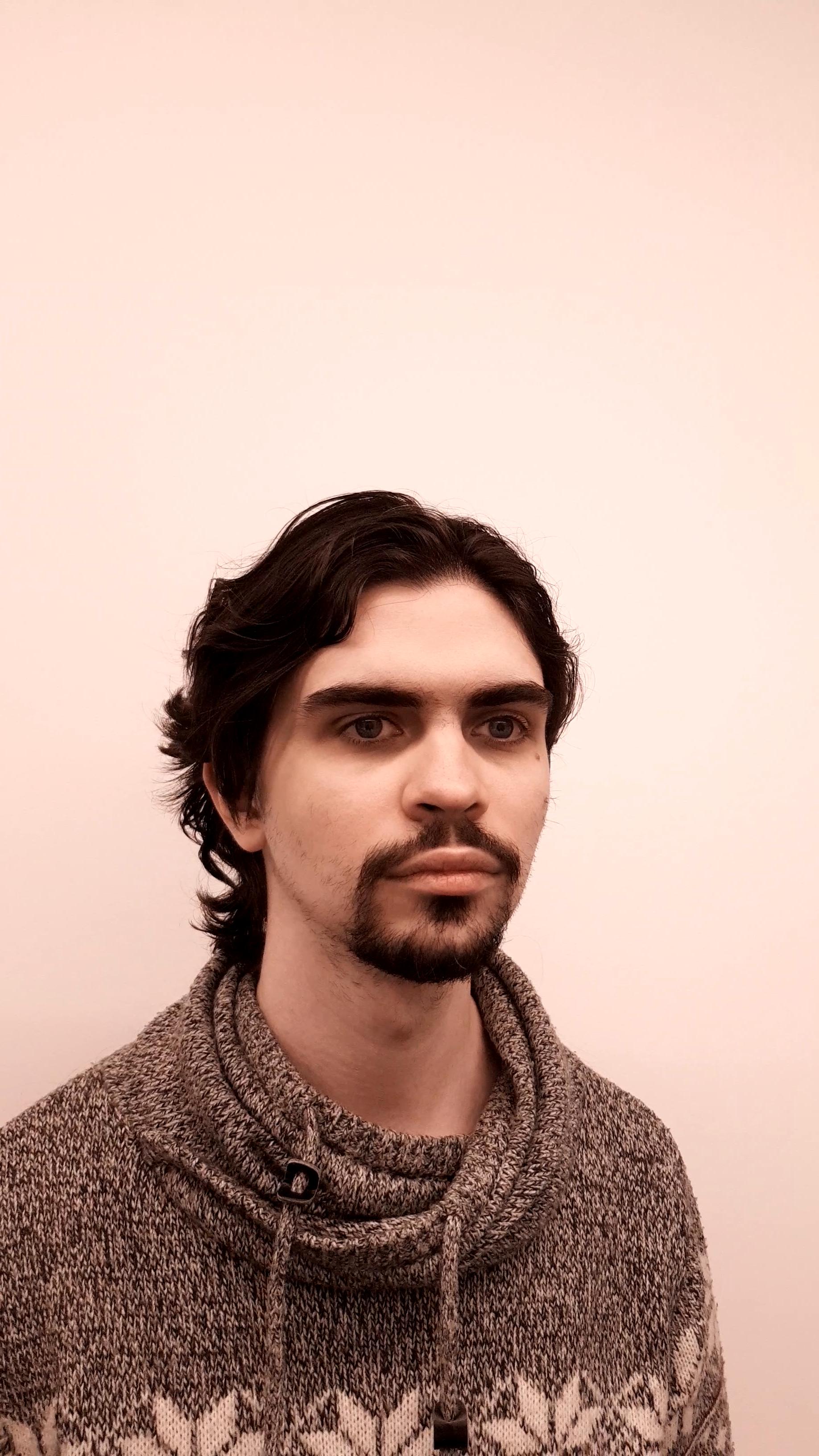}
    \end{subfigure}
    \hfill
    \begin{subfigure}{.11\textwidth}
        \centering
        \includegraphics[width=\textwidth]{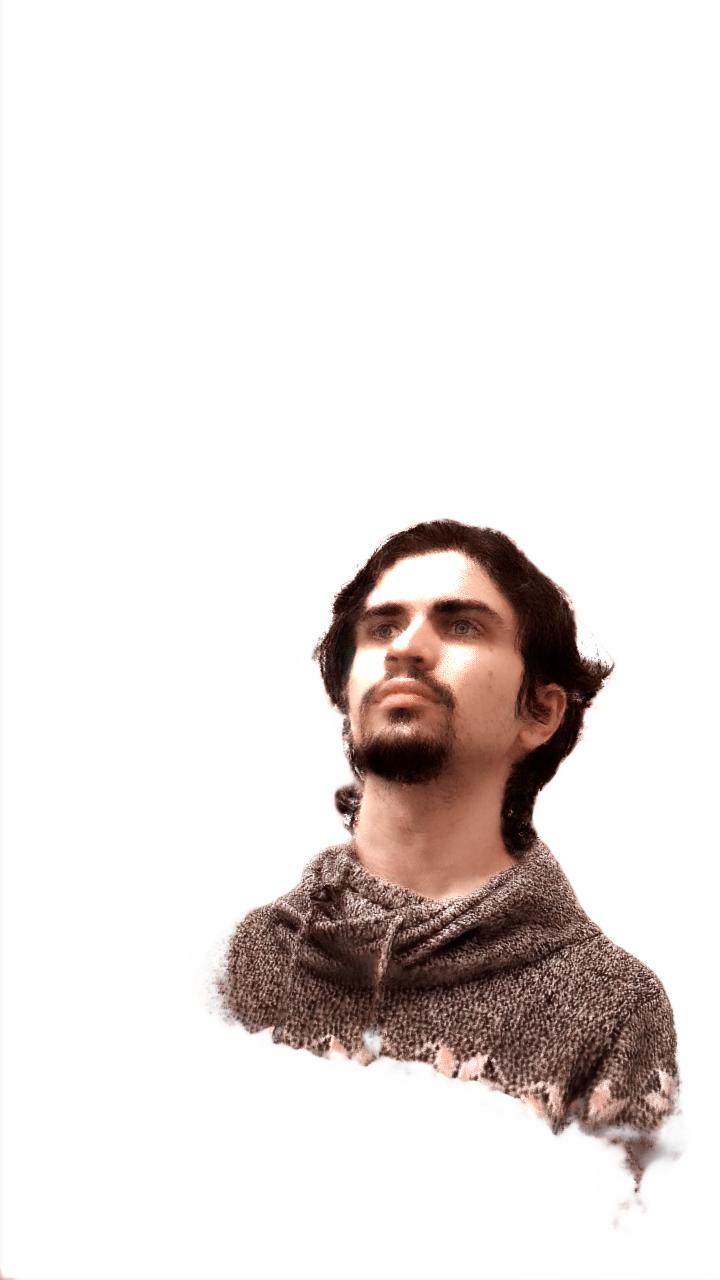}
    \end{subfigure}
    \begin{subfigure}{.11\textwidth}
        \centering
        \includegraphics[width=\textwidth]{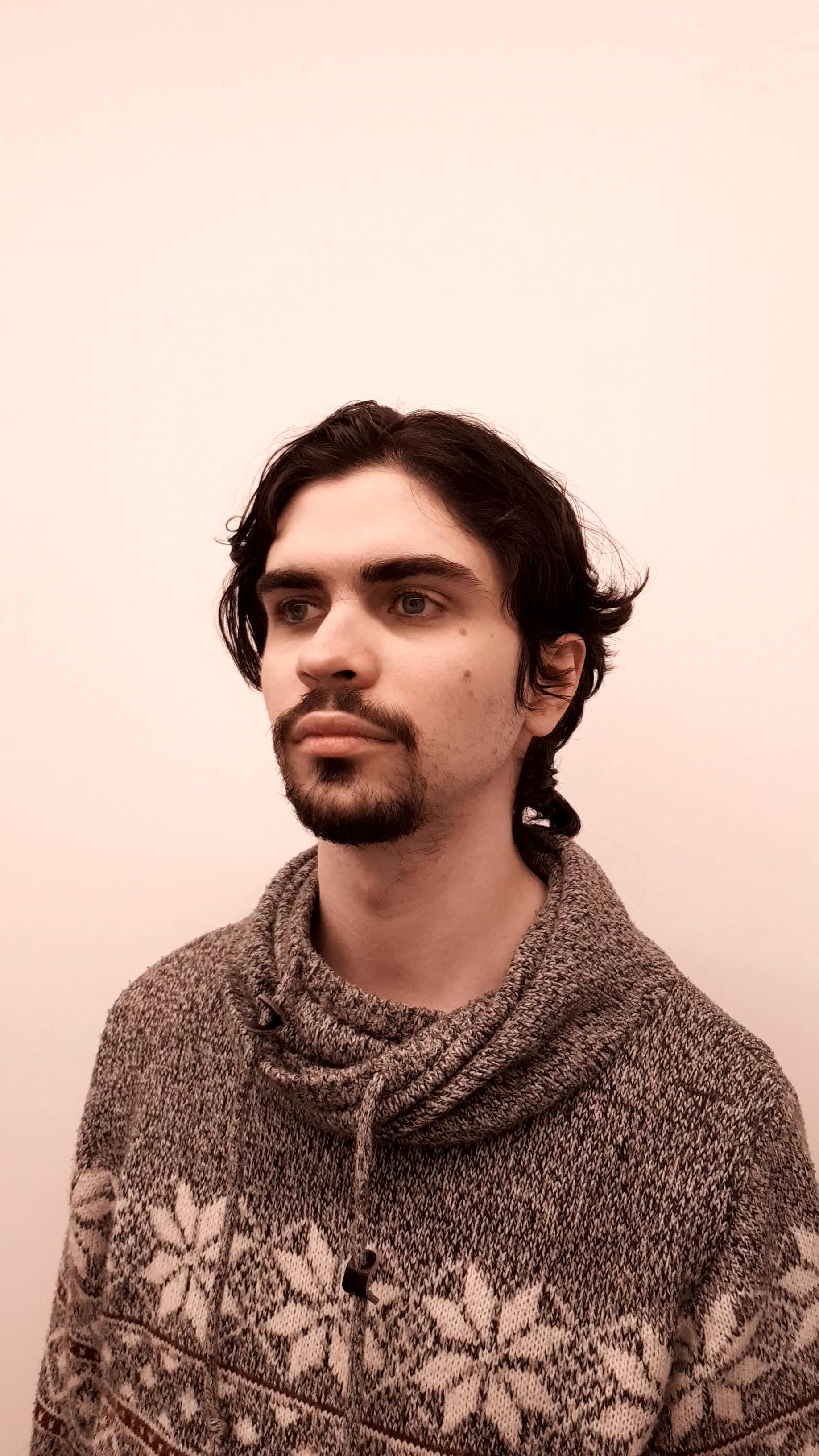}
    \end{subfigure}
    \hfill
    \begin{subfigure}{.11\textwidth}
        \centering
        \includegraphics[width=\textwidth]{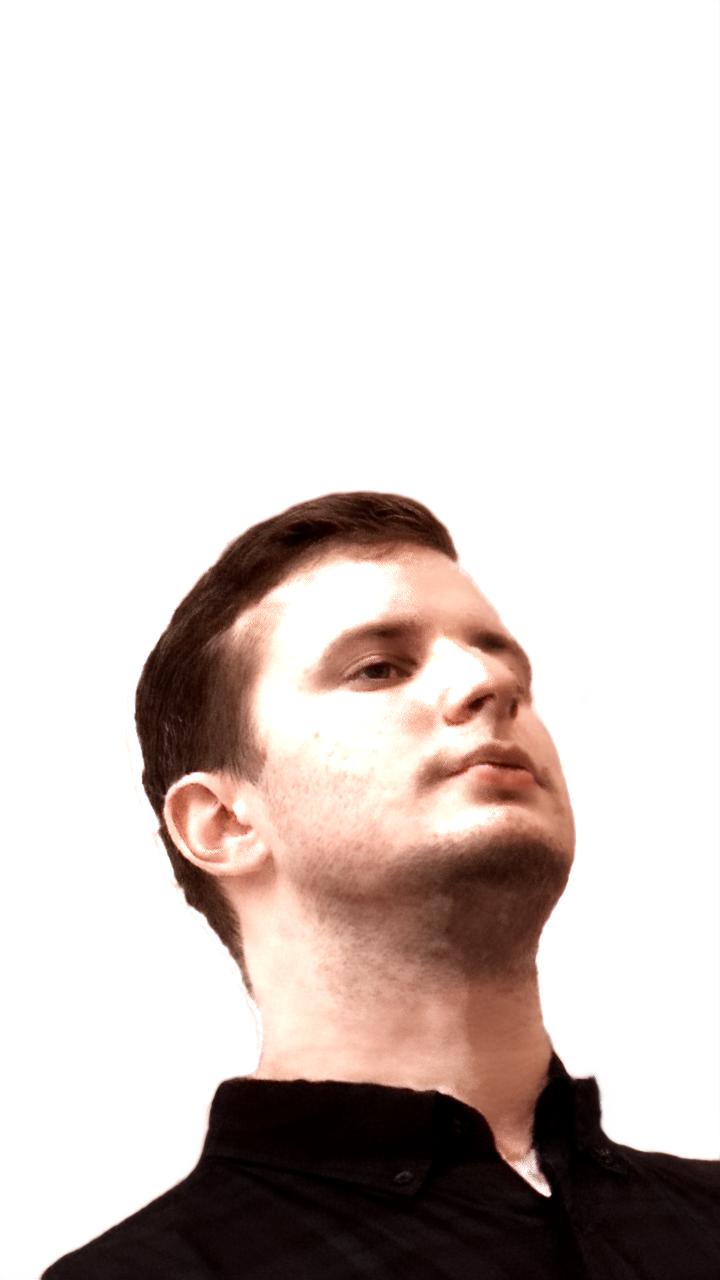}
    \end{subfigure}
    \begin{subfigure}{.11\textwidth}
        \centering
        \includegraphics[width=\textwidth]{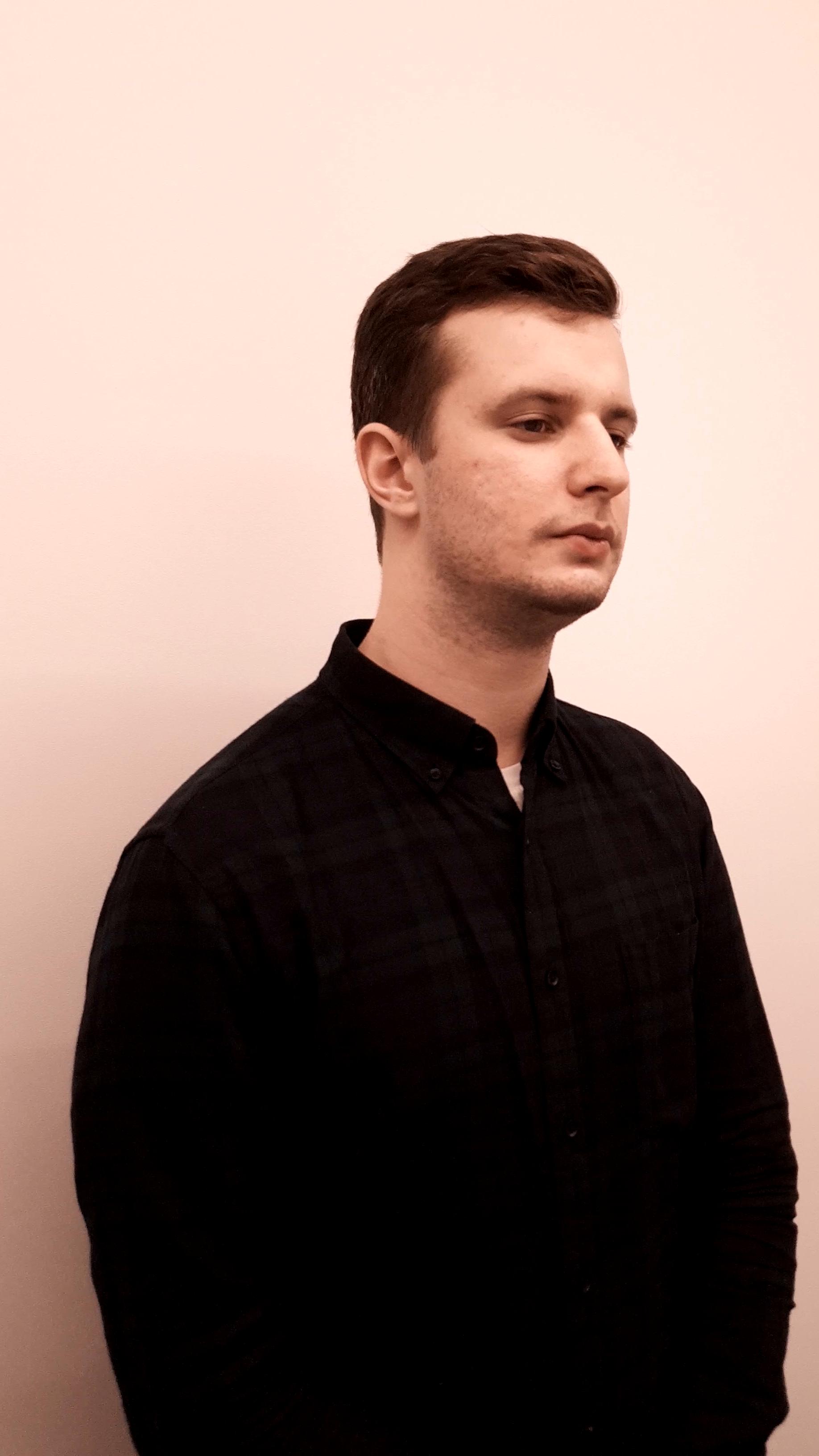}
    \end{subfigure}
    \hfill
    \begin{subfigure}{.11\textwidth}
        \centering
        \includegraphics[width=\textwidth]{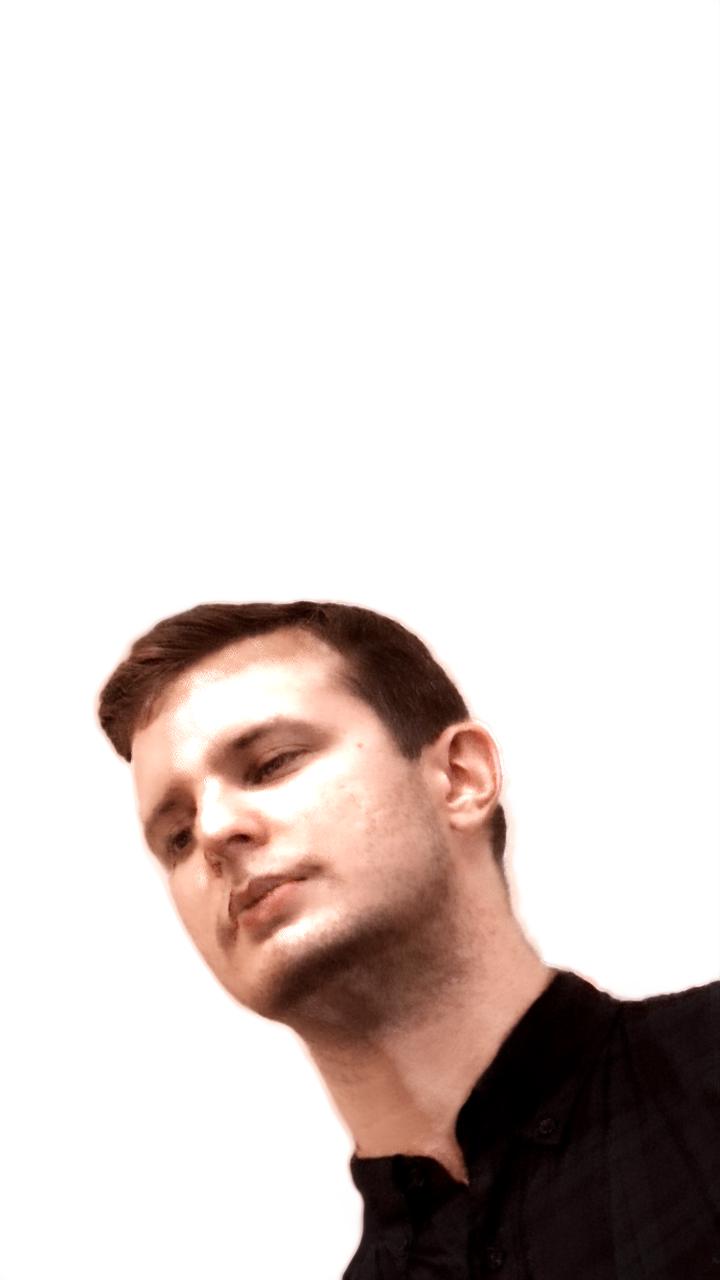}
    \end{subfigure}
    \begin{subfigure}{.11\textwidth}
        \centering
        \includegraphics[width=\textwidth]{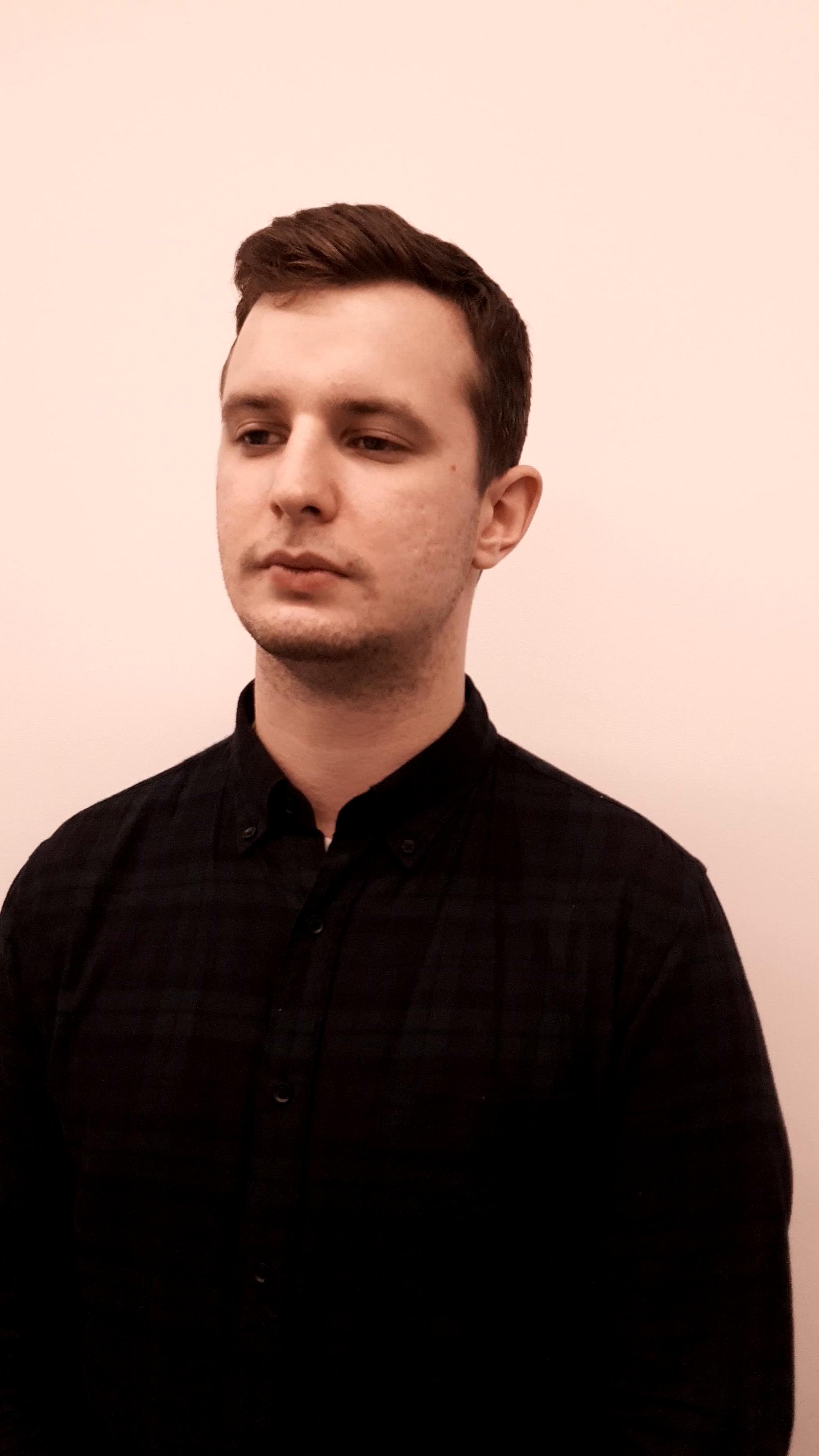}
    \end{subfigure}
    
    \begin{subfigure}{.11\textwidth}
        \centering
        \includegraphics[width=\textwidth]{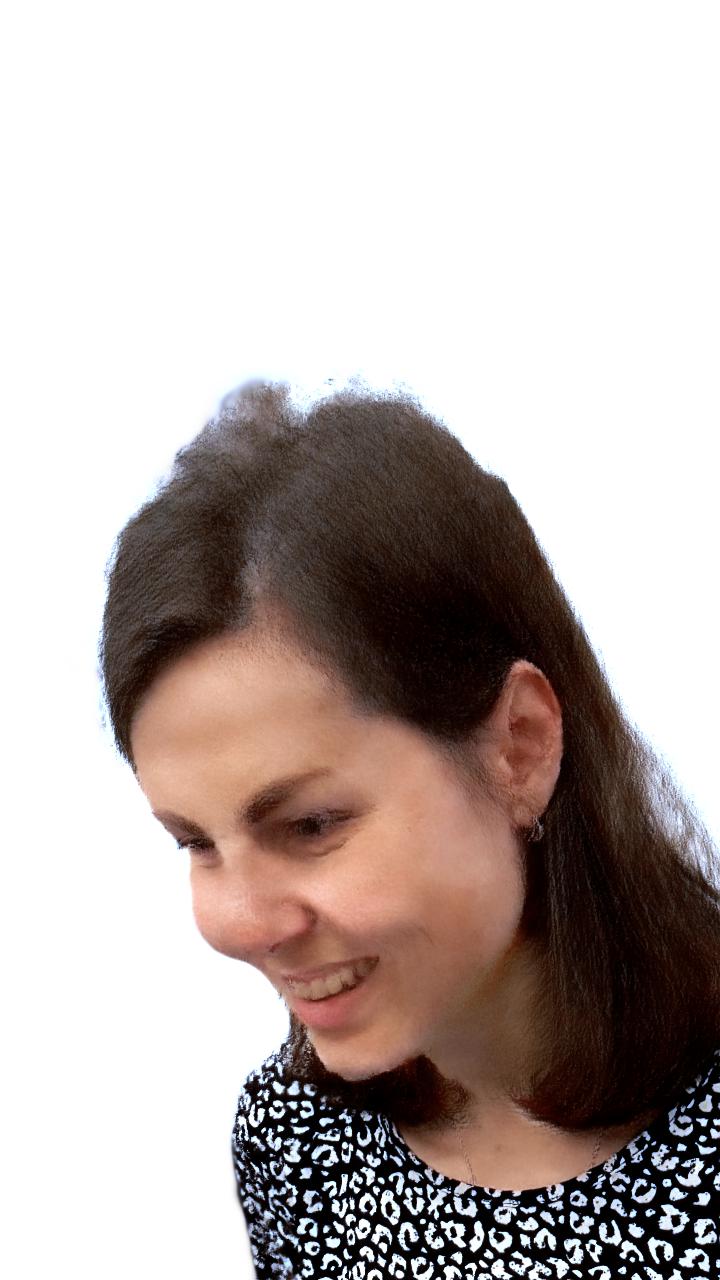}
    \end{subfigure}
    \begin{subfigure}{.11\textwidth}
        \centering
        \includegraphics[width=\textwidth]{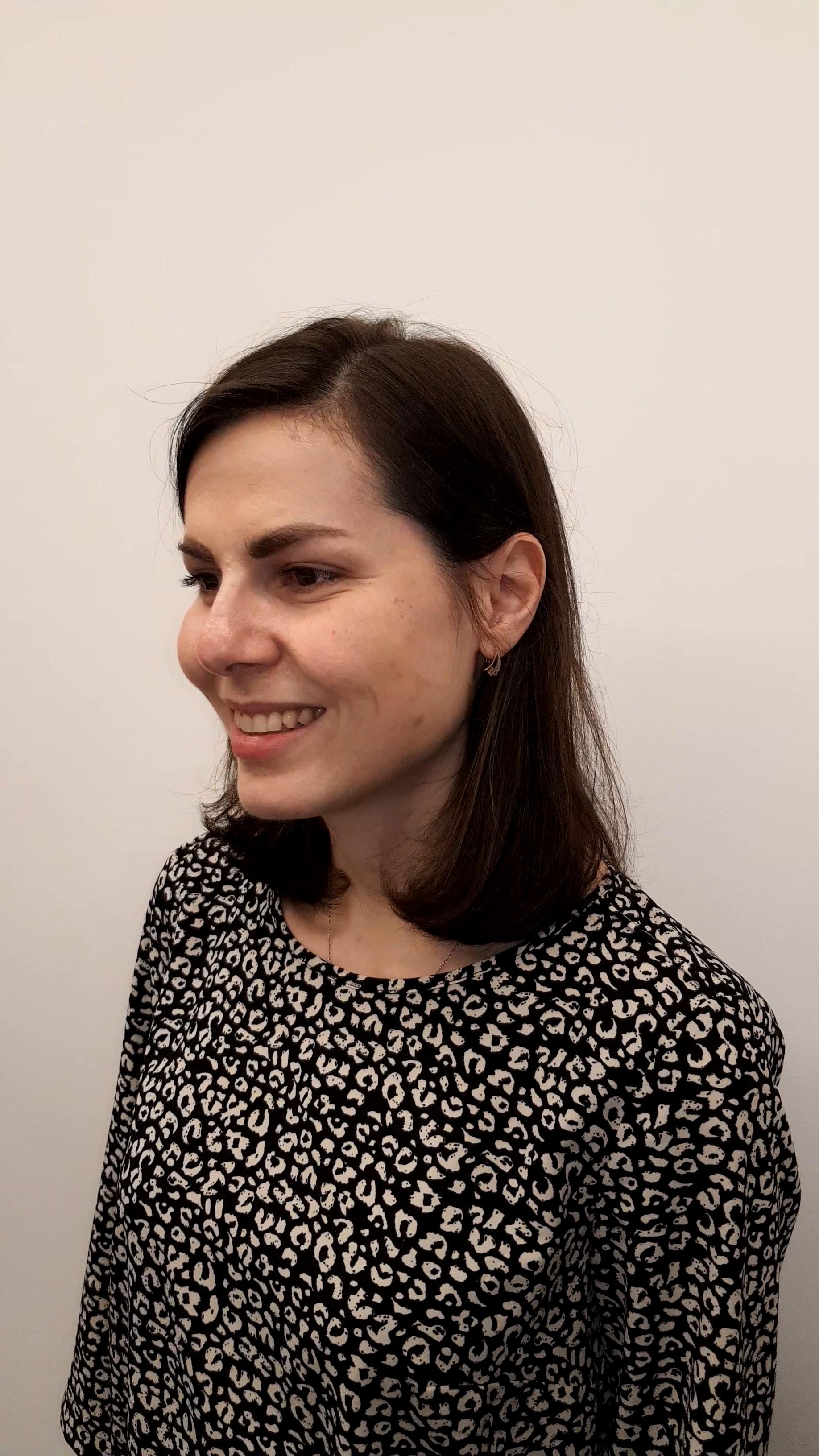}
    \end{subfigure}
    \hfill
    \begin{subfigure}{.11\textwidth}
        \centering
        \includegraphics[width=\textwidth]{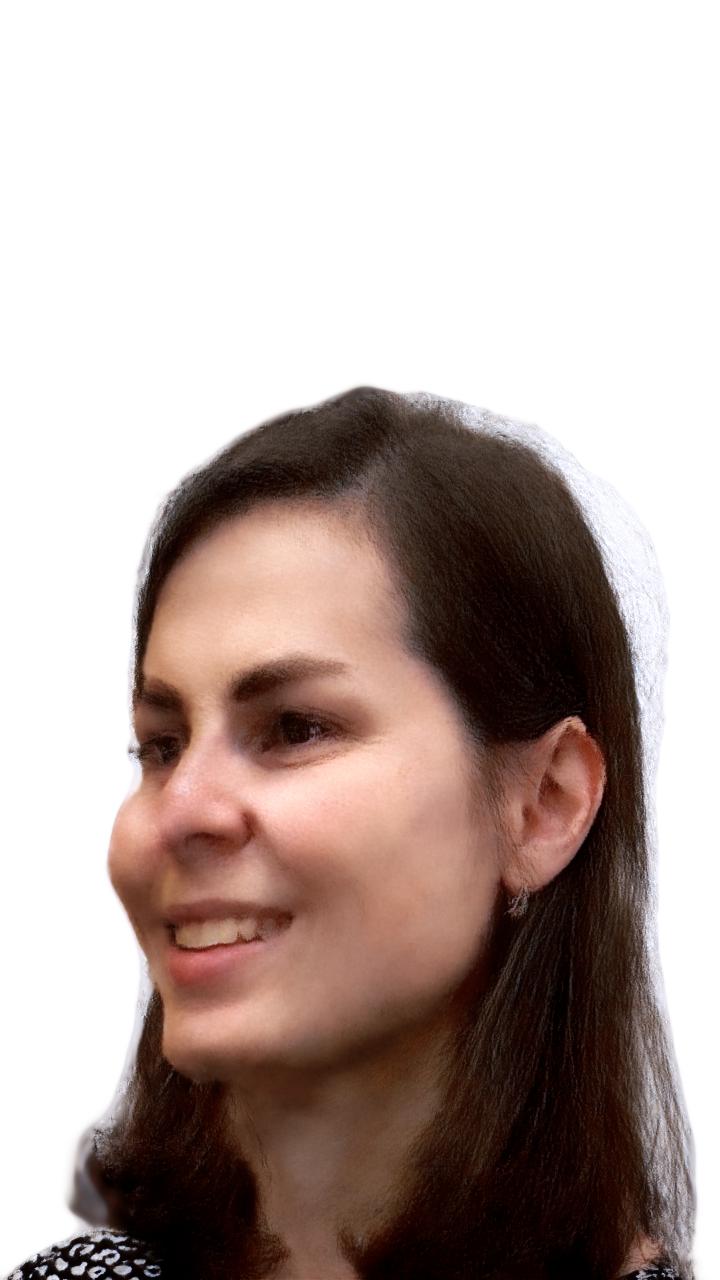}
    \end{subfigure}
    \begin{subfigure}{.11\textwidth}
        \centering
        \includegraphics[width=\textwidth]{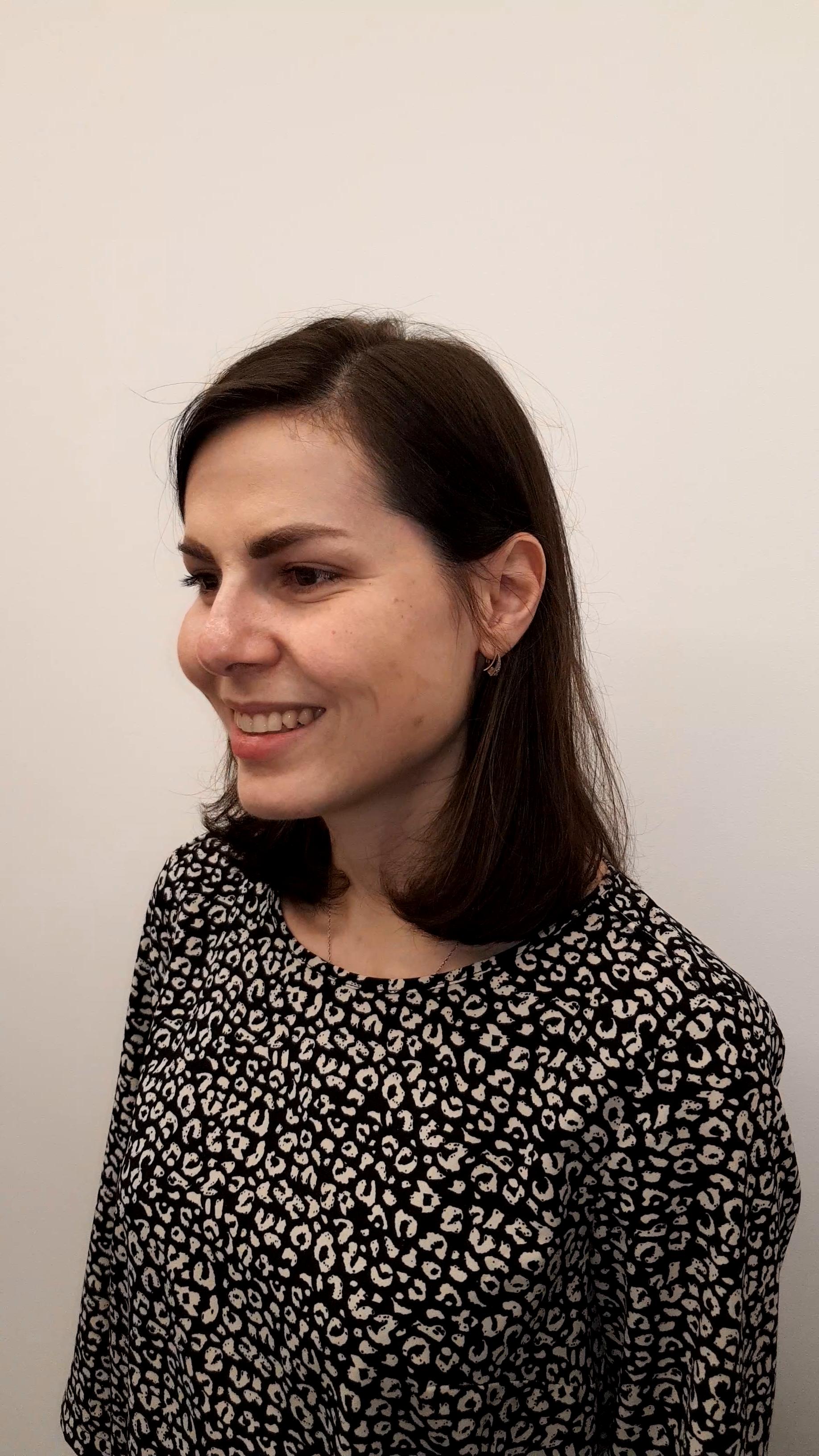}
    \end{subfigure}
    \hfill
    \begin{subfigure}{.11\textwidth}
        \centering
        \includegraphics[width=\textwidth]{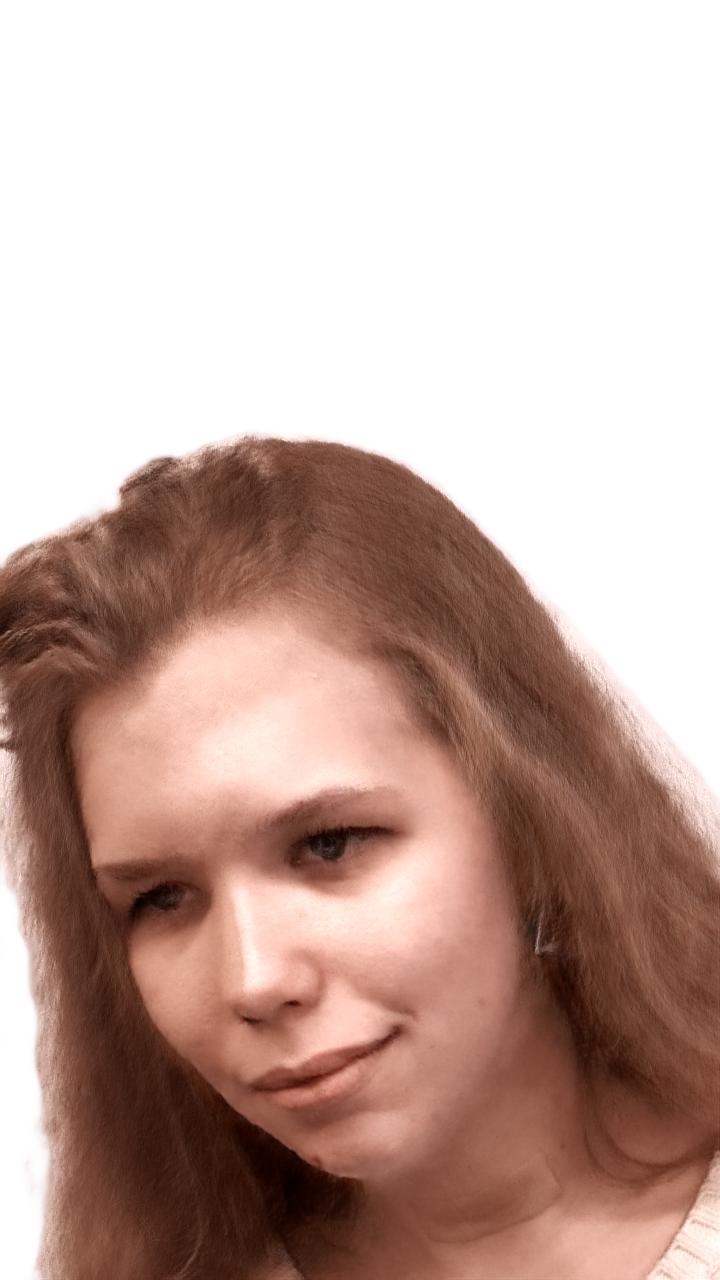}
    \end{subfigure}
    \begin{subfigure}{.11\textwidth}
        \centering
        \includegraphics[width=\textwidth]{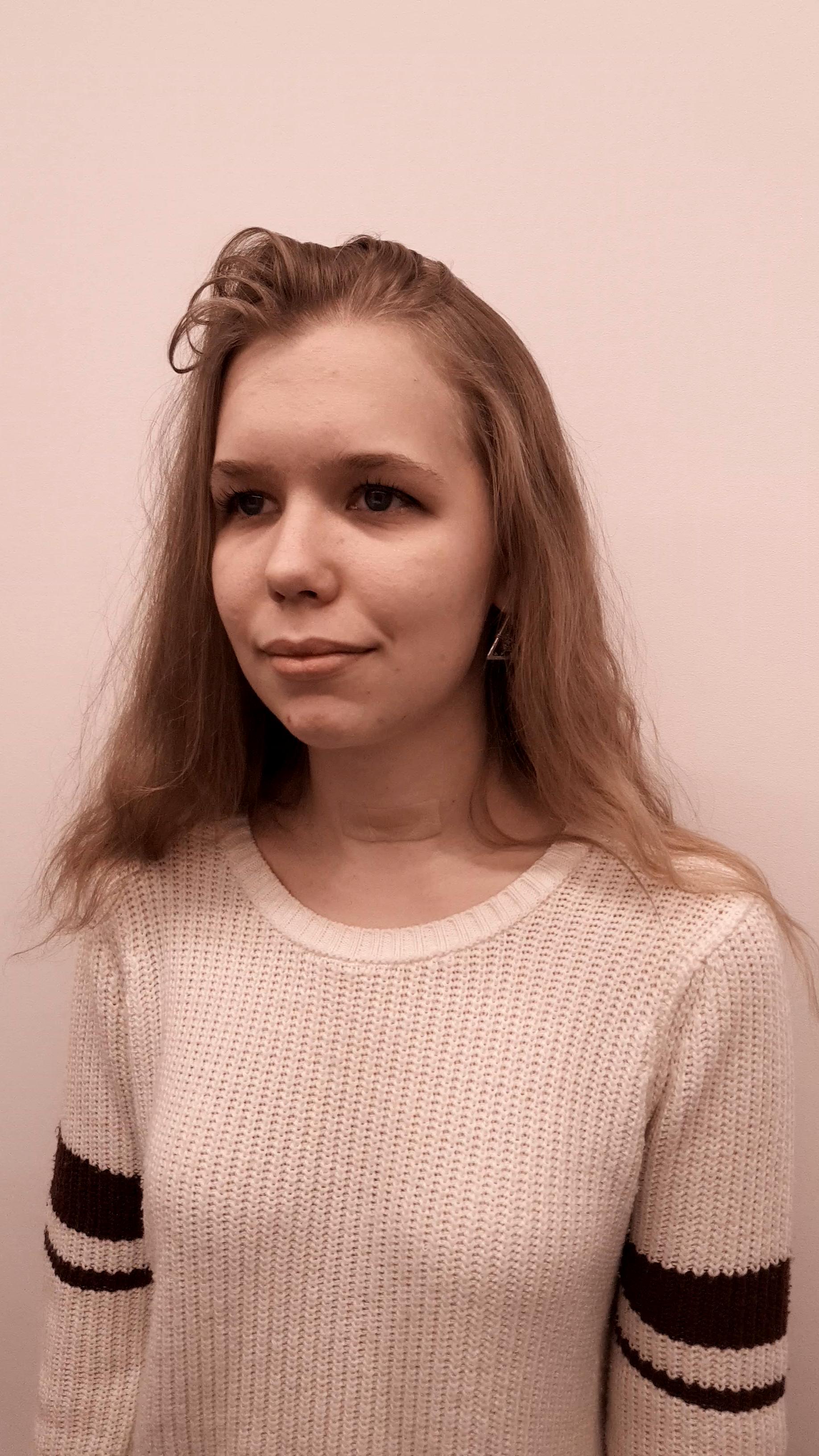}
    \end{subfigure}
    \hfill
    \begin{subfigure}{.11\textwidth}
        \centering
        \includegraphics[width=\textwidth]{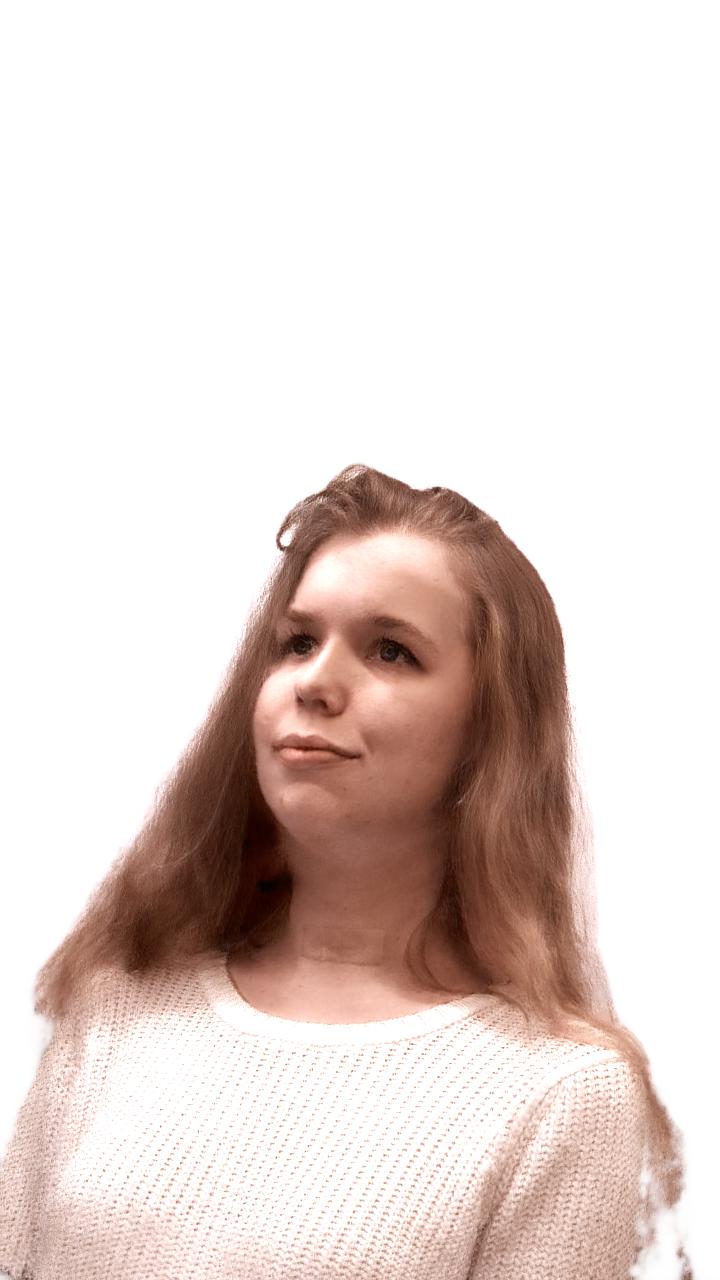}
    \end{subfigure}
    \begin{subfigure}{.11\textwidth}
        \centering
        \includegraphics[width=\textwidth]{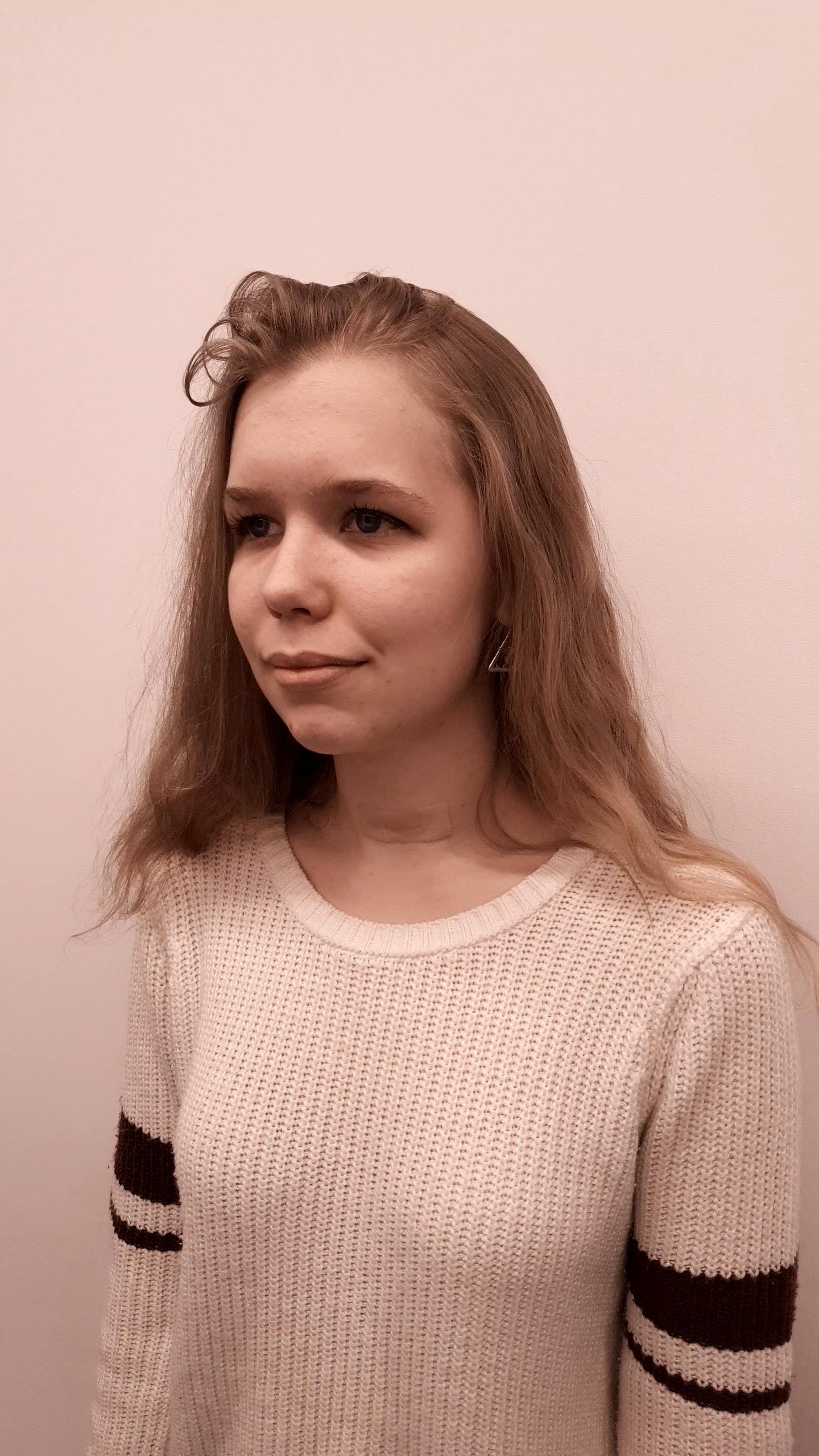}
    \end{subfigure}
    \caption{Our models rendered from viewpoints selected far from the trajectory of capture. In each case, the nearest frame from the captured video is shown alongside. The nearest frame is selected by taking 5\% of train viewpoints with the most similar view angle and then selecting the view with the closest viewpoint among those. The models are rendered with additional randomly-directed light. The segmentation artefacts near the top of the head can be attributed to the sparseness of point clouds in that region resulting from our capture protocol. \textit{Electronic zoom-in recommended.}}
    \label{fig:real_far_viewpoints}
\end{figure*}

In Fig.~\ref{fig:real_SH}, the results of the comparison with a recently released DPR method~\cite{Zhou19} are presented. To make the comparison feasible, we relight the head by a combination of ambient light and $1^\mathrm{st}$-order Spherical Harmonics (SH)~\cite{Ramamoorthi01}. DPR is based on an end-to-end neural network that receives a single image, a new SH lighting to be applied, and returns a relighted image. We cropped inputs to DPR by a face localization network~\cite{Zhang16} and enlargening the bounding boxes to make them closer to the cropping of faces in the training set of DPR (CelebA-HQ \cite{Karras17}). % In addition, we replace ambient light by the captured environmental lighting of the room, similarly to the additional lighting scheme employed above, and report the results in the \textit{Ours (add. lighting)} row. We found that this kind of relighting often yields to more plausible results, and at the same time resembles the reshading behaviour of DPR more closely. 
In contrast to our method, DPR requires only one image and is able to realistically relight it, though it can be seen that some areas can be incorrectly reshaded at the side views. Our method operates on a video covering all head parts, but produces more consistent lighting and can additionally render a head from novel viewpoints. Performance of relighting with simultaneous view resynthesis is showcased in Fig.~\ref{fig:real_far_viewpoints}. There, the renderings in additional lighting setting are showcased along with the nearest views from the captured video.  The nearest frame is selected by taking 5\% of train viewpoints with the most similar view angle and then selecting the view with the closest viewpoint among those. Since random zoom-in/zoom-out and random cropping are used at the training stage, angle deviation is more critical when comparing the predicted result to the nearest frame.

% In the Supplementary Material, we provide additional qualitative and quantitative comparison on synthetic data (using the human models from \cite{RenderPeople}).

% \begin{table}[]
% \centering
%     \begin{tabular}{l|c|c|c}
%                  & VGG                   & FID                   & \multicolumn{1}{c|}{IS} \\ \hline
%     Ours       &                       &                       &  \multicolumn{1}{c|}{}   \\ \hline
    
%     \end{tabular}
%     \newline
%     \caption{Comparison of albedo, normals estimation and final rendering quality over 5 smartphone-captured head portraits. To report the results, we use 3 perceptual metrics commonly used in the field, evaluated only on holdout flashlighted images.}
%     \label{tab:my_label}
% \end{table}

\subsection{Synthetically created people}

In addition to the main experiments with smartphone videos, we conducted evaluations on synthetically rendered data. This comparison is based on the upper-body part of three full-body models from RenderPeople~\cite{RenderPeople}. Each model is defined as a mesh with diffuse texture, normal texture, and other parameters. Using Blender Eevee rendering engine, we closely simulate the flash - no flash acquisition scheme employed for real portraits. The half-circular camera trajectory is drawn in front of the subject, with each view covering the head and shoulders of the model, and is evenly divided into 100 viewpoints. For every fifth viewpoint, a directional light source of constant power is added to the scene, co-located with the camera. The lighted subject is rendered for each of the 100 viewpoints in $3000 \times 3000$ resolution and 32-bit floating point HDR format without compression. Point clouds with 7 million points are uniformly sampled from the models meshes.

Similarly to the smartphone videos preprocessing pipeline, here we process each rendered frame with PRNet~\cite{Feng18}, following the same procedure conducted for real portraits (it performs at the similar level of accuracy on synthetic data from RenderPeople). Additionally, we precompute the renderings of each model colored with its diffuse texture (albedo render) and normals texture (normal render with world-space XYZ colored as RGB). Among 100 frames, we selected every $20^\mathrm{th}$ (a flashlighted one), as well as one preceding and one subsequent frame for every one of them for validation (resulting in 30 frames total), while the rest were used for training (fitting). In contrast to the real head portraits, here we initialize the albedo texture $\mathcal{T}_A$ not by the median flashlighted texture $\mathcal{T}_F$, but by the median \textbf{non}-flashlighted texture $\mathcal{T}_{NF}$:

\begin{equation}
    \begin{aligned}
    \mathcal{T}_{NF} &= \mathrm{median}(I_{p_1} \odot \mathrm{Posmap}_{p_1},\, \dots,\, I_{p_T} \odot \mathrm{Posmap}_{p_T}), \\
    & \{p_1, \dots, p_T\} \cup \{s_1, \dots, s_M\} = \{1, \dots, P\},\\
    & \{p_1, \dots, p_T\} \cap \{s_1, \dots, s_M\} = \varnothing
    \end{aligned}
\end{equation}

\noindent (less formally, $p_1, \dots, p_T$ denote the indices of non-flashlighted captured frames). This is done here to make the comparison between the estimated albedo and the ground truth albedo more precise by correctly matching their color temperature.

\begin{figure*}[h]
    \centering
    % trim: left lower right upper
    % ---------------- Carla
    \textit{Person 1 (Carla)}
    
    \rotatebox{90}{ }
    \hspace{0.25cm}
    \adjincludegraphics[trim={{.26\width} {.44\height} {.61\width} {.35\width}},clip,width=.135\linewidth]{images/schemes/arrowsX.pdf}
    \adjincludegraphics[trim={{.26\width} {.44\height} {.61\width} {.35\width}},clip,width=.135\linewidth]{images/schemes/arrowsY.pdf}
    \adjincludegraphics[trim={{.26\width} {.44\height} {.61\width} {.35\width}},clip,width=.135\linewidth]{images/schemes/arrowsZ.pdf}
    % \hfill
    \adjincludegraphics[trim={{.26\width} {.44\height} {.61\width} {.35\width}},clip,width=.135\linewidth]{images/schemes/arrowsX.pdf}
    \adjincludegraphics[trim={{.26\width} {.44\height} {.61\width} {.35\width}},clip,width=.135\linewidth]{images/schemes/arrowsY.pdf}
    % \hfill
    \adjincludegraphics[trim={{.26\width} {.44\height} {.61\width} {.35\width}},clip,width=.135\linewidth]{images/schemes/arrowsX.pdf}
    \adjincludegraphics[trim={{.26\width} {.44\height} {.61\width} {.35\width}},clip,width=.135\linewidth]{images/schemes/arrowsY.pdf}
    % \hfill
    
    \rotatebox{90}{\hspace{0.4cm} Ours}
    \adjincludegraphics[trim={0 0 0 0},clip,width=.135\linewidth]{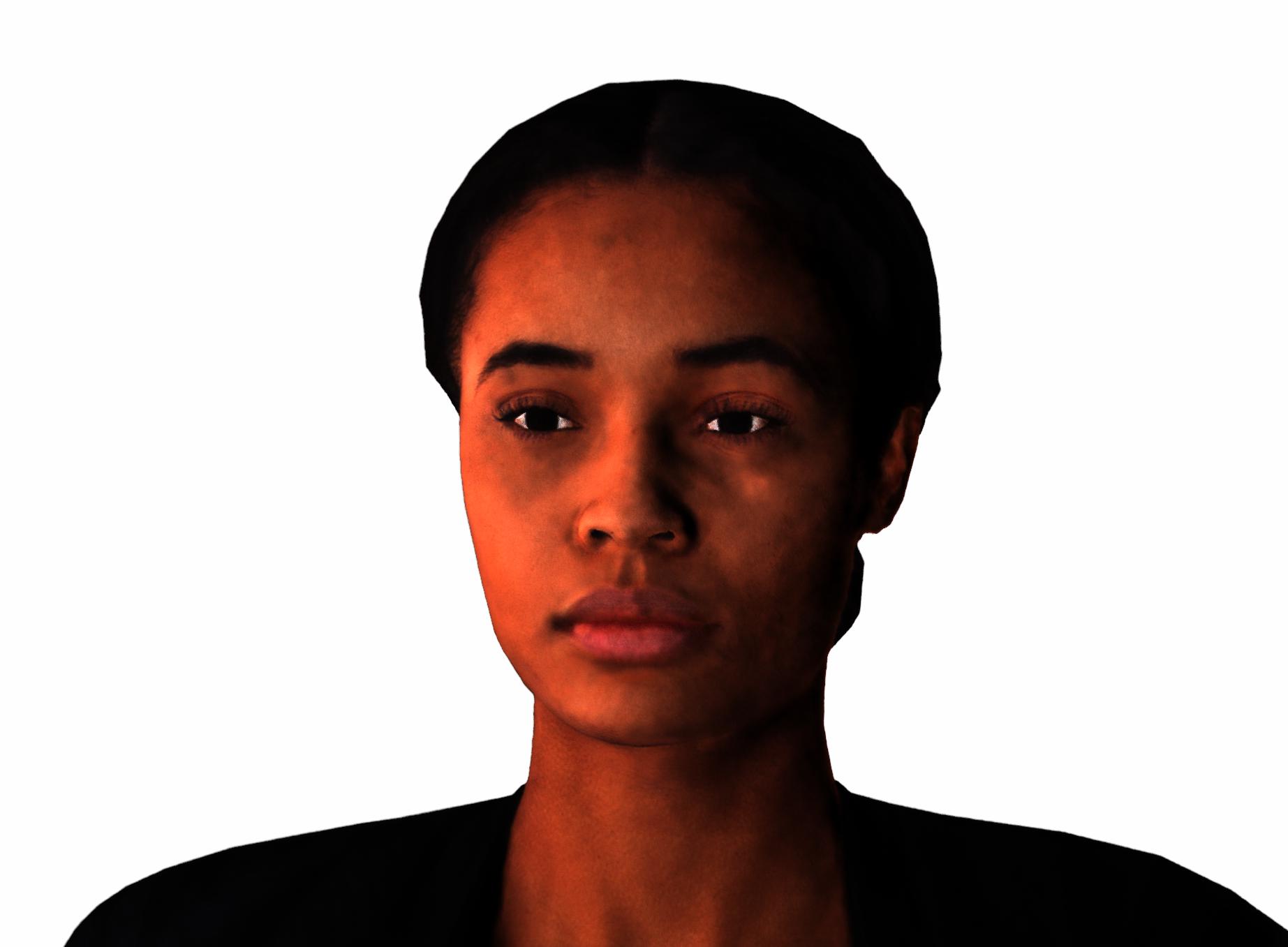}
    \adjincludegraphics[trim={0 0 0 0},clip,width=.135\linewidth]{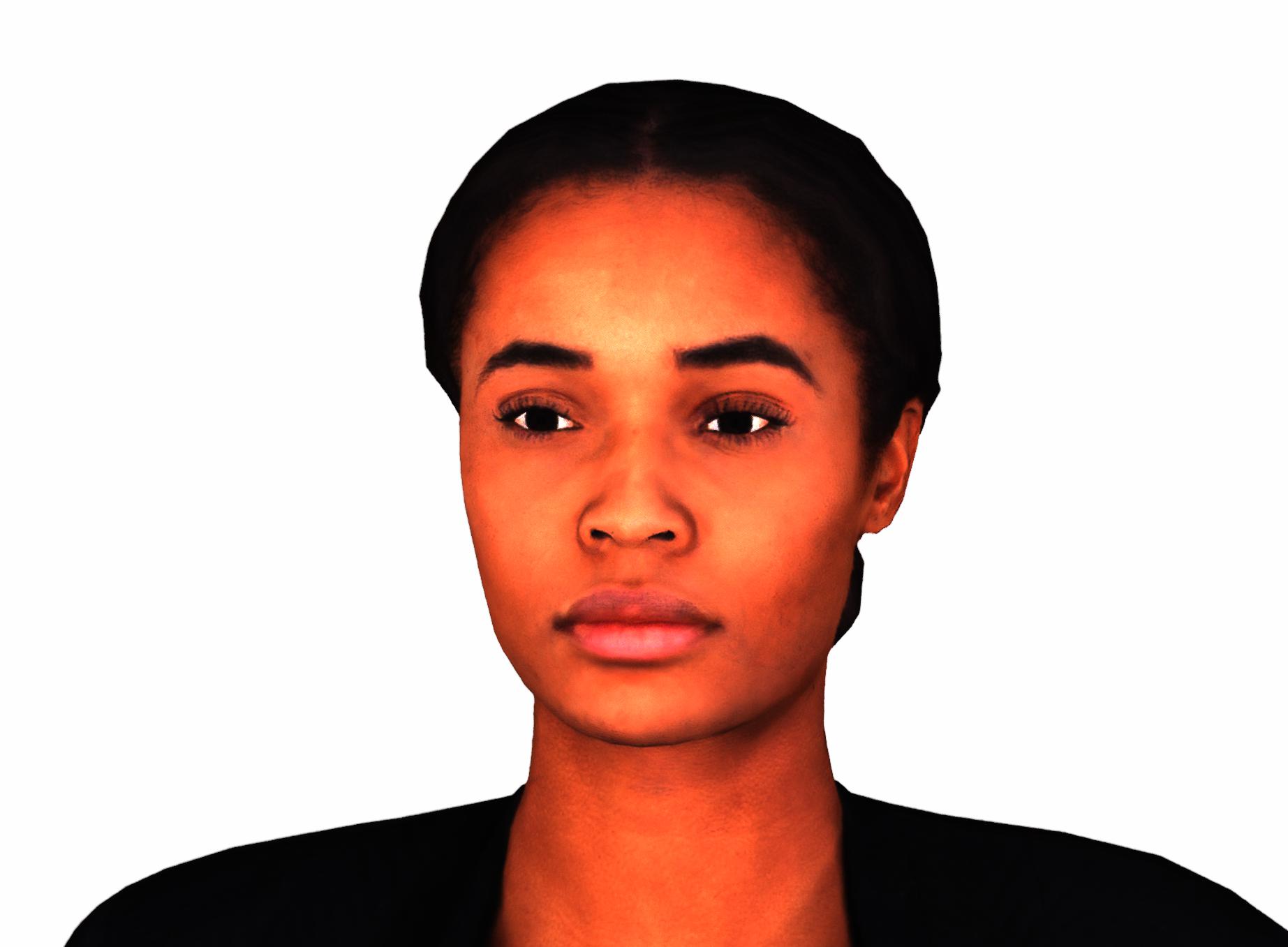}
    \adjincludegraphics[trim={0 0 0 0},clip,width=.135\linewidth]{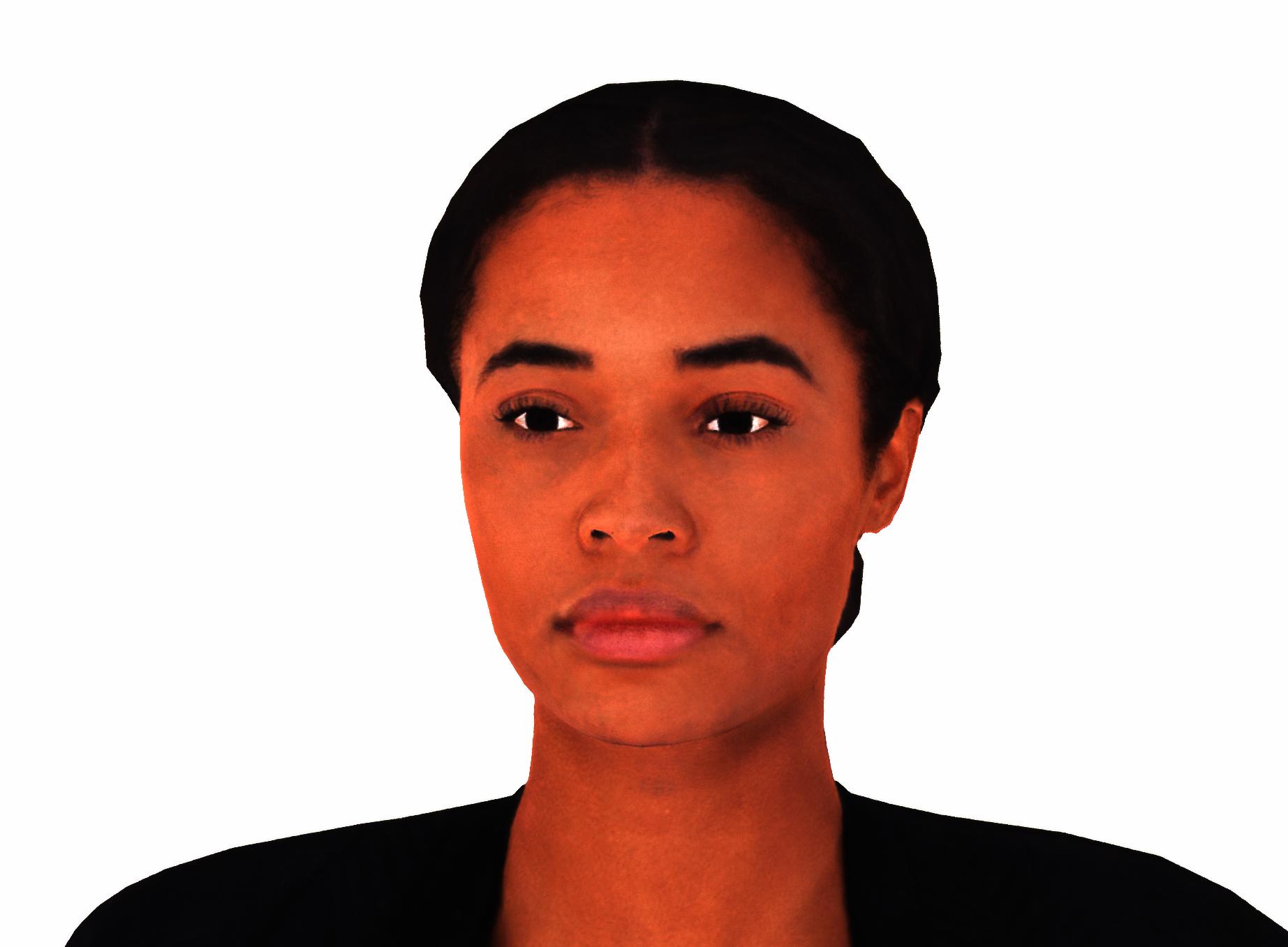}
    \hspace{0.5cm}
    % \hfill
    \adjincludegraphics[trim={0 0 {.25\width} 0},clip,width=.1\linewidth]{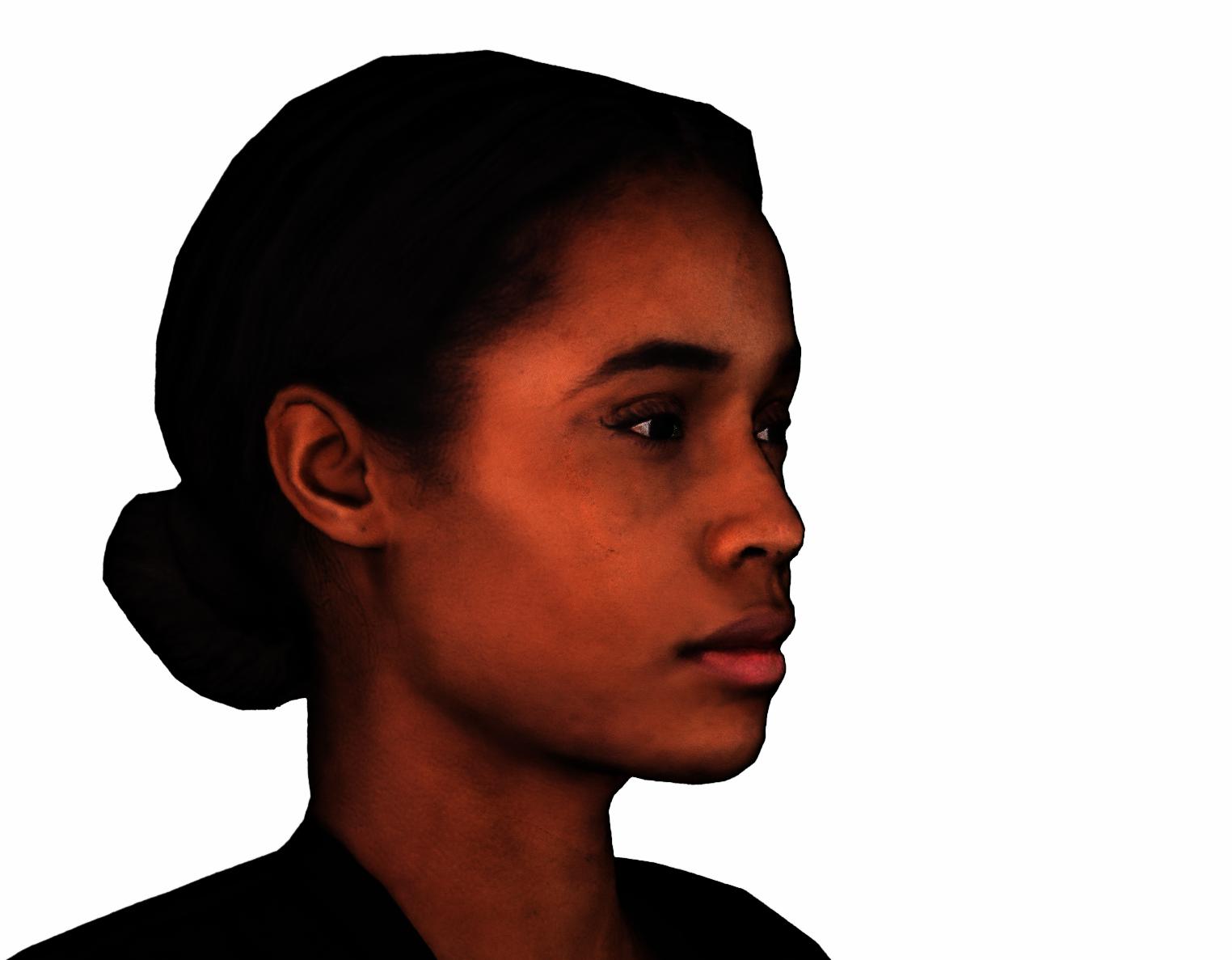}
    \hspace{0.25cm}
    \adjincludegraphics[trim={0 0 {.25\width} 0},clip,width=.1\linewidth]{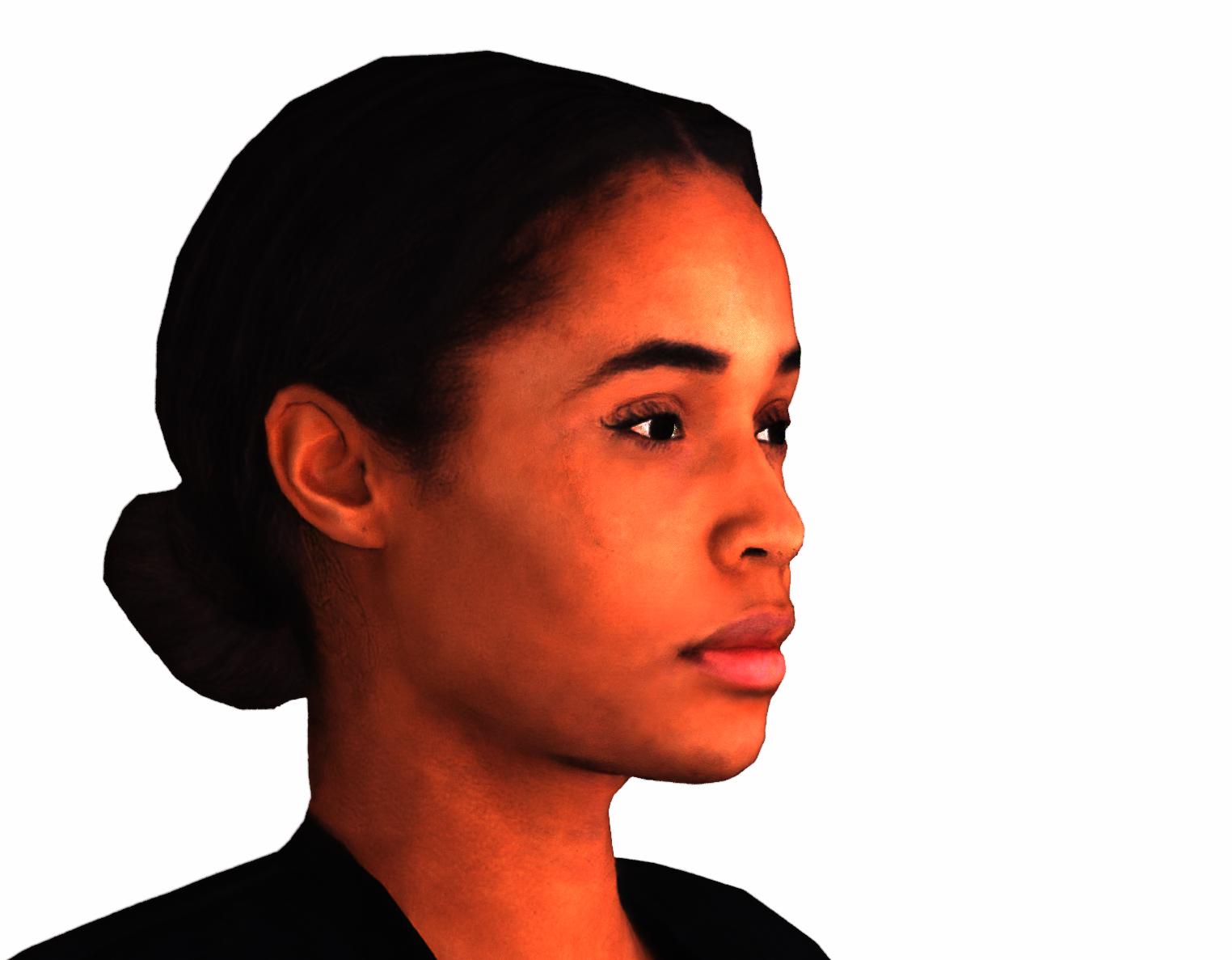}
    % \hfill
    \adjincludegraphics[trim={0 0 0 0},clip,width=.135\linewidth]{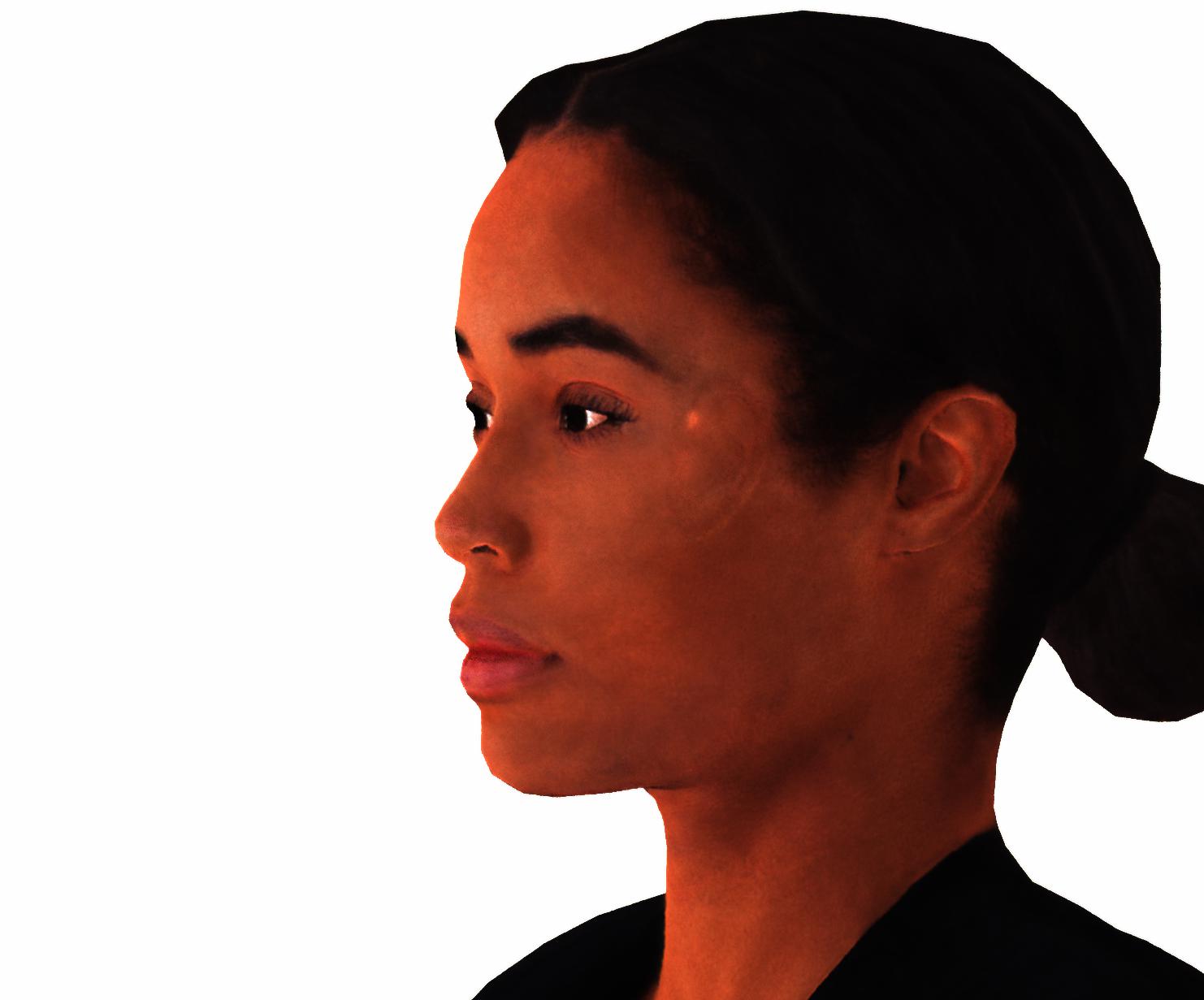}
    \adjincludegraphics[trim={0 0 0 0},clip,width=.135\linewidth]{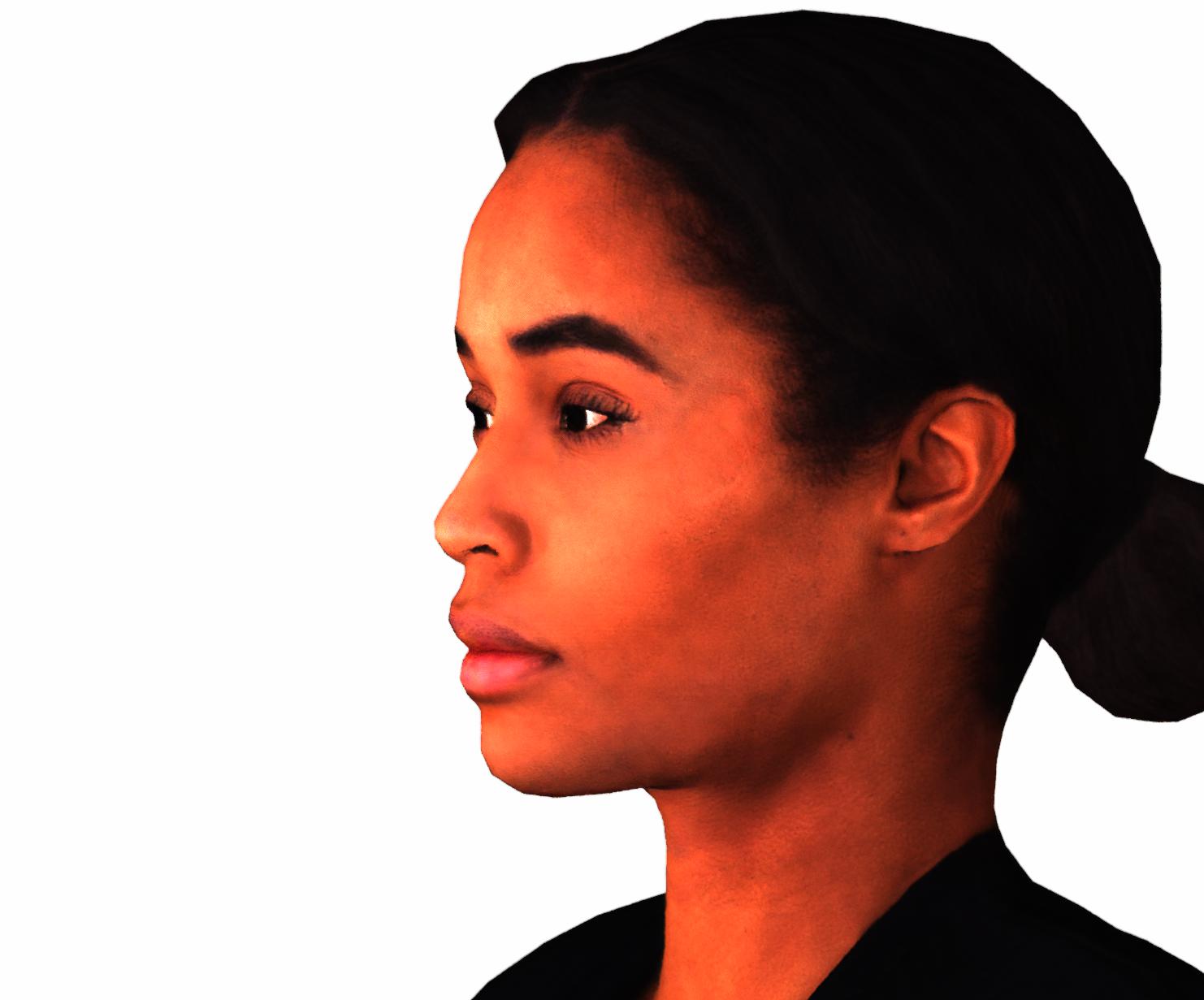}
    
    \rotatebox{90}{\hspace{0.25cm} DPR~\cite{Zhou19}}
    \adjincludegraphics[trim={0 0 0 0},clip,width=.135\linewidth]{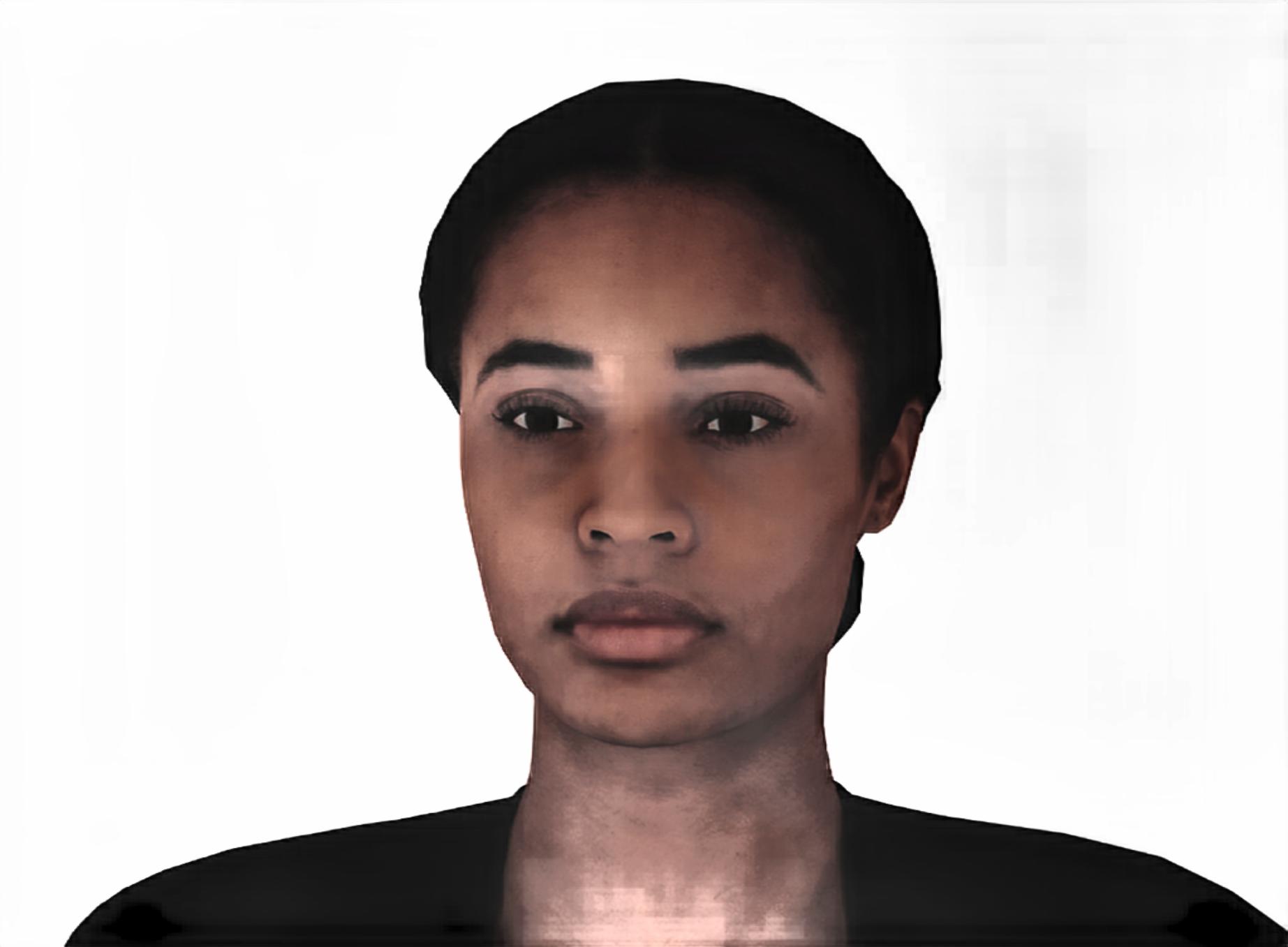}
    \adjincludegraphics[trim={0 0 0 0},clip,width=.135\linewidth]{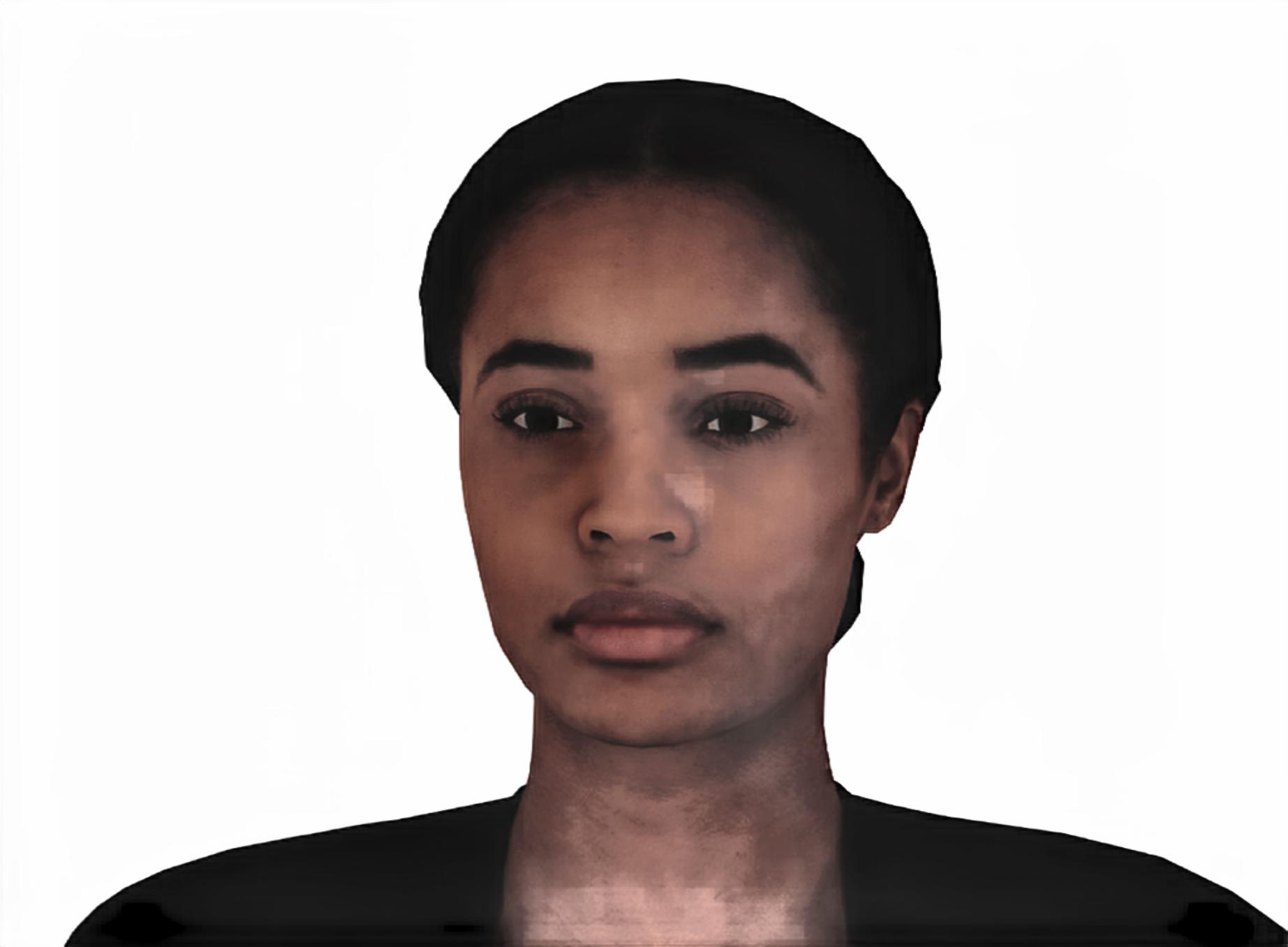}
    \adjincludegraphics[trim={0 0 0 0},clip,width=.135\linewidth]{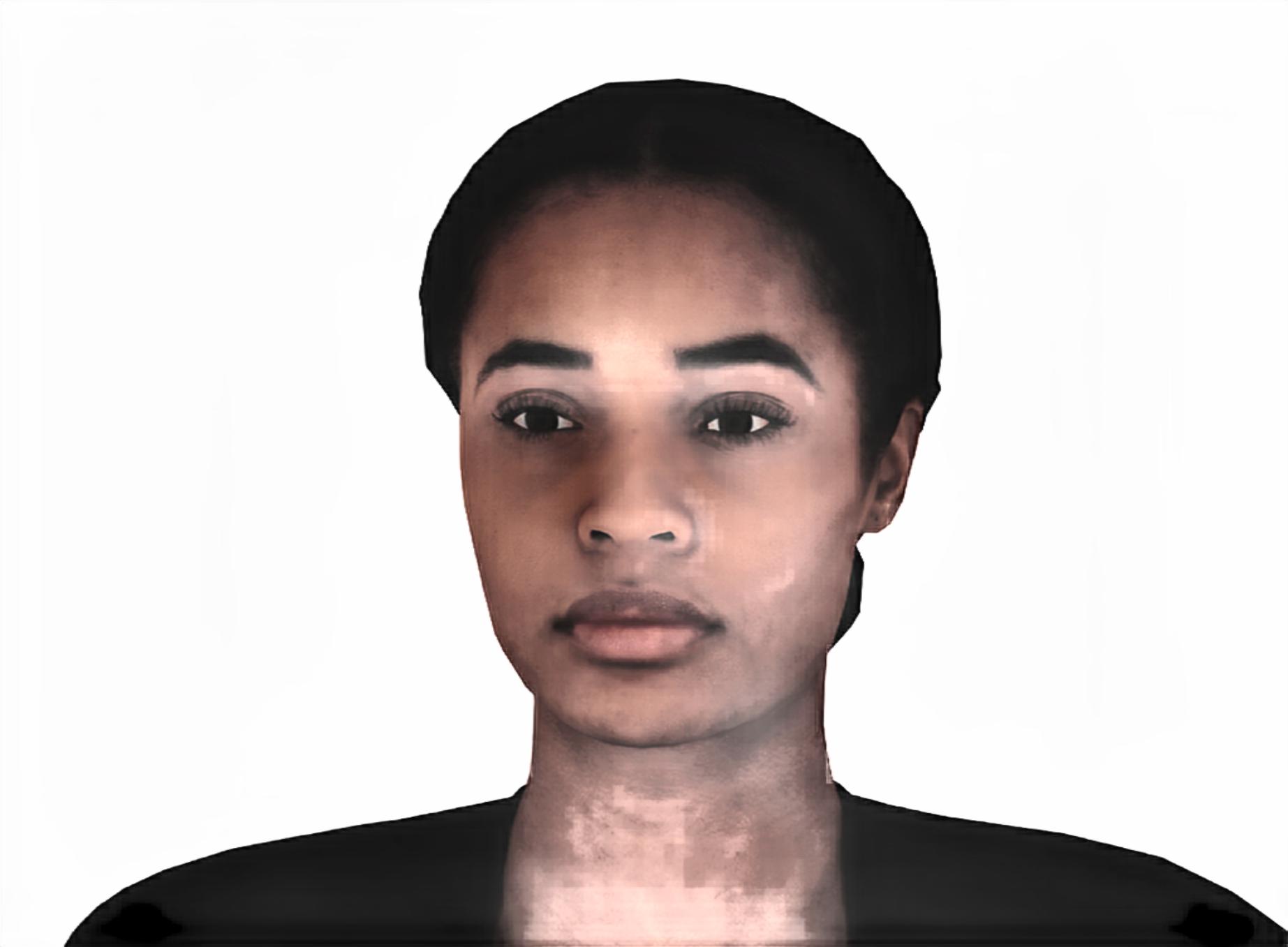}
    \hspace{0.5cm}
    % \hfill
    \adjincludegraphics[trim={0 0 {.25\width} 0},clip,width=.1\linewidth]{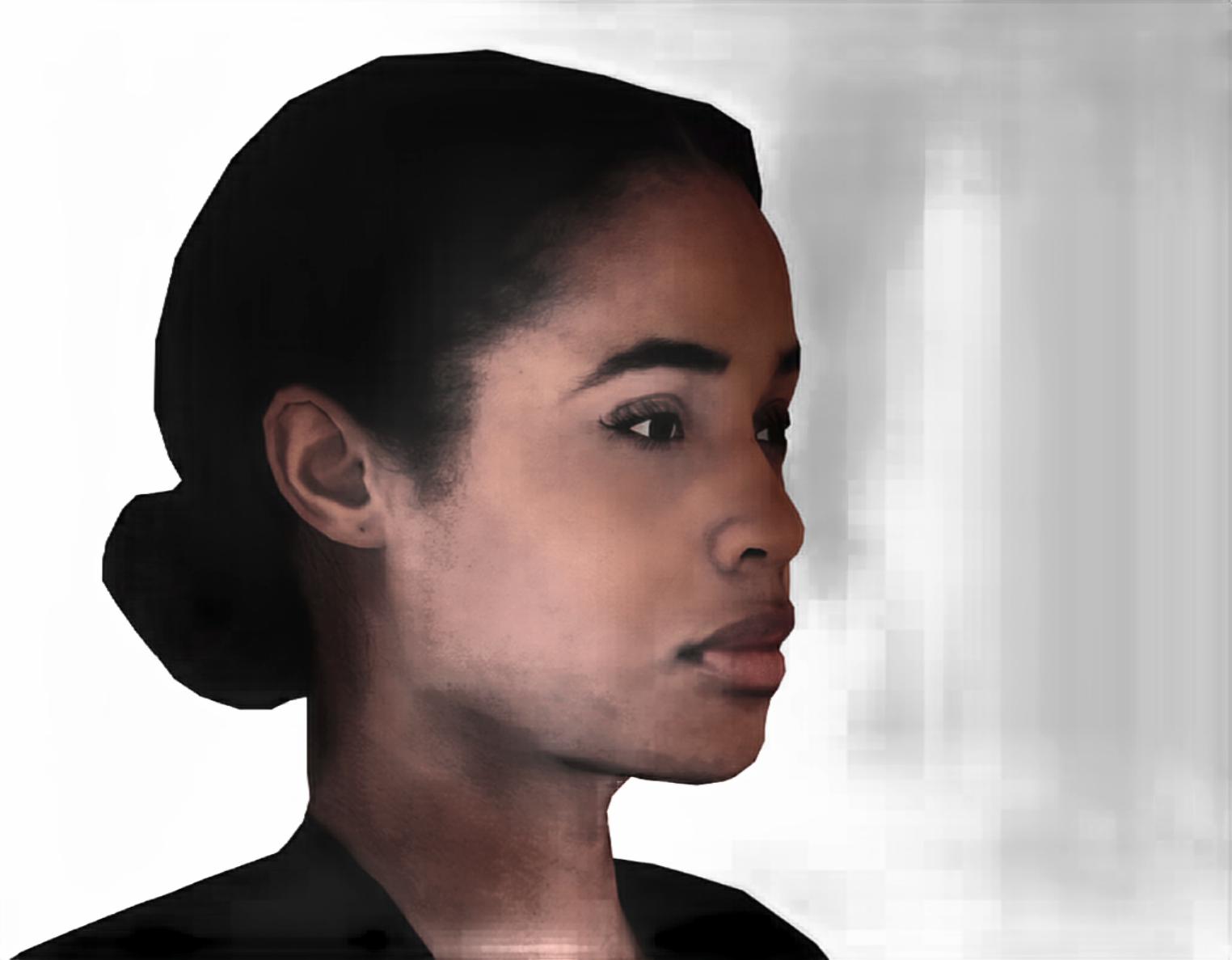}
    \hspace{0.25cm}
    \adjincludegraphics[trim={0 0 {.25\width} 0},clip,width=.1\linewidth]{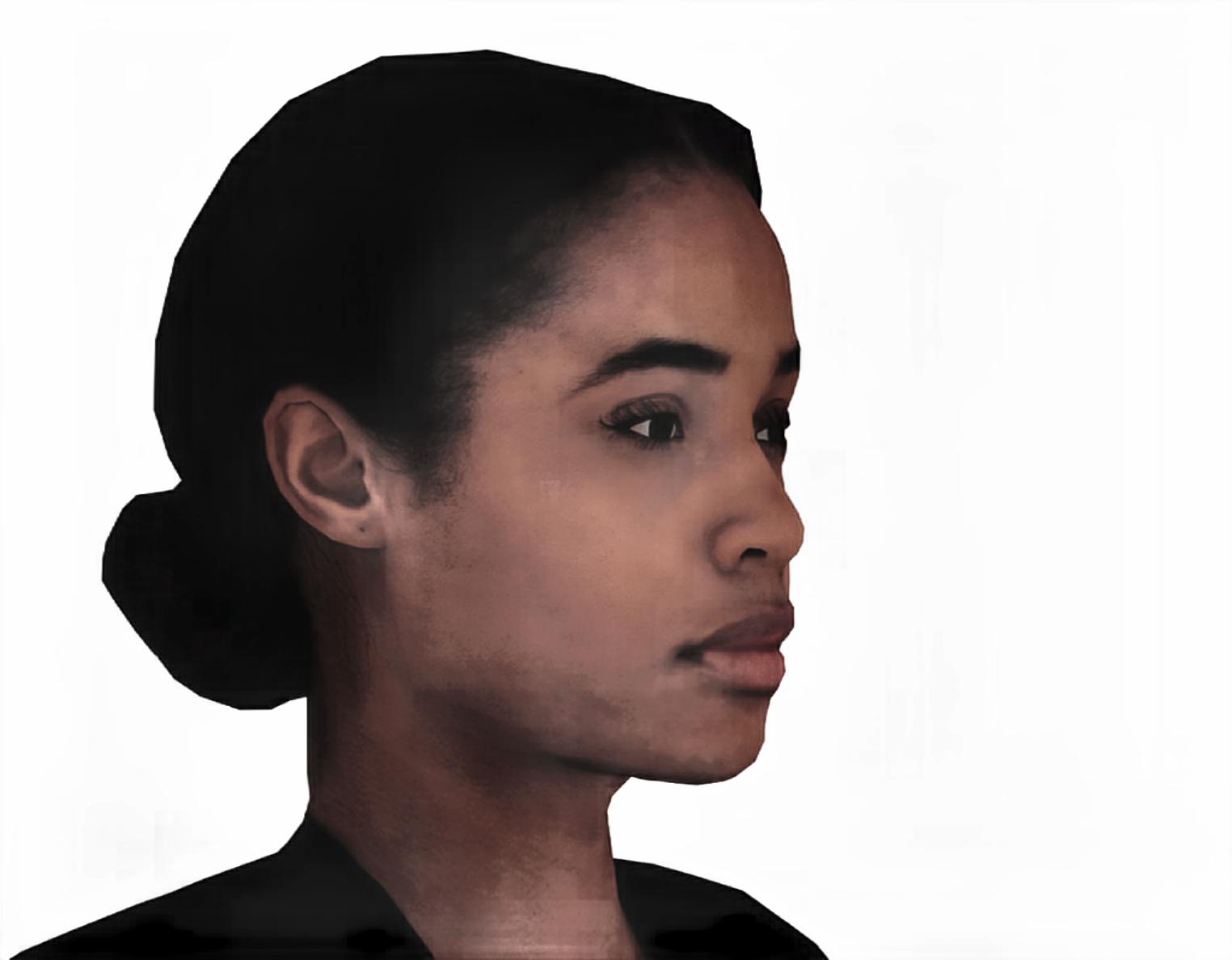}
    % \hfill
    \adjincludegraphics[trim={0 0 0 0},clip,width=.135\linewidth]{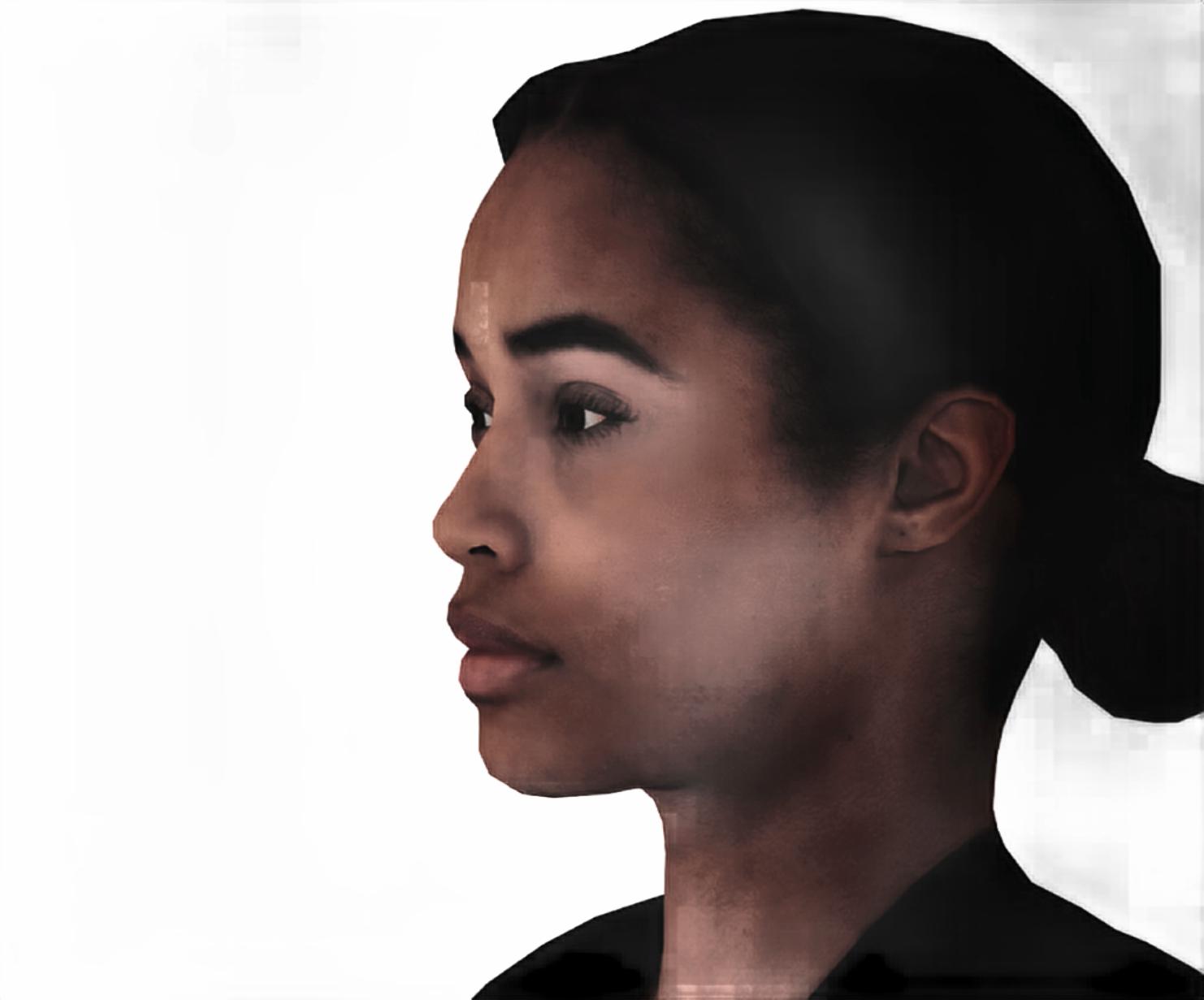}
    \adjincludegraphics[trim={0 0 0 0},clip,width=.135\linewidth]{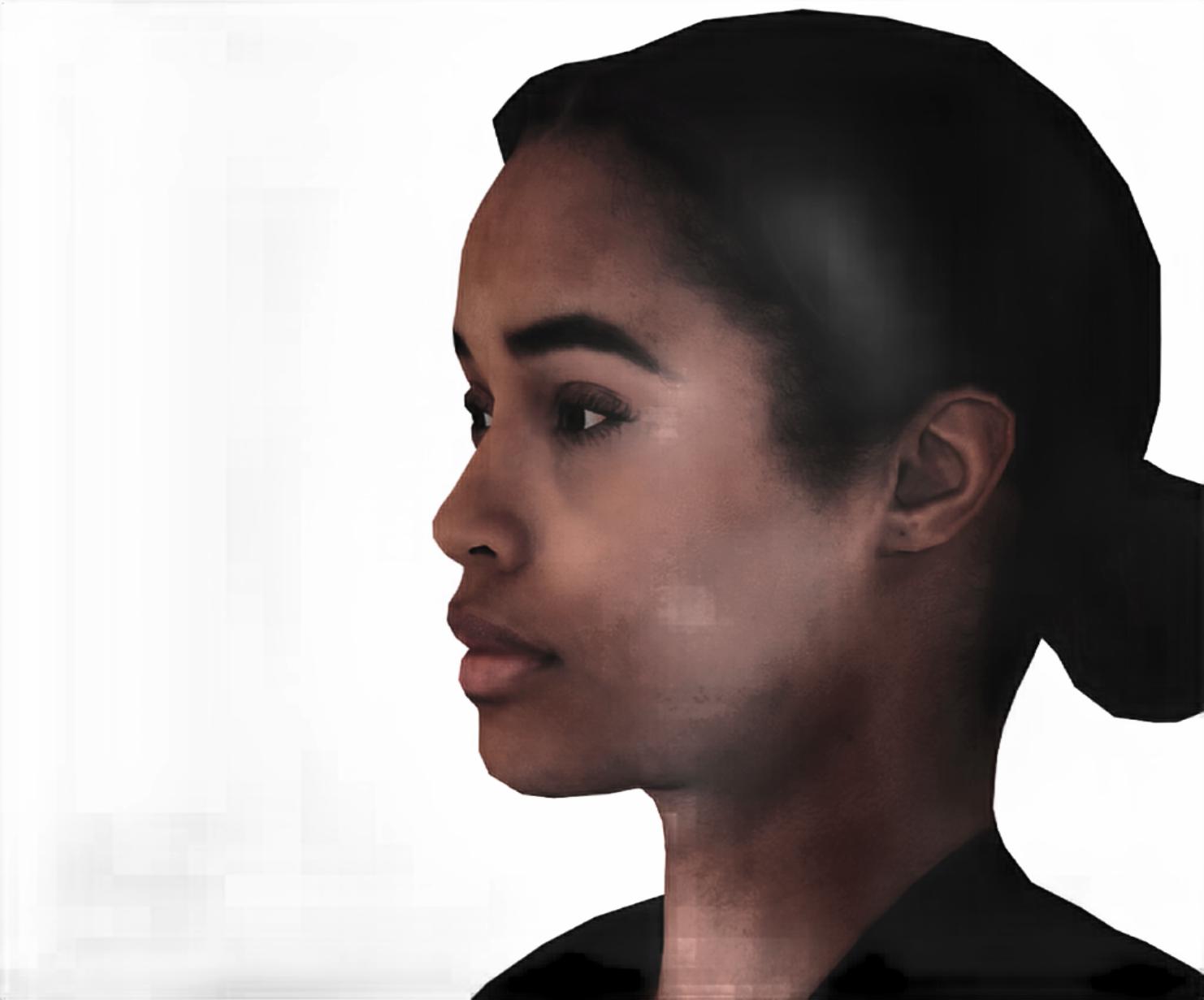}
    
    \rotatebox{90}{\hspace{-0.15cm} Ground truth}
    \adjincludegraphics[trim={0 0 0 0},clip,width=.135\linewidth]{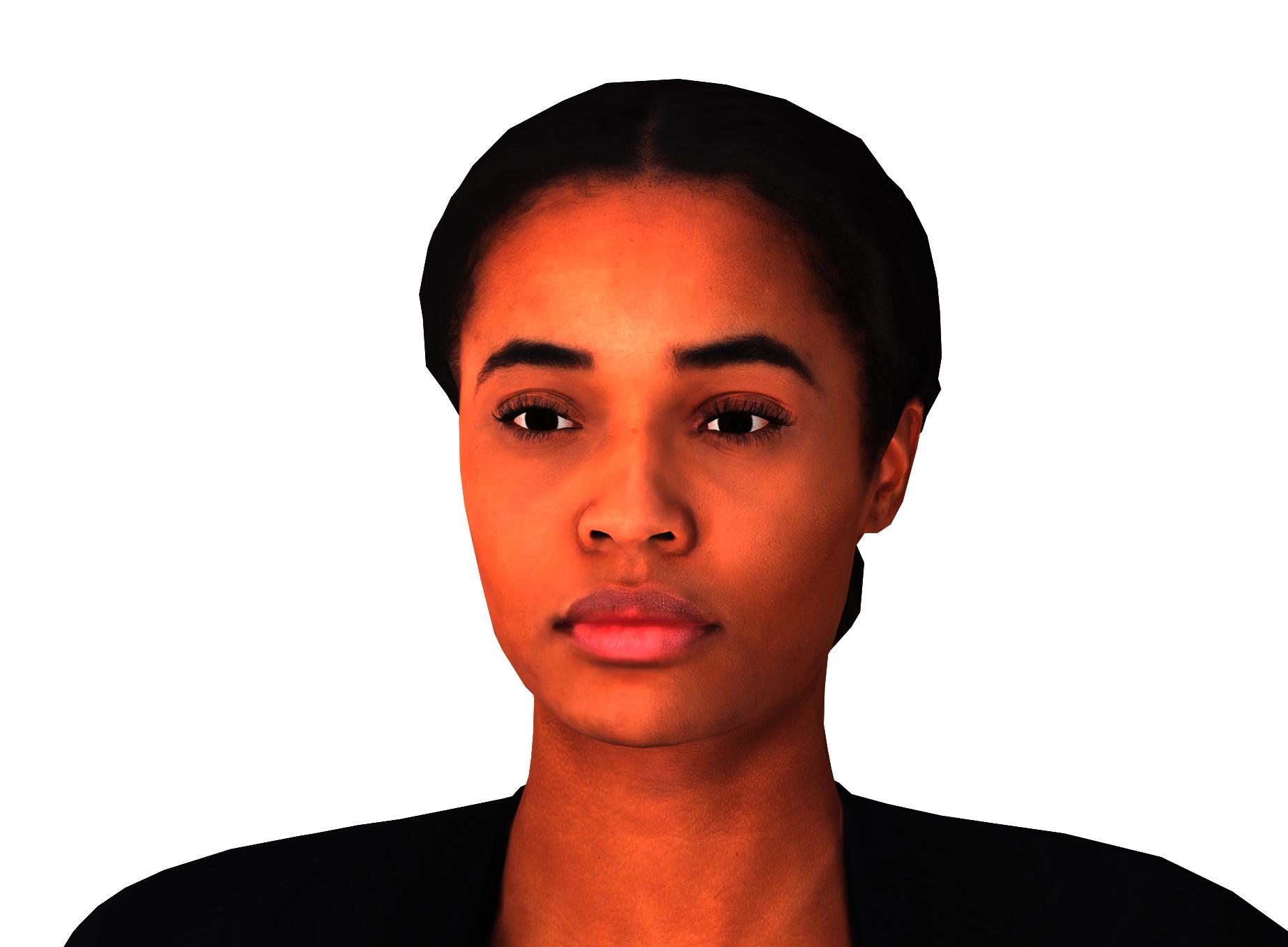}
    \adjincludegraphics[trim={0 0 0 0},clip,width=.135\linewidth]{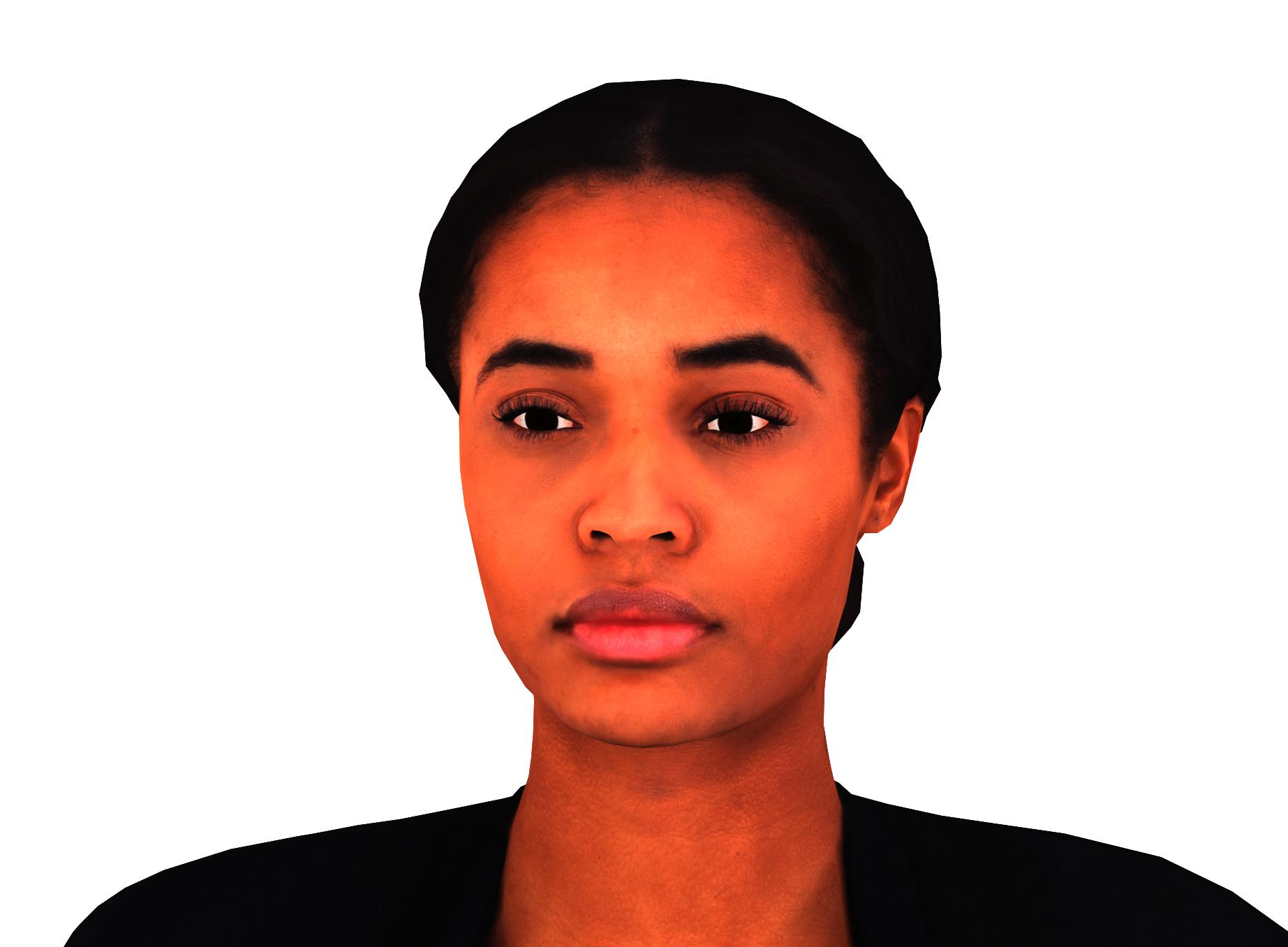}
    \adjincludegraphics[trim={0 0 0 0},clip,width=.135\linewidth]{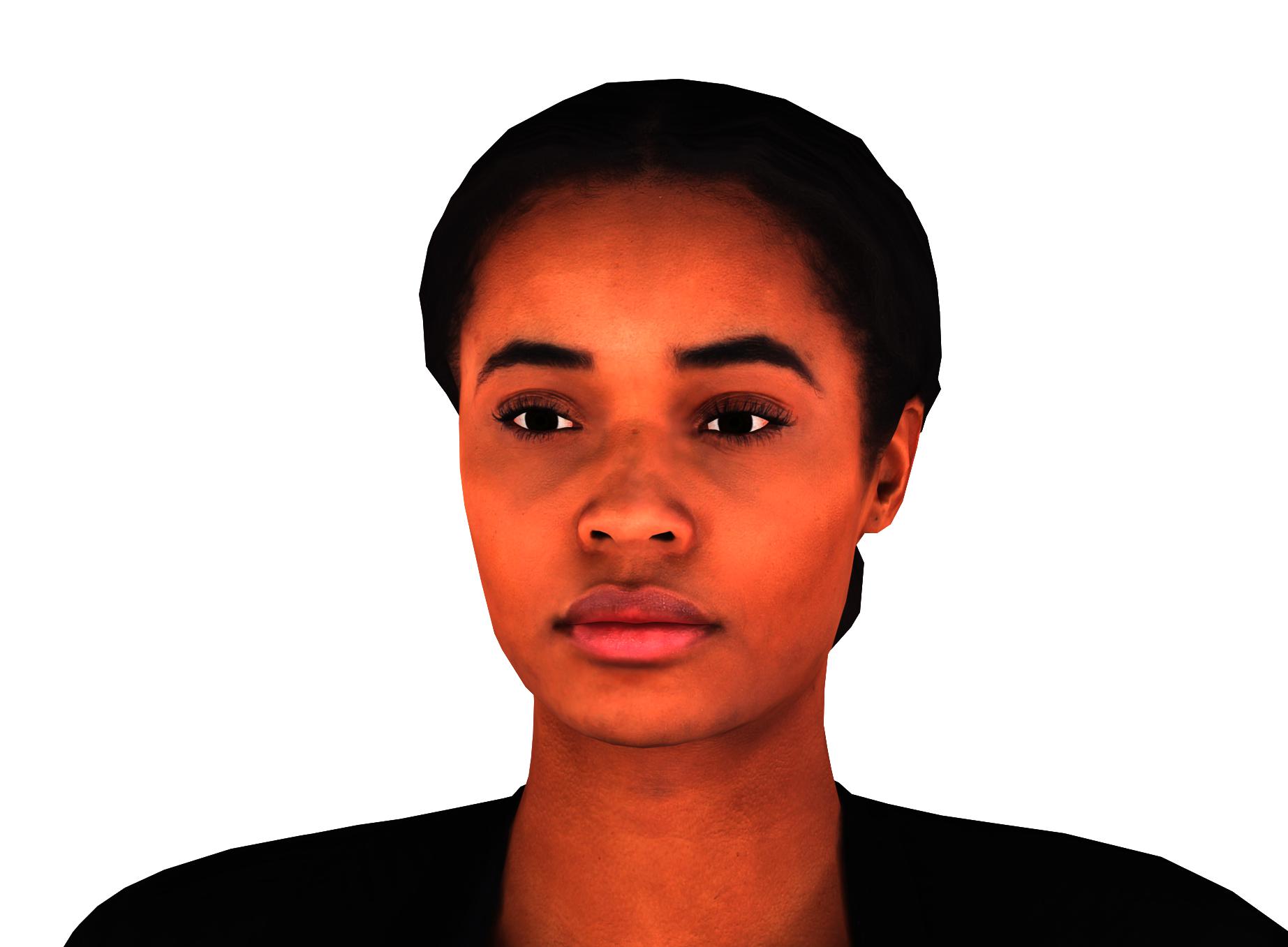}
    \hspace{0.5cm}
    % \hfill
    \adjincludegraphics[trim={0 0 {.25\width} 0},clip,width=.1\linewidth]{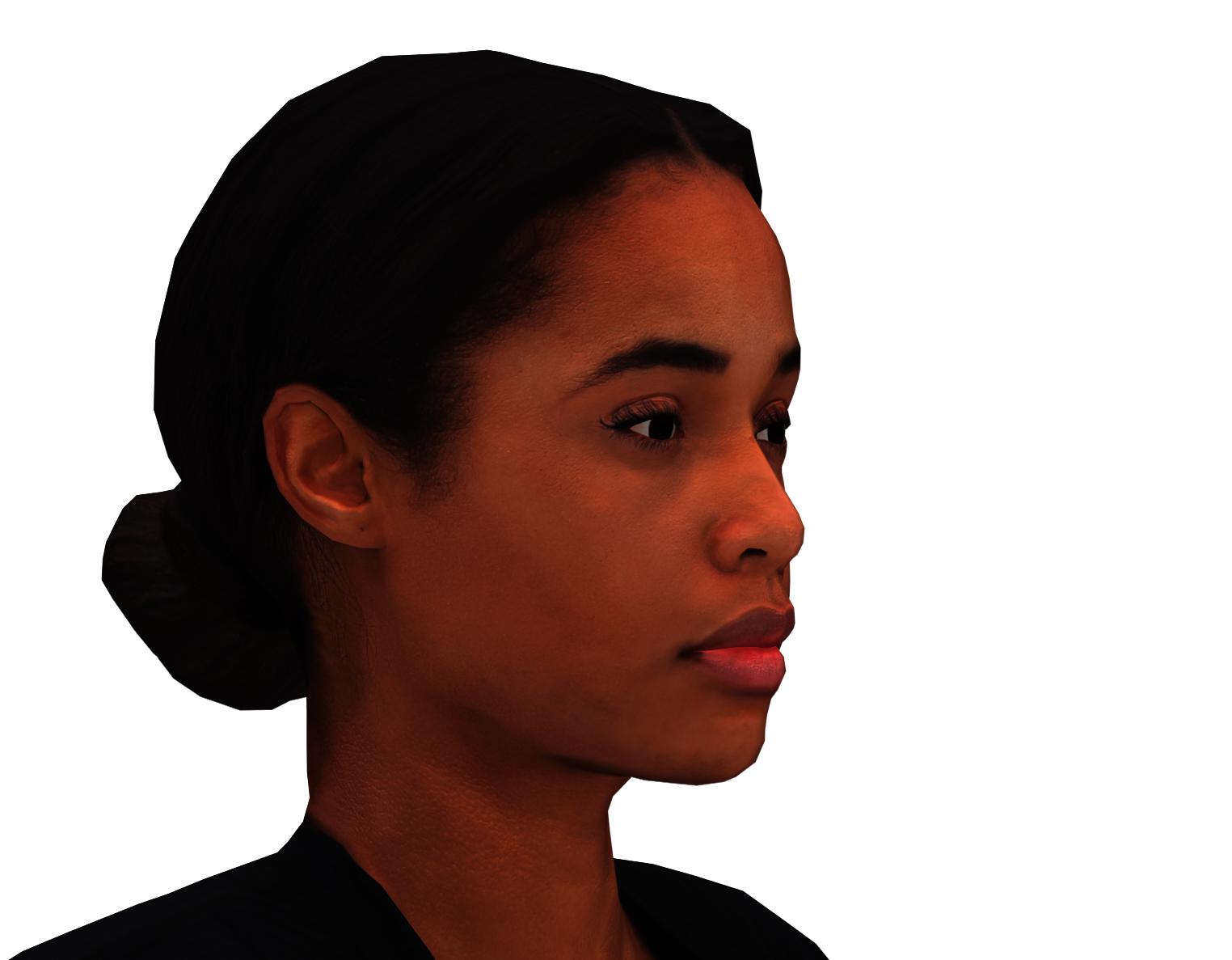}
    \hspace{0.25cm}
    \adjincludegraphics[trim={0 0 {.25\width} 0},clip,width=.1\linewidth]{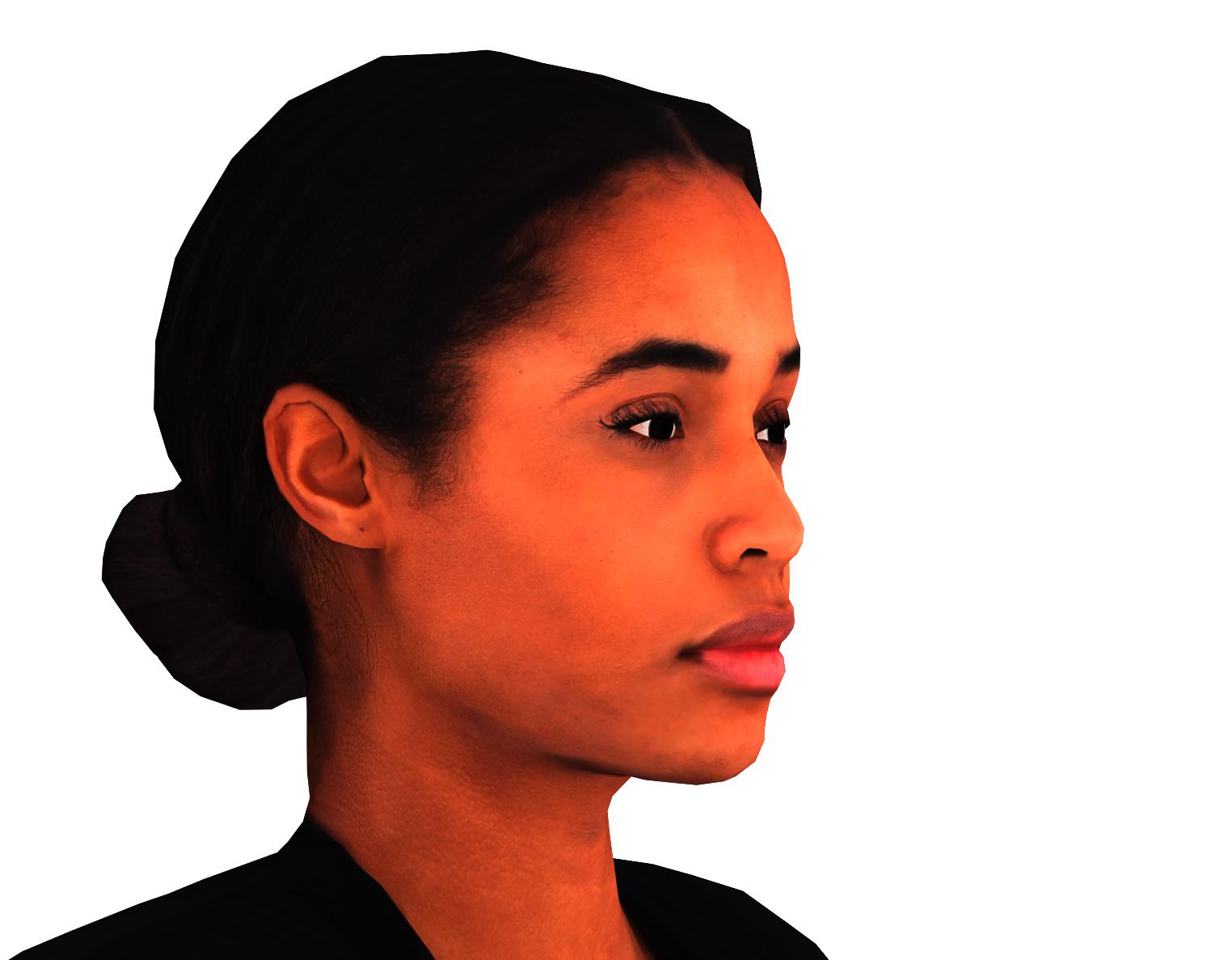}
    % \hfill
    \adjincludegraphics[trim={0 0 0 0},clip,width=.135\linewidth]{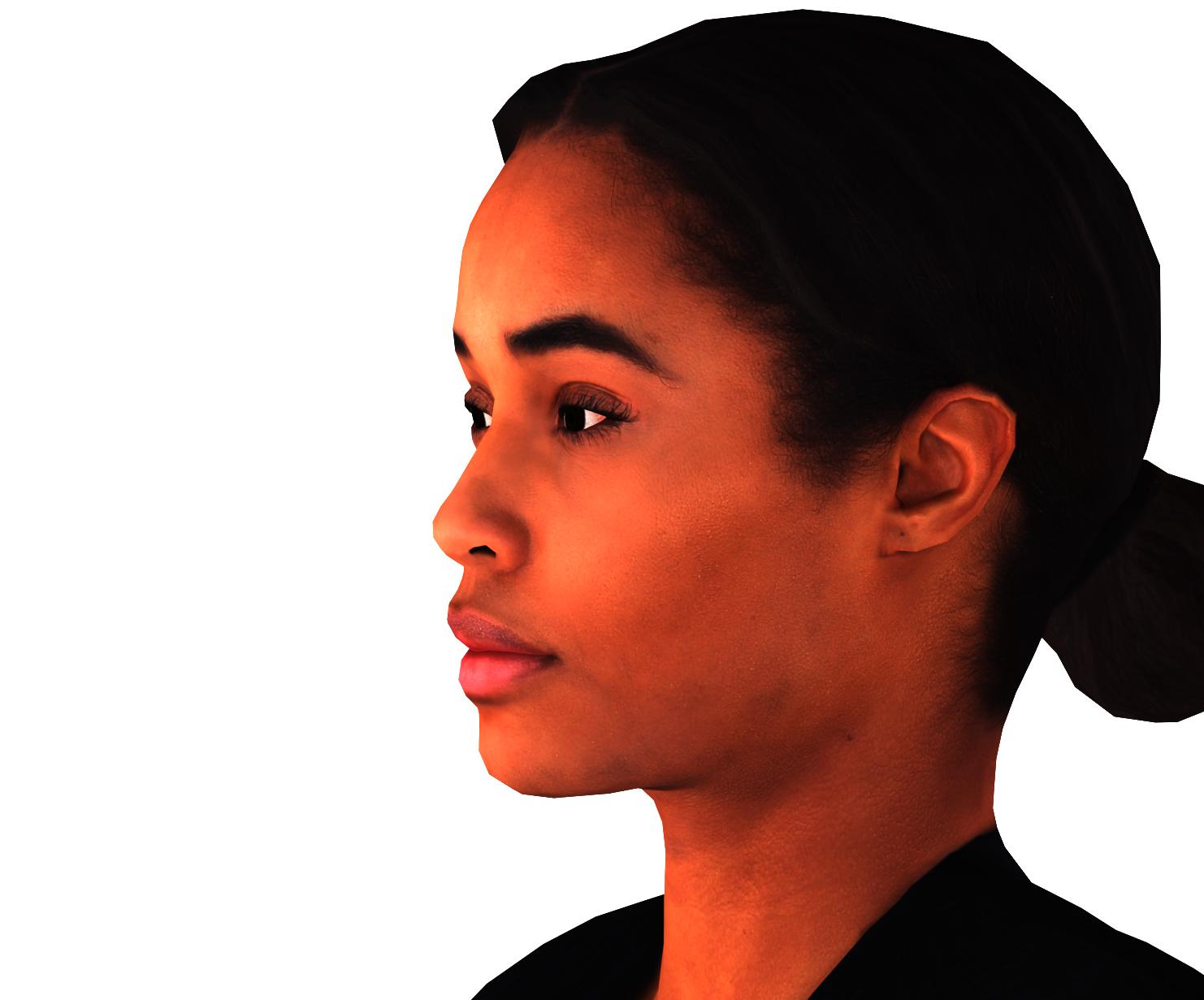}
    \adjincludegraphics[trim={0 0 0 0},clip,width=.135\linewidth]{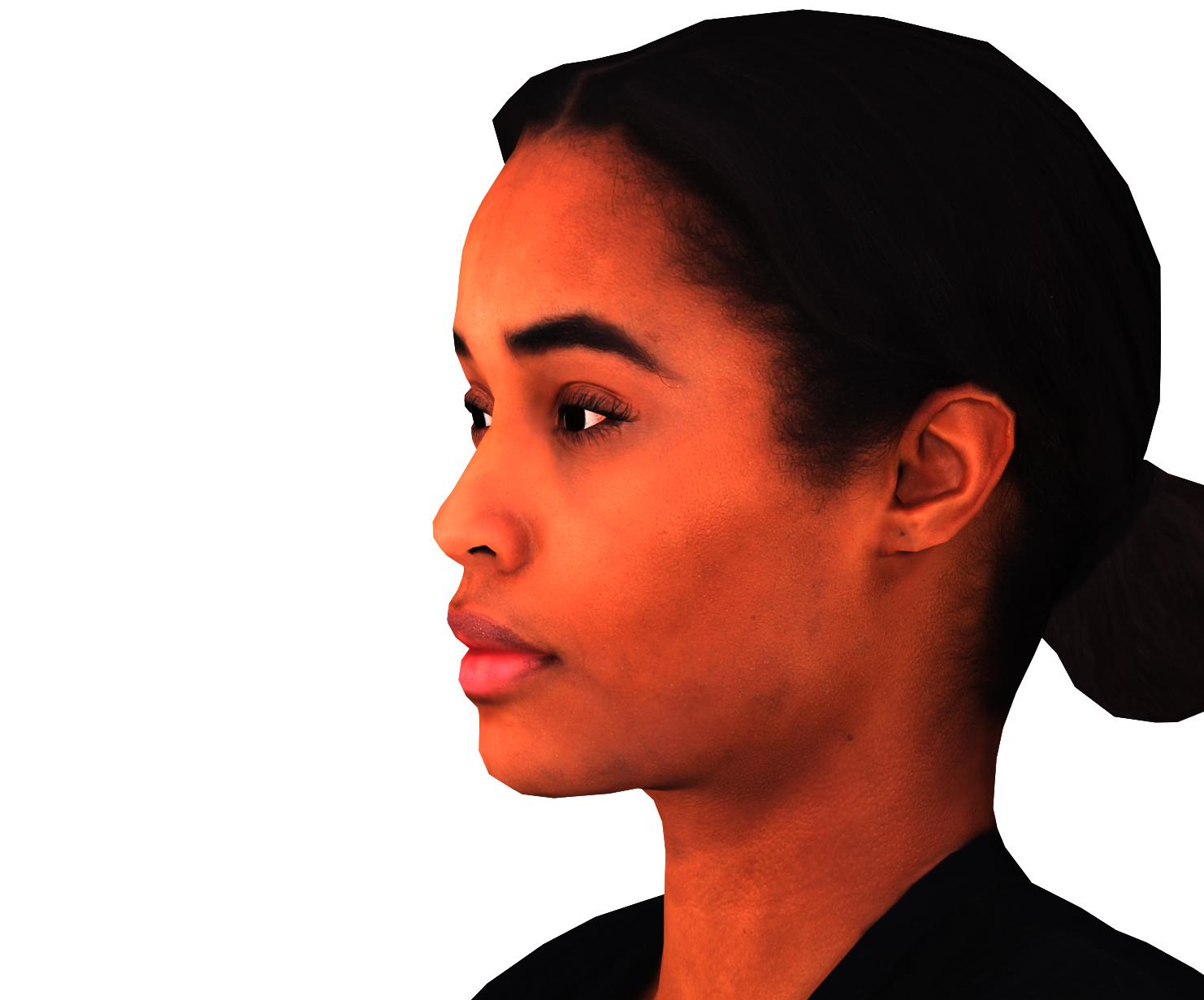}
    
    % ---------------- Claudia
    \vspace{0.5cm}
    
    \textit{Person 2 (Claudia)}
    
    \rotatebox{90}{\hspace{0.4cm} Ours}
    \adjincludegraphics[trim={0 0 0 0},clip,width=.135\linewidth]{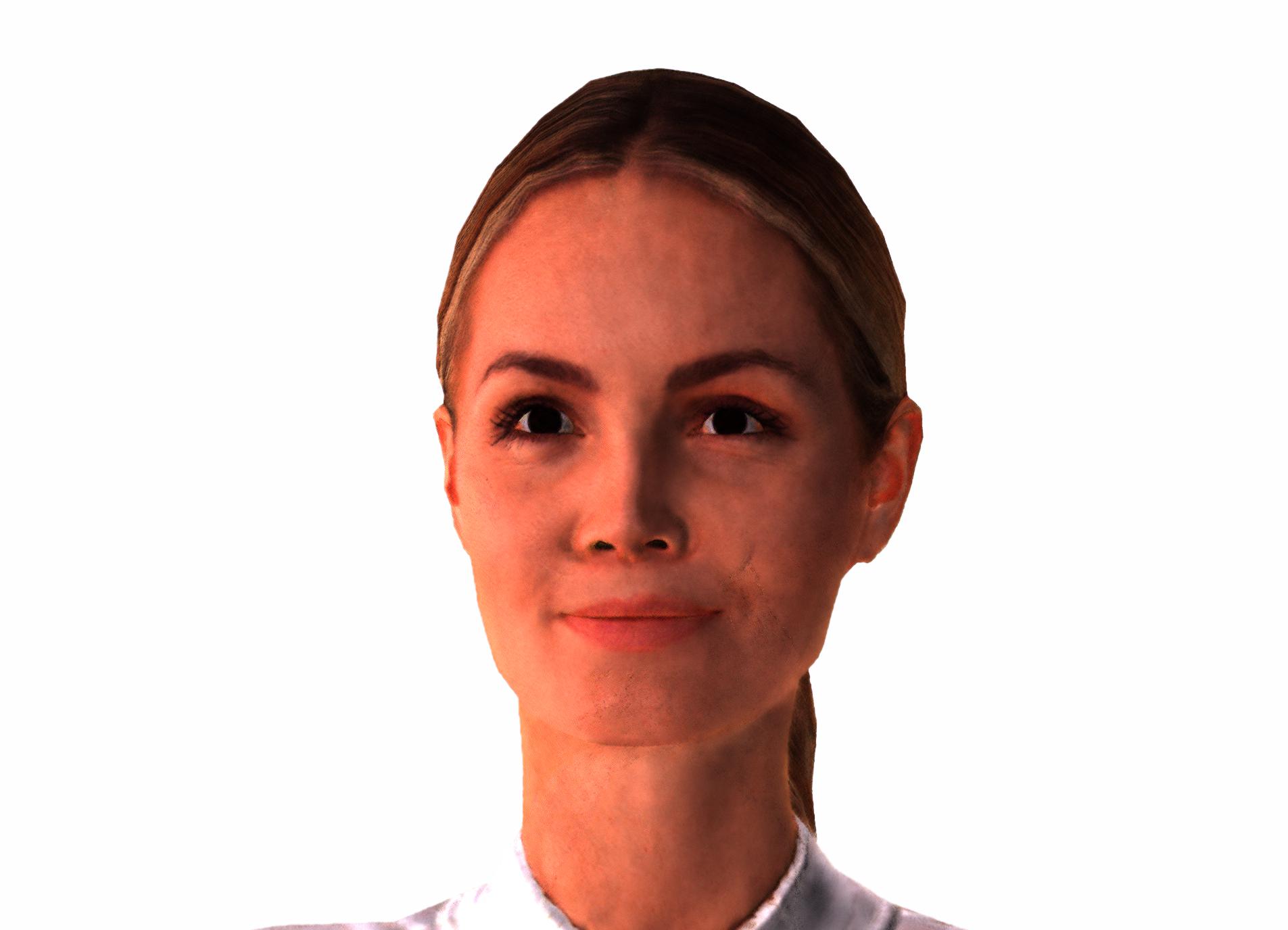}
    \adjincludegraphics[trim={0 0 0 0},clip,width=.135\linewidth]{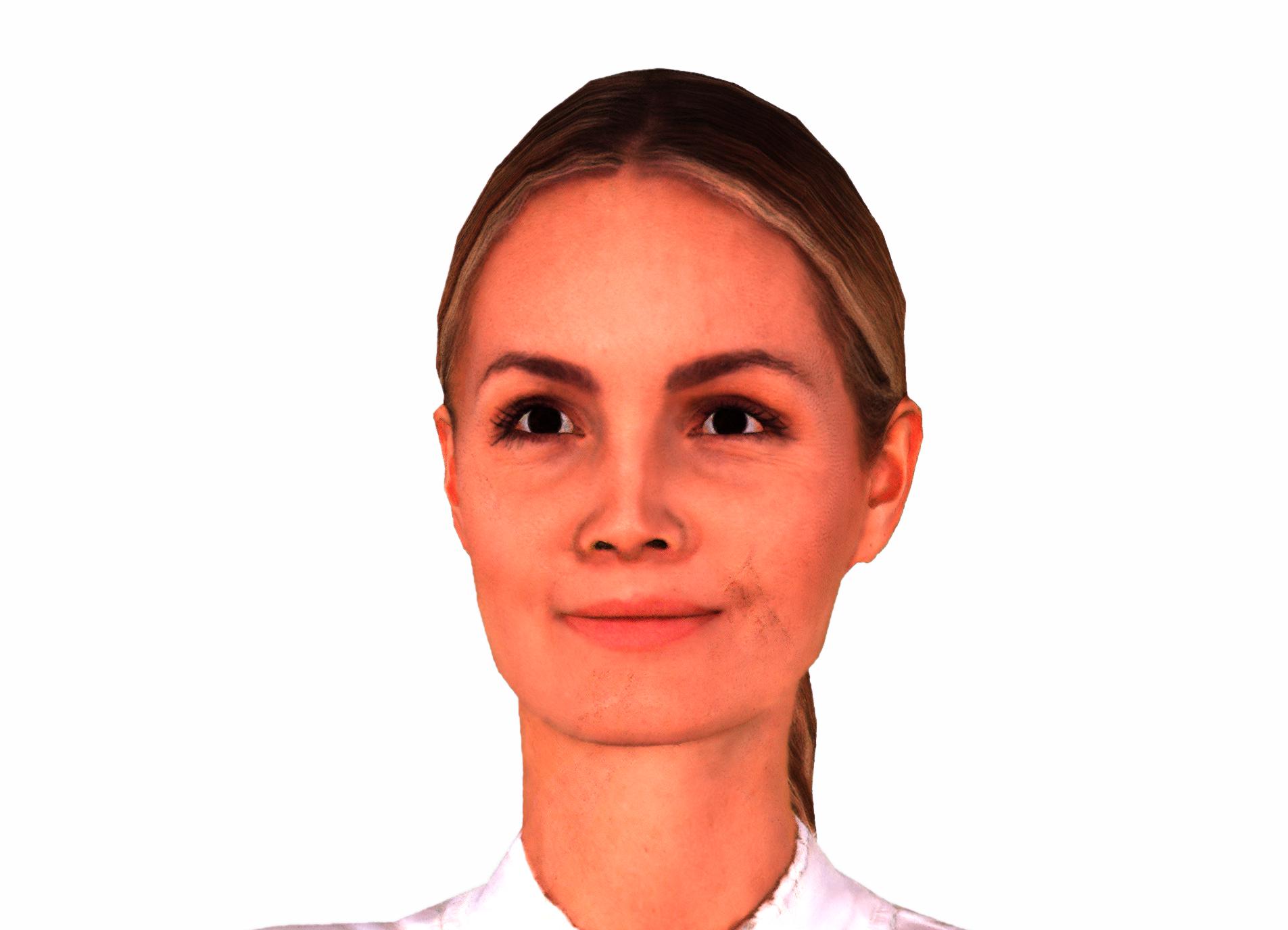}
    \adjincludegraphics[trim={0 0 0 0},clip,width=.135\linewidth]{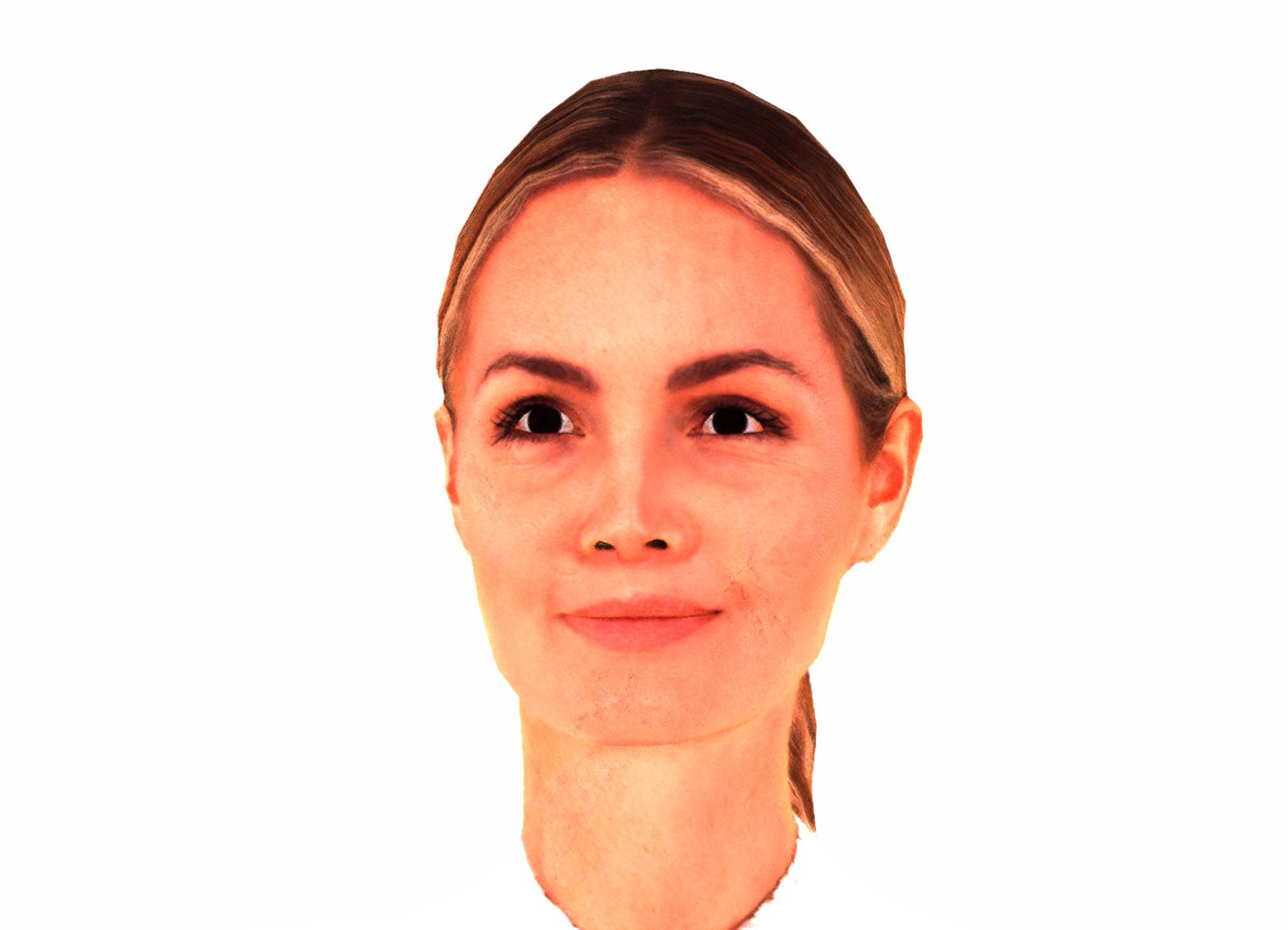}
    \hspace{0.5cm}
    % \hfill
    \adjincludegraphics[trim={0 0 {.25\width} 0},clip,width=.1\linewidth]{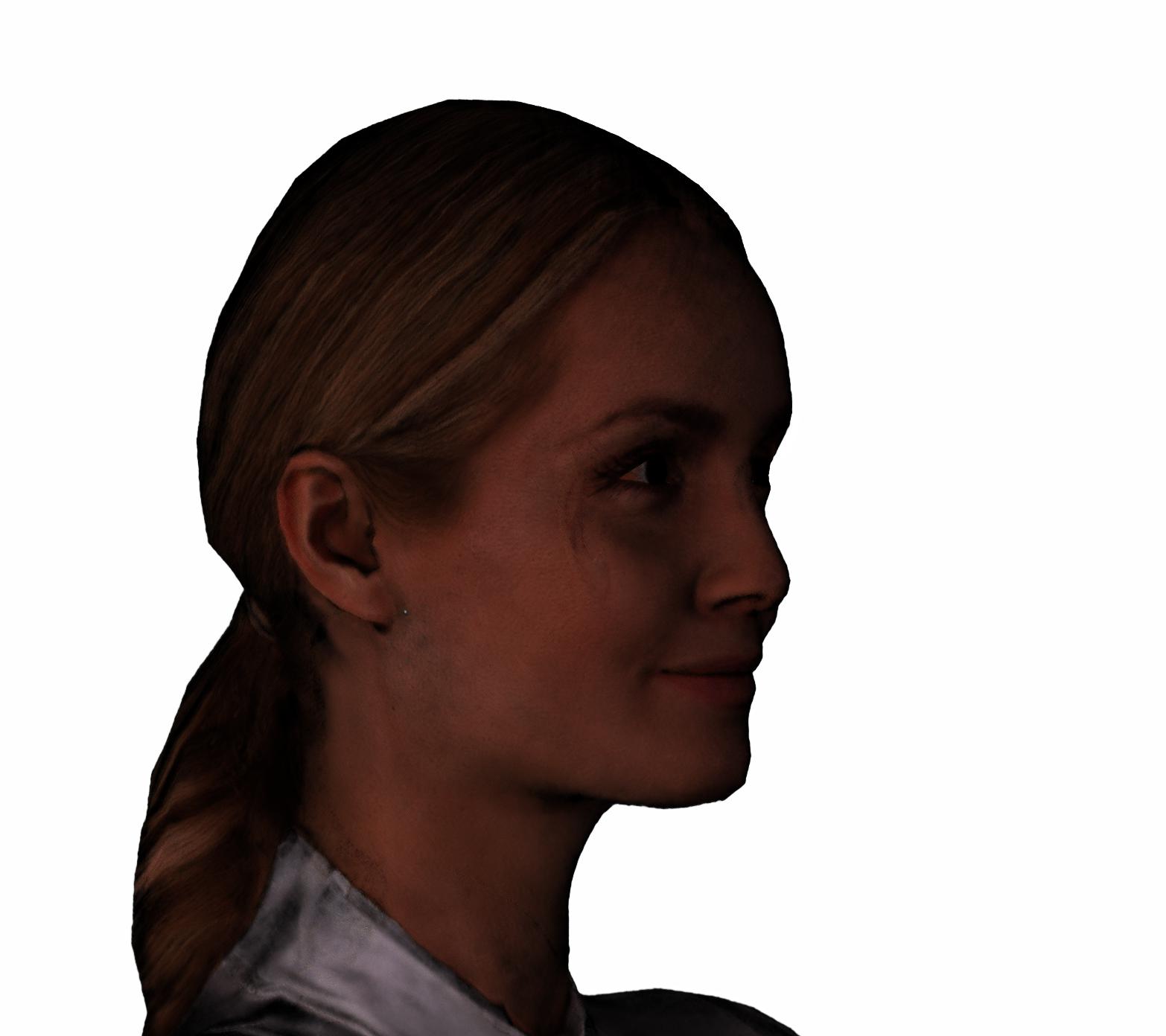}
    \hspace{0.25cm}
    \adjincludegraphics[trim={0 0 {.25\width} 0},clip,width=.1\linewidth]{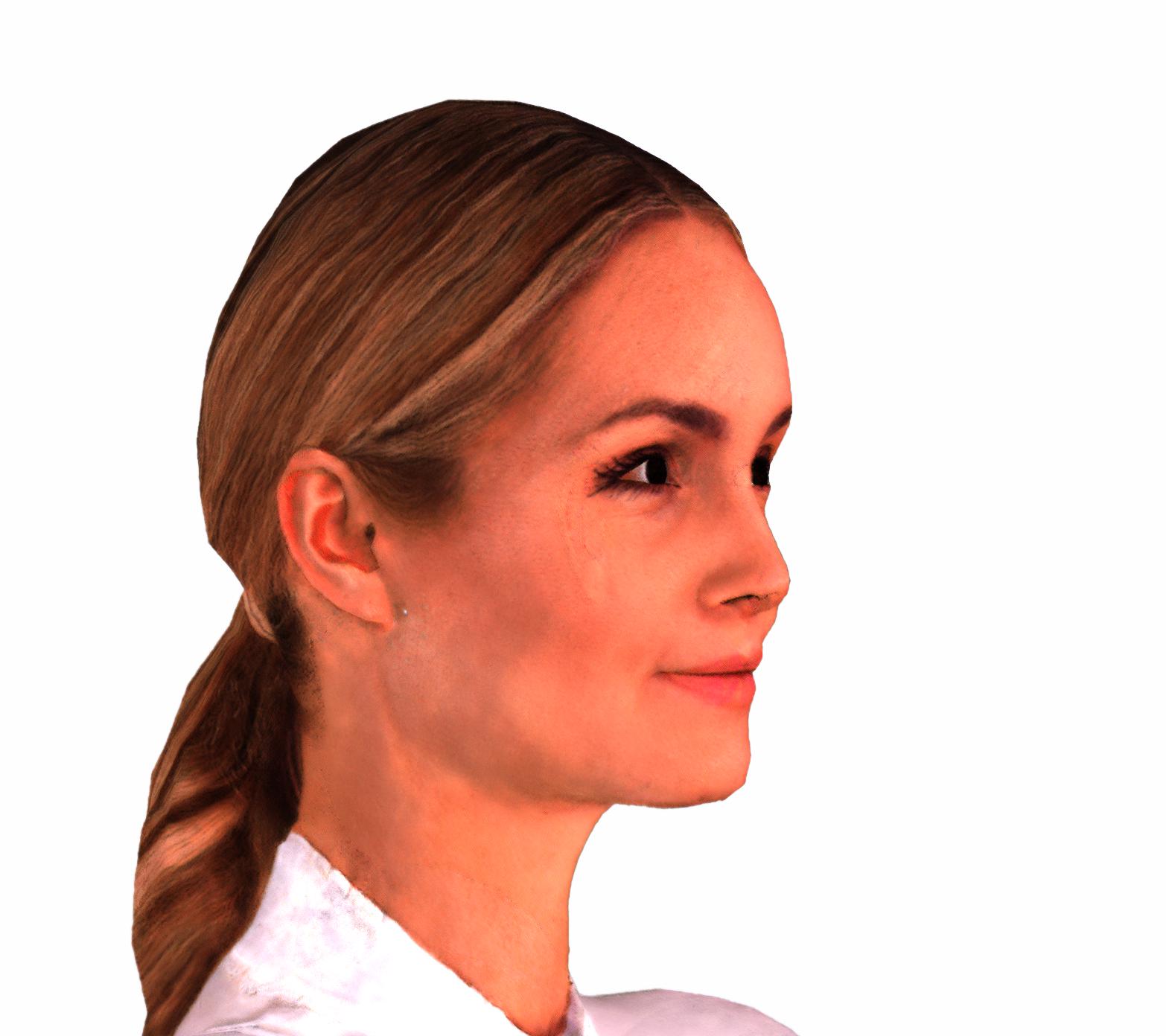}
    % \hfill
    \adjincludegraphics[trim={0 0 0 0},clip,width=.135\linewidth]{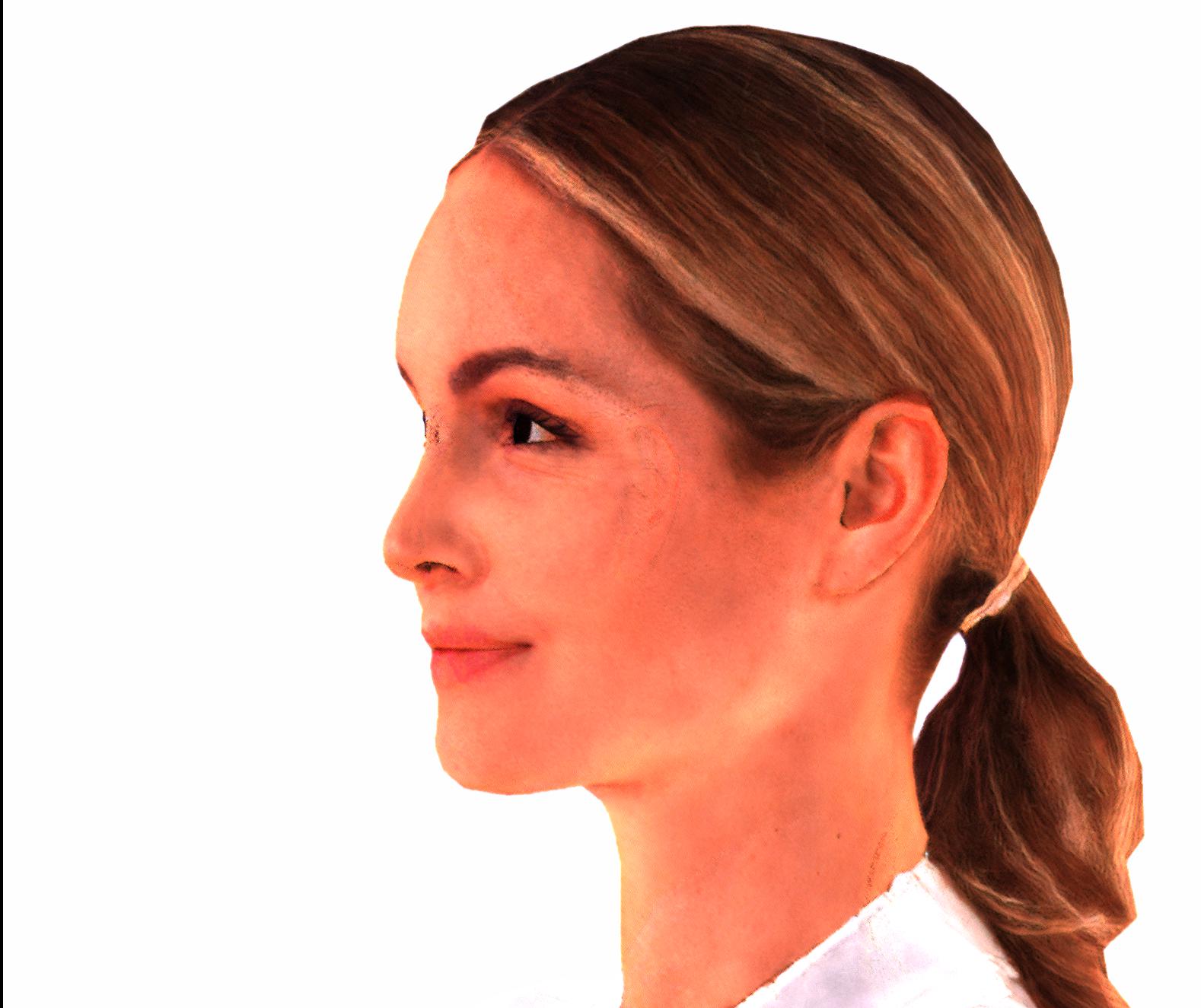}
    \adjincludegraphics[trim={0 0 0 0},clip,width=.135\linewidth]{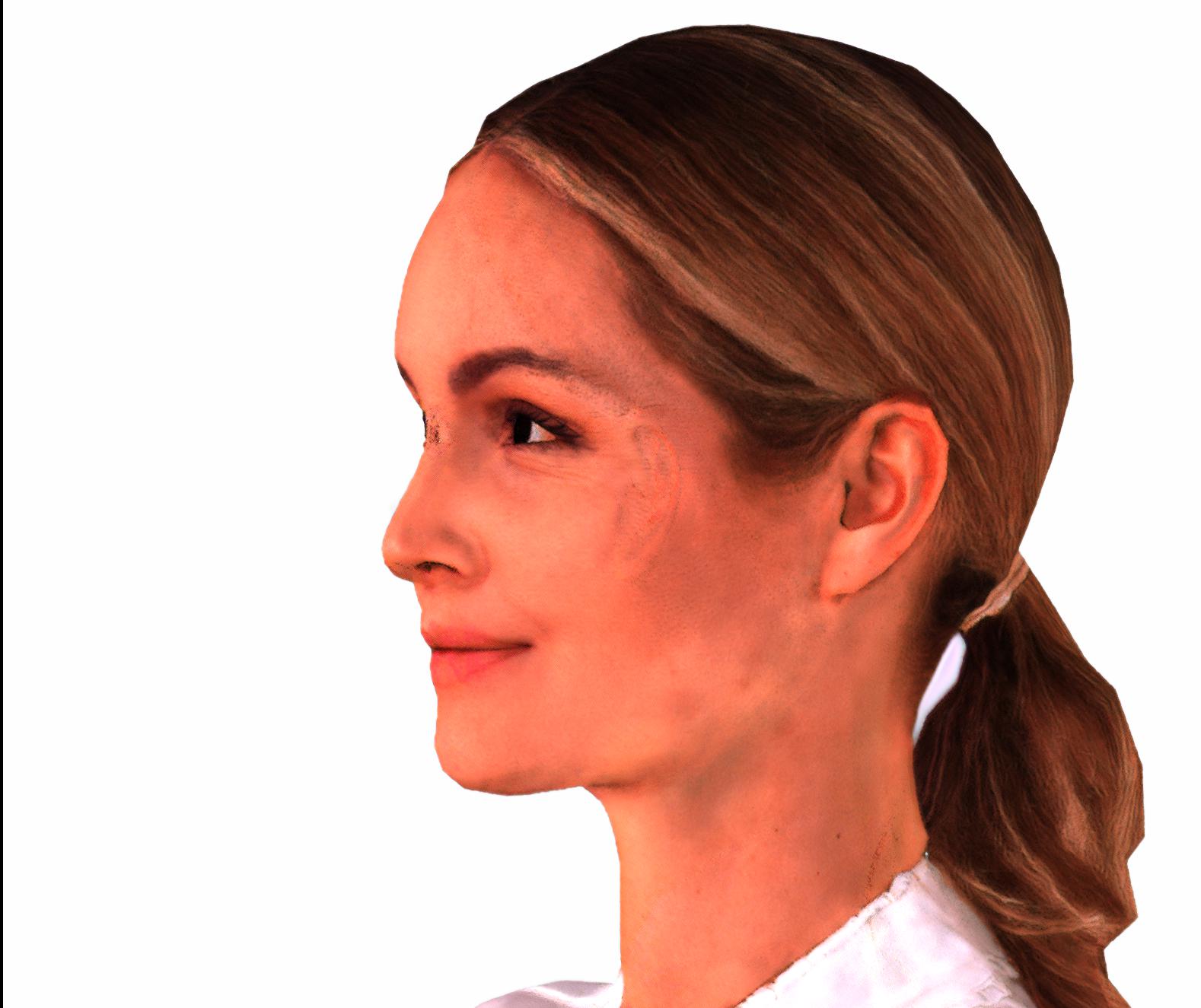}
    
    \rotatebox{90}{\hspace{0.25cm} DPR~\cite{Zhou19}}
    \adjincludegraphics[trim={0 0 0 0},clip,width=.135\linewidth]{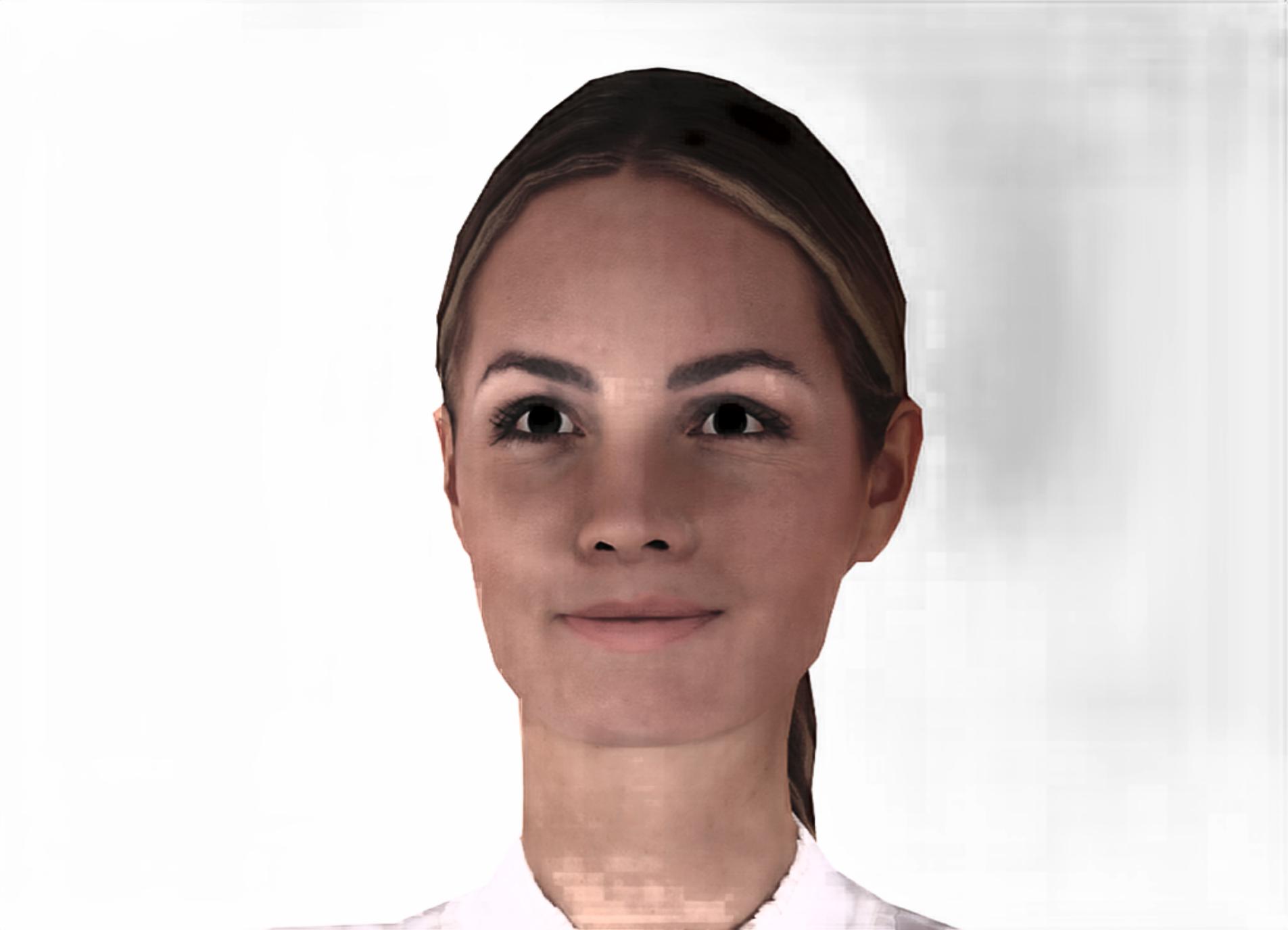}
    \adjincludegraphics[trim={0 0 0 0},clip,width=.135\linewidth]{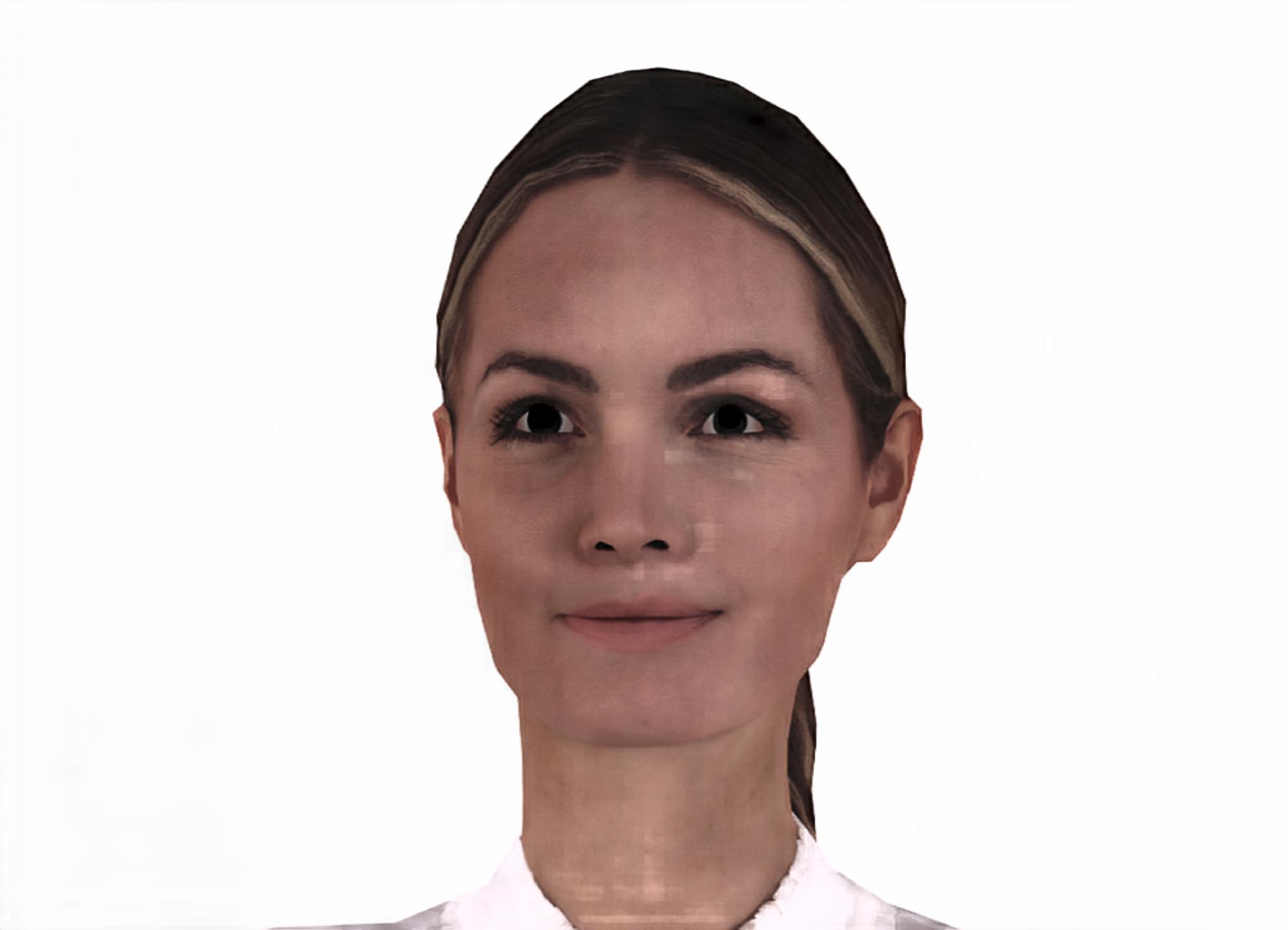}
    \adjincludegraphics[trim={0 0 0 0},clip,width=.135\linewidth]{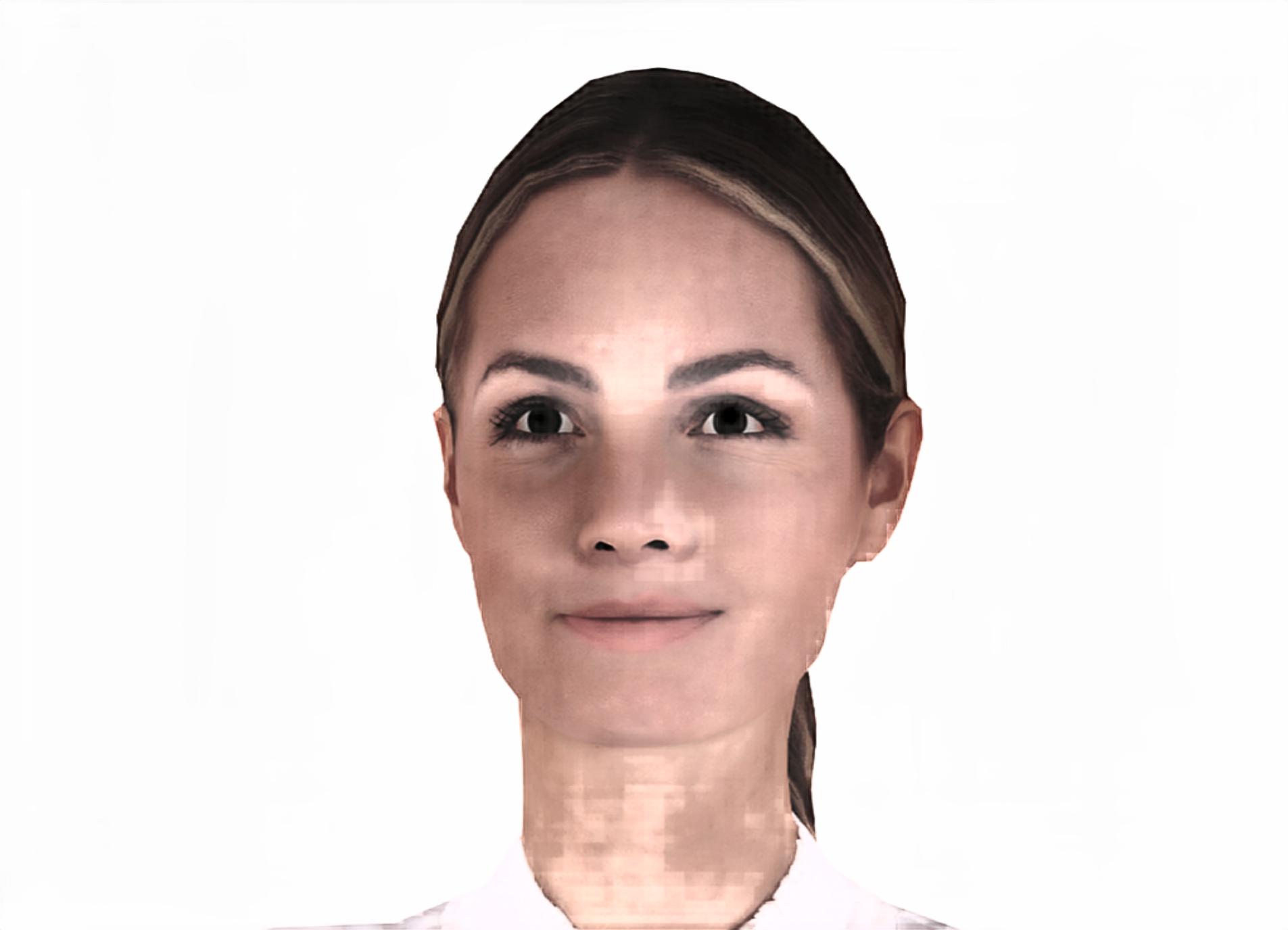}
    \hspace{0.5cm}
    % \hfill
    \adjincludegraphics[trim={0 0 {.25\width} 0},clip,width=.1\linewidth]{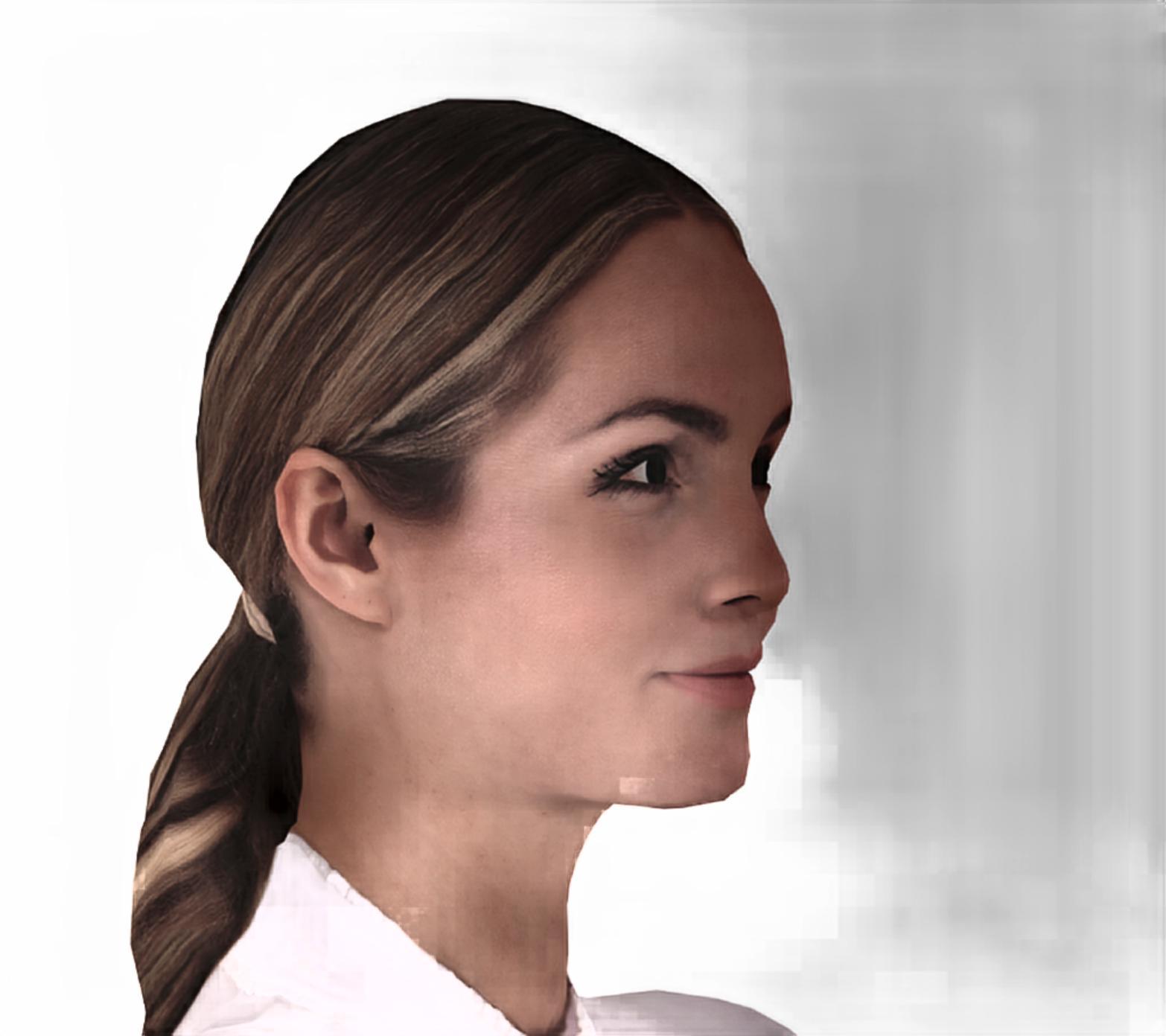}
    \hspace{0.25cm}
    \adjincludegraphics[trim={0 0 {.25\width} 0},clip,width=.1\linewidth]{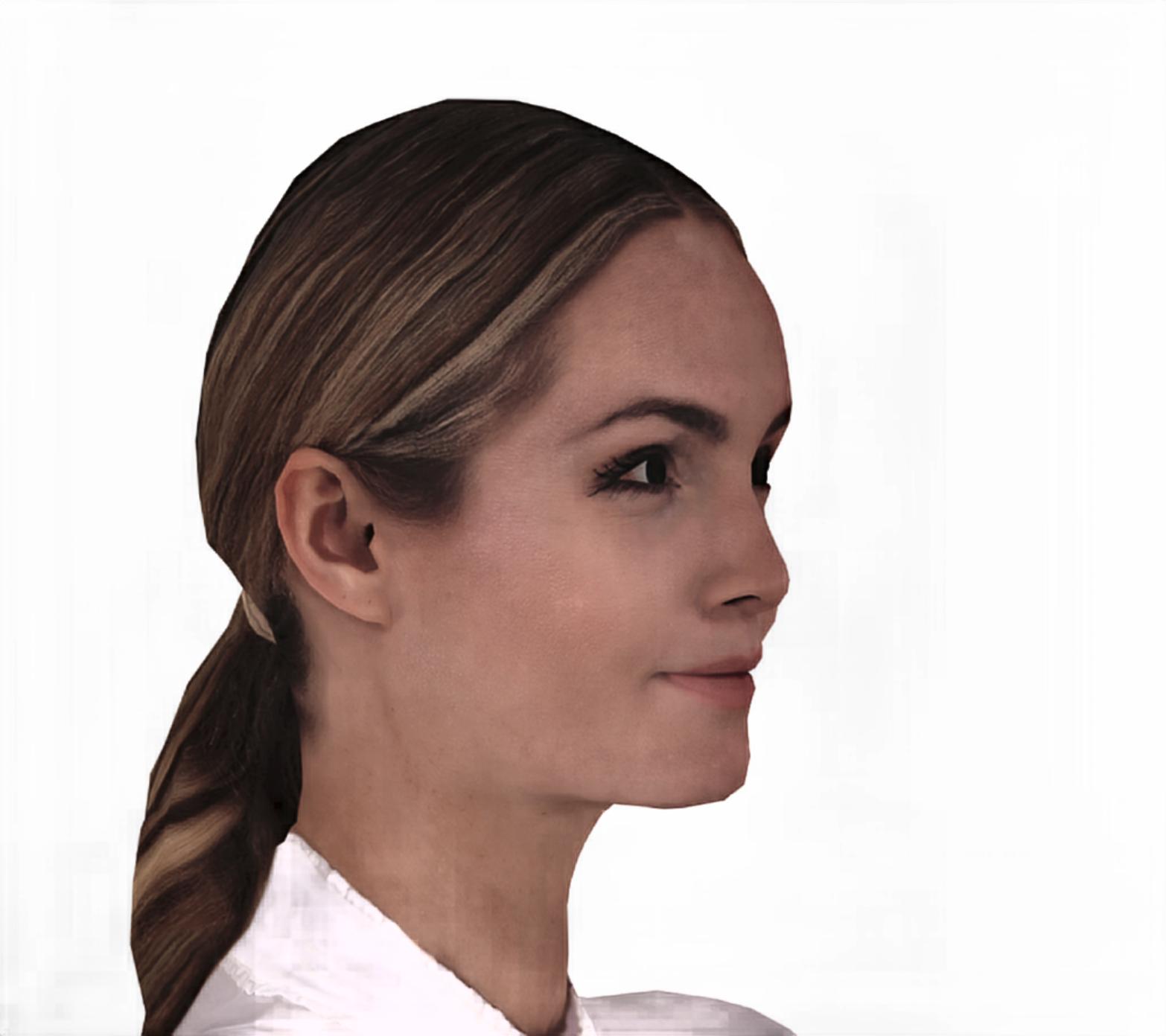}
    % \hfill
    \adjincludegraphics[trim={0 0 0 0},clip,width=.135\linewidth]{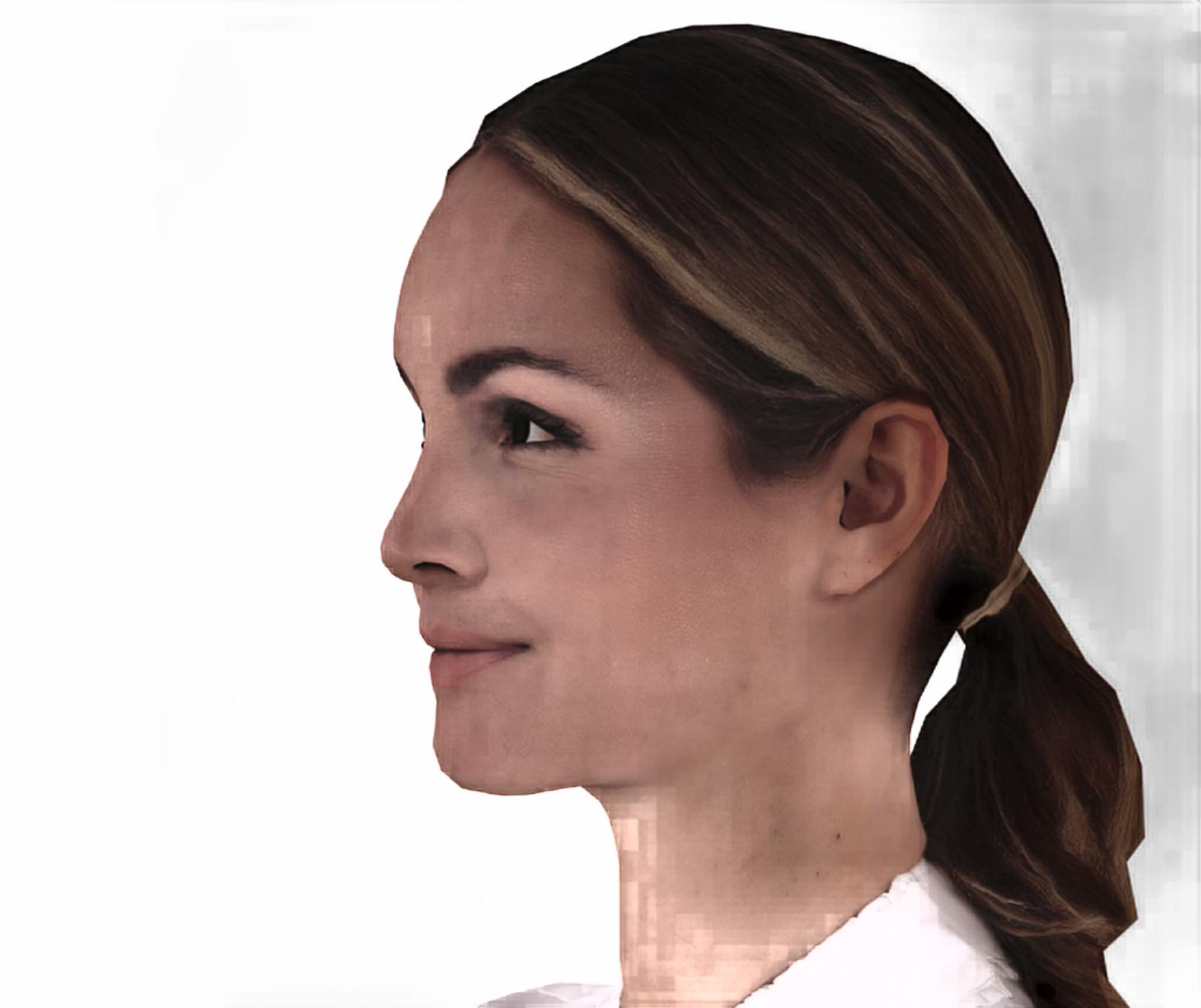}
    \adjincludegraphics[trim={0 0 0 0},clip,width=.135\linewidth]{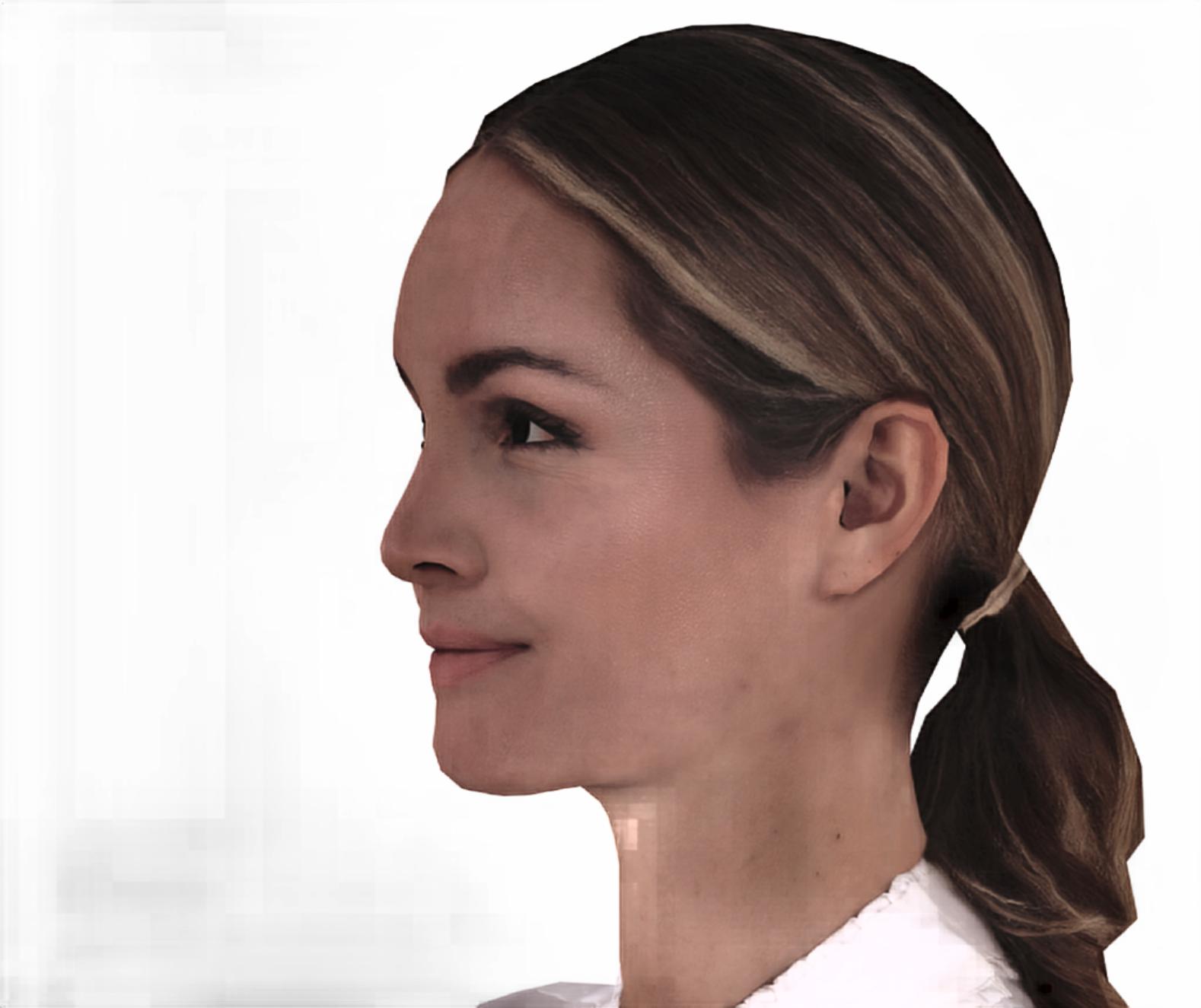}
    
    \rotatebox{90}{\hspace{-0.15cm} Ground truth}
    \adjincludegraphics[trim={0 0 0 0},clip,width=.135\linewidth]{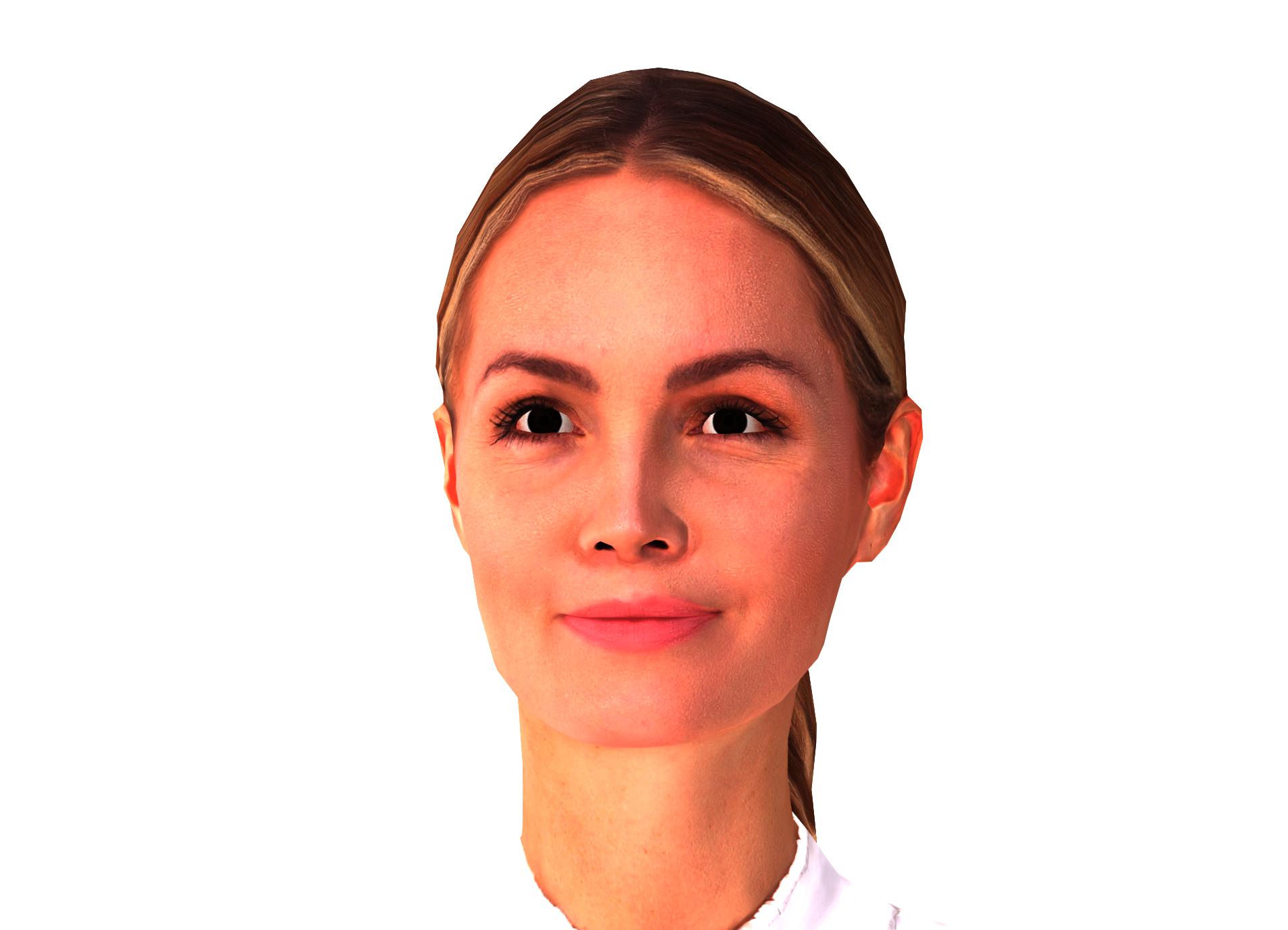}
    \adjincludegraphics[trim={0 0 0 0},clip,width=.135\linewidth]{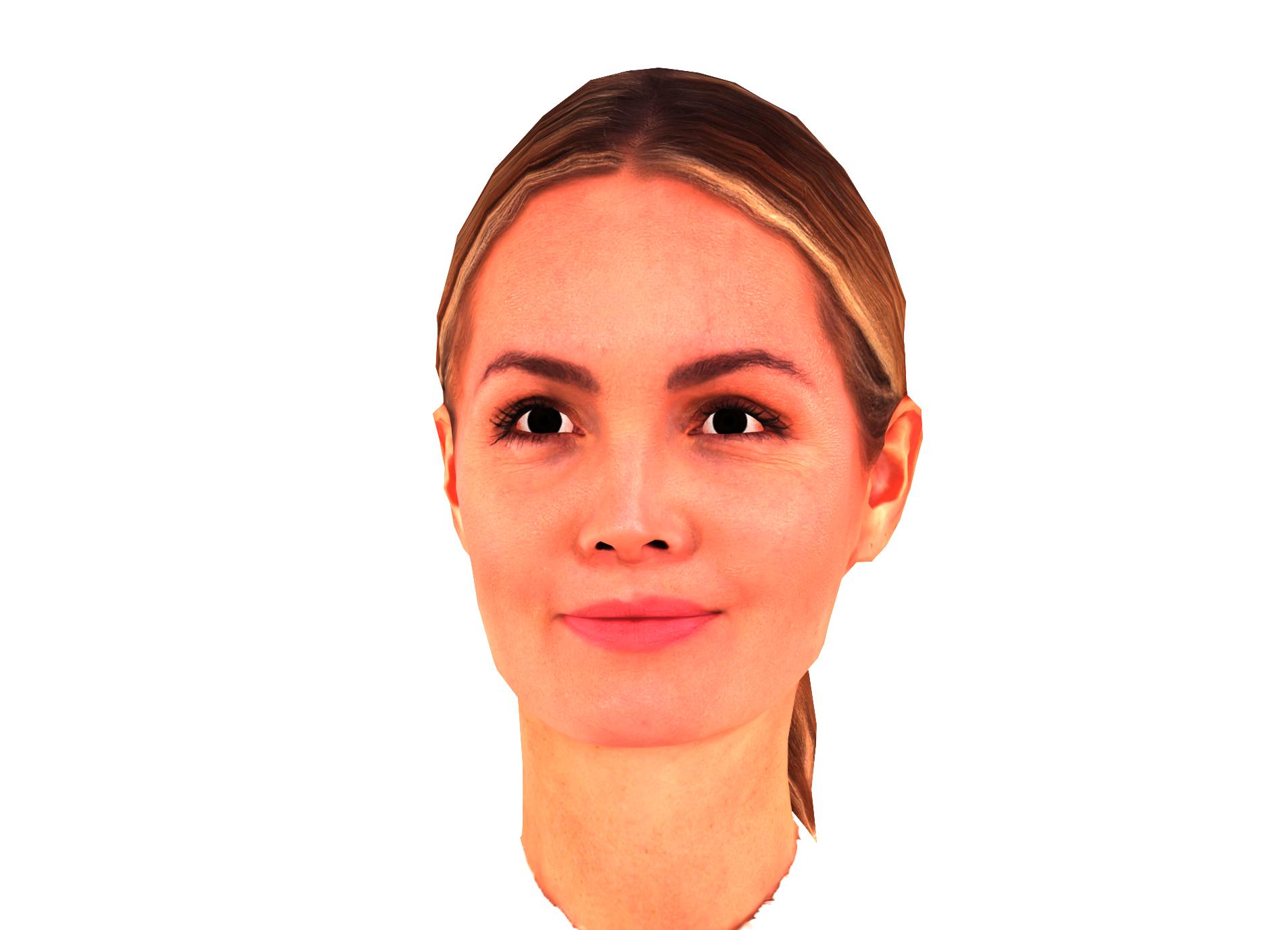}
    \adjincludegraphics[trim={0 0 0 0},clip,width=.135\linewidth]{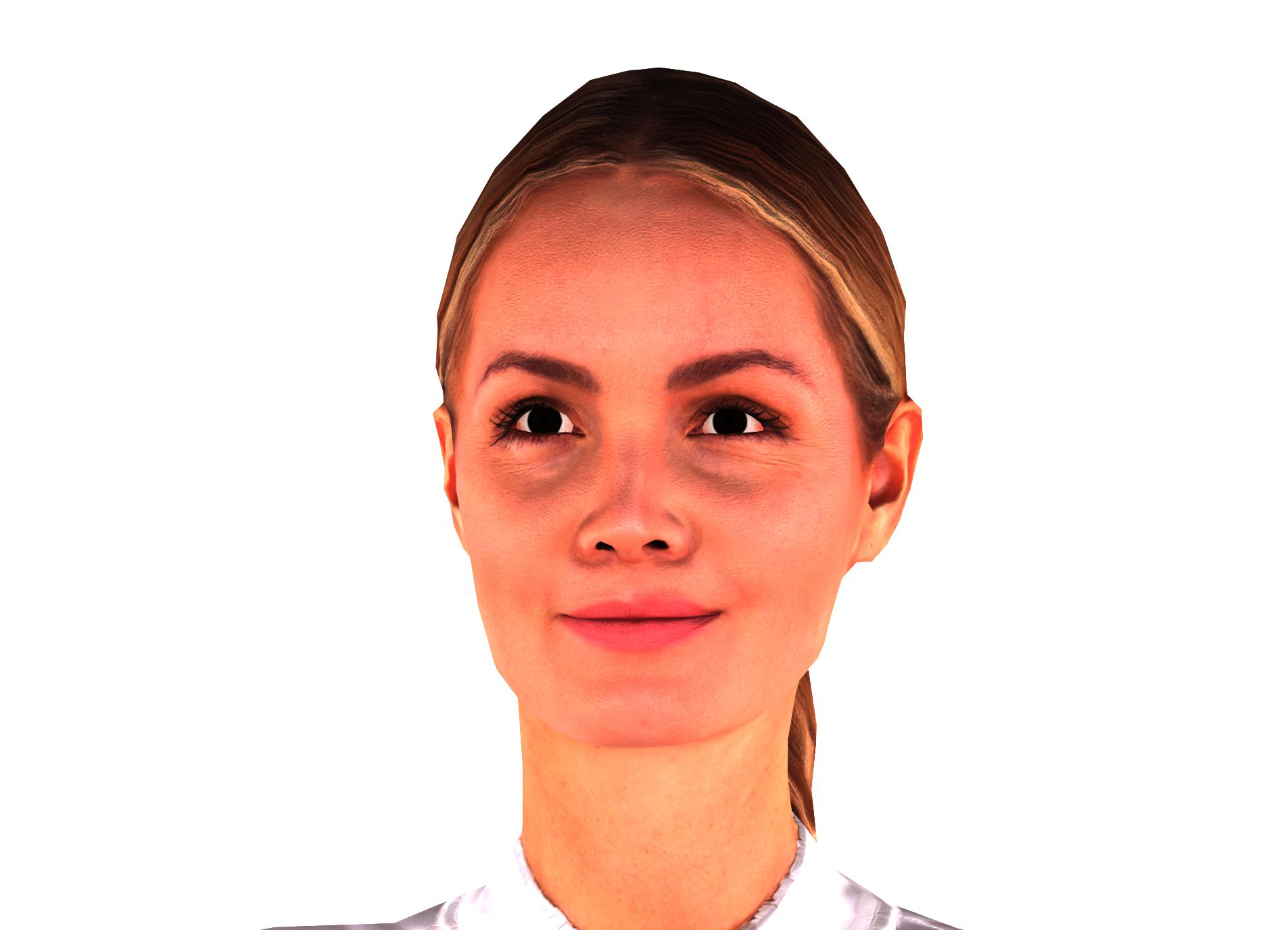}
    \hspace{0.5cm}
    % \hfill
    \adjincludegraphics[trim={0 0 {.25\width} 0},clip,width=.1\linewidth]{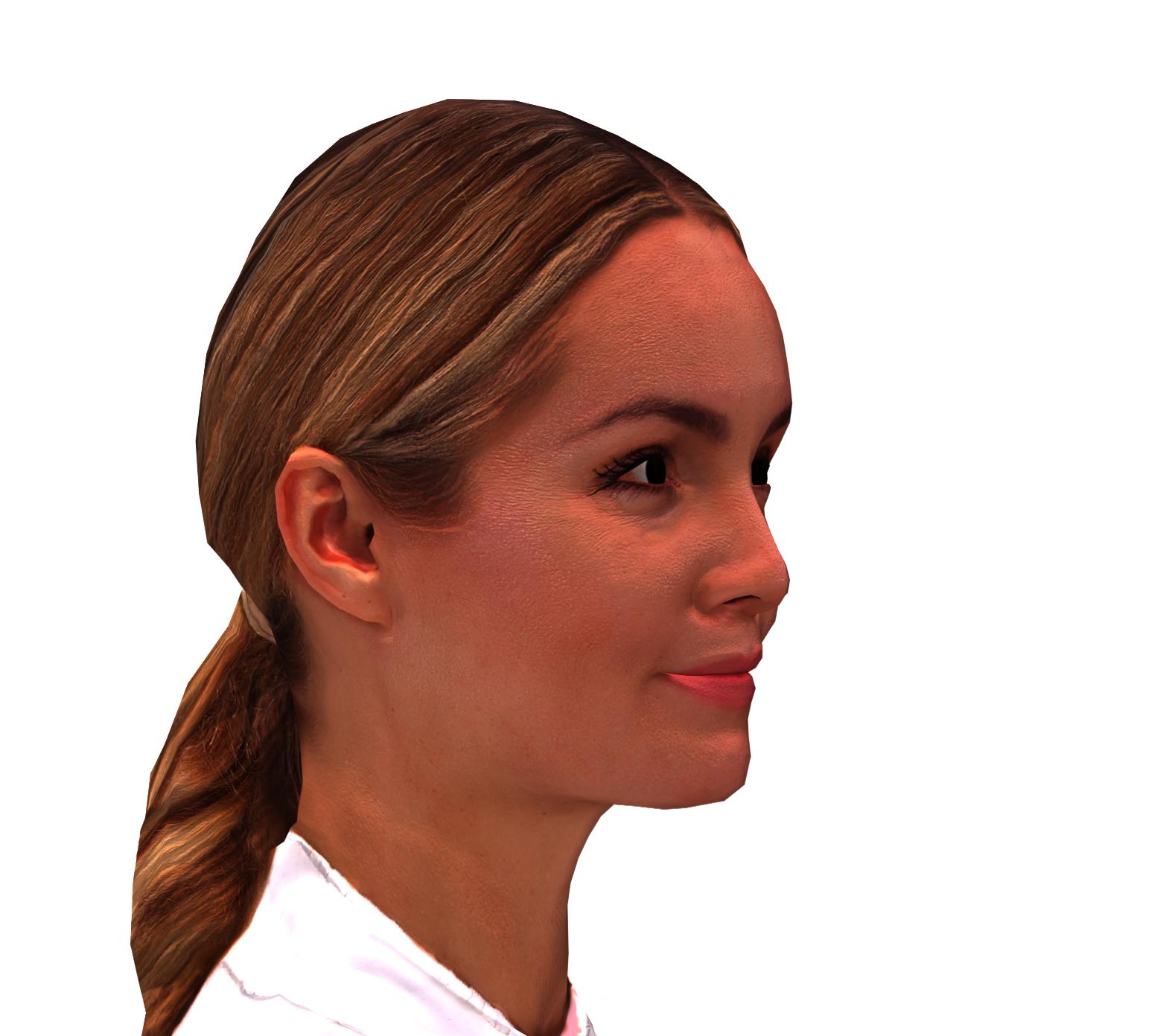}
    \hspace{0.25cm}
    \adjincludegraphics[trim={0 0 {.25\width} 0},clip,width=.1\linewidth]{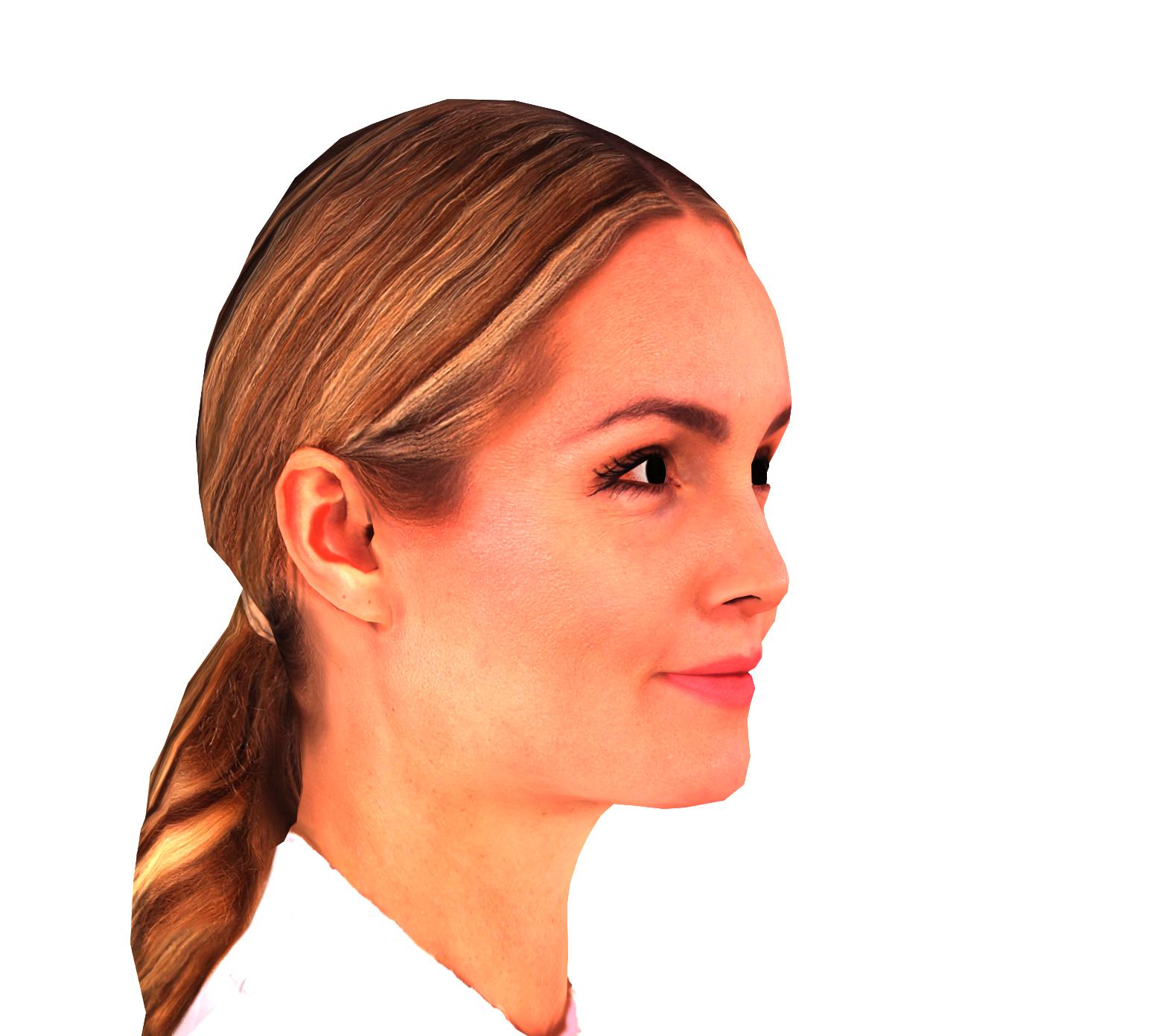}
    % \hfill
    \adjincludegraphics[trim={0 0 0 0},clip,width=.135\linewidth]{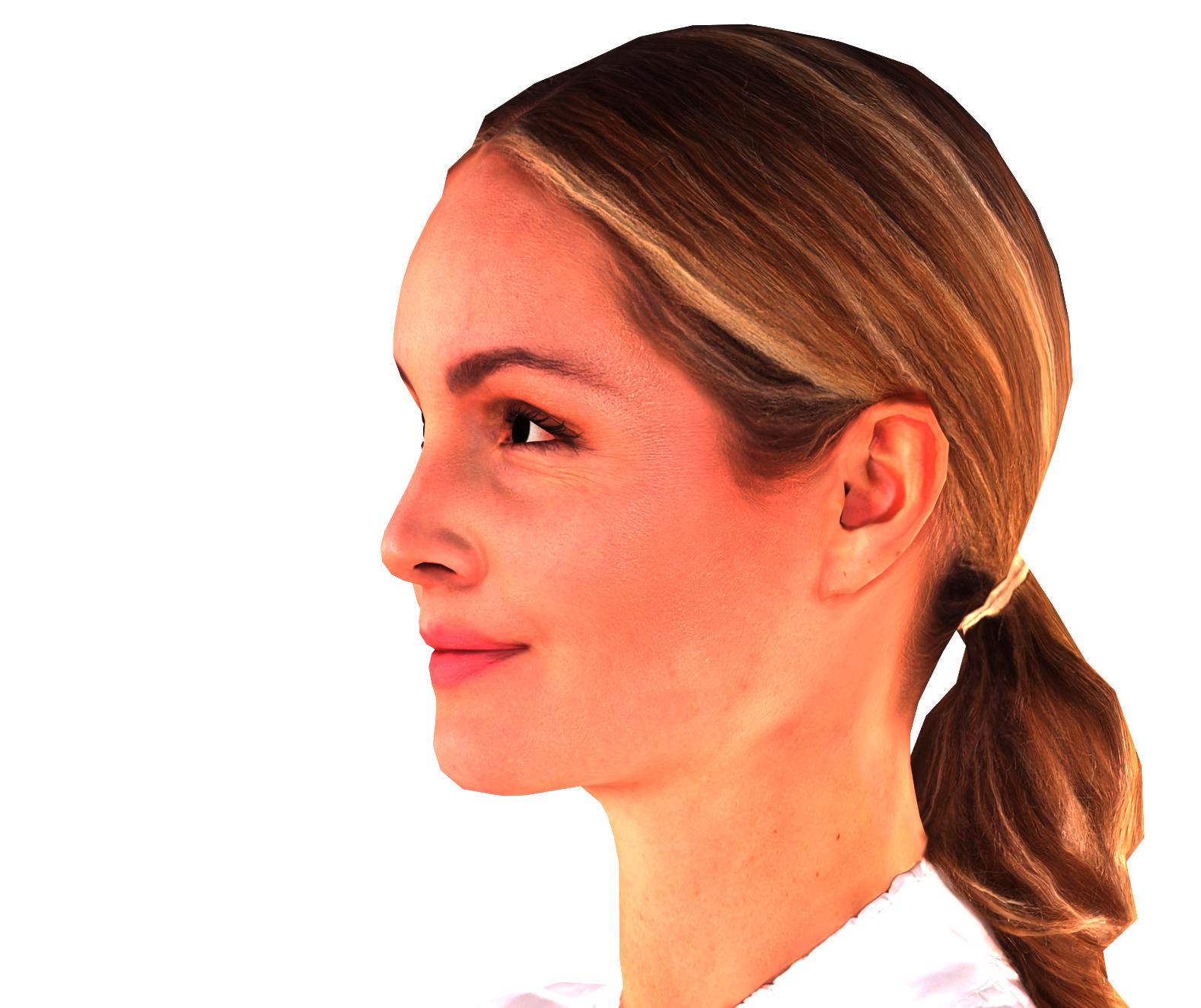}
    \adjincludegraphics[trim={0 0 0 0},clip,width=.135\linewidth]{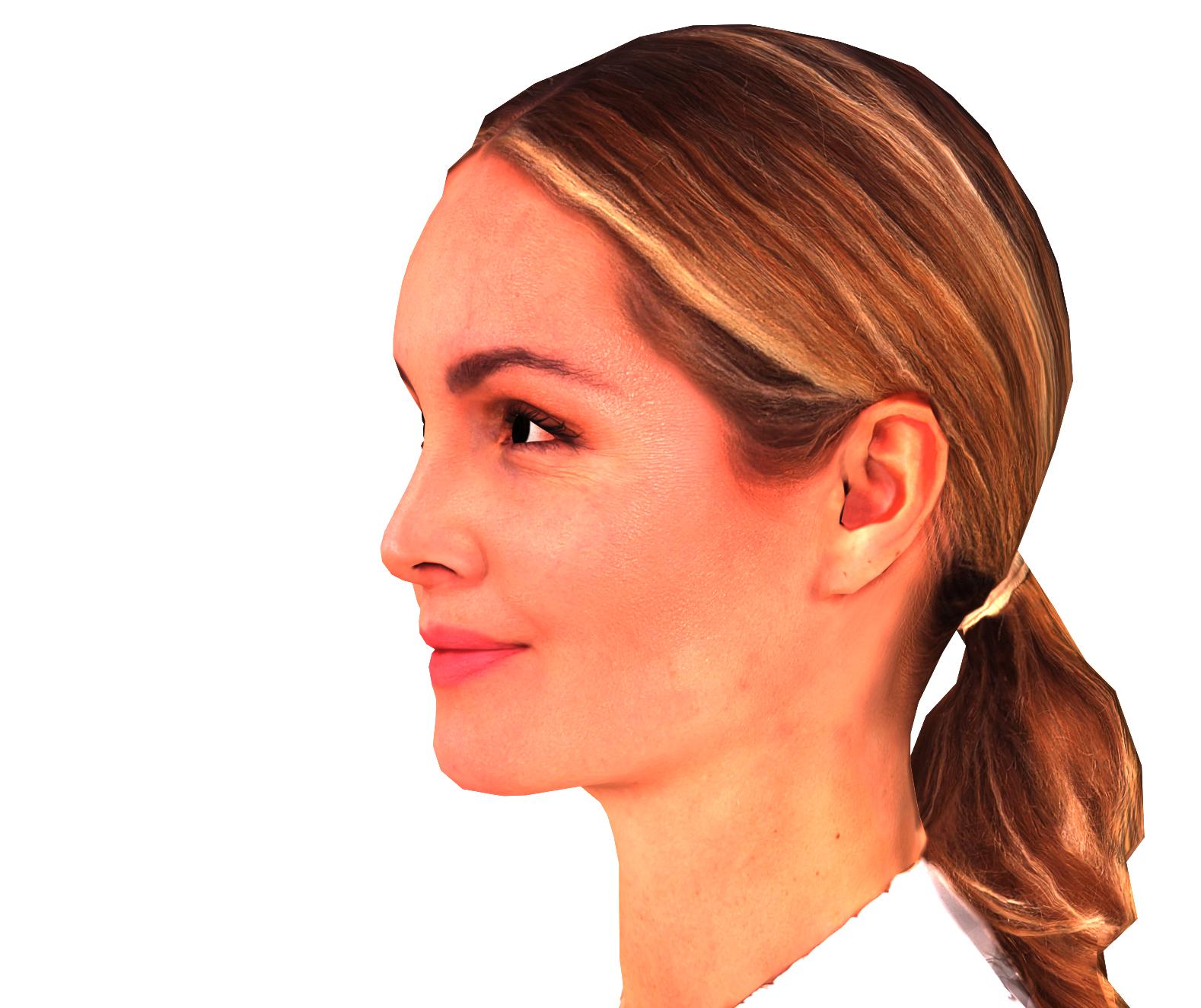}
    
    % ---------------- Eric
    \vspace{0.5cm}
    
    \textit{Person 3 (Eric)}
    
    \rotatebox{90}{\hspace{0.4cm} Ours}
    \adjincludegraphics[trim={0 0 0 0},clip,width=.135\linewidth]{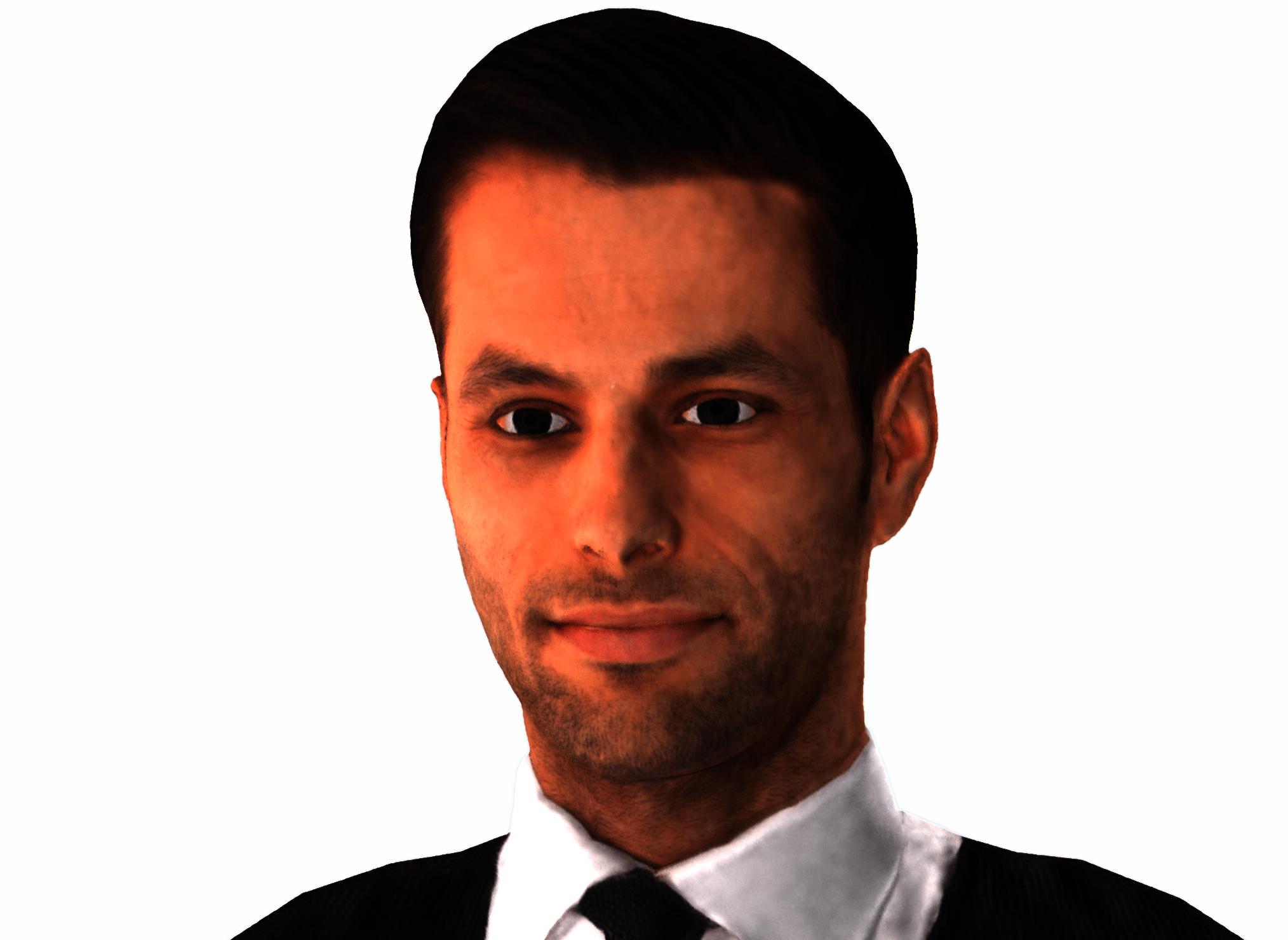}
    \adjincludegraphics[trim={0 0 0 0},clip,width=.135\linewidth]{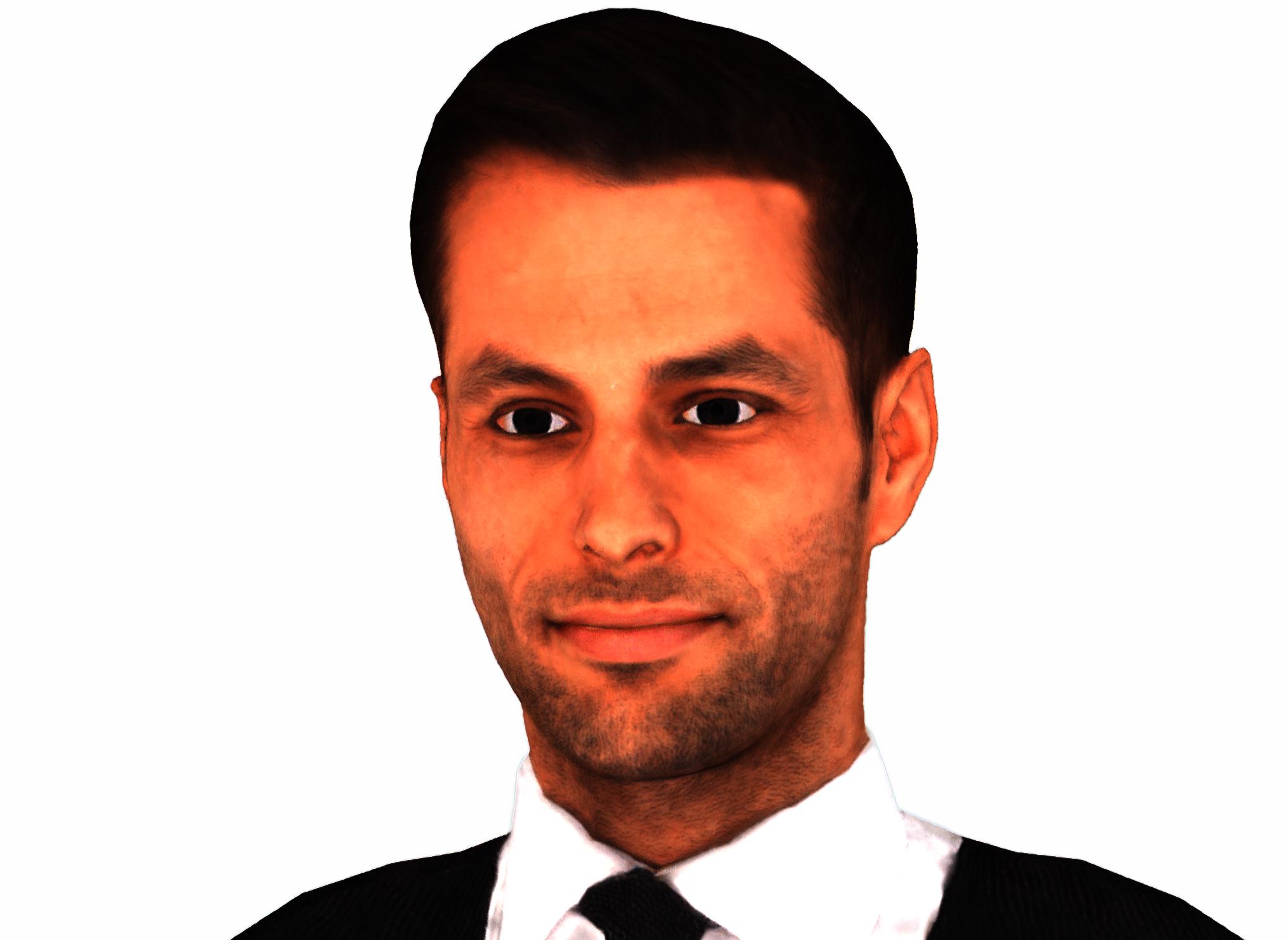}
    \adjincludegraphics[trim={0 0 0 0},clip,width=.135\linewidth]{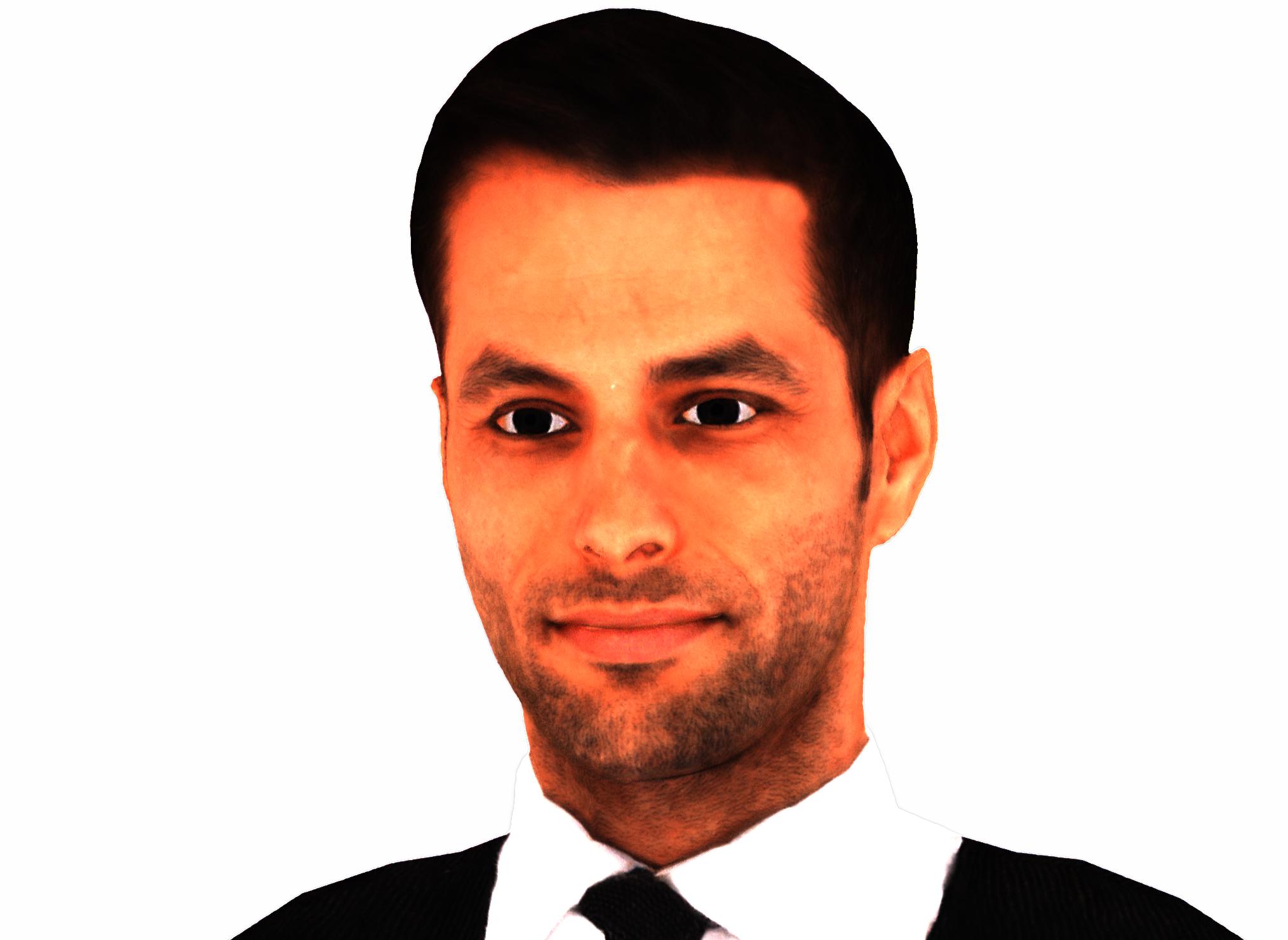}
    \hspace{0.5cm}
    % \hfill
    \adjincludegraphics[trim={0 0 {.25\width} 0},clip,width=.1\linewidth]{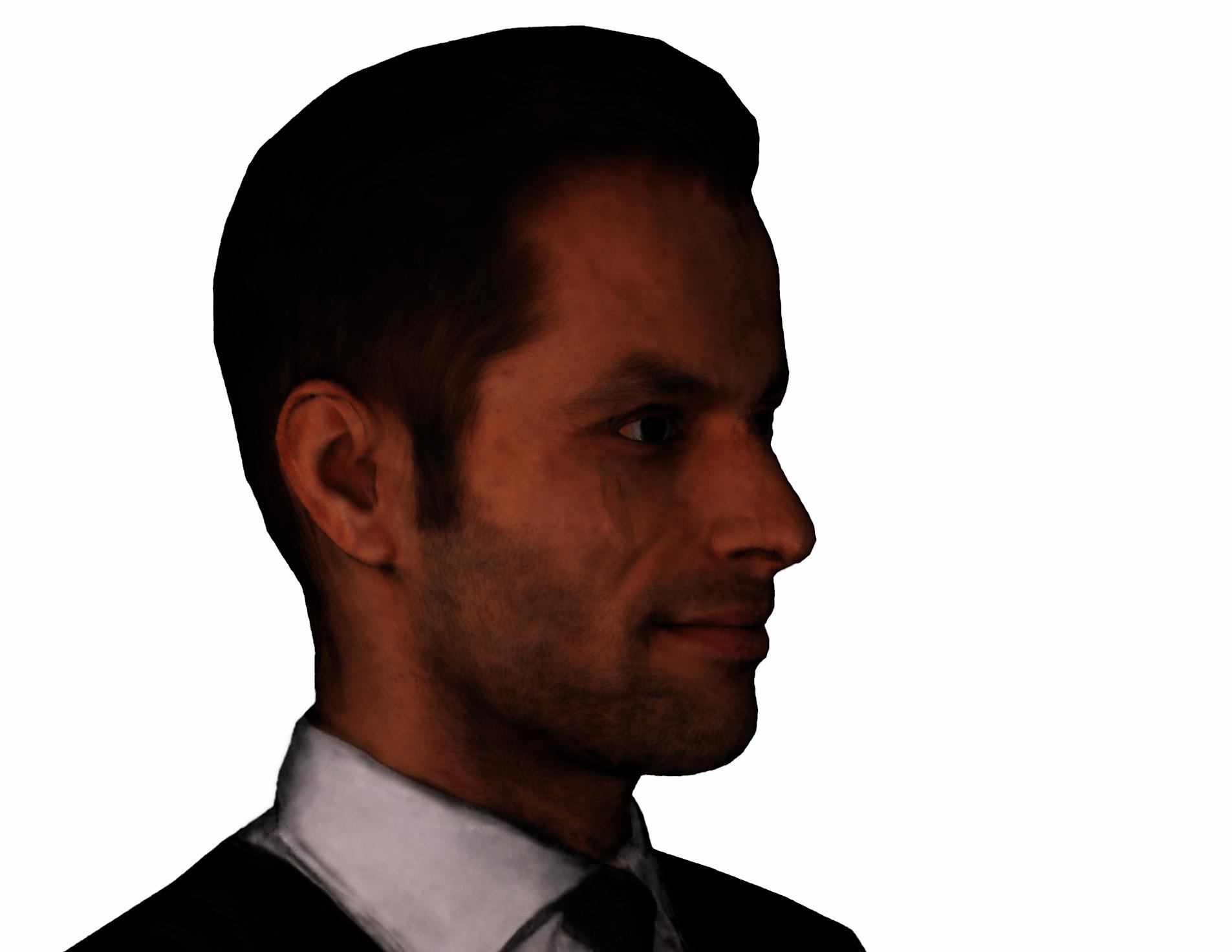}
    \hspace{0.25cm}
    \adjincludegraphics[trim={0 0 {.25\width} 0},clip,width=.1\linewidth]{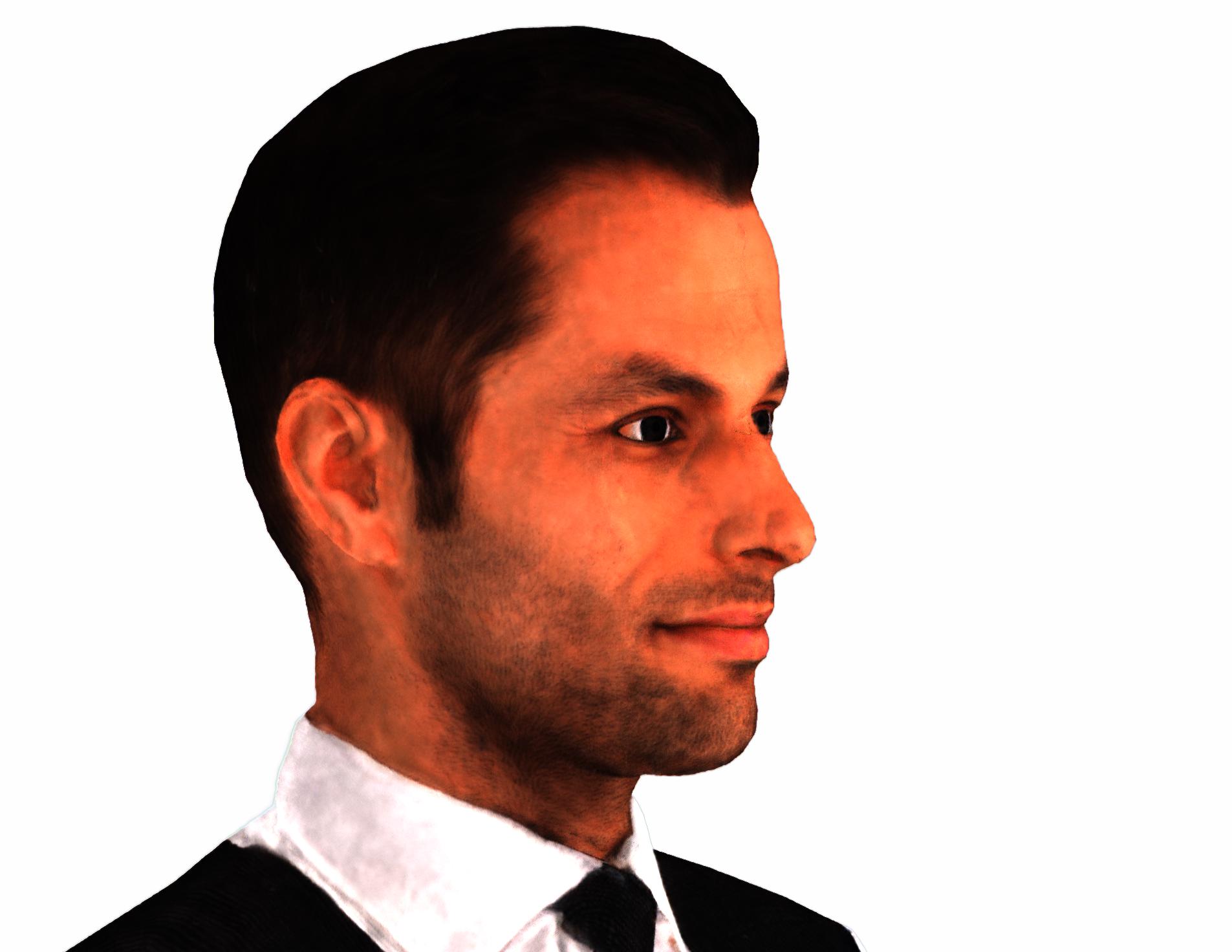}
    % \hfill
    \adjincludegraphics[trim={0 0 0 0},clip,width=.135\linewidth]{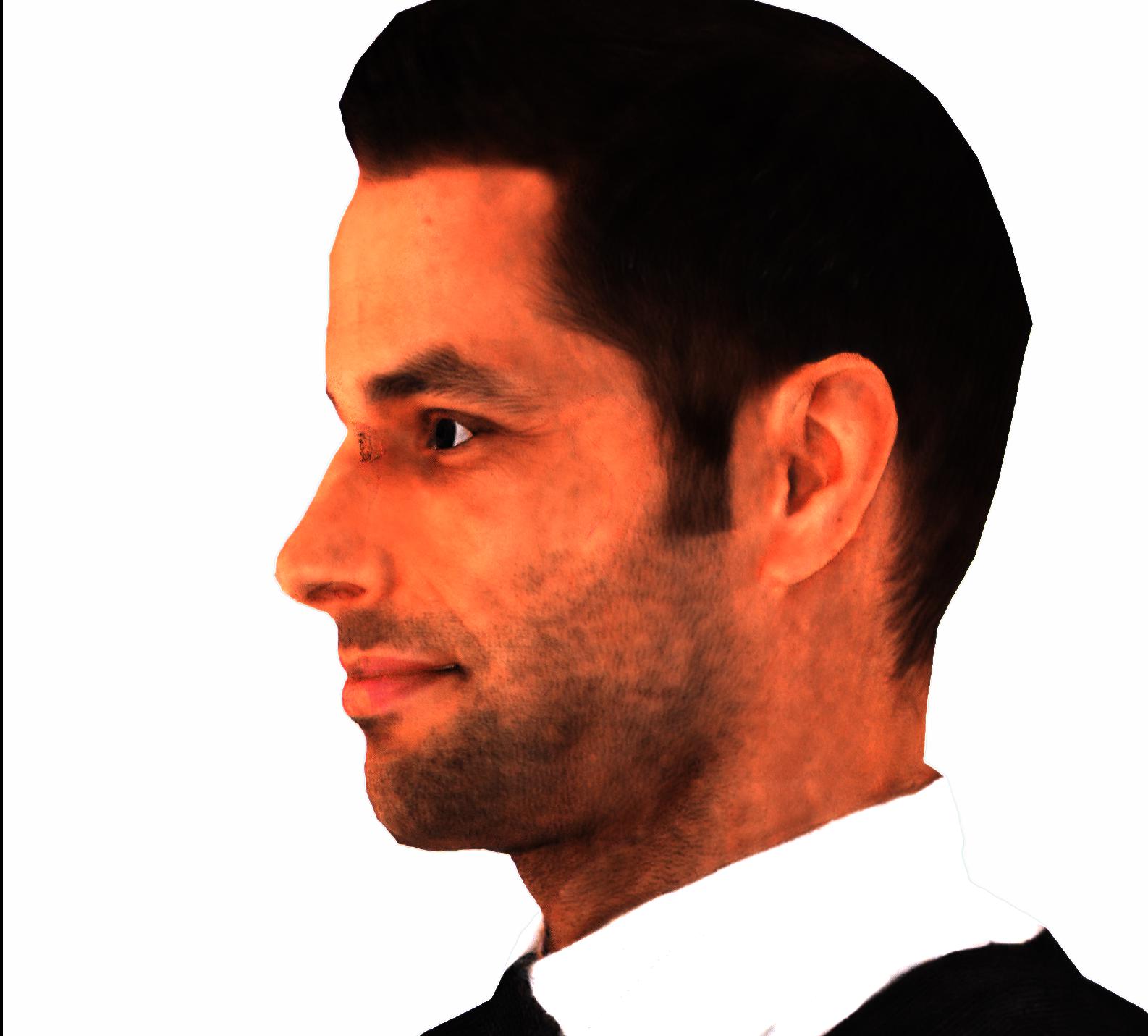}
    \adjincludegraphics[trim={0 0 0 0},clip,width=.135\linewidth]{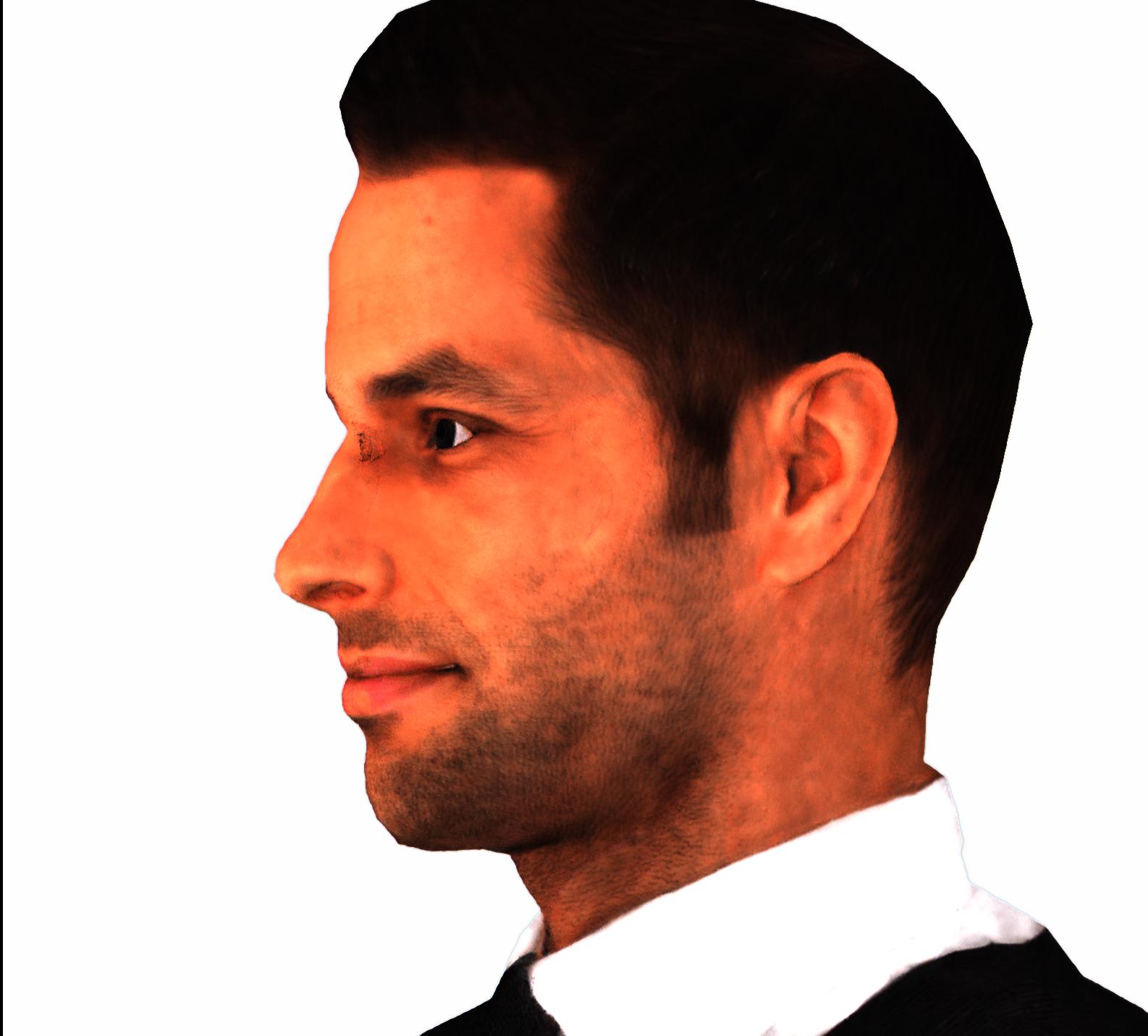}
    
    \rotatebox{90}{\hspace{0.25cm} DPR~\cite{Zhou19}}
    \adjincludegraphics[trim={0 0 0 0},clip,width=.135\linewidth]{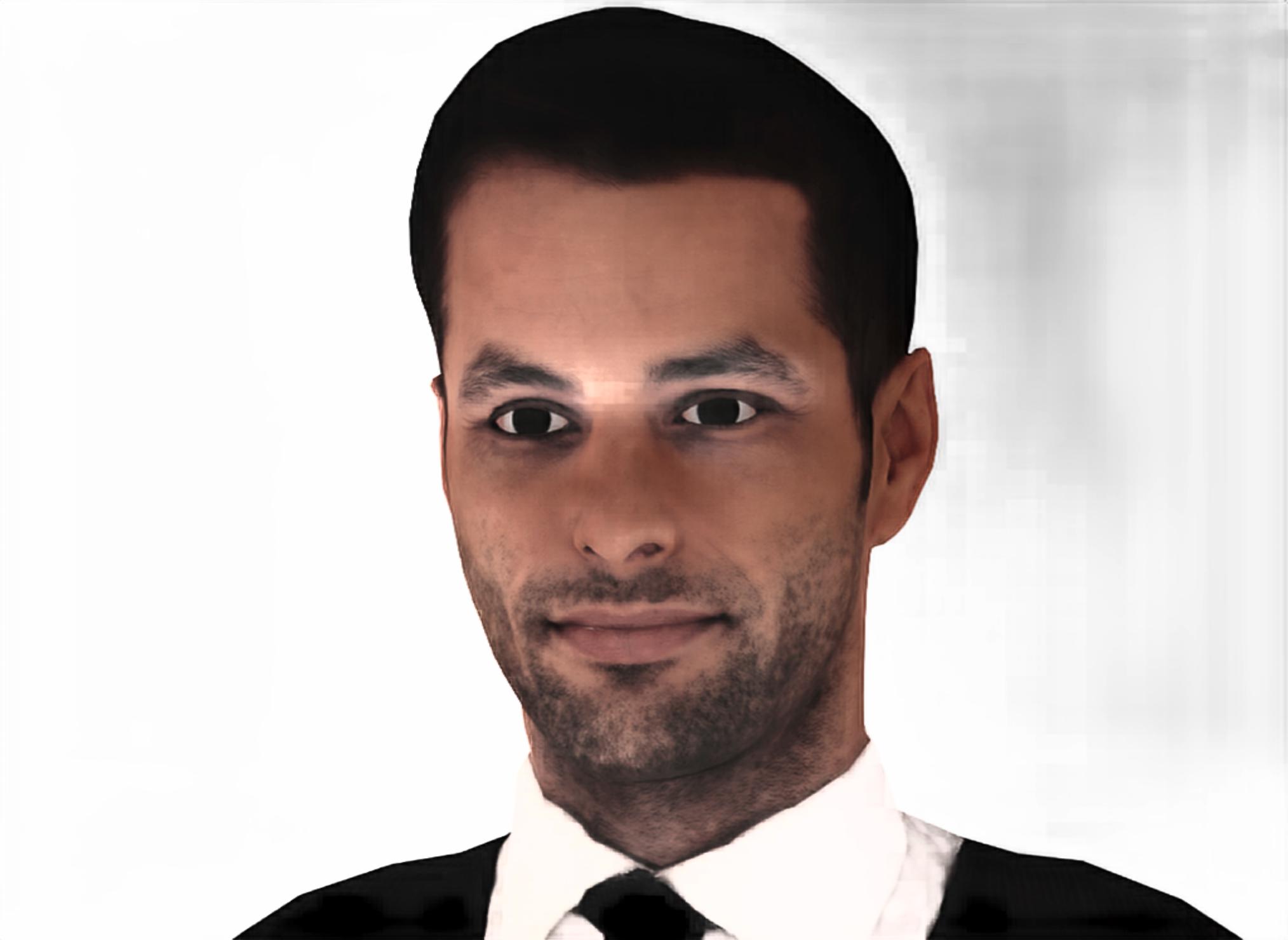}
    \adjincludegraphics[trim={0 0 0 0},clip,width=.135\linewidth]{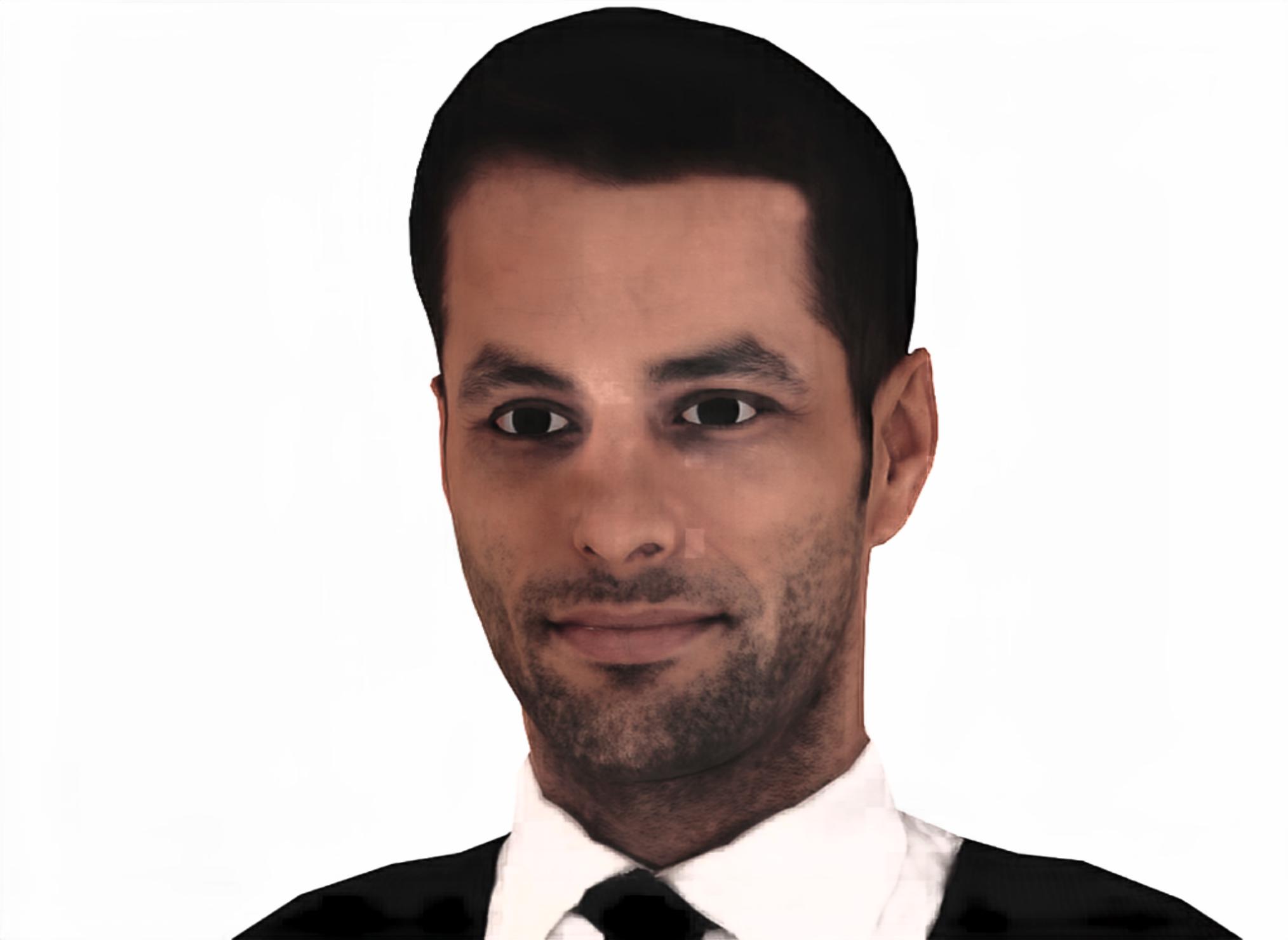}
    \adjincludegraphics[trim={0 0 0 0},clip,width=.135\linewidth]{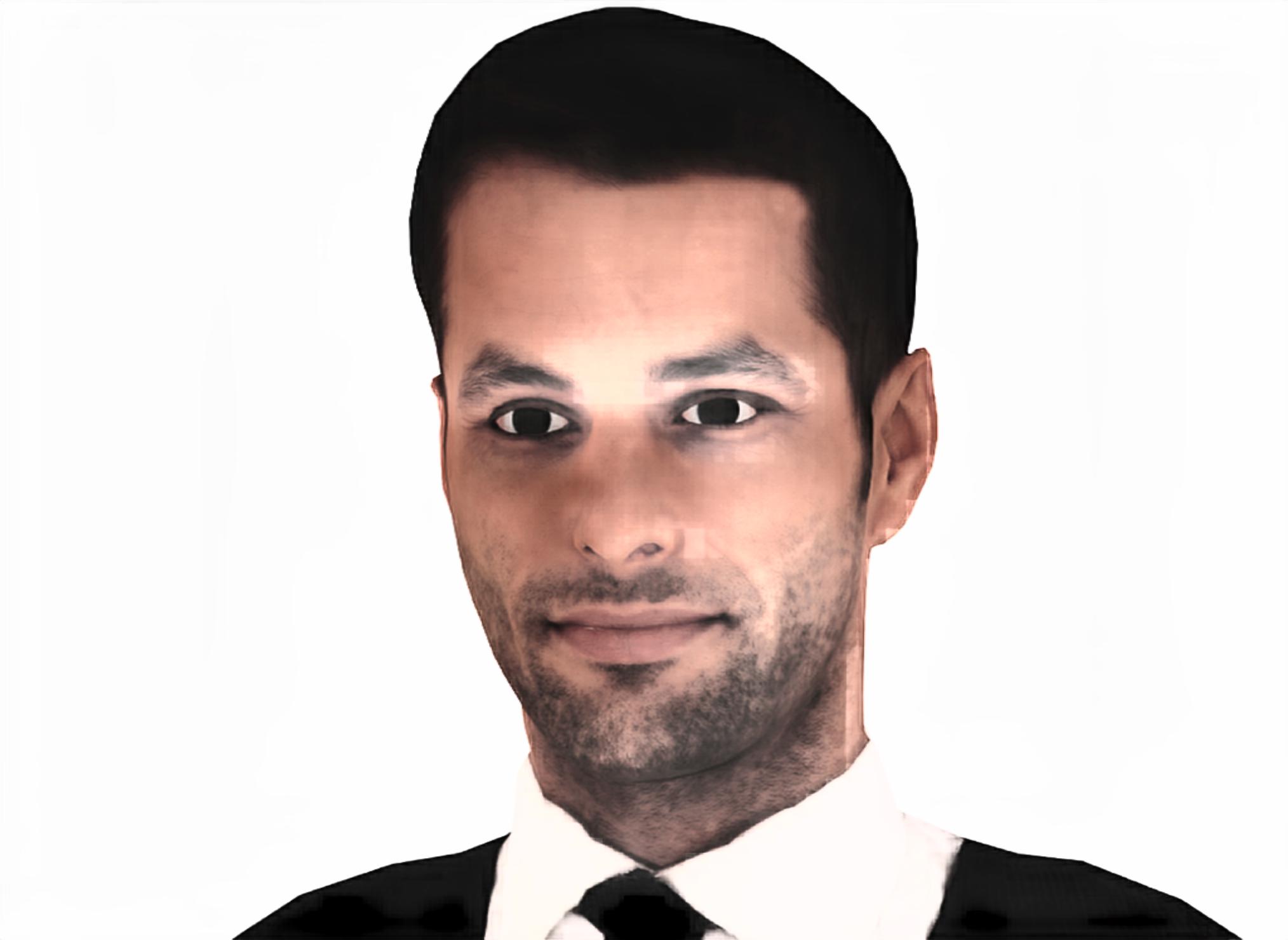}
    \hspace{0.5cm}
    % \hfill
    \adjincludegraphics[trim={0 0 {.25\width} 0},clip,width=.1\linewidth]{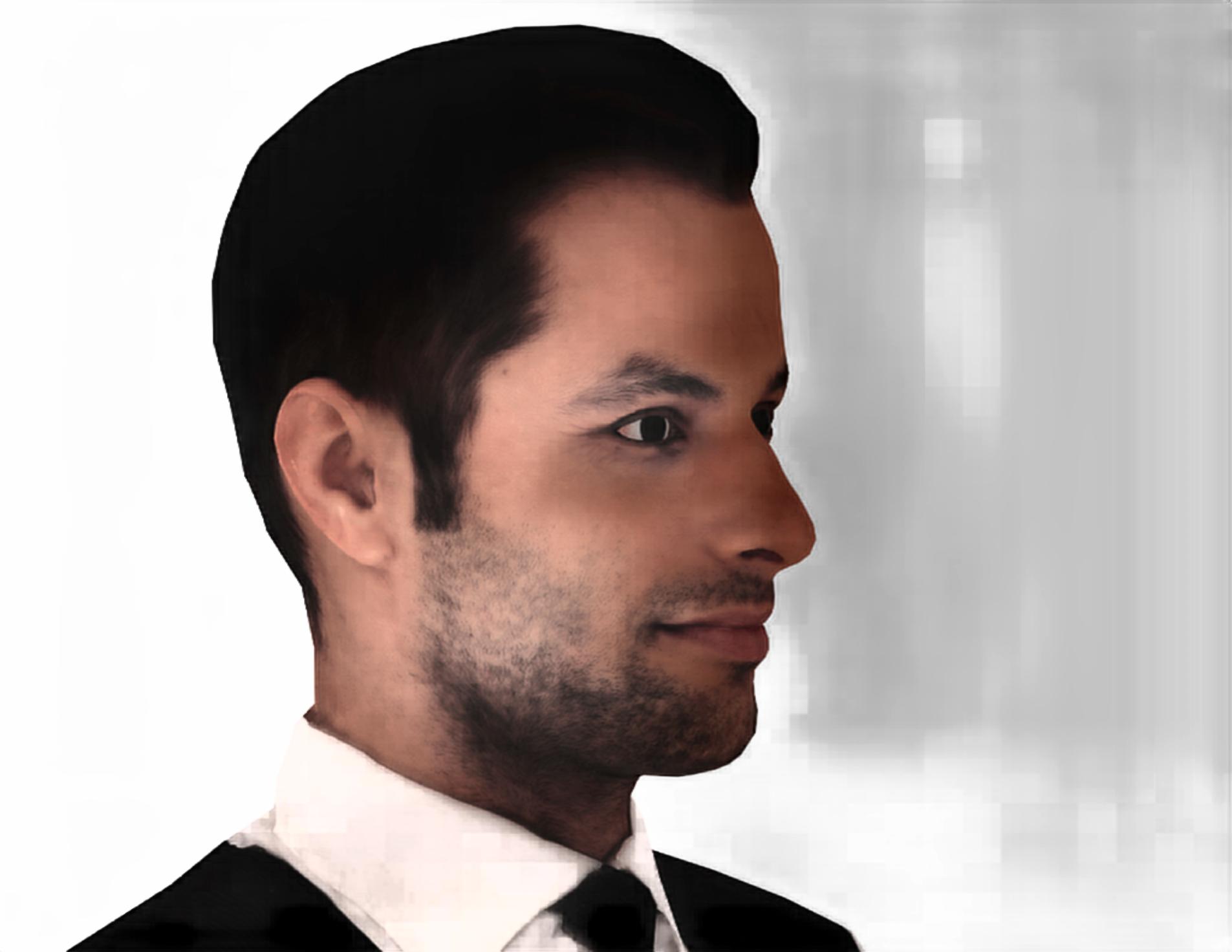}
    \hspace{0.25cm}
    \adjincludegraphics[trim={0 0 {.25\width} 0},clip,width=.1\linewidth]{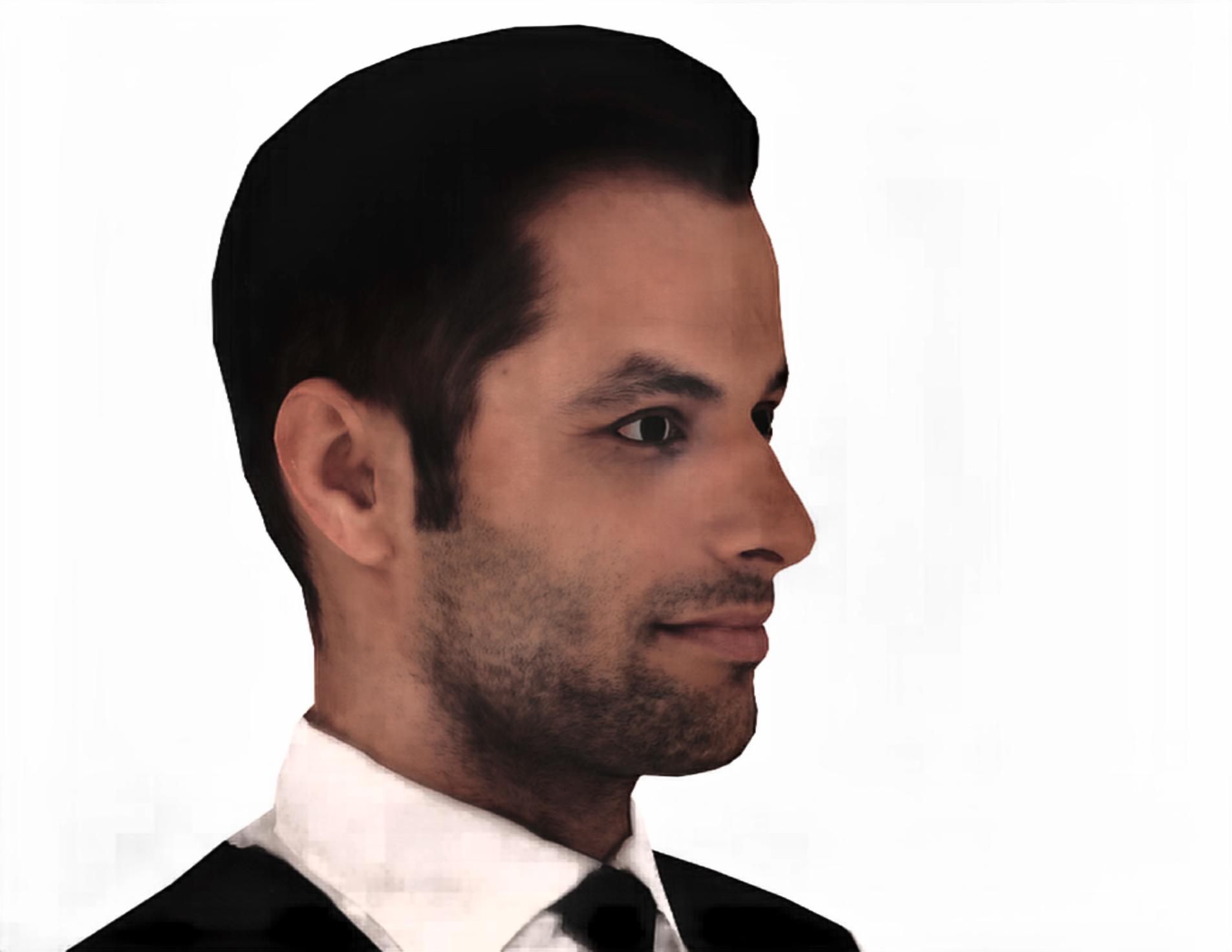}
    % \hfill
    \adjincludegraphics[trim={0 0 0 0},clip,width=.135\linewidth]{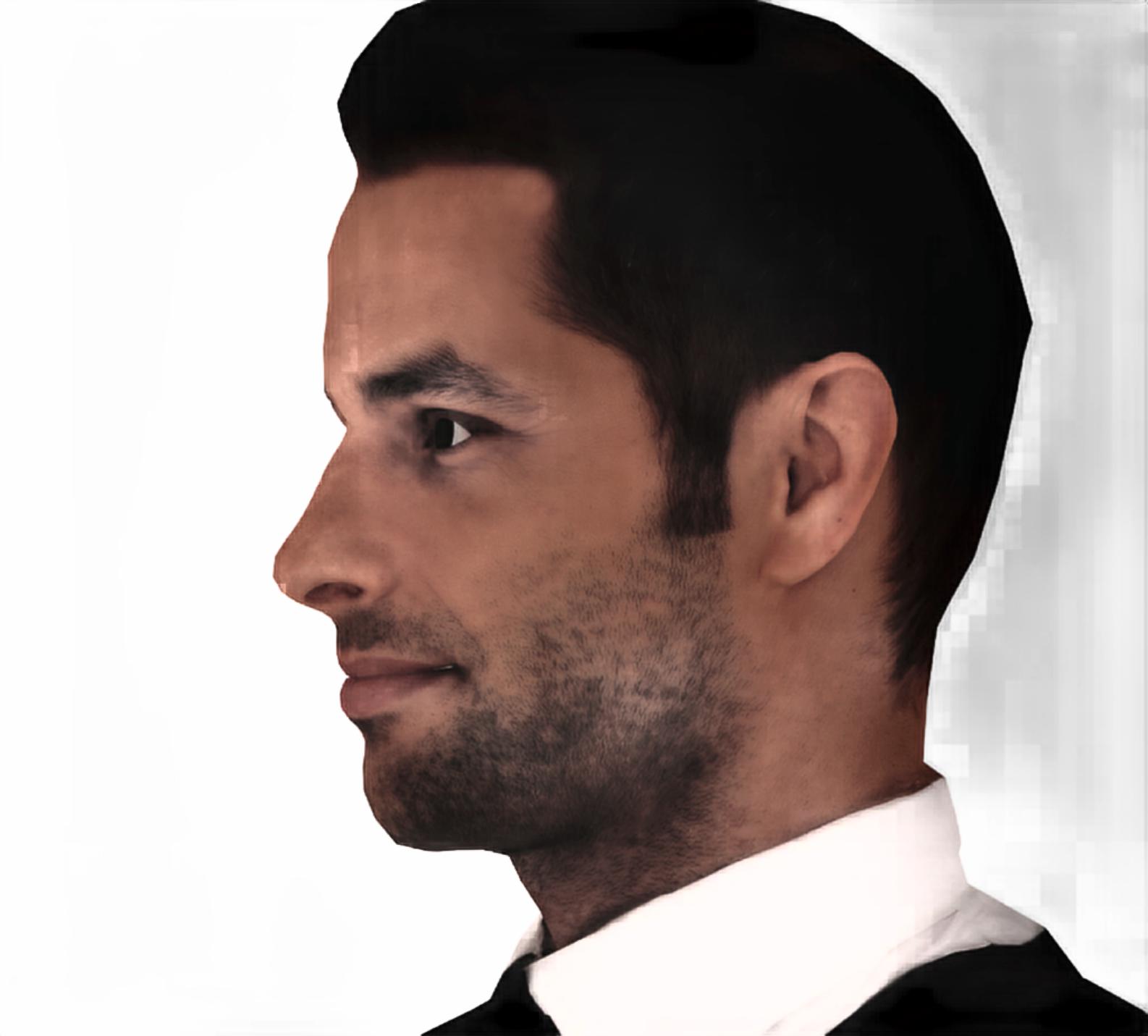}
    \adjincludegraphics[trim={0 0 0 0},clip,width=.135\linewidth]{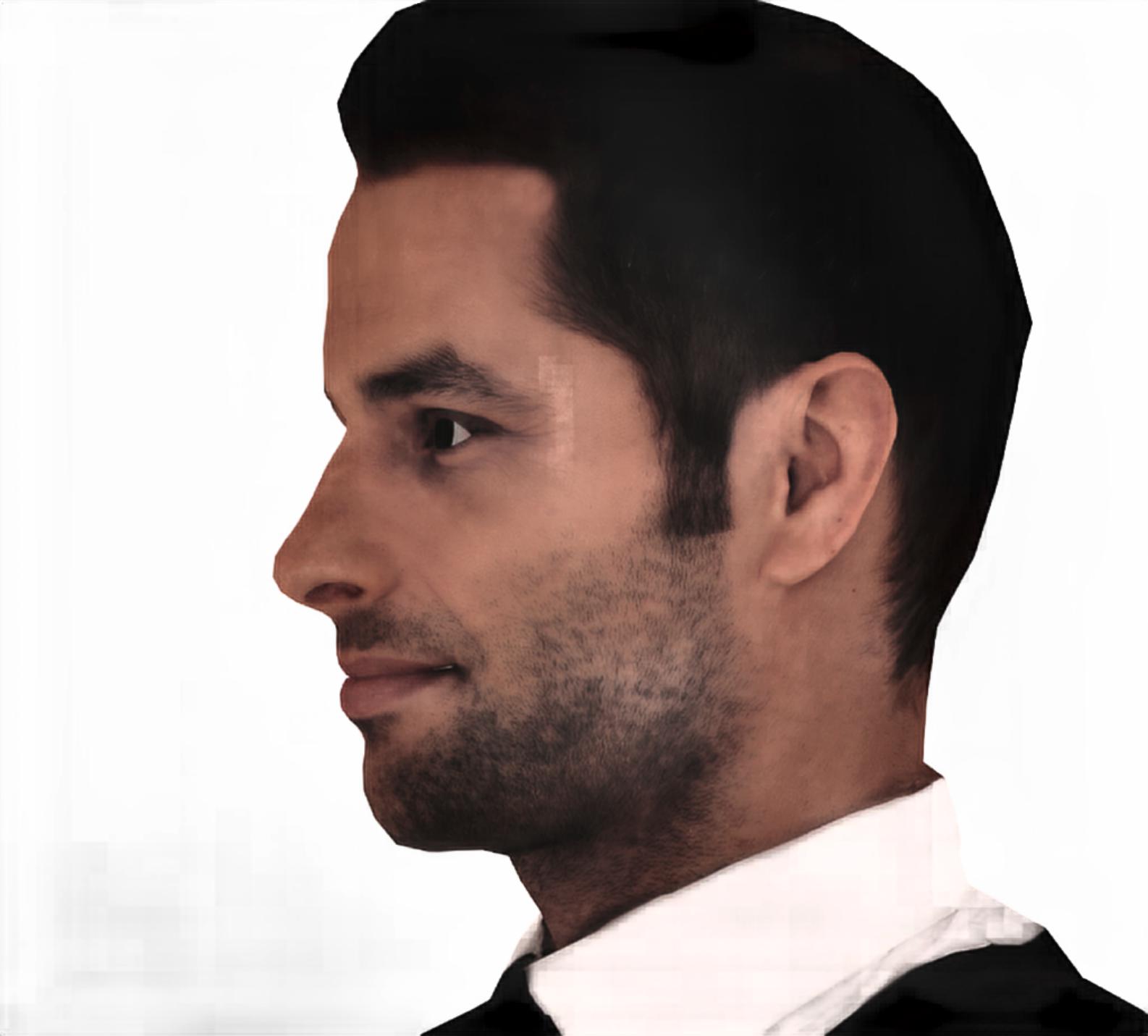}
    
    \rotatebox{90}{\hspace{-0.15cm} Ground truth}
    \adjincludegraphics[trim={0 0 0 0},clip,width=.135\linewidth]{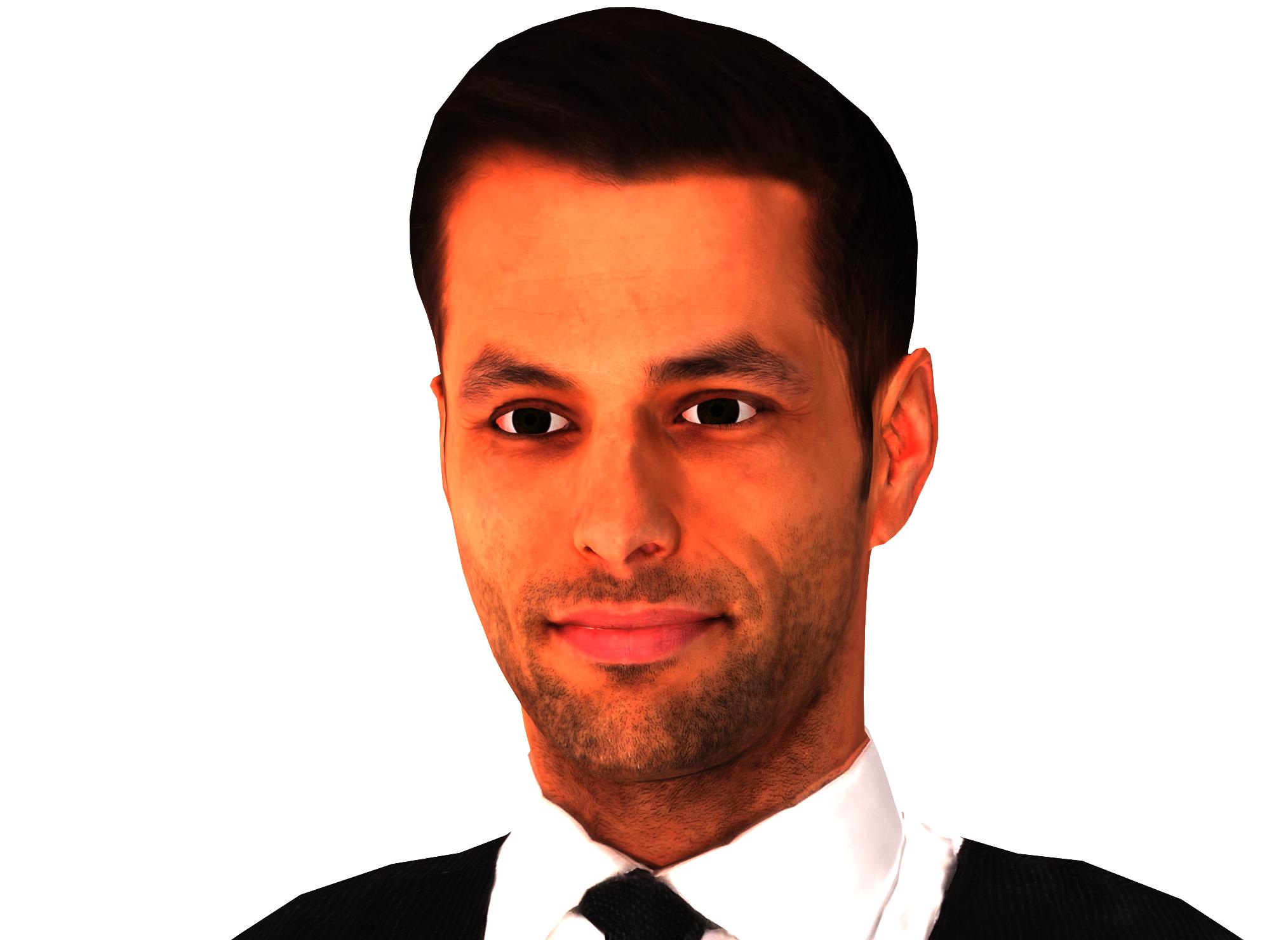}
    \adjincludegraphics[trim={0 0 0 0},clip,width=.135\linewidth]{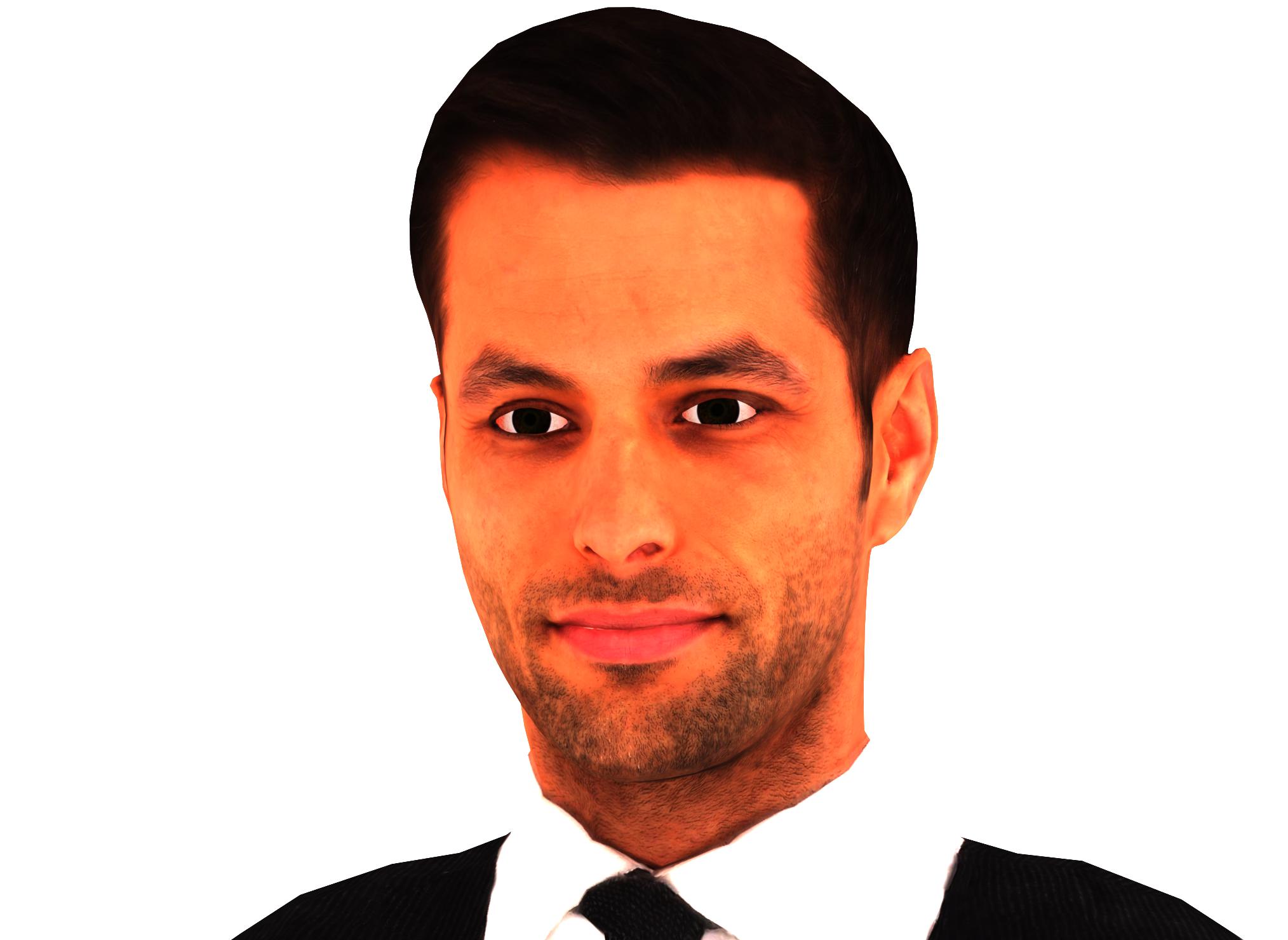}
    \adjincludegraphics[trim={0 0 0 0},clip,width=.135\linewidth]{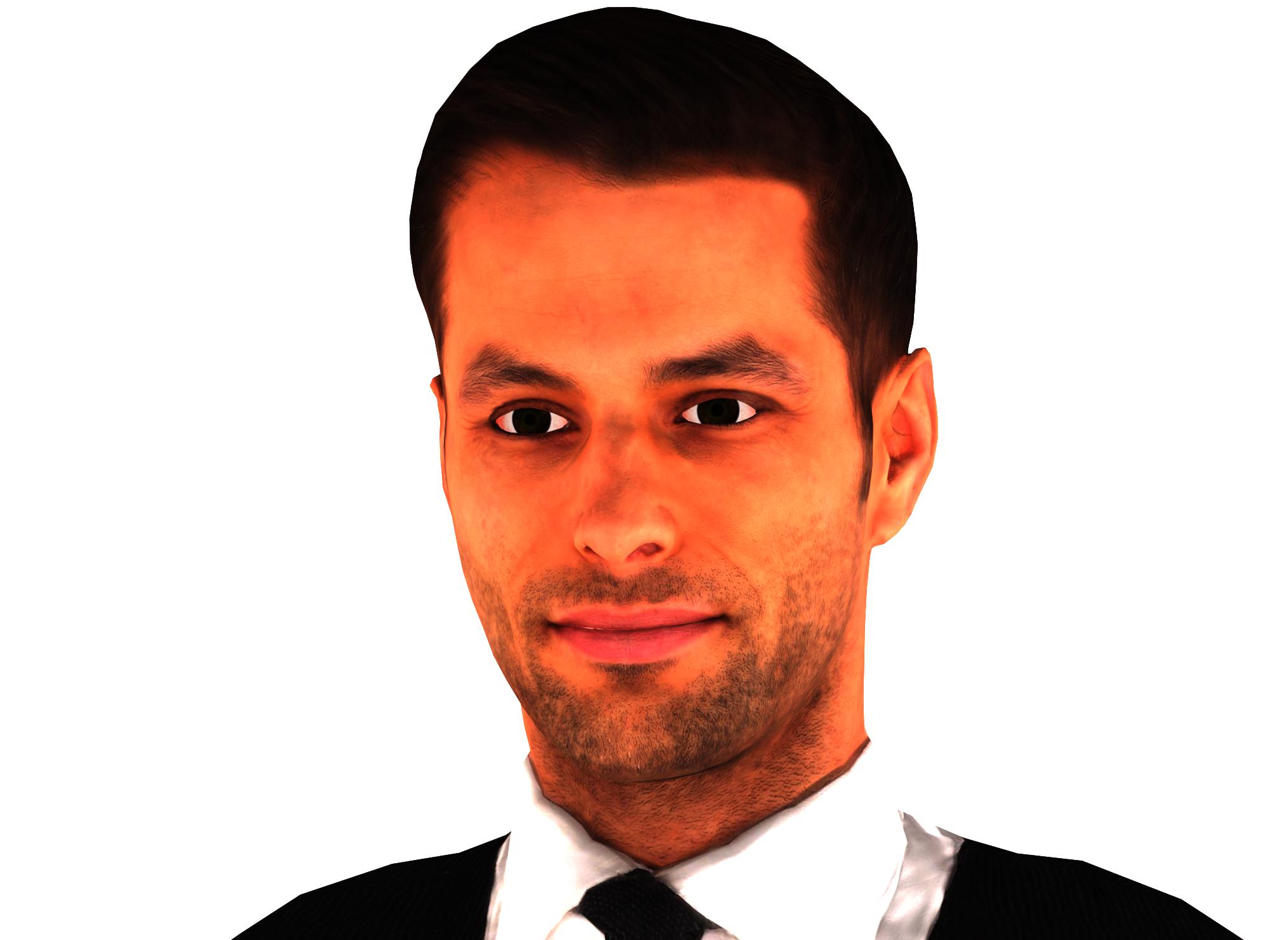}
    \hspace{0.5cm}
    % \hfill
    \adjincludegraphics[trim={0 0 {.25\width} 0},clip,width=.1\linewidth]{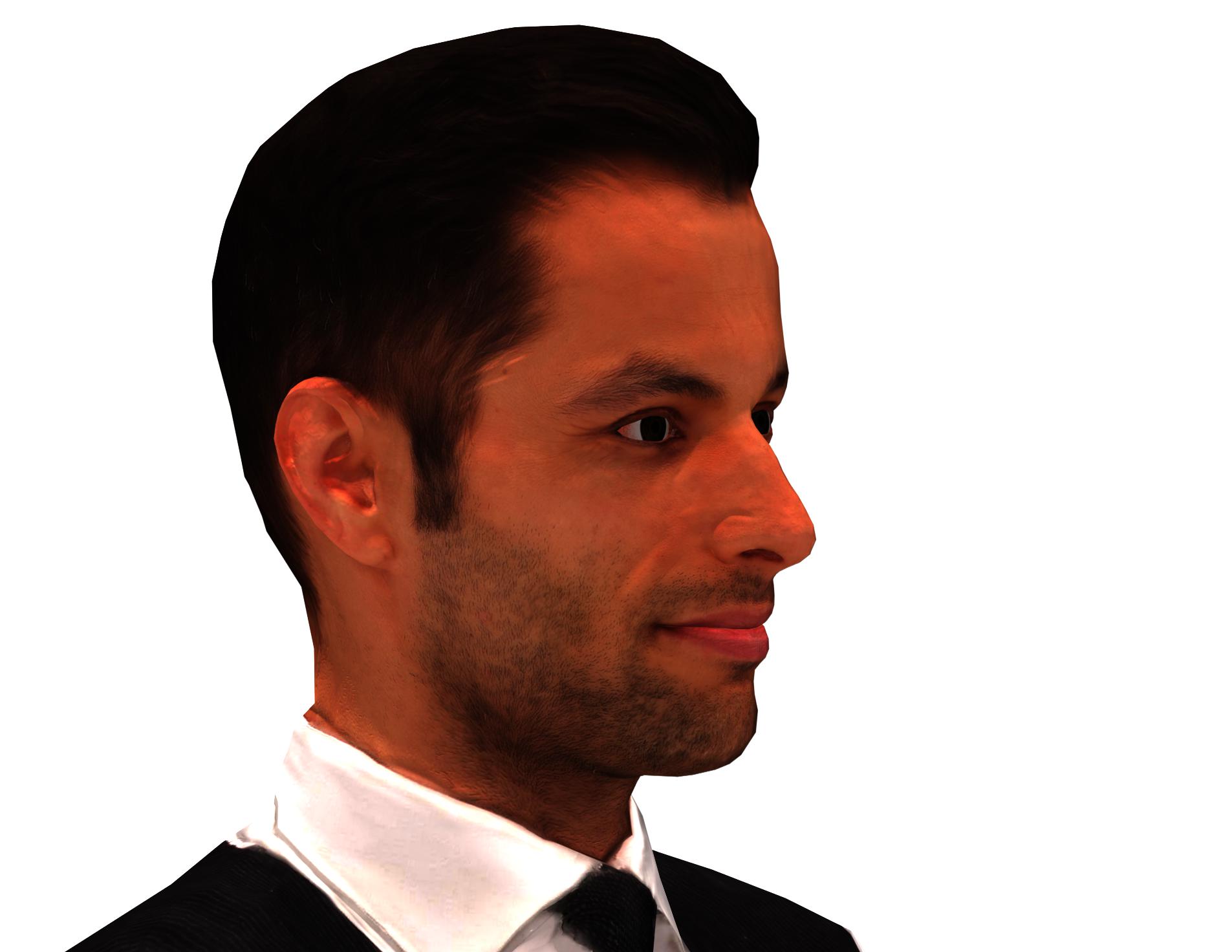}
    \hspace{0.25cm}
    \adjincludegraphics[trim={0 0 {.25\width} 0},clip,width=.1\linewidth]{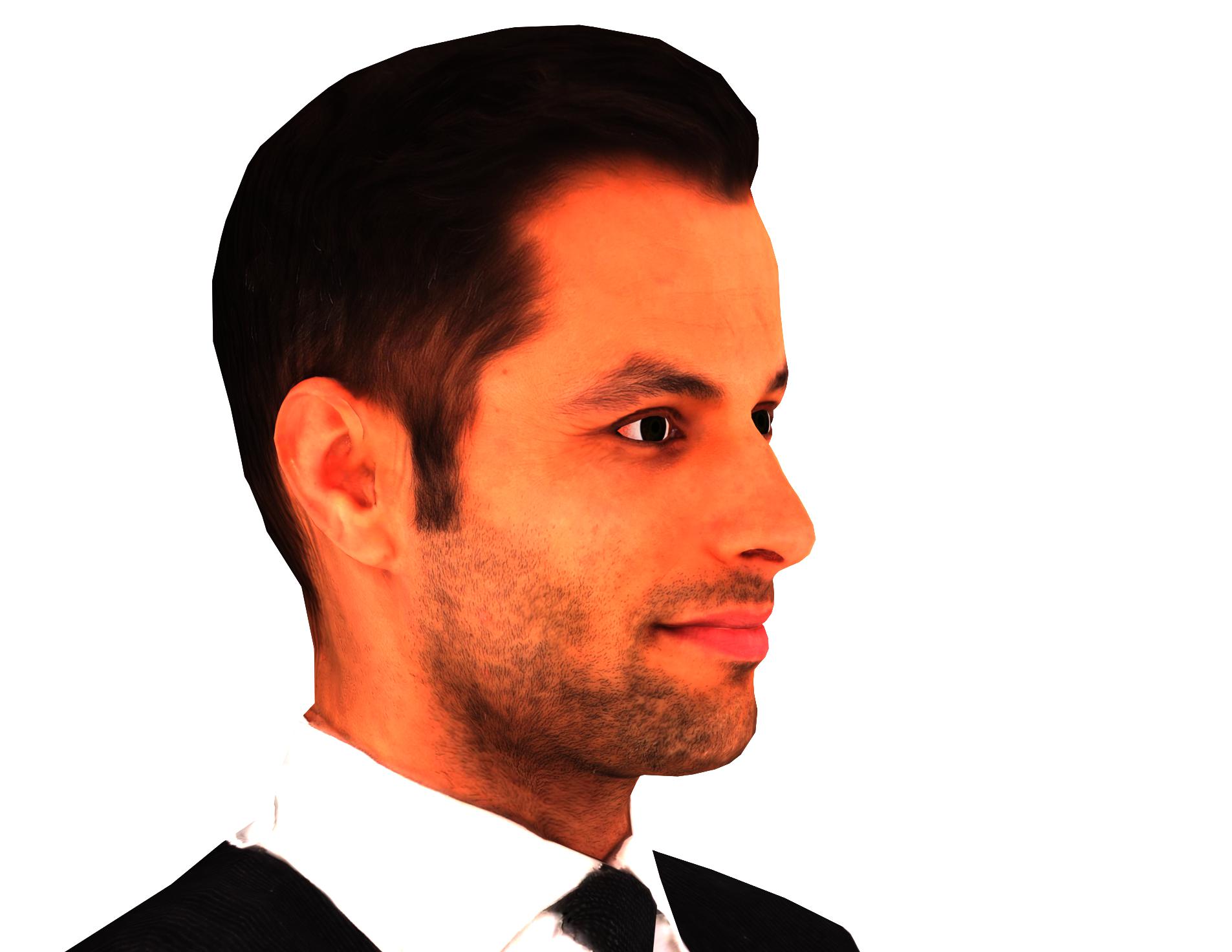}
    % \hfill
    \adjincludegraphics[trim={0 0 0 0},clip,width=.135\linewidth]{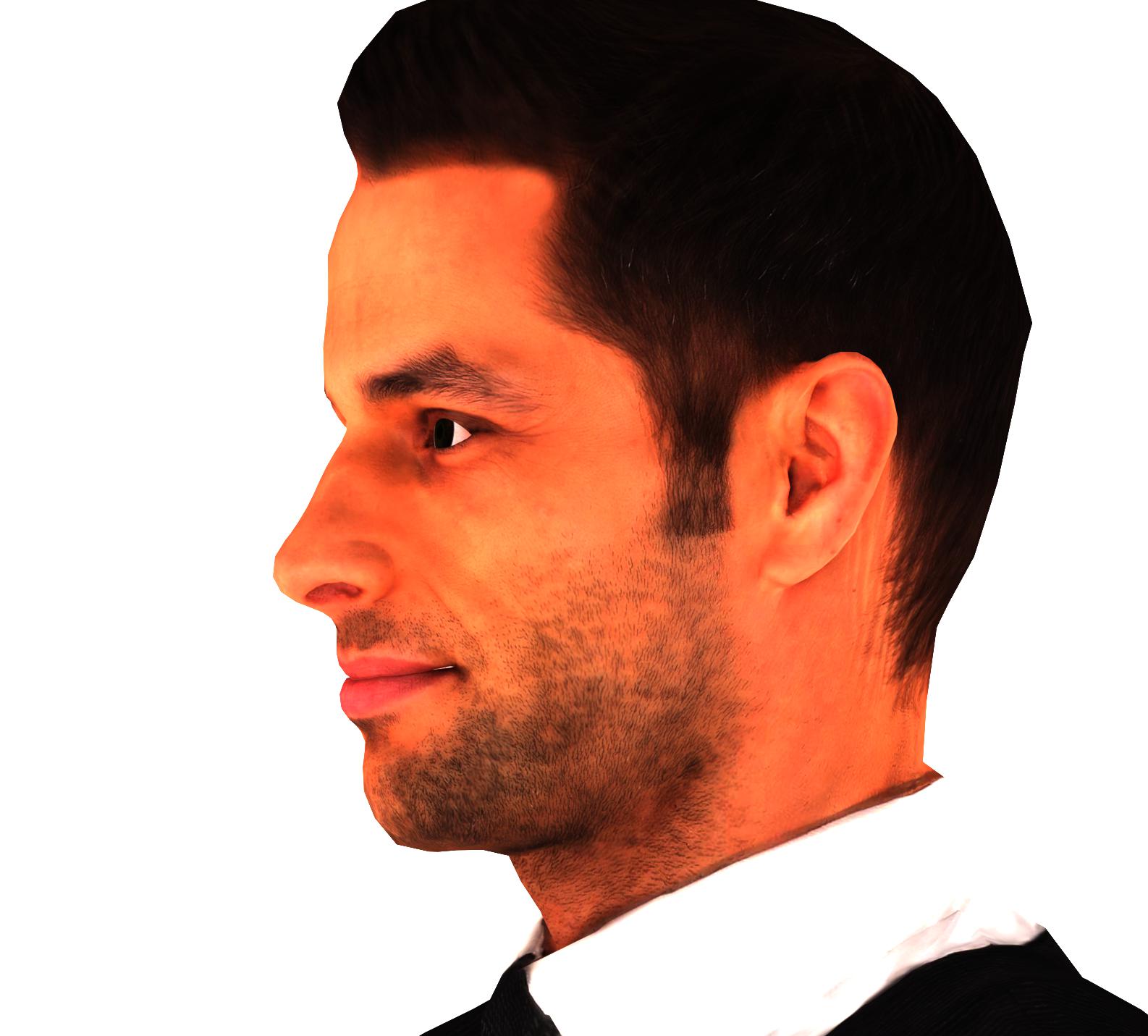}
    \adjincludegraphics[trim={0 0 0 0},clip,width=.135\linewidth]{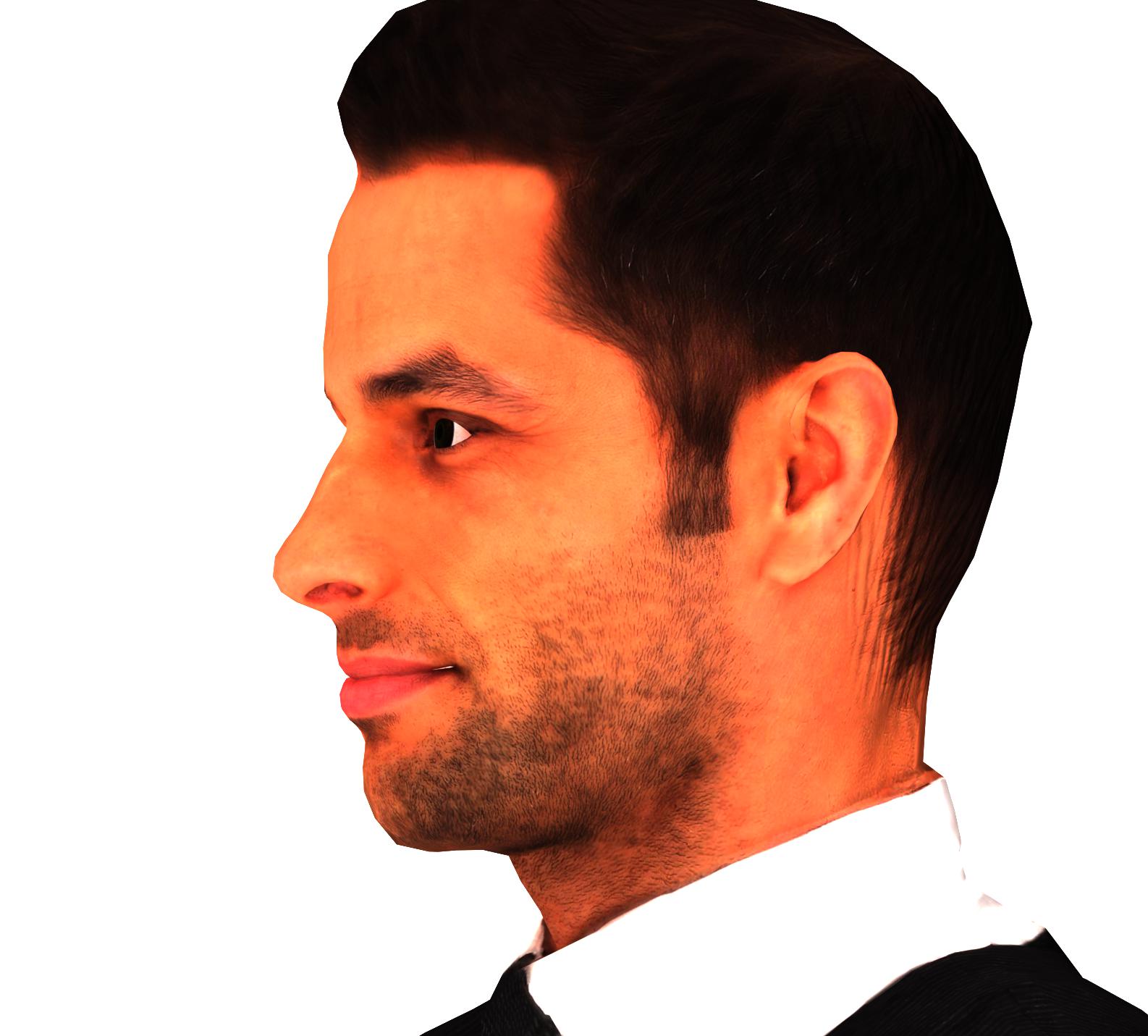}
    
    \caption{Qualitative comparison of the rendering synthetic models from RenderPeople under several viewpoints from validation and ambient+spherical harmonics lighting (see caption of Table~\ref{tab:synth_comparison} for more detail). The light directions are given in the \textbf{world space} and are aligned with the coordinate axes associated with the frontalized head.}
    \label{fig:synth_relighting}
\end{figure*}

The comparison of the relighted image with the ground truth on validation is presented in Table~\ref{tab:synth_comparison}. To make the comparison with the DPR baseline~\cite{Zhou19} feasible, the images for the metrics evaluation in the Table~\ref{tab:synth_comparison} were relighted by a combination of ambient light and $1^\mathrm{st}$-order Spherical Harmonics~\cite{Ramamoorthi01} (see the subsection~\ref{sec:exp_real}). Since DPR requires an image to be relighted, we only used non-flashlighted frames for the metrics evaluation. We cropped all images by a face localization network~\cite{Zhang16} and enlargened the bounding boxes to make them closer to the cropping of the faces in the training set of DPR (CelebA-HQ \cite{Karras17}). The predicted renderings of our method were cropped by the same bounding boxes for the comparison. Furthermore, we found that DPR significantly changes the color temperature of the original photo and alters the contribution of the direction light compared to the ambient light. Because of that, for the sake of fair comparison, the contribution of ambient and directional components of spherical harmonics passed to DPR were manually reweighted to match the color temperature of the ground truth as much as possible. The evaluation is based on three commonly used perceptual metrics (VGG~\cite{Johnson16}, LPIPS~\cite{Zhang18}, FID~\cite{Heusel17}) calculated over images with the masked background (for the predictions of both methods and the ground truth). The results are averaged over three people, ten non-flashlighted validation frames for each of them, and three directions of $1^\mathrm{st}$-order SH (\textit{left-to-right, up-to-down, forward-to-backward} in the world space associated with the frontalized head). The examples of the renderings by our method and DPR under these conditions are presented in Fig.~\ref{fig:synth_relighting}. 

\begin{figure*}
    \centering
    \hspace{0.2cm}
    \begin{subfigure}{.19\linewidth}
        \centering 
        Albedo (predicted)
    \end{subfigure} 
    \hfill
    \begin{subfigure}{.19\linewidth}
        \centering 
        Albedo (g.t.)
    \end{subfigure}
    \hfill
    \begin{subfigure}{.19\linewidth}
        \centering 
        Normals (predicted)
    \end{subfigure}
    \hfill
    \begin{subfigure}{.19\linewidth}
        \centering 
        Normals (PRNet) 
    \end{subfigure}
    \hfill
    \begin{subfigure}{.19\linewidth}
        \centering 
        Normals (g.t.)
    \end{subfigure}
    %\hspace{1.0cm} Albedo (pred.) \hspace{0.4cm} \hfill Albedo (g.t.) \hfill \hspace{0.4cm} Normals (pred.) \hfill \hspace{0.3cm} Normals (PRNet) \hfill \hspace{0.3cm} Normals (g.t.) \hspace{1.0cm} 
    
    \rotatebox{90}{\hspace{0.3cm} \textit{Person 1 (Carla)}}
    \adjincludegraphics[trim={0 0 0 0},clip,width=.19\linewidth]{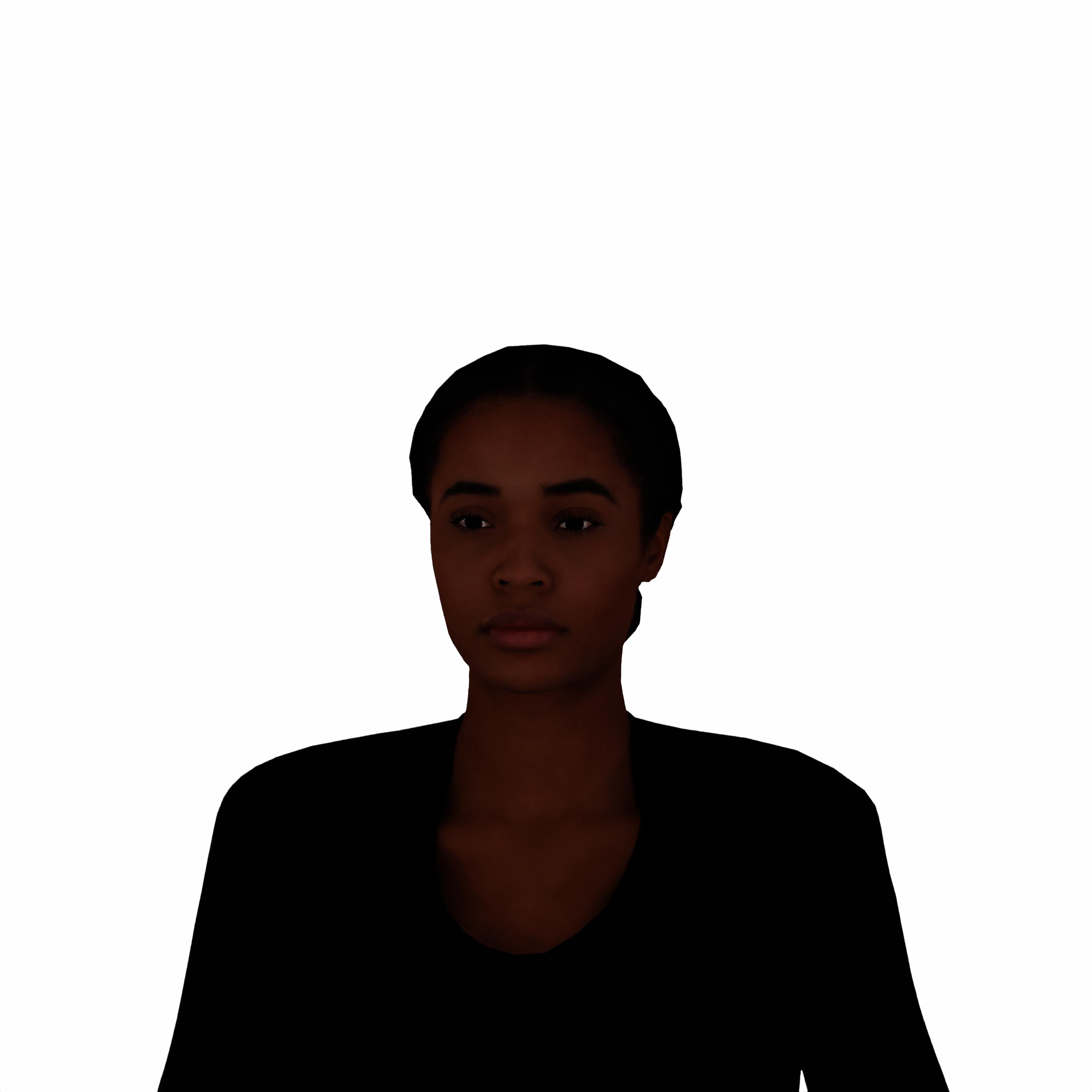}
    \hfill
    \adjincludegraphics[trim={0 0 0 0},clip,width=.19\linewidth]{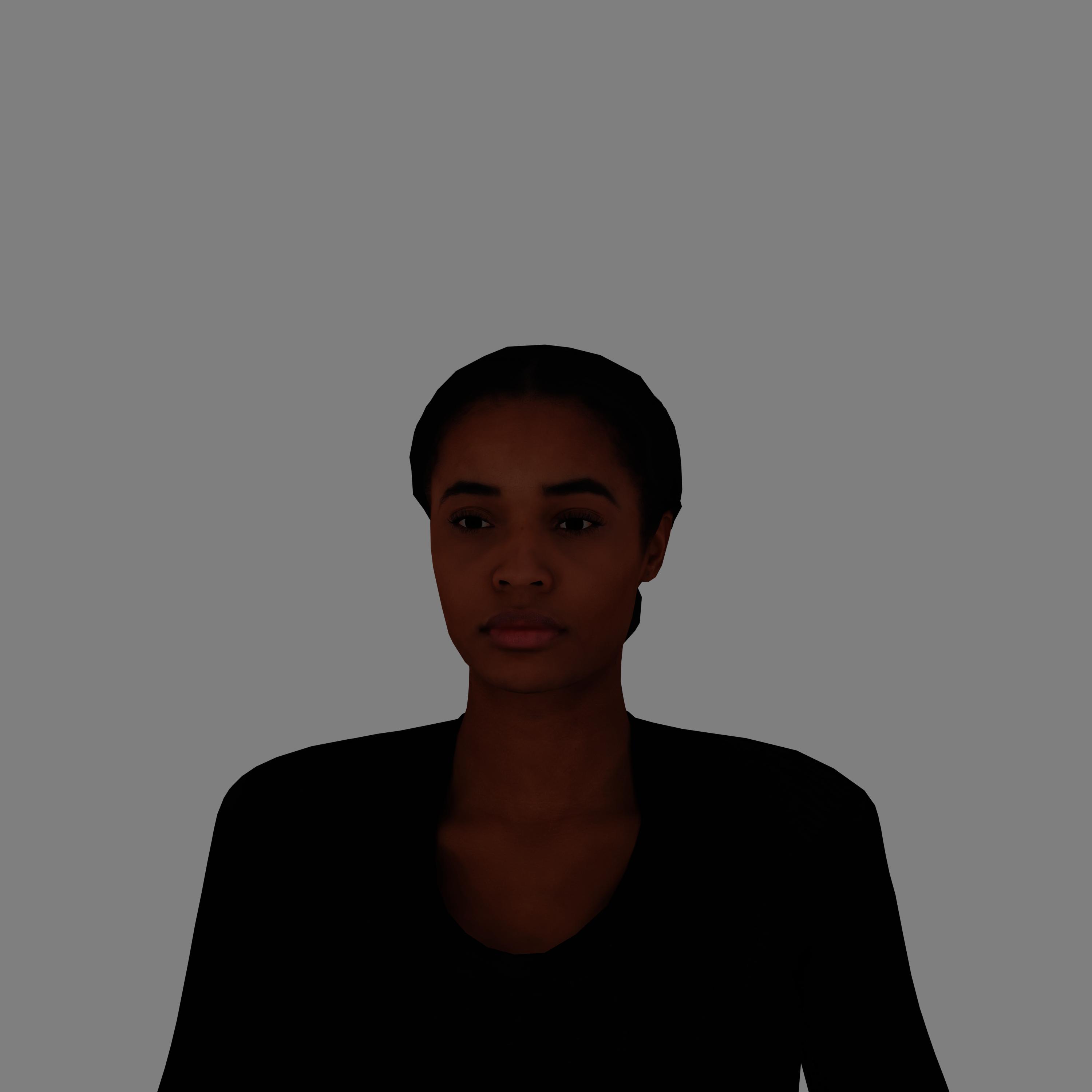}
    \hfill
    \adjincludegraphics[trim={0 0 0 0},clip,width=.19\linewidth]{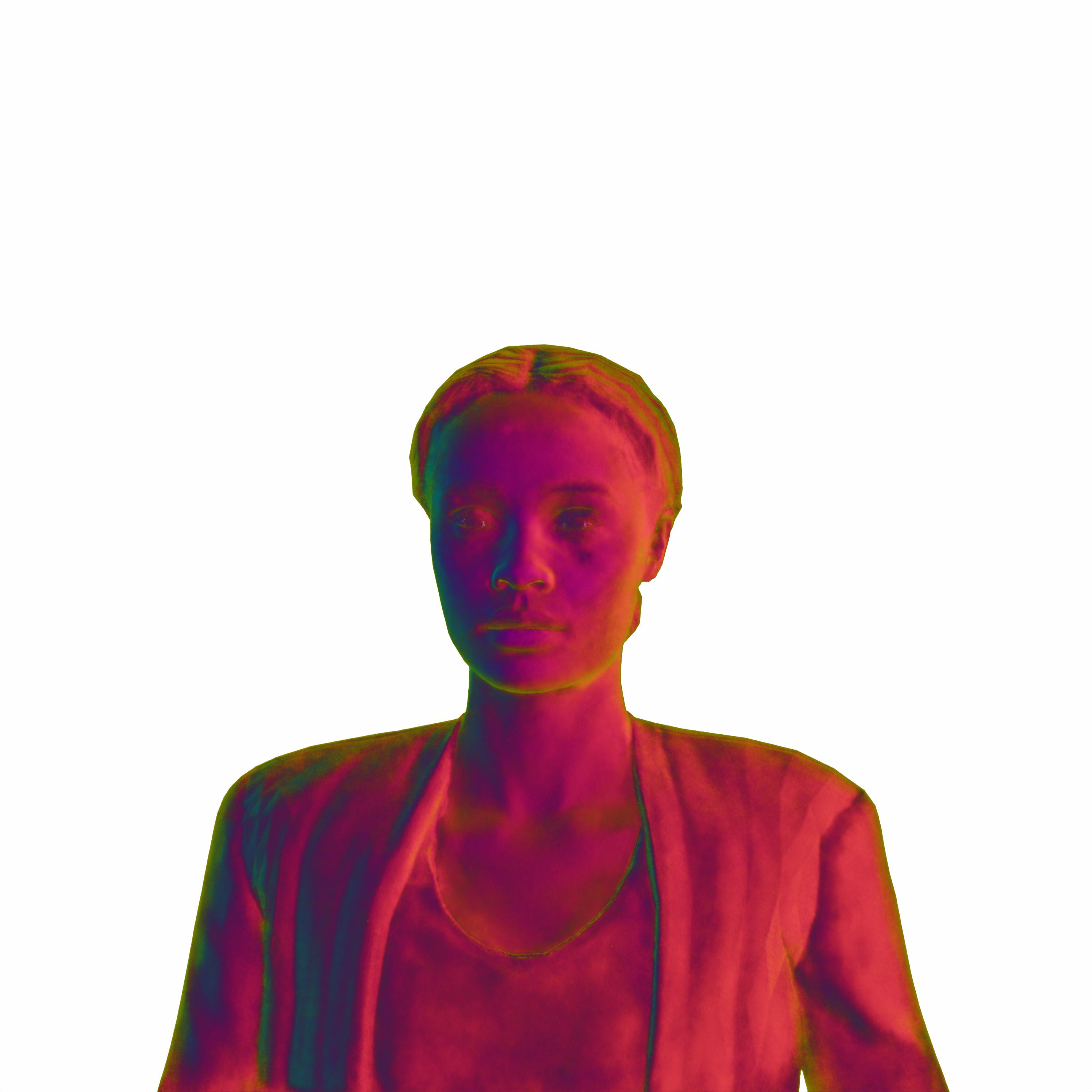}
    \hfill
    \adjincludegraphics[trim={0 0 0 0},clip,width=.19\linewidth]{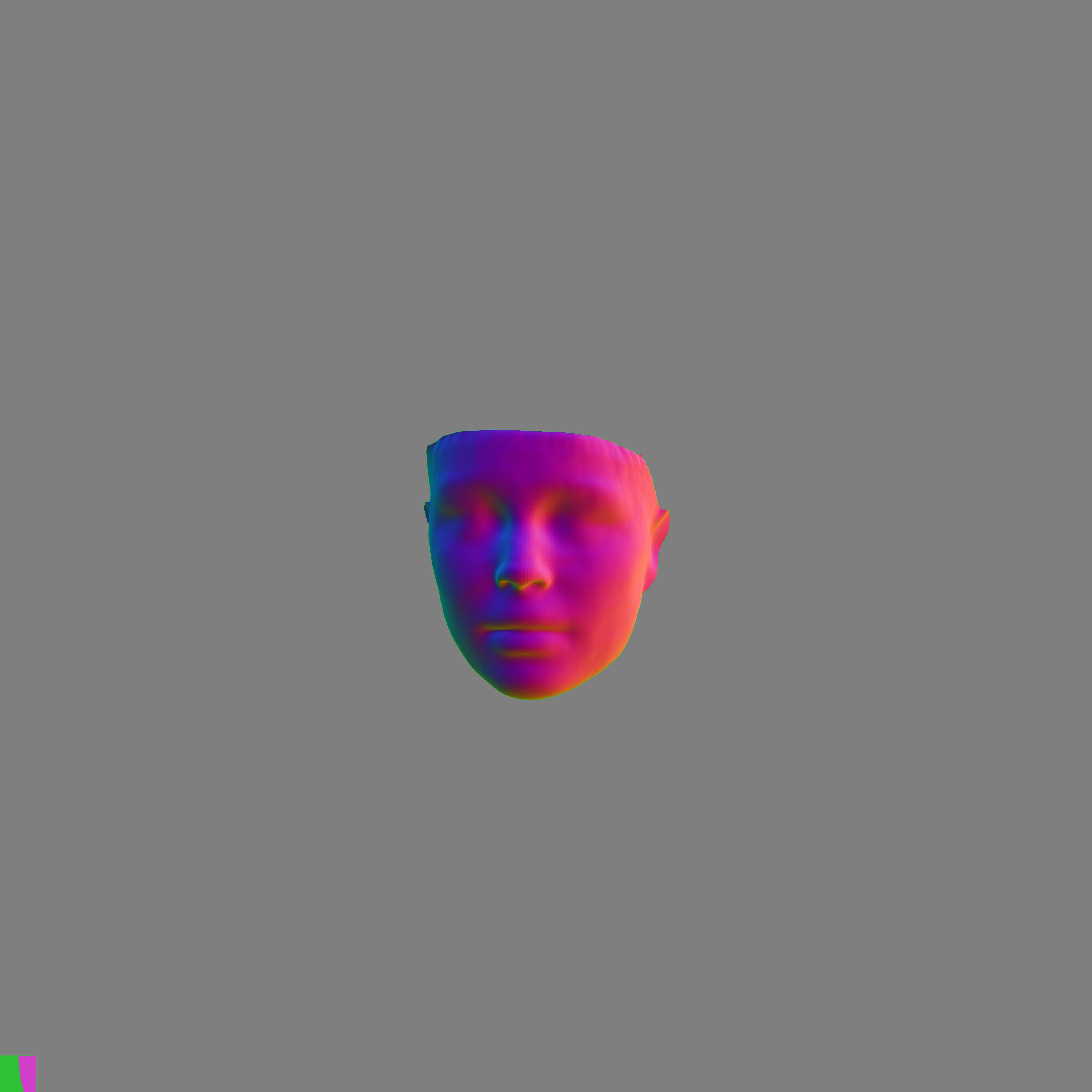}
    \hfill
    \adjincludegraphics[trim={0 0 0 0},clip,width=.19\linewidth]{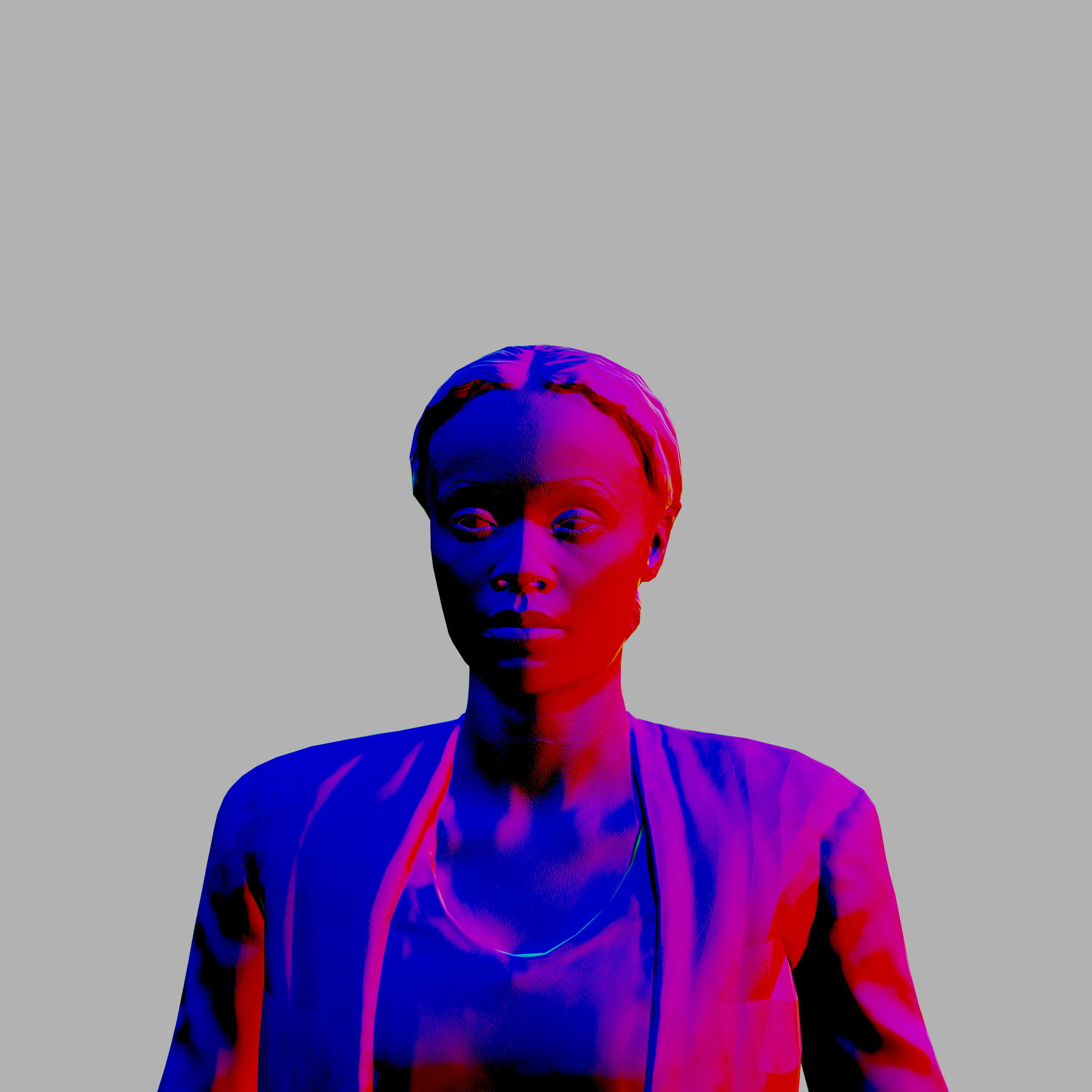}
    
    \rotatebox{90}{\hspace{0.2cm} \textit{Person 2 (Claudia)}}
    \adjincludegraphics[trim={0 0 0 0},clip,width=.19\linewidth]{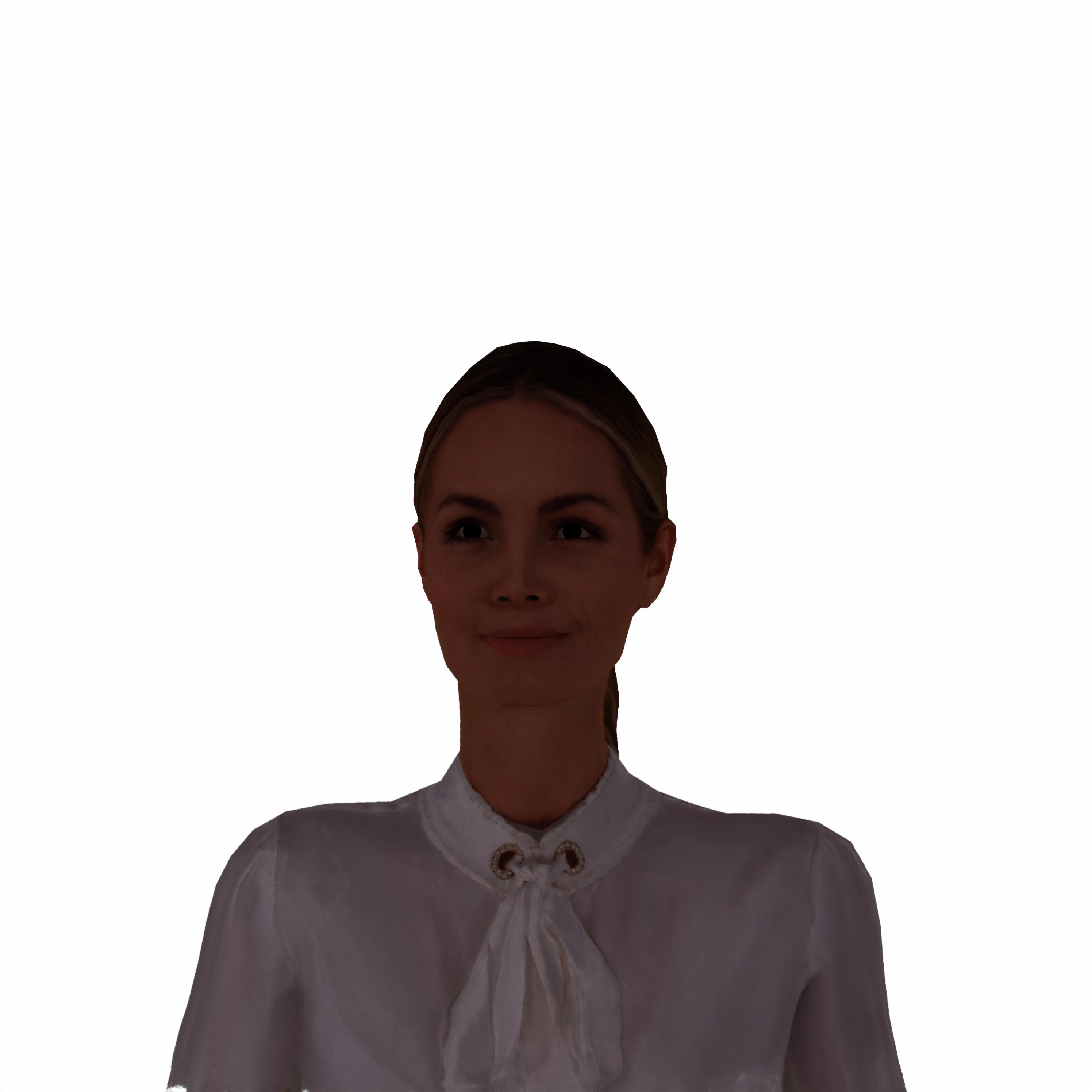}
    \hfill
    \adjincludegraphics[trim={0 0 0 0},clip,width=.19\linewidth]{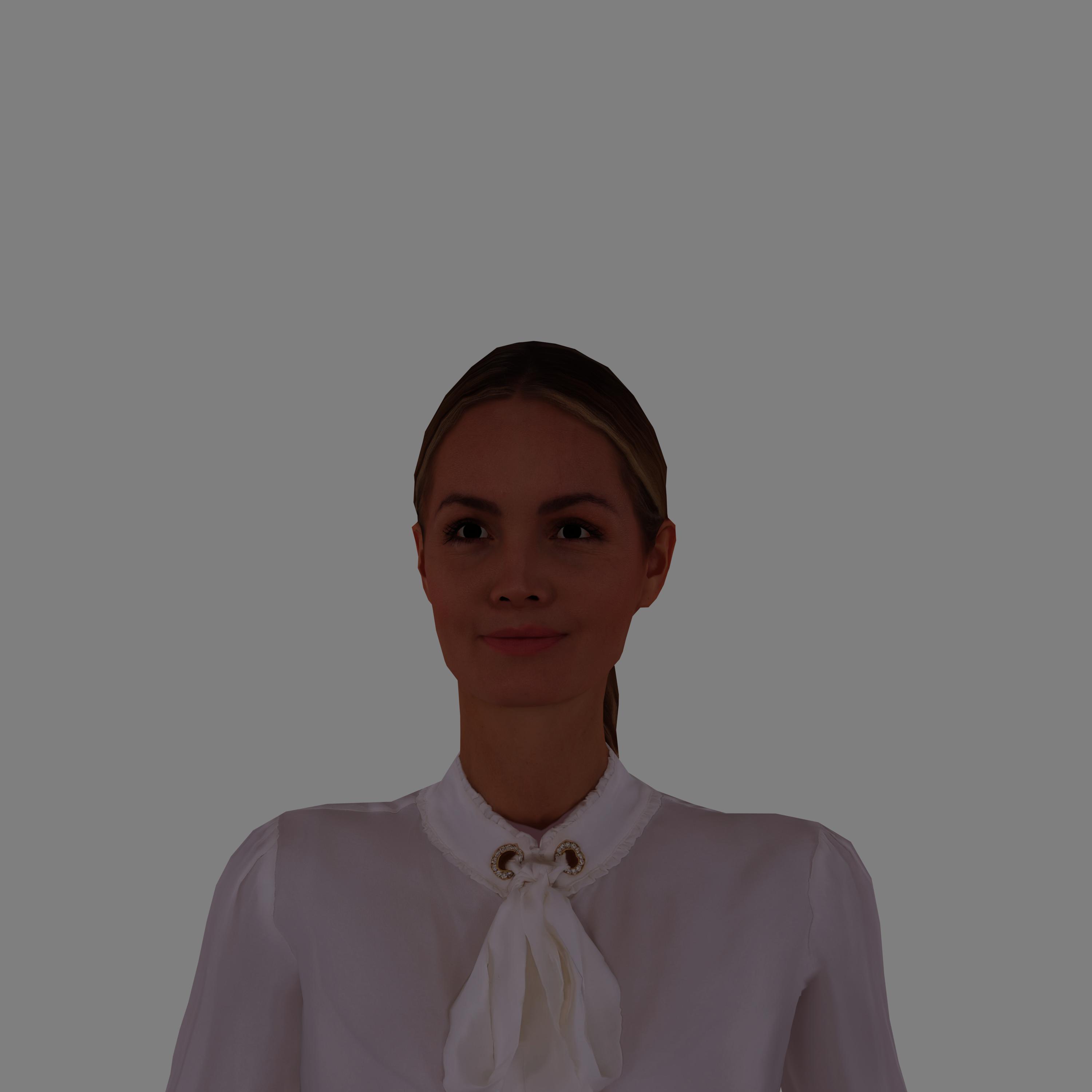}
    \hfill
    \adjincludegraphics[trim={0 0 0 0},clip,width=.19\linewidth]{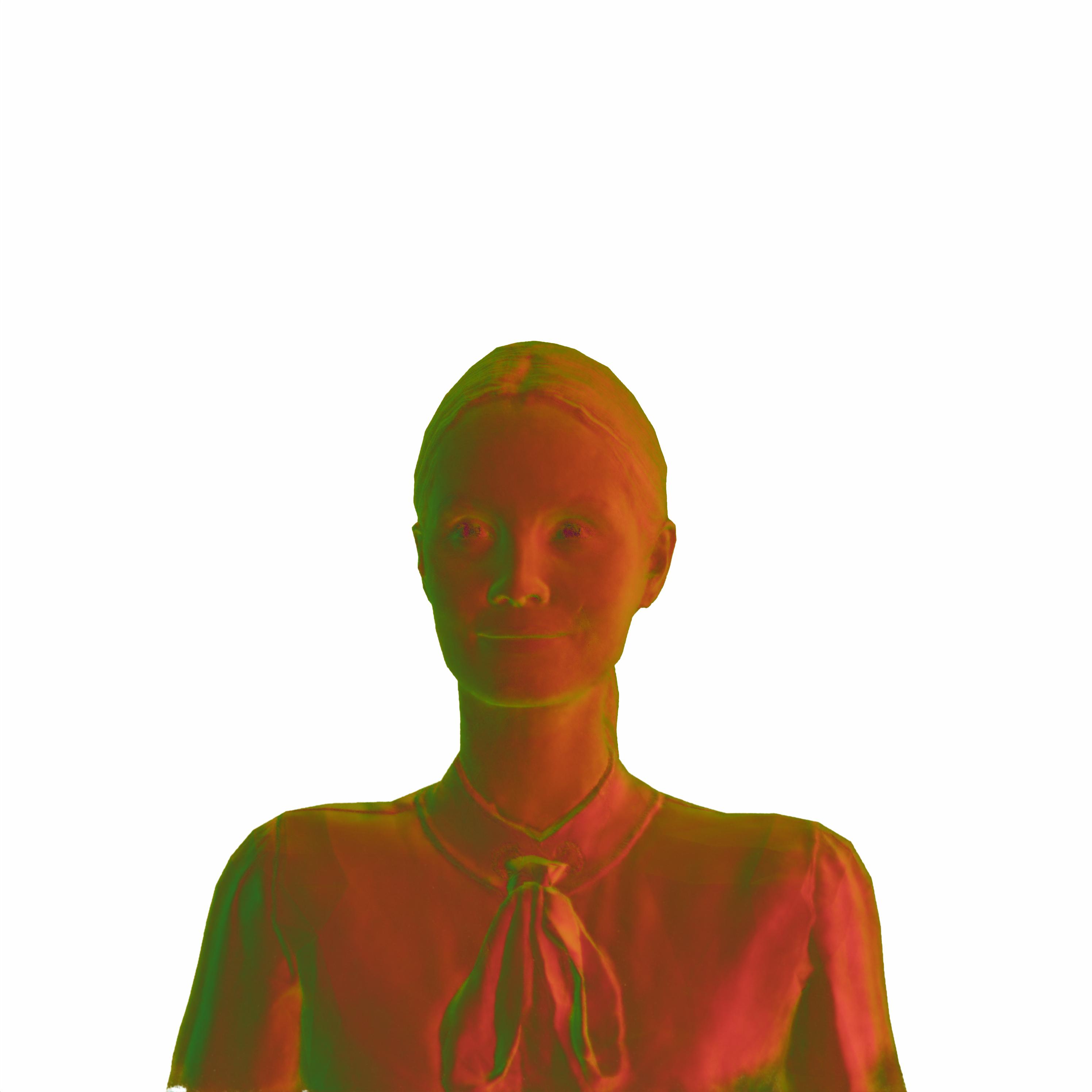}
    \hfill
    \adjincludegraphics[trim={0 0 0 0},clip,width=.19\linewidth]{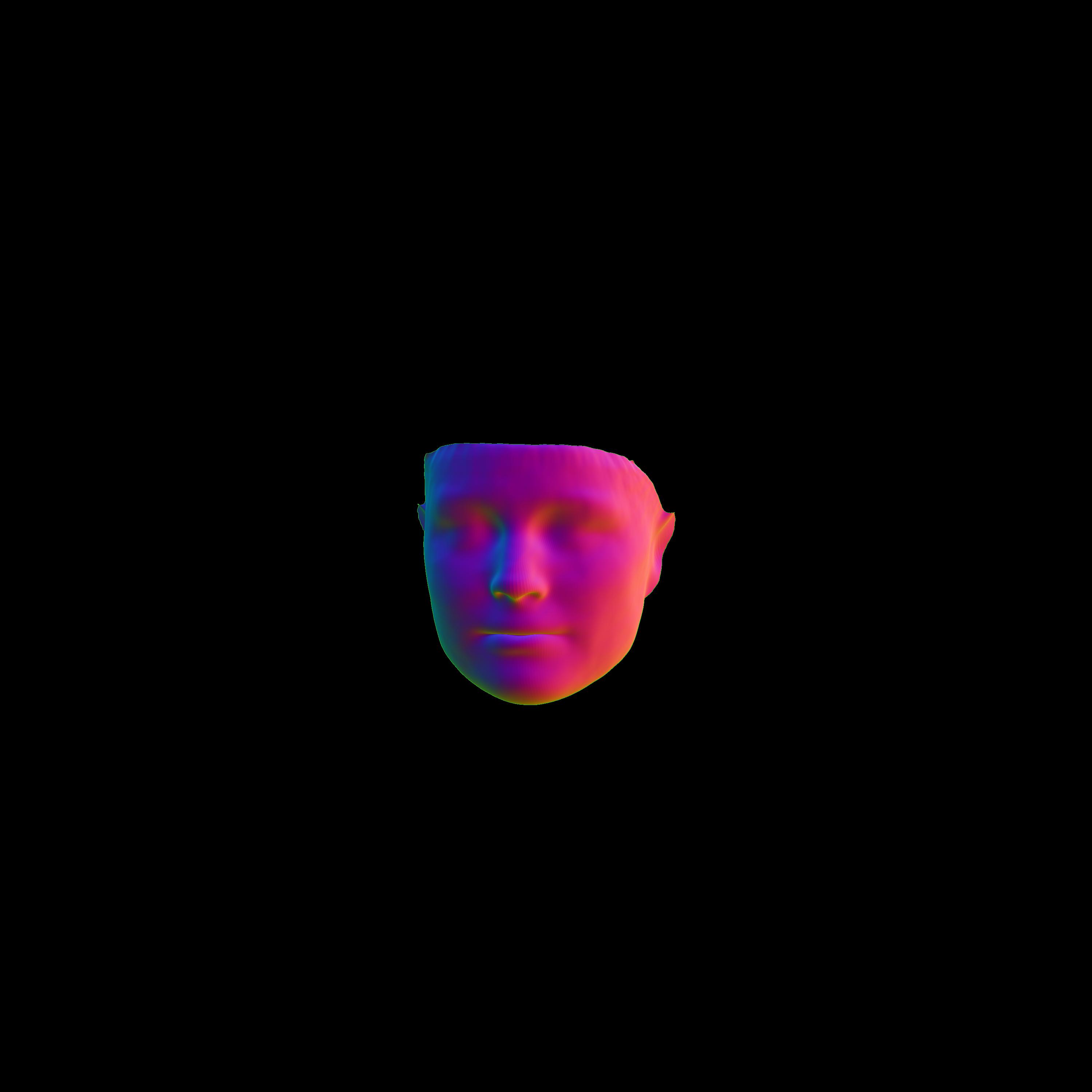}
    \hfill
    \adjincludegraphics[trim={0 0 0 0},clip,width=.19\linewidth]{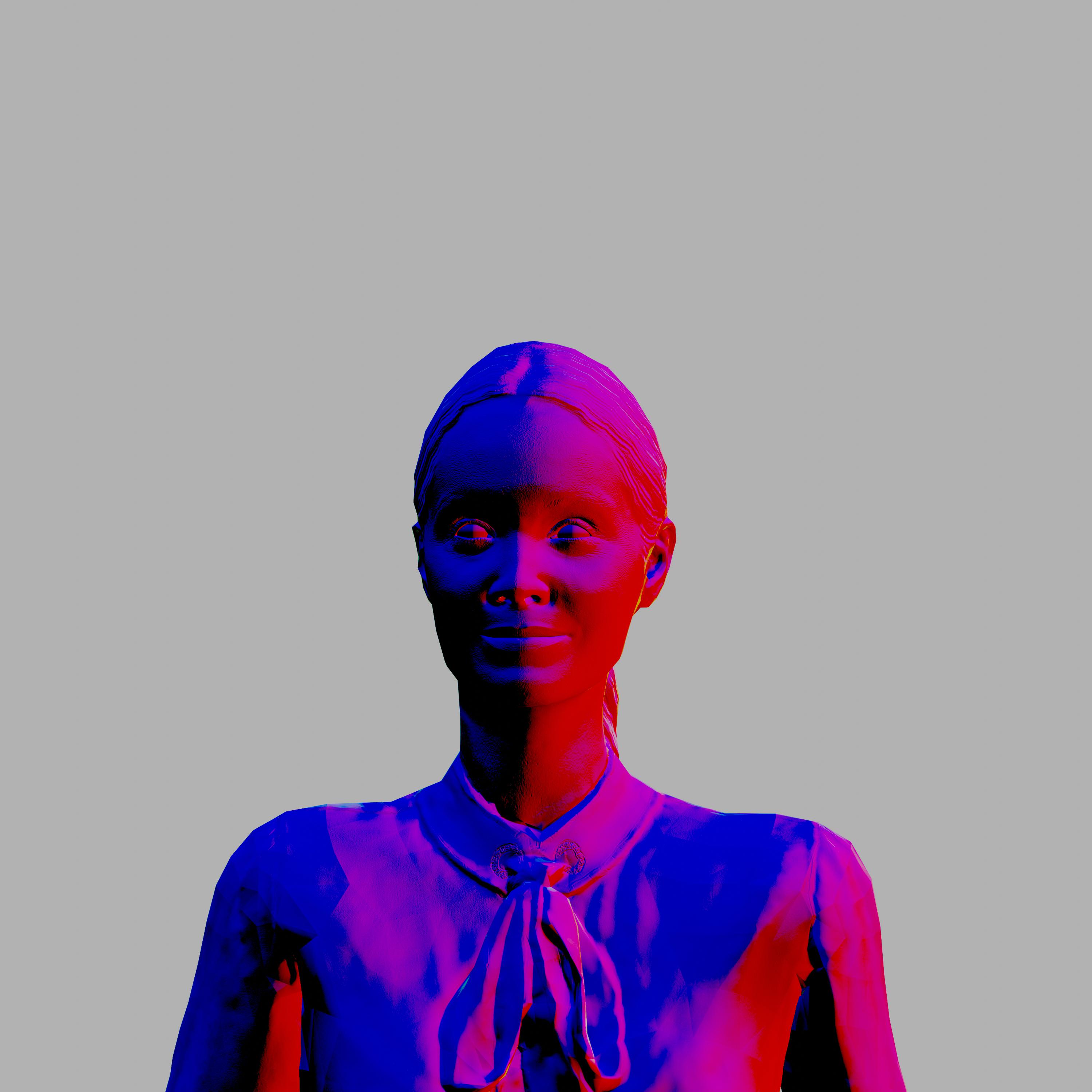}
    
    \rotatebox{90}{\hspace{0.4cm} \textit{Person 3 (Eric)}}
    \adjincludegraphics[trim={0 0 0 0},clip,width=.19\linewidth]{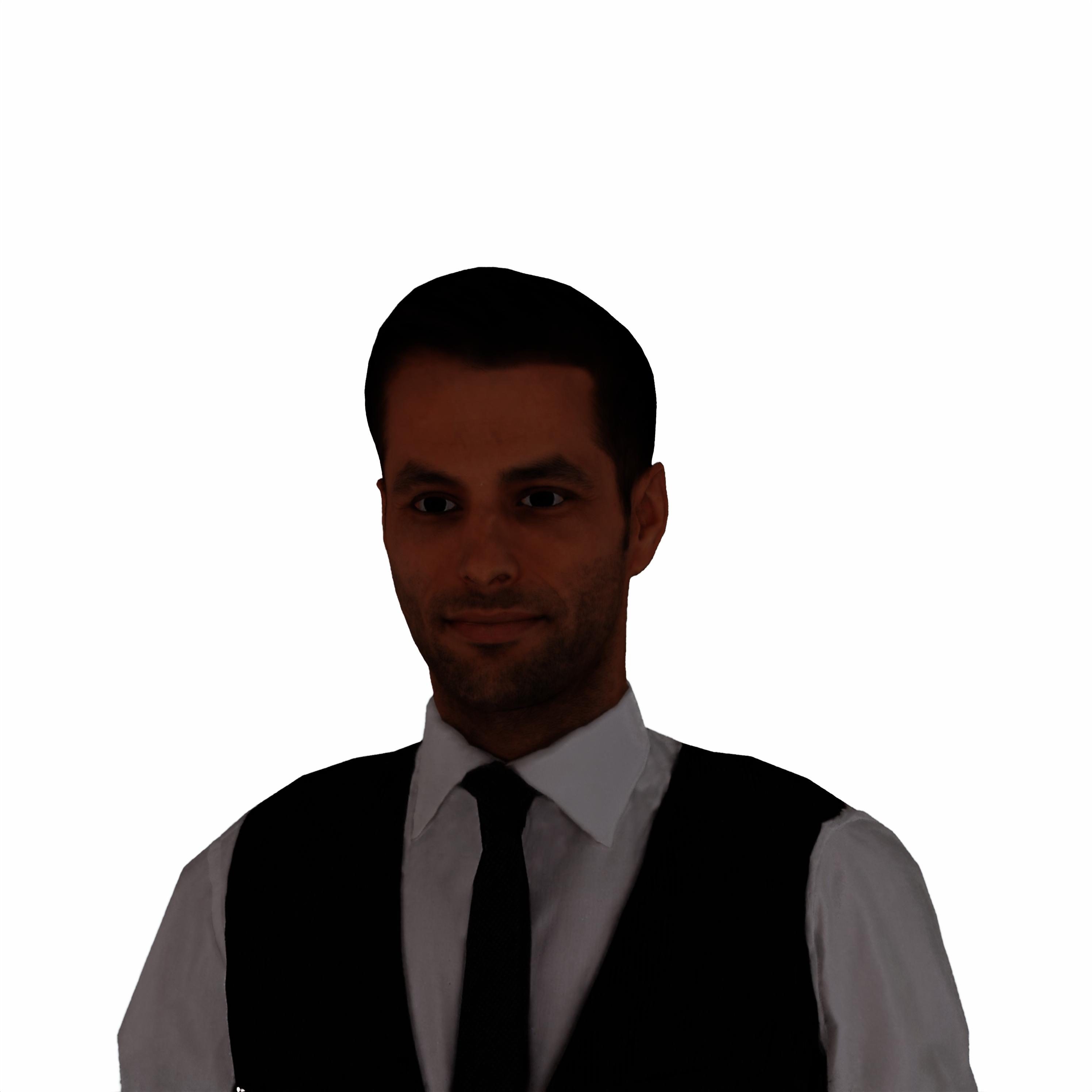}
    \hfill
    \adjincludegraphics[trim={0 0 0 0},clip,width=.19\linewidth]{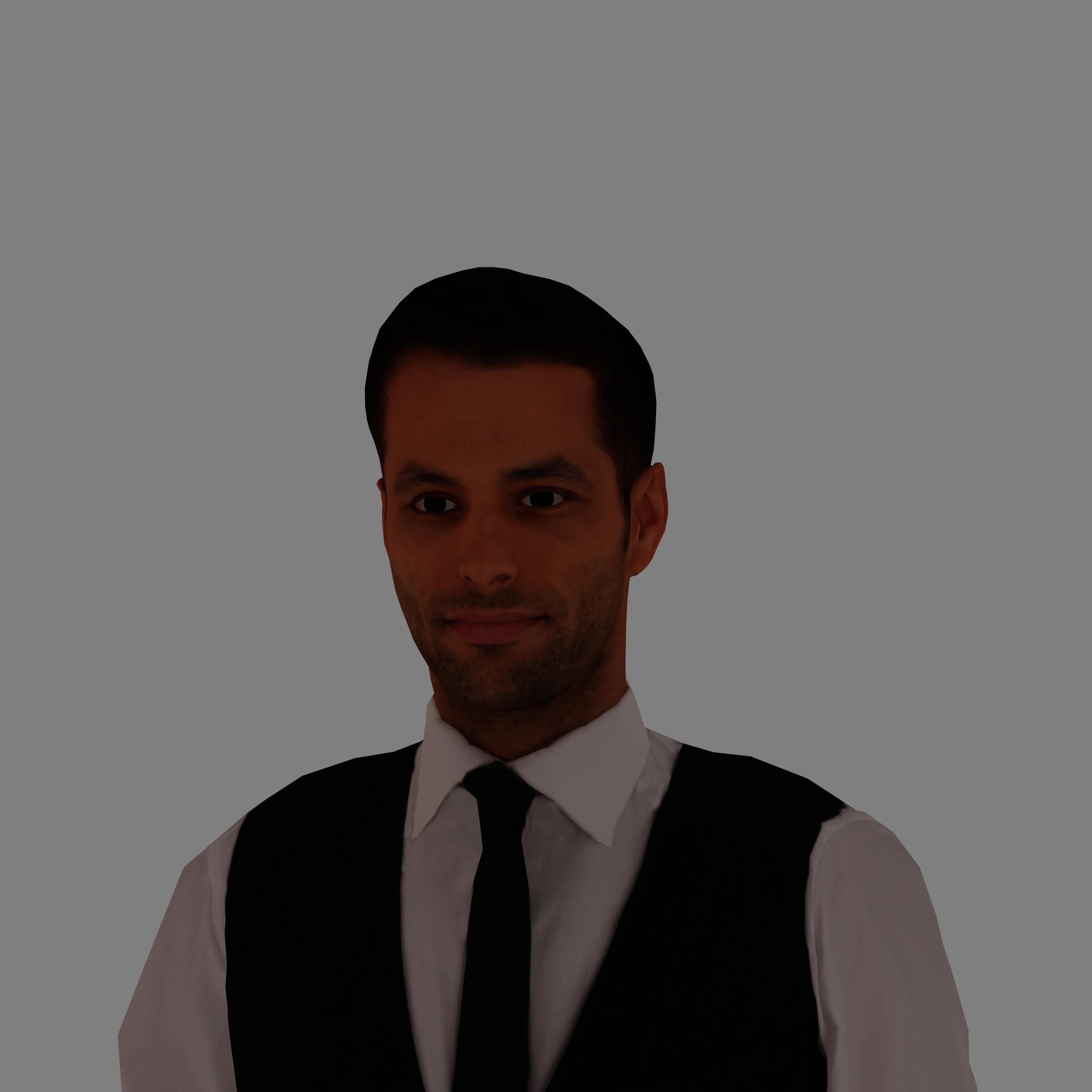}
    \hfill
    \adjincludegraphics[trim={0 0 0 0},clip,width=.19\linewidth]{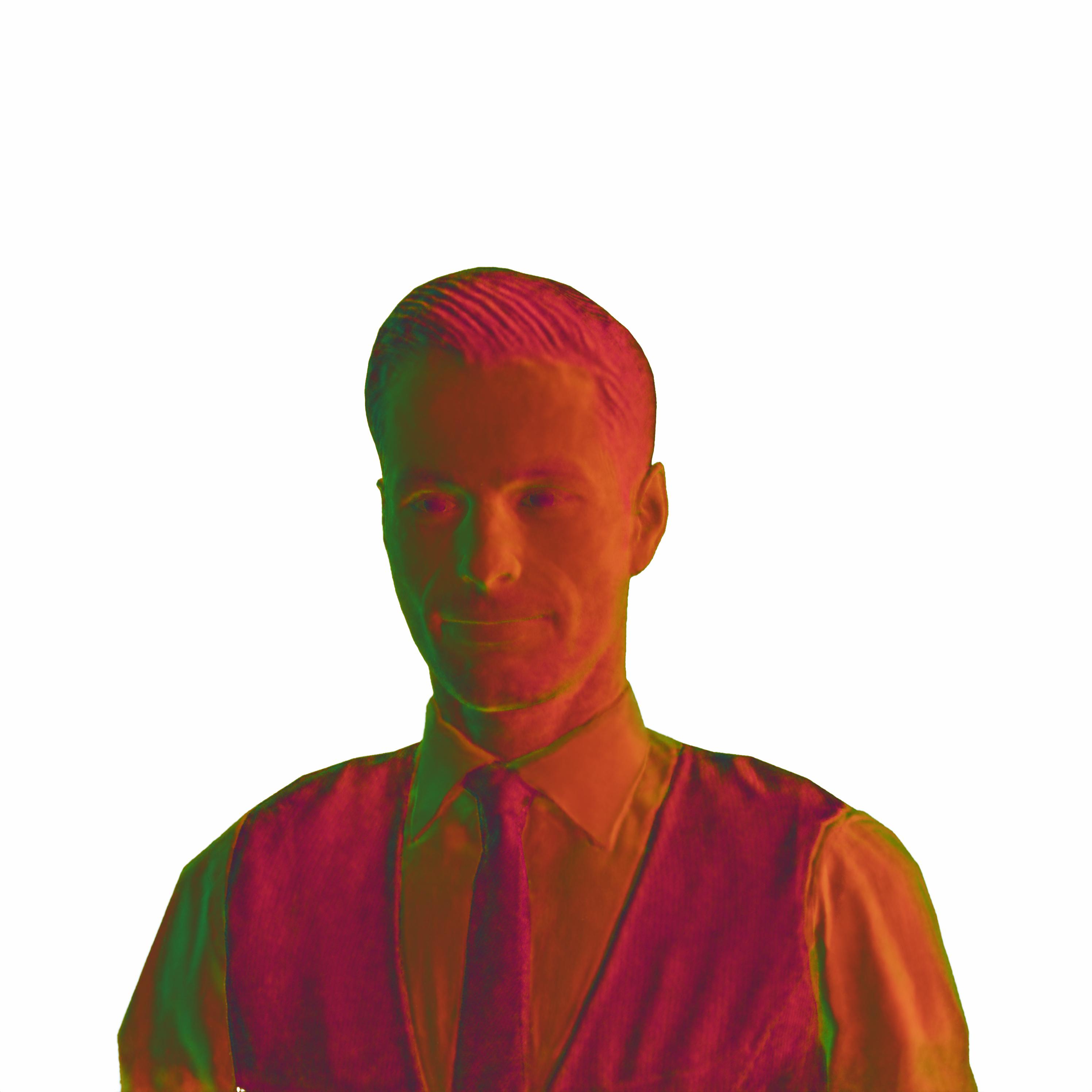}
    \hfill
    \adjincludegraphics[trim={0 0 0 0},clip,width=.19\linewidth]{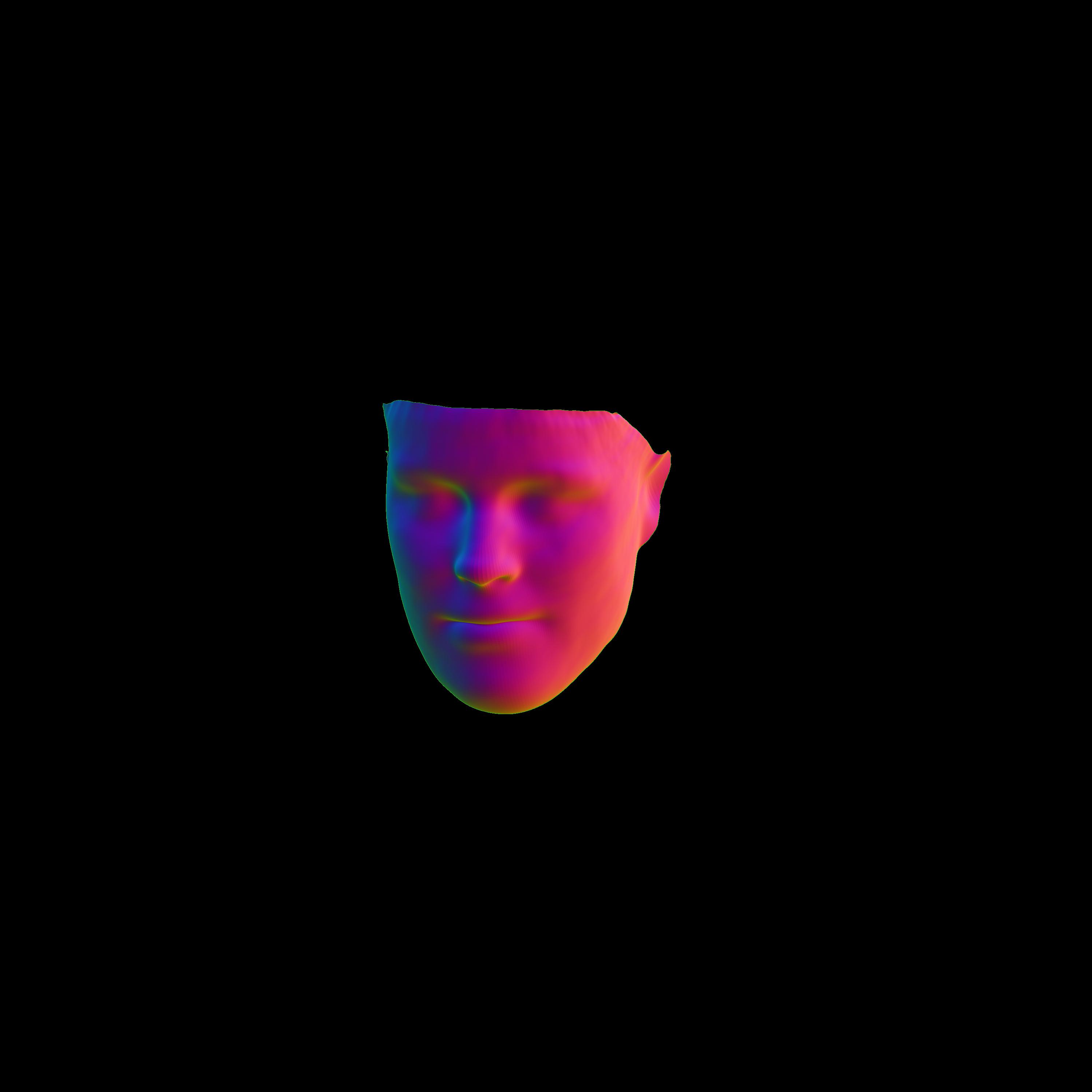}
    \hfill
    \adjincludegraphics[trim={0 0 0 0},clip,width=.19\linewidth]{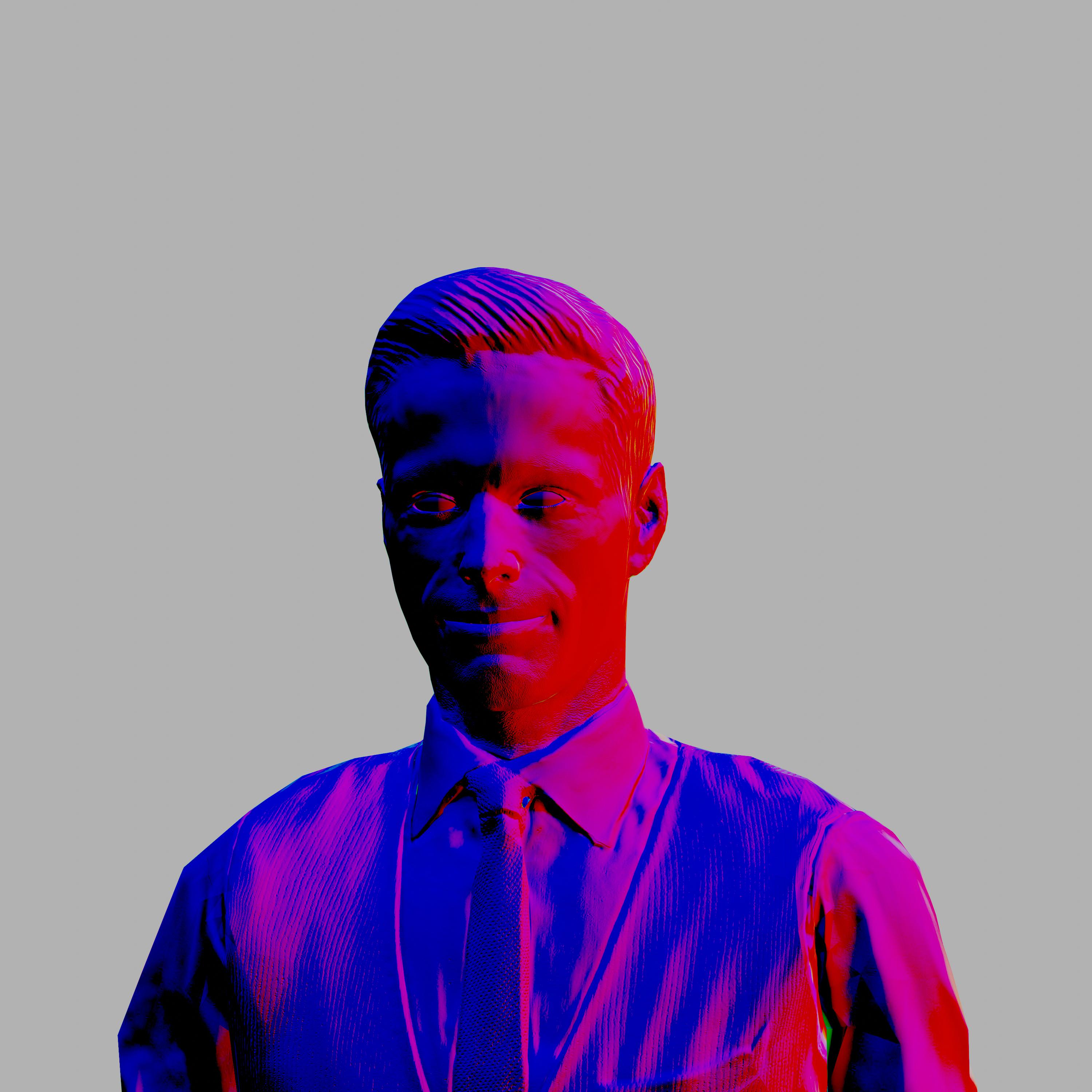}
    
    \caption{The comparison of predicted albedo rendering $A$ and normal rendering $N$ with the ground truth \textit{(g.t.)} albedo and normals, respectively, rendered from the textured meshes in Blender. Additionally, we include the normals for the facial region obtained from face meshes predicted by PRNet~\cite{Feng18}. Since normals from PRNet are the only source of direct normals supervision, their difference from the ground truth ones partially explains the systematic difference between the predicted and ground truth normals. The network is capable of extrapolating the learned normals from the face region to the head and upper body.}
    \label{fig:albedo_normals_compn}
\end{figure*}

Fig.~\ref{fig:albedo_normals_compn} depicts the difference between the predicted albedo and normals and the ground truth ones. In addition, the rendered normals for the facial part obtained from PRNet~\cite{Feng18} meshes are presented. Normals from PRNet are the only source of direct supervision for normals during the pipeline training. They are predicted by PRNet with a certain error (especially for the side pixels at each view), which partially explains the general difference between the predicted normals and the ground truth normals. This effect can also be attributed to the non-uniqueness of normals, required to fit the flashlighted training images, which might happen due to the limited supervision and co-located lighting used during training. Note, however, that the network is able to propagate the normals learned for the face region to the head and upper body (to a certain extent).

\begin{table}[]
\centering
    \begin{tabular}{l|c|c|c}
                        & \multicolumn{3}{c}{\textit{Person 1 (Carla)}} \\
                        & VGG $\downarrow$                   & FID $\downarrow$                   & LPIPS $\downarrow$ \\ \hline
        % Ours (Albedo)       &                       &                       &  \multicolumn{1}{c|}{}   \\ \hline
        % Ours (Normals)      &                       &                       & \multicolumn{1}{c|}{}   \\ \hline \hline
        Ours (Final result) & \textbf{174.59}  & \textbf{61.664}  & \textbf{0.0815} \\ \hline
        DPR~\cite{Zhou19} (Final result)  & 305.96  & 69.263 & 0.129    \\
        \multicolumn{3}{}{} \\
                        & \multicolumn{3}{c}{\textit{Person 2 (Claudia)}} \\
                        & VGG $\downarrow$                   & FID $\downarrow$                   & LPIPS $\downarrow$ \\ \hline
        % Ours (Albedo)       &                       &                       &  \multicolumn{1}{c|}{}   \\ \hline
        % Ours (Normals)      &                       &                       & \multicolumn{1}{c|}{}   \\ \hline \hline
        Ours (Final result) & \textbf{286.94}  & \textbf{58.713} & \textbf{0.1380} \\ \hline
        DPR~\cite{Zhou19} (Final result)  & 331.88 & 59.791 & 0.1441   \\
        \multicolumn{3}{}{} \\
                        & \multicolumn{3}{c}{\textit{Person 3 (Eric)}} \\
                        & VGG $\downarrow$                   & FID $\downarrow$                   & LPIPS $\downarrow$ \\ \hline
        % Ours (Albedo)       &                       &                       &  \multicolumn{1}{c|}{}   \\ \hline
        % Ours (Normals)      &                       &                       & \multicolumn{1}{c|}{}   \\ \hline \hline
        Ours (Final result) & \textbf{259.14}  & 51.158 & \textbf{0.1259} \\ \hline
        DPR~\cite{Zhou19} (Final result)  & 335.07 & \textbf{31.467}  & 0.1418   \\  
    \end{tabular}
    \newline
    \caption{Quantitative comparison of the relighted images for 3 synthetic people from RenderPeople dataset. For each of the people, the final result is the image relighted by a combination of ambient light and $1^\mathrm{st}$-order Spherical Harmonics (SH). In this setting, the comparison with the DPR method~\cite{Zhou19} is possible. The ground truth was constructed from albedo and normals rendered from mesh in Blender. To report the results, we use three perceptual metrics evaluated on images for ten validation viewpoints corresponding to the non-flashlighted images from the validation. The result is averaged across all of these viewpoints and three ambient+spherical harmonics lightings (with first-order spherical harmonics simulating \textit{left-to-right}, \textit{up-to-down}, \textit{forward-to-backward} directional light). The predictions under these lighting conditions are presented in Fig.~\ref{fig:synth_relighting}.}
    \label{tab:synth_comparison}
\end{table}

\subsection{Ablation study}

Fig.~\ref{fig:real_ablation} reflects the comparison with several ablations conducted on a real head portrait, created by: removing single loss term, excluding the point cloud filtering stage from the preprocessing pipeline, or leaving only VGG term in $\Delta$ mismatch. Note the difference in normals reconstruction without $L_\mathrm{normal}$ (in this case, the network predicts normals closer to average due to the lack of geometric information) and with only VGG term in $\Delta$ (results in worse color matching of normals). The albedo becomes more blurry without $L_\mathrm{TV}$ (zoom-in is recommended), while the shadows are captured less accurately without $L_\mathrm{symm}$.

% L(&\phi, \D, C^\mathrm{room}, C^\mathrm{flash}, \mathcal{T}_A) = L_\mathrm{final}(\phi, \D, C^\mathrm{room}, C^\mathrm{flash})  + \alpha_\mathrm{normal} L_\mathrm{normal}(\phi, \D) + \alpha_\mathrm{symm} L_\mathrm{symm}(\phi, \D, \mathcal{T}_A) + F \cdot \alpha_\mathrm{cm} L_\mathrm{cm}(\phi, \D) + \alpha_\mathrm{TV} L_\mathrm{TV}(\phi, \D) + \alpha_\mathrm{mask} L_\mathrm{mask}(\phi, \D)%

\begin{table*}
    \centering
    \setlength\tabcolsep{1pt}
    \begin{tabular}{cccccccc}
        &
        Ours &
        w/o $L_\mathrm{normal}$ &
        w/o $L_\mathrm{symm}$ &
        w/o $L_\mathrm{cm}$ &
        w/o $L_\mathrm{TV}$ &
        w/o cloud filtering &
        only VGG in $\Delta$ \\
        
        \multirow{1}{*}{\rotatebox{90}{Rendered $\mathcal{I}$}} &
        \raisebox{-0.5\height}{\adjincludegraphics[width=.14\textwidth,clip,trim={0 {.4\height} 0 0}]{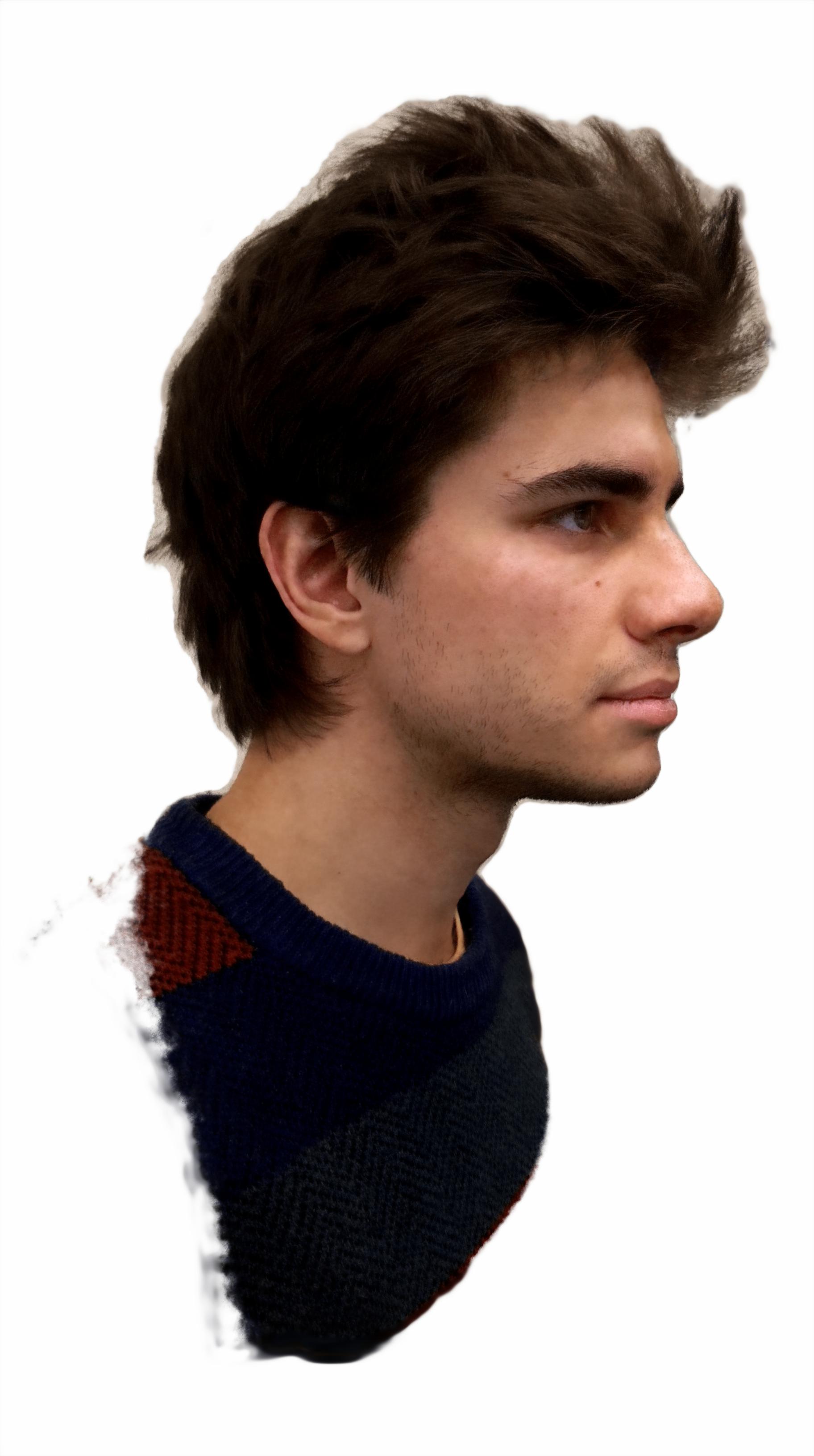}} & 
        \raisebox{-0.5\height}{\adjincludegraphics[clip,trim={0 {.4\height} 0 0},width=.14\textwidth]{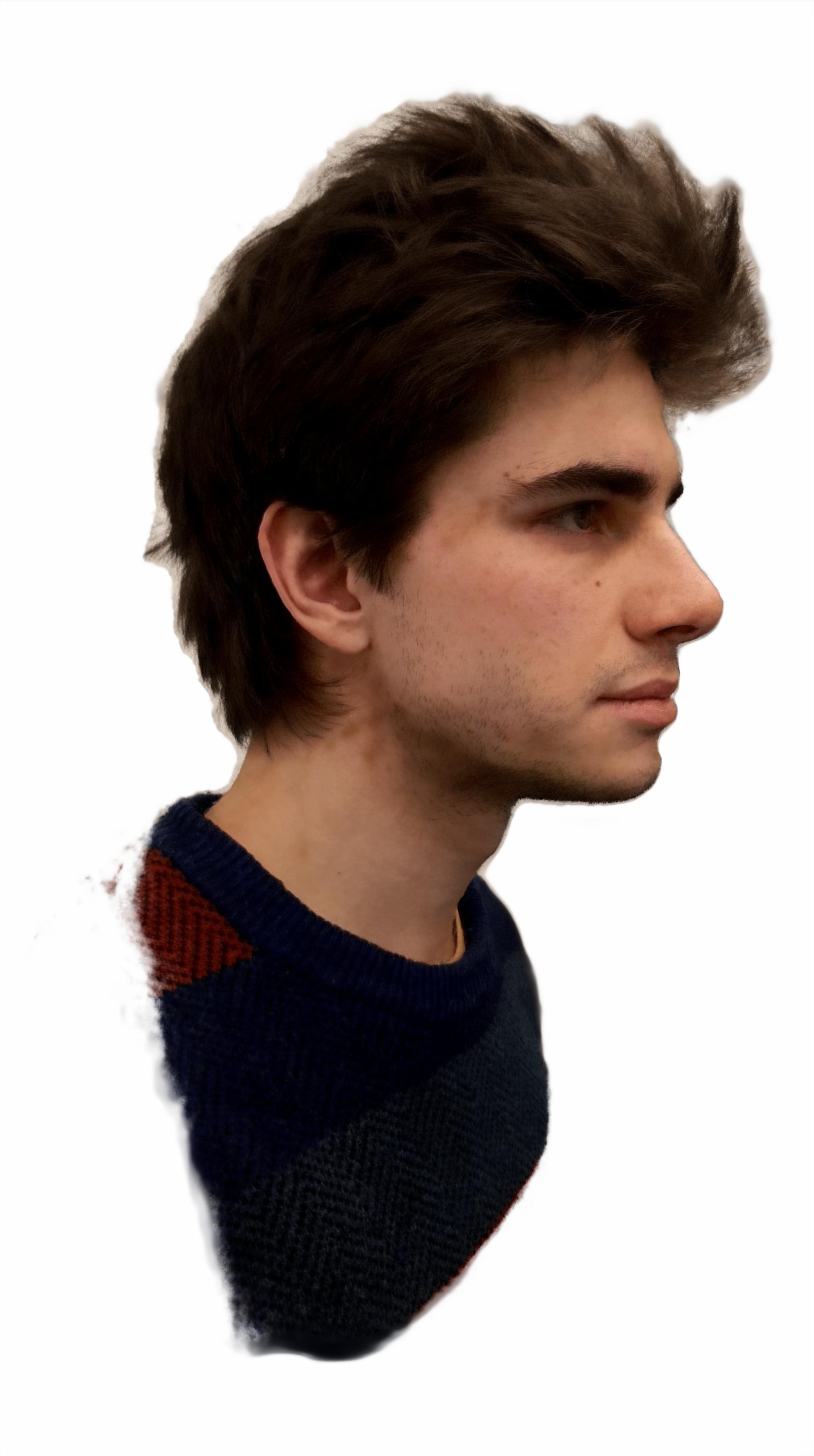}} &
        \raisebox{-0.5\height}{\adjincludegraphics[clip,trim={0 {.4\height} 0 0},width=.14\textwidth]{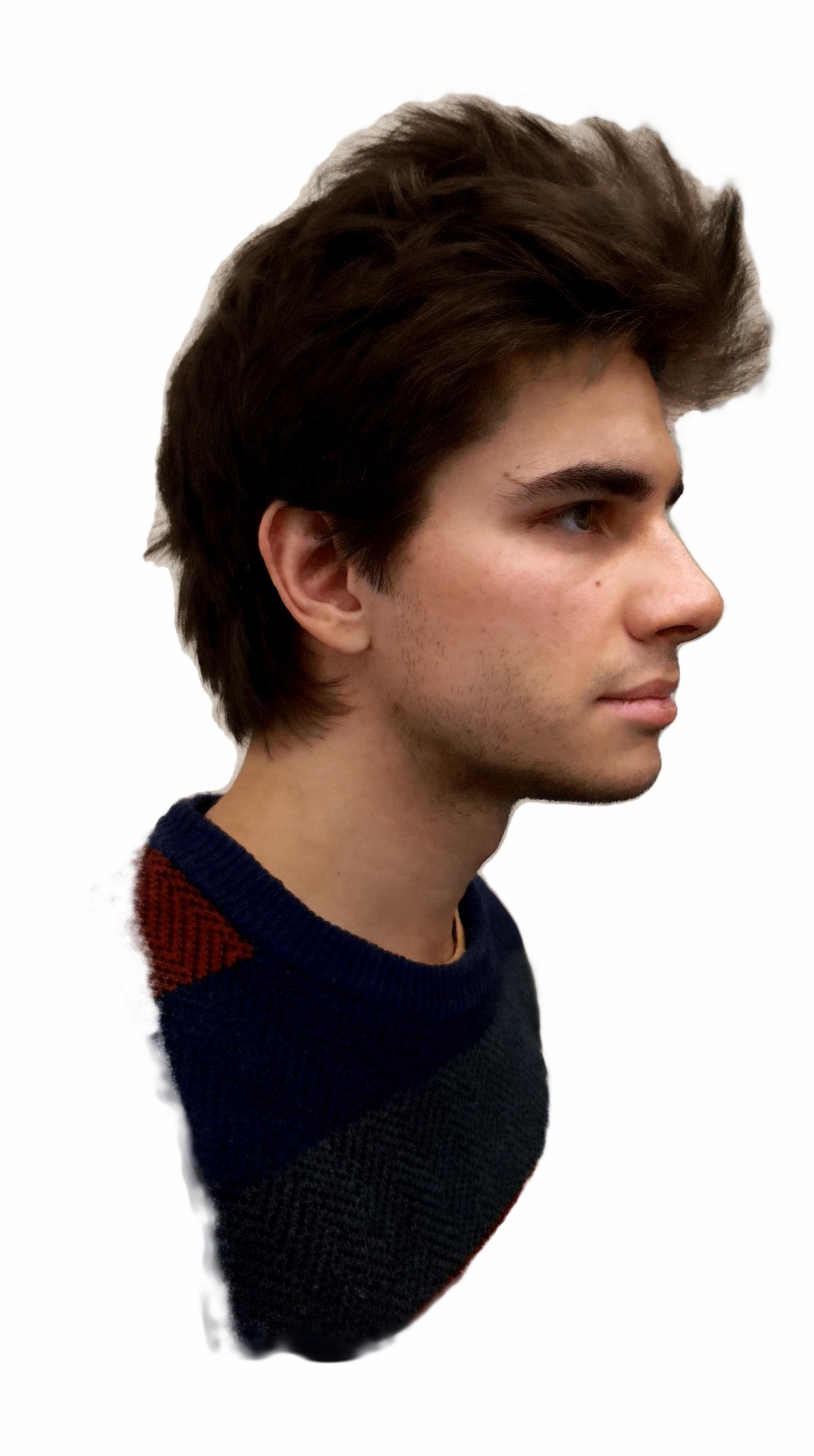}} &
        \raisebox{-0.5\height}{\adjincludegraphics[clip,trim={0 {.4\height} 0 0},width=.14\textwidth]{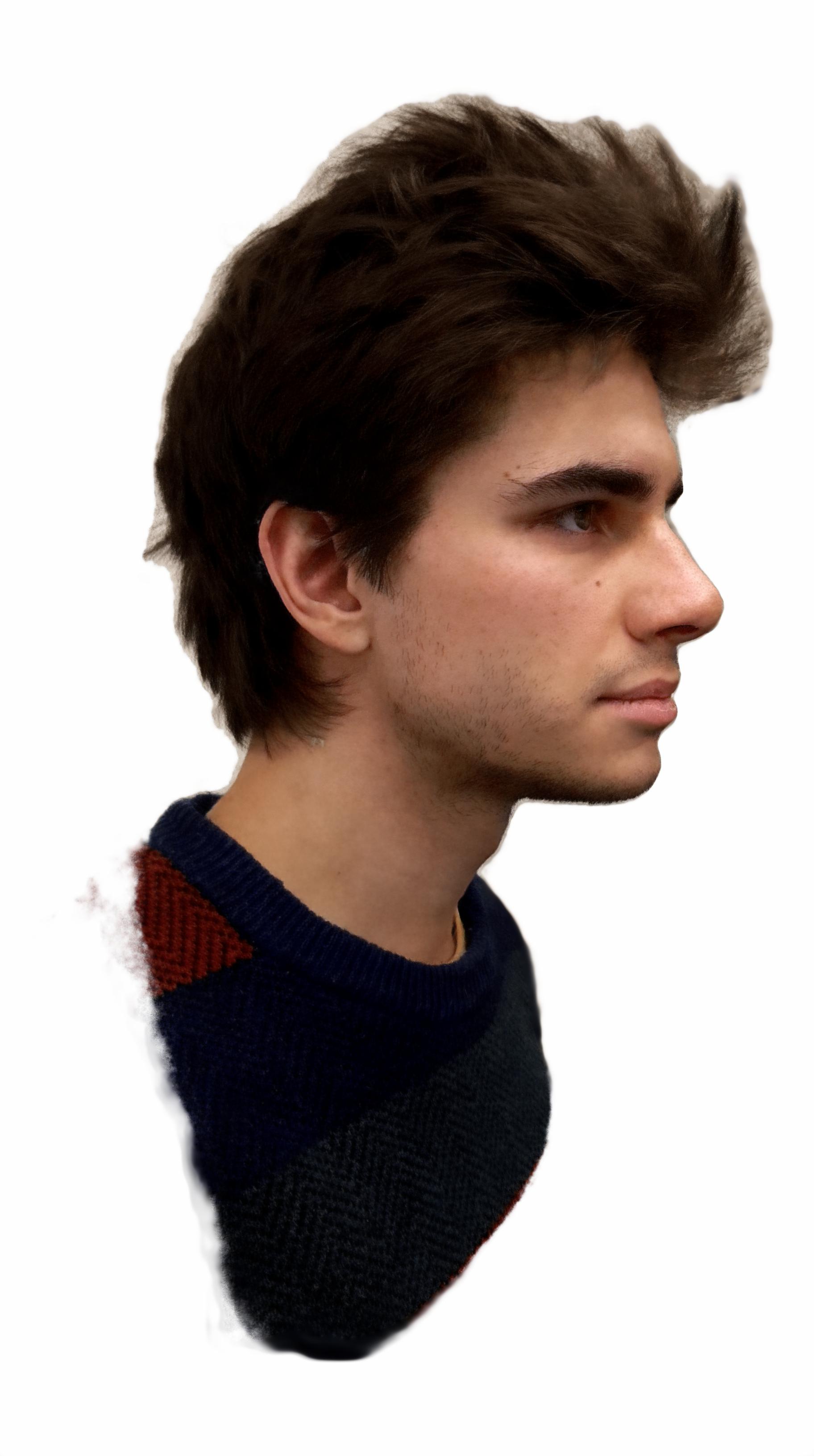}} &
        \raisebox{-0.5\height}{\adjincludegraphics[clip,trim={0 {.4\height} 0 0},width=.14\textwidth]{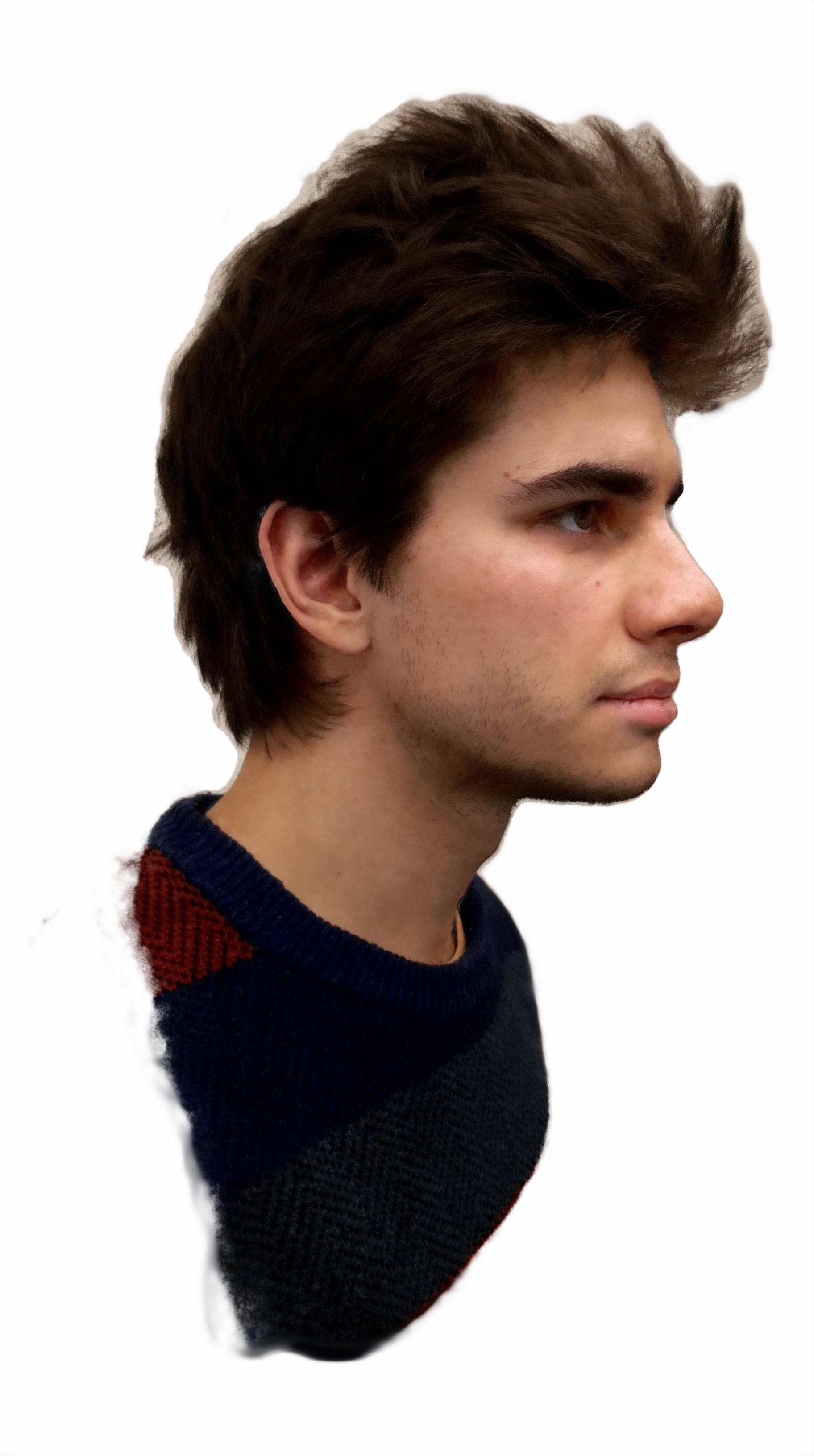}} &
        \raisebox{-0.5\height}{\adjincludegraphics[clip,trim={0 {.4\height} 0 0},width=.14\textwidth]{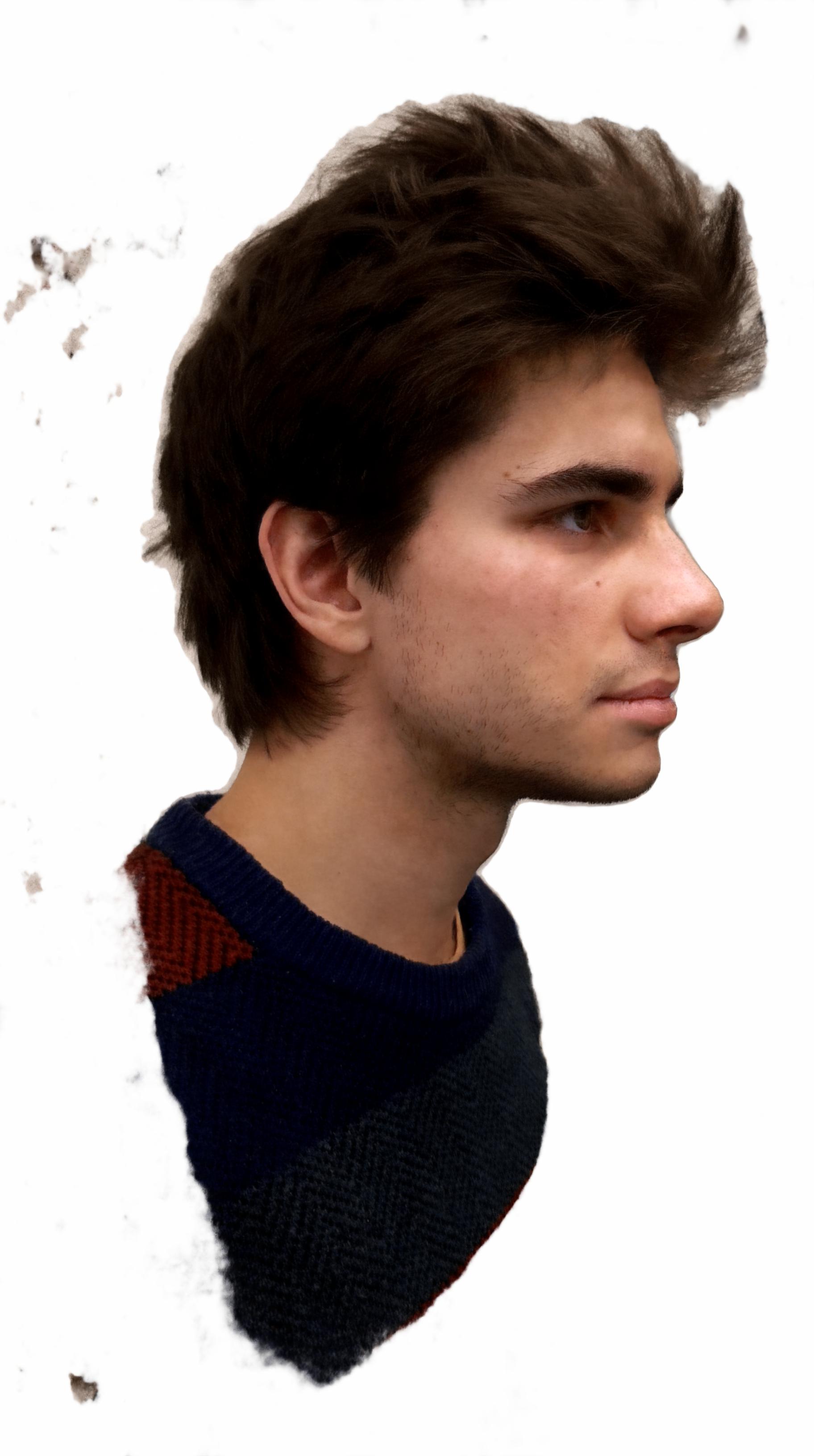}} &
        \raisebox{-0.5\height}{\adjincludegraphics[clip,trim={0 {.4\height} 0 0},width=.14\textwidth]{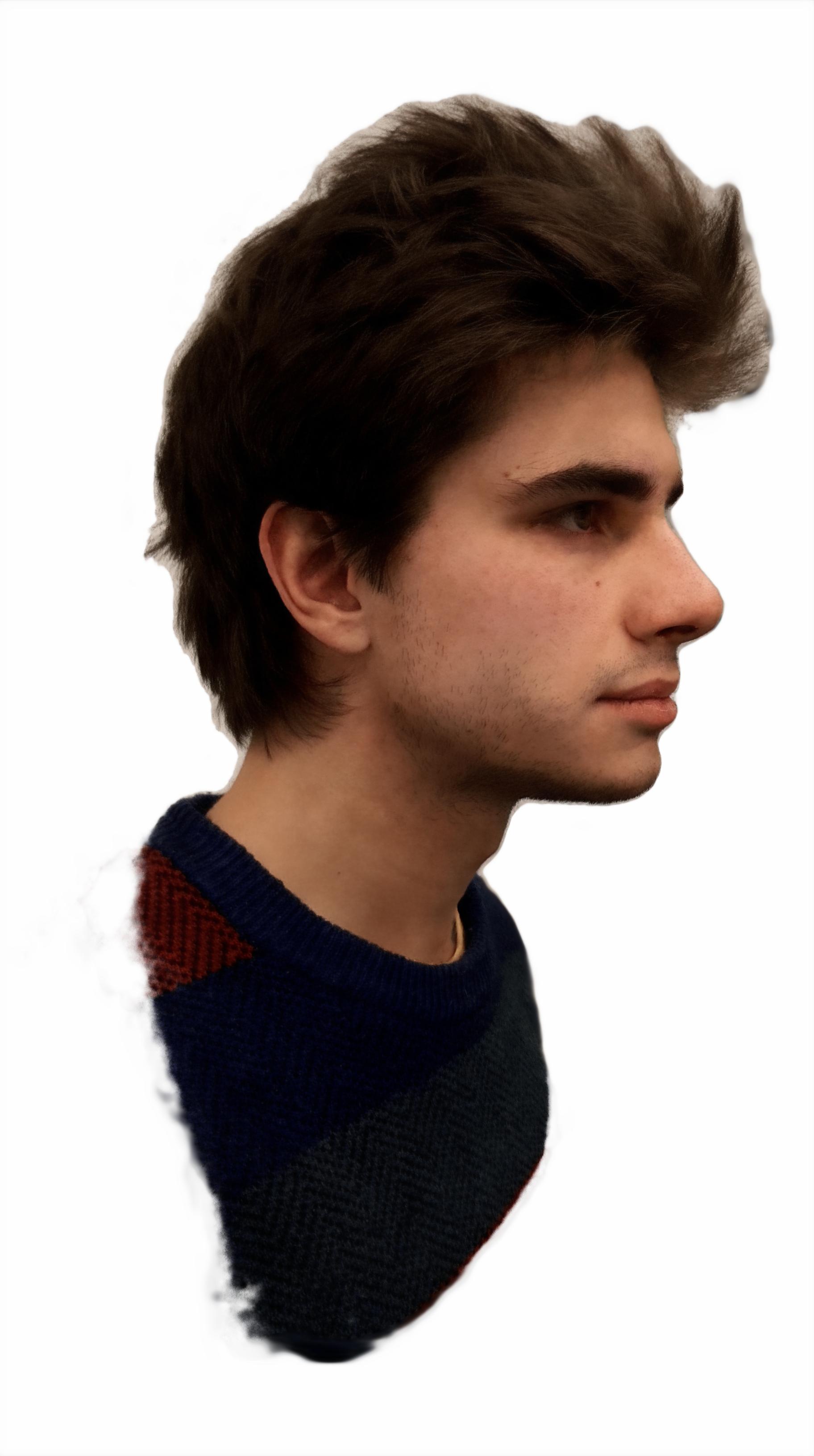}} \\
        
        \multirow{1}{*}{\rotatebox{90}{Albedo $A$}} &
        \raisebox{-0.5\height}{\adjincludegraphics[clip,trim={0 {.4\height} 0 0},width=.14\textwidth]{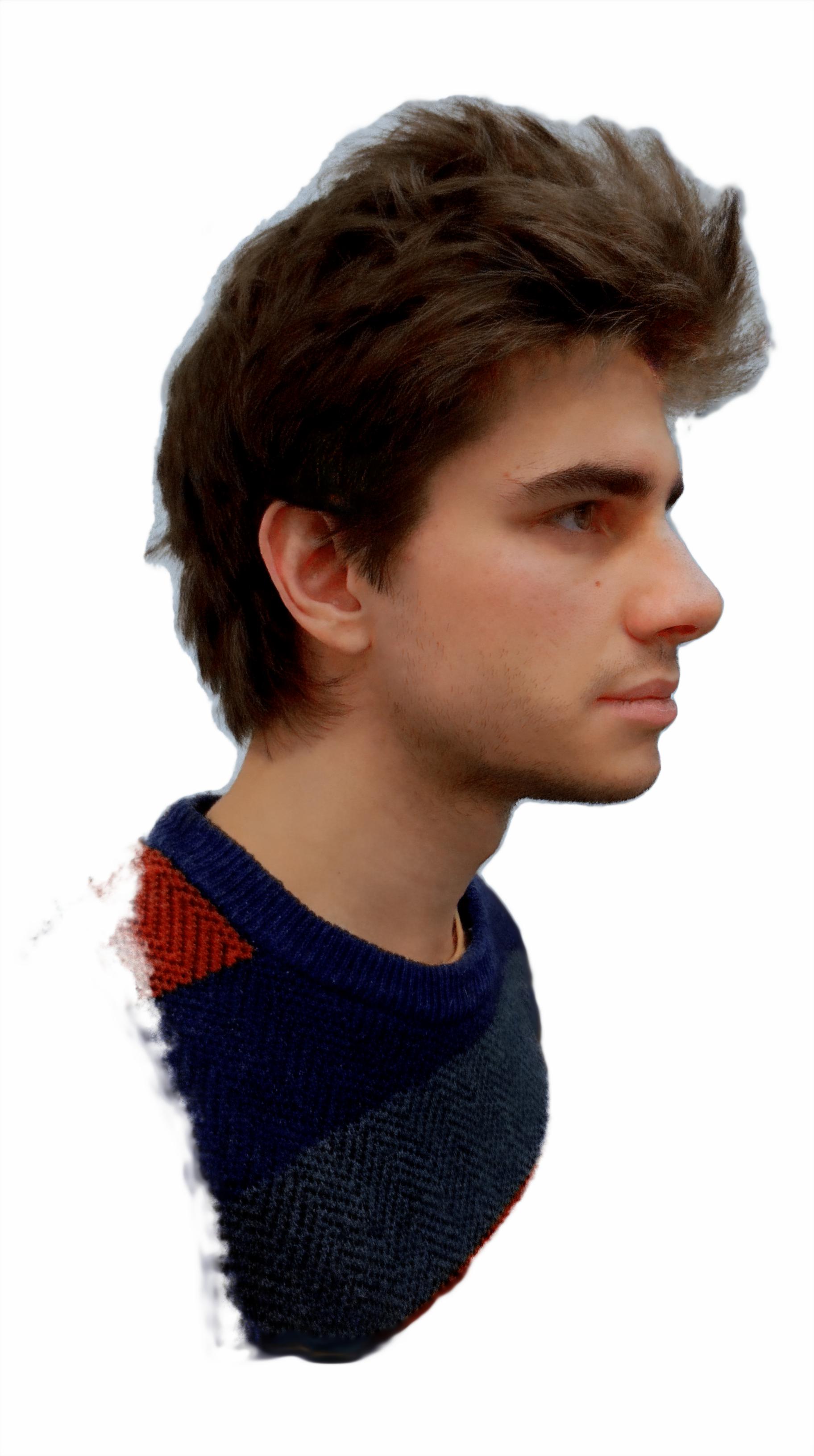}} & 
        \raisebox{-0.5\height}{\adjincludegraphics[clip,trim={0 {.4\height} 0 0},width=.14\textwidth]{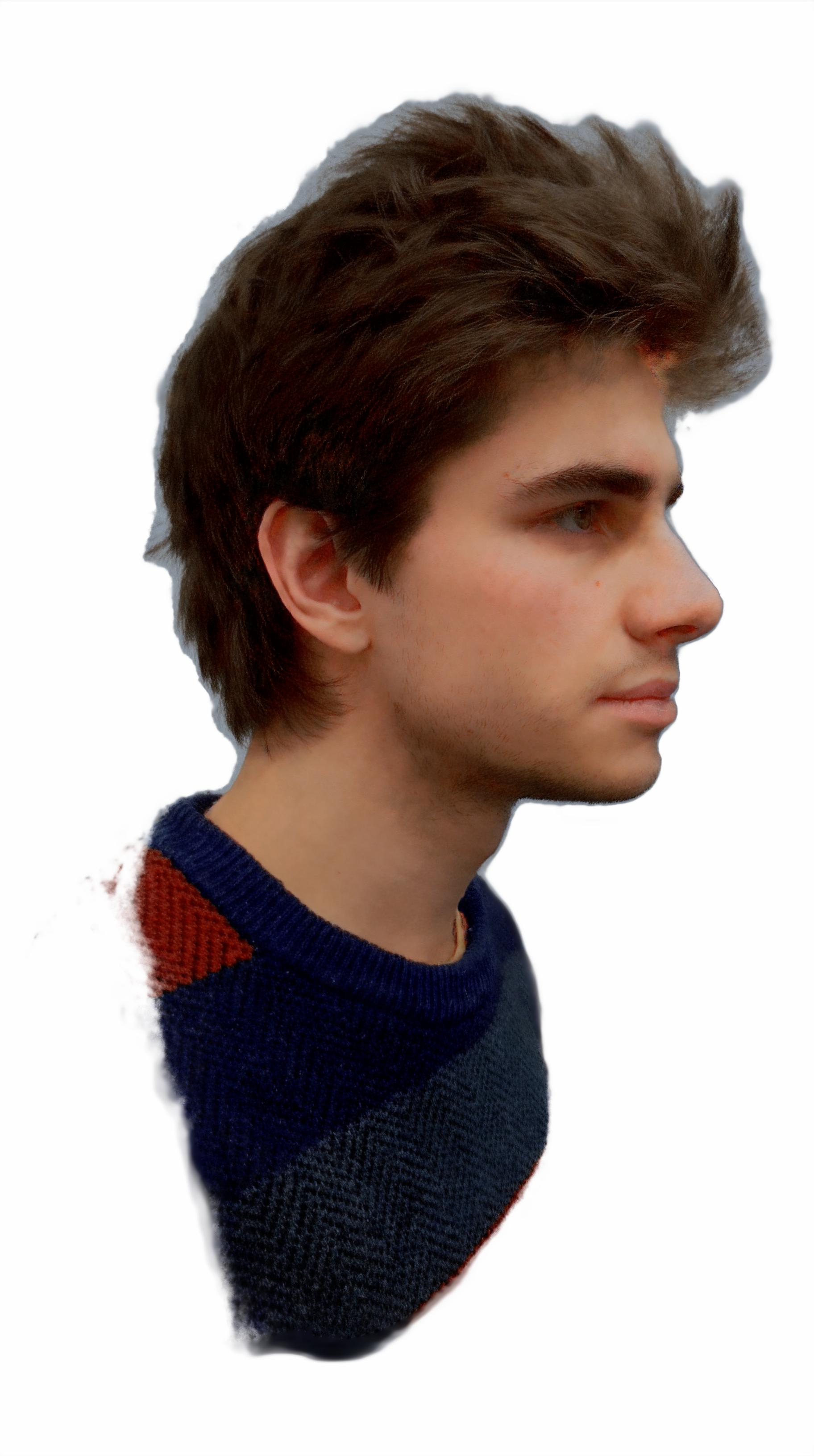}} &
        \raisebox{-0.5\height}{\adjincludegraphics[clip,trim={0 {.4\height} 0 0},width=.14\textwidth]{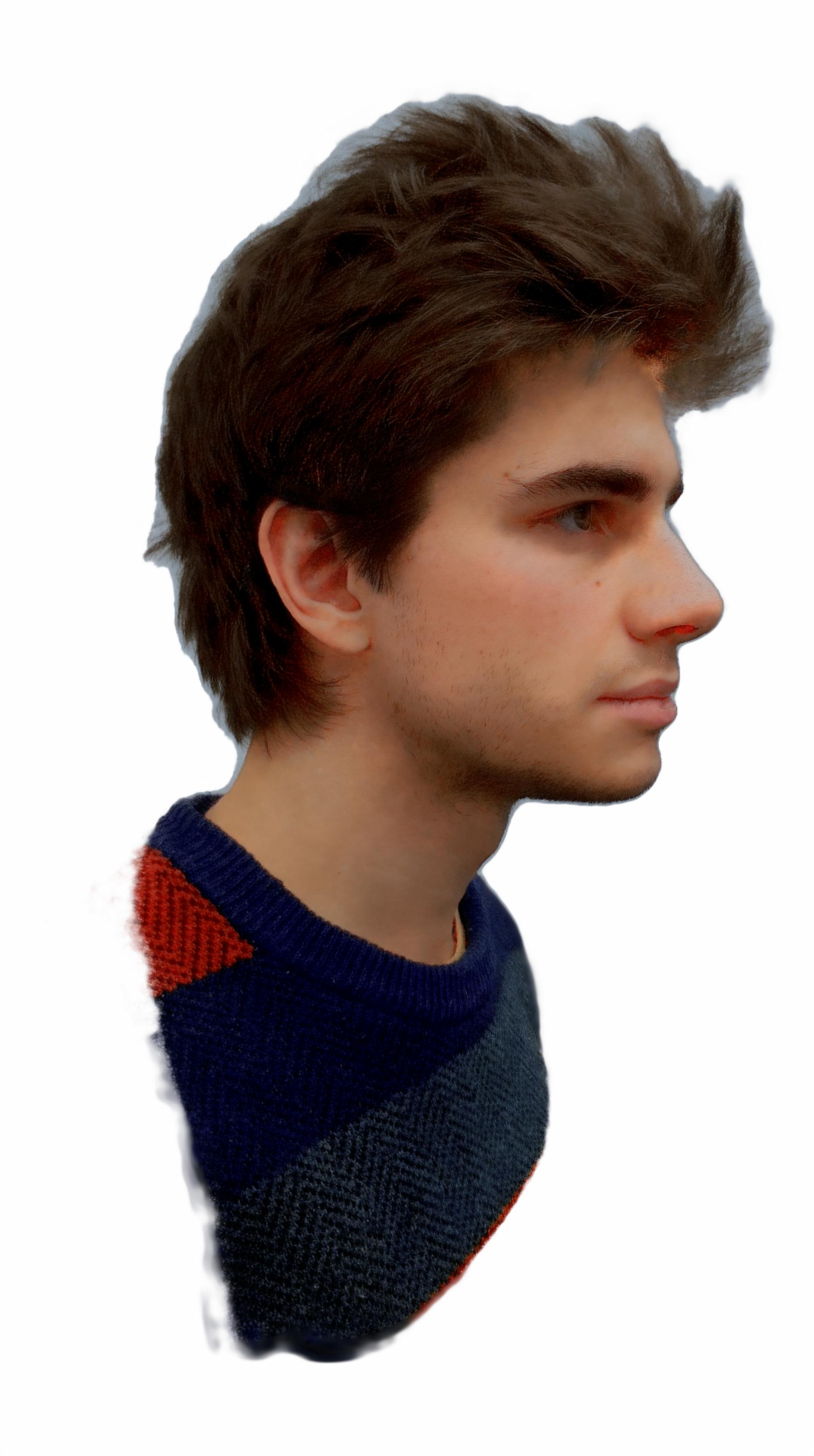}} &
        \raisebox{-0.5\height}{\adjincludegraphics[clip,trim={0 {.4\height} 0 0},width=.14\textwidth]{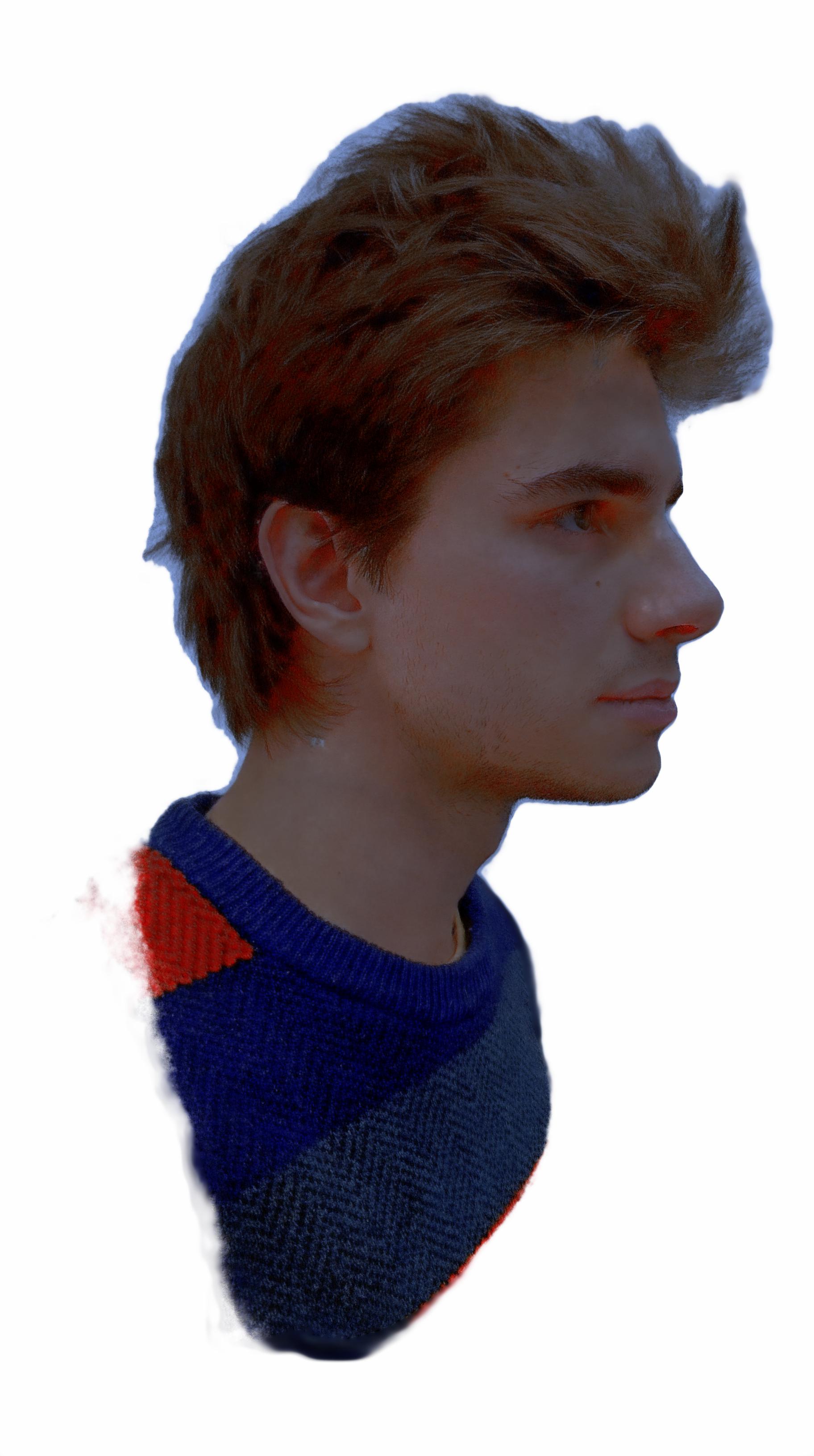}} &
        \raisebox{-0.5\height}{\adjincludegraphics[clip,trim={0 {.4\height} 0 0},width=.14\textwidth]{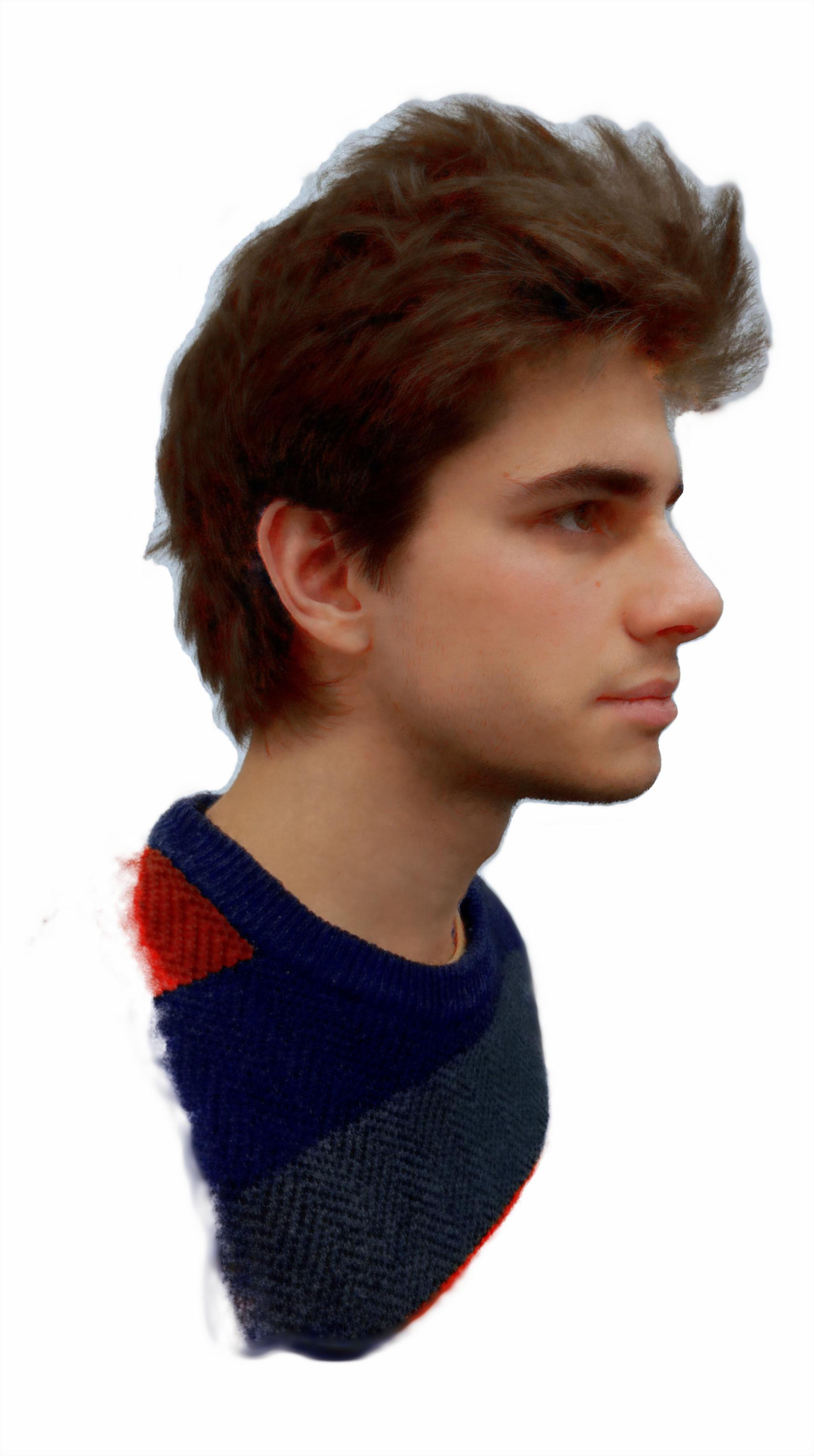}} &
        \raisebox{-0.5\height}{\adjincludegraphics[clip,trim={0 {.4\height} 0 0},width=.14\textwidth]{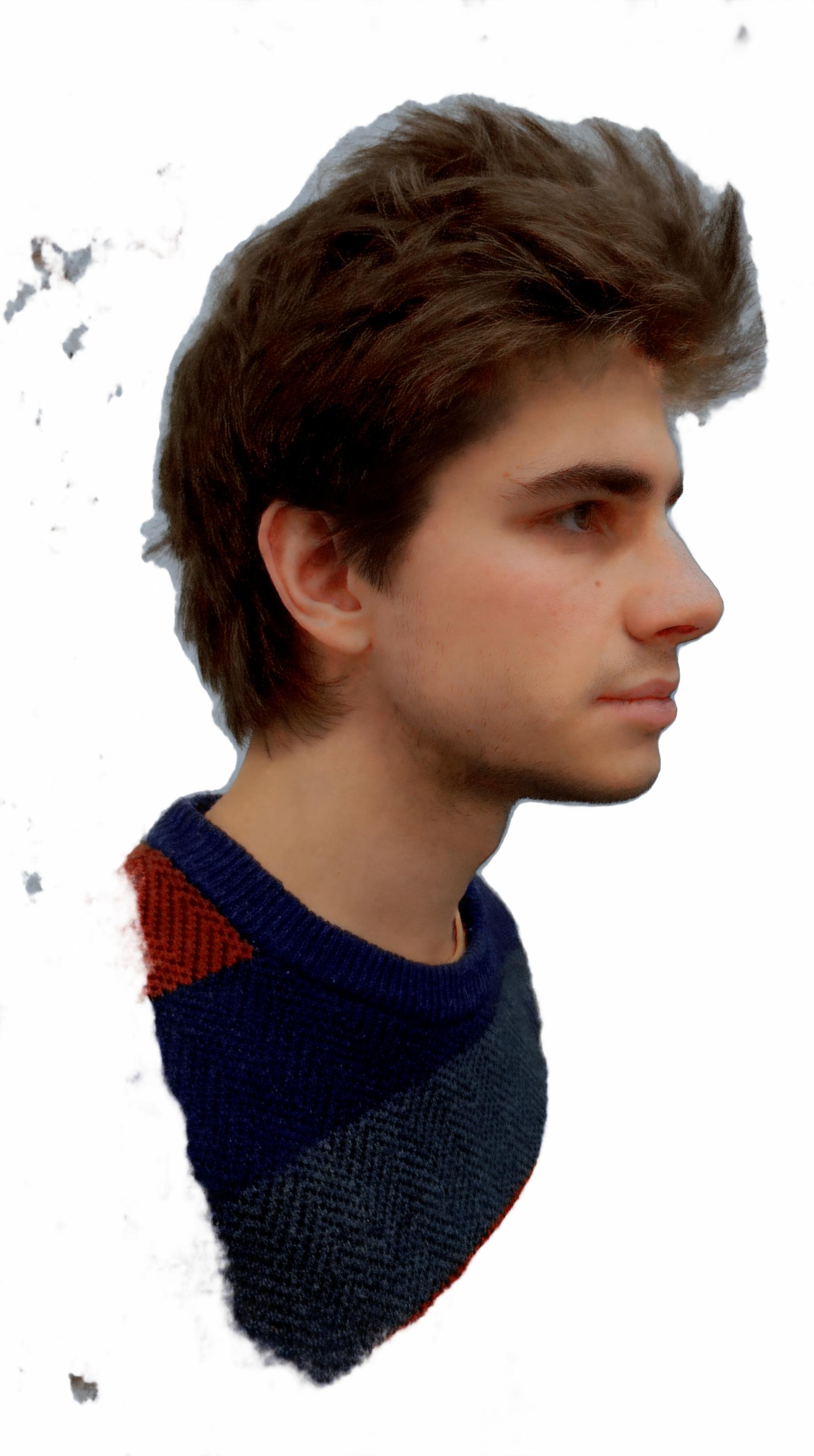}} &
        \raisebox{-0.5\height}{\adjincludegraphics[clip,trim={0 {.4\height} 0 0},width=.14\textwidth]{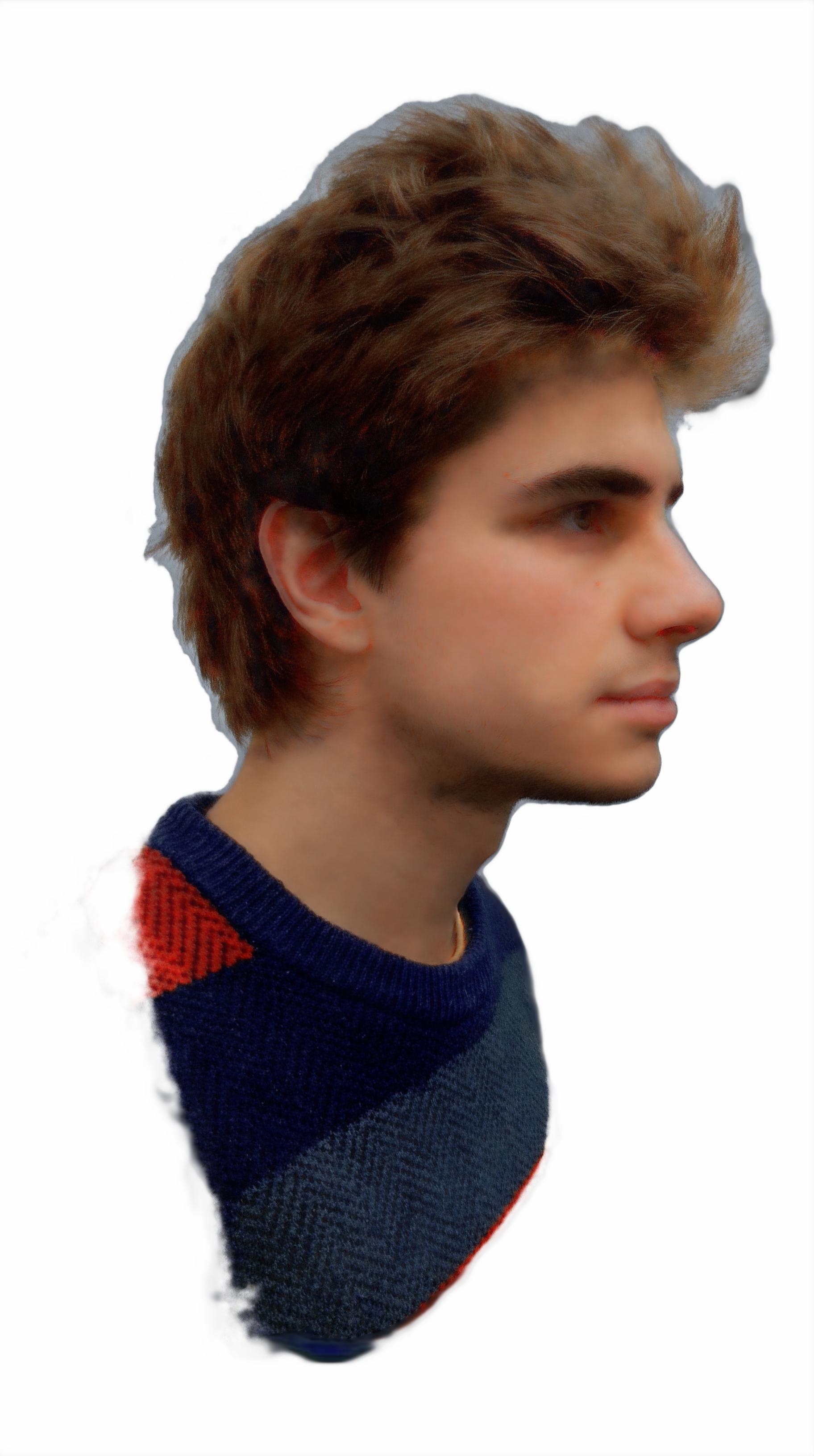}} \\
        
        \multirow{1}{*}{\rotatebox{90}{Normals $N$}} &
        \raisebox{-0.5\height}{\adjincludegraphics[clip,trim={0 {.4\height} 0 0},width=.14\textwidth]{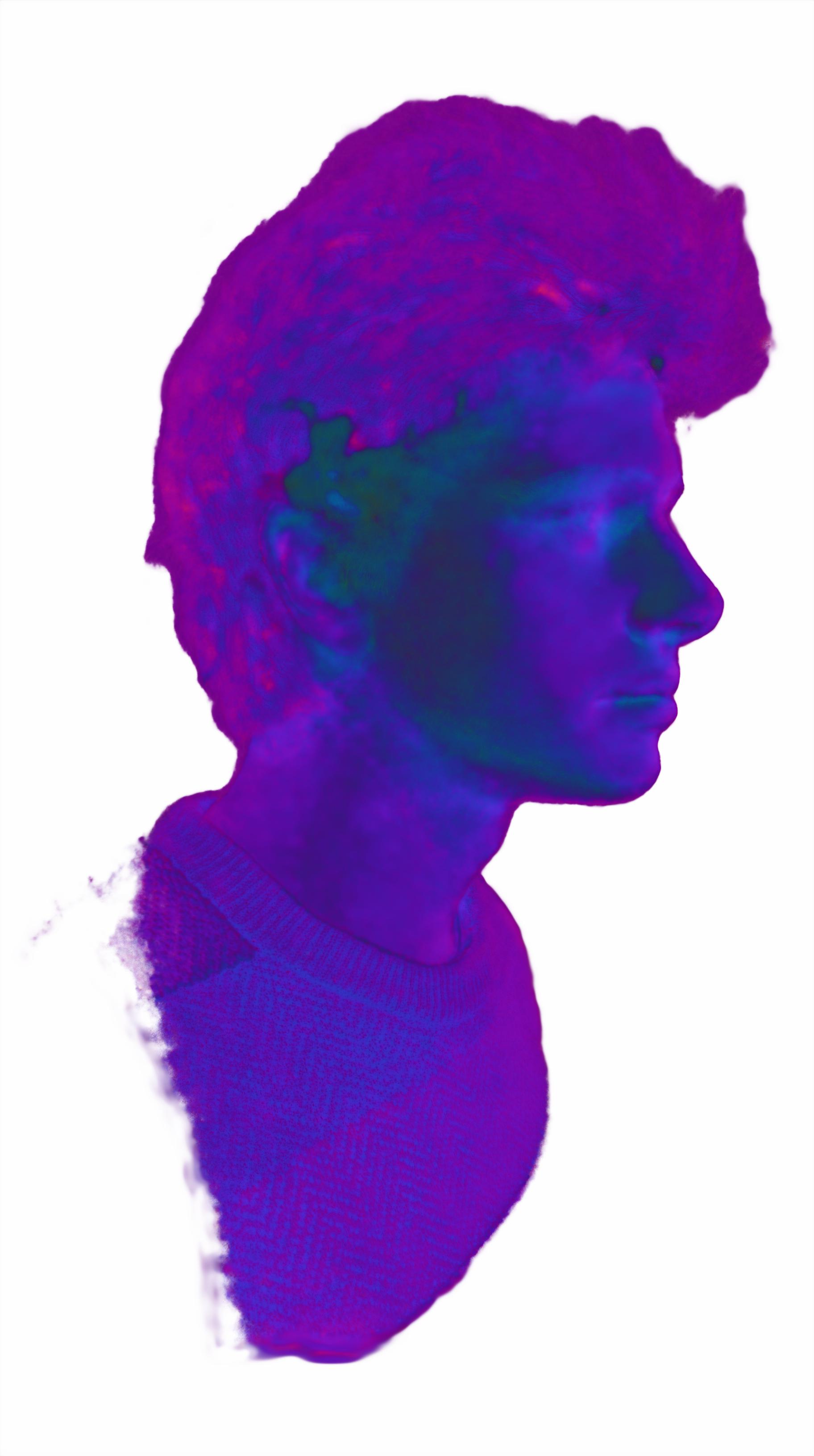}} & 
        \raisebox{-0.5\height}{\adjincludegraphics[clip,trim={0 {.4\height} 0 0},width=.14\textwidth]{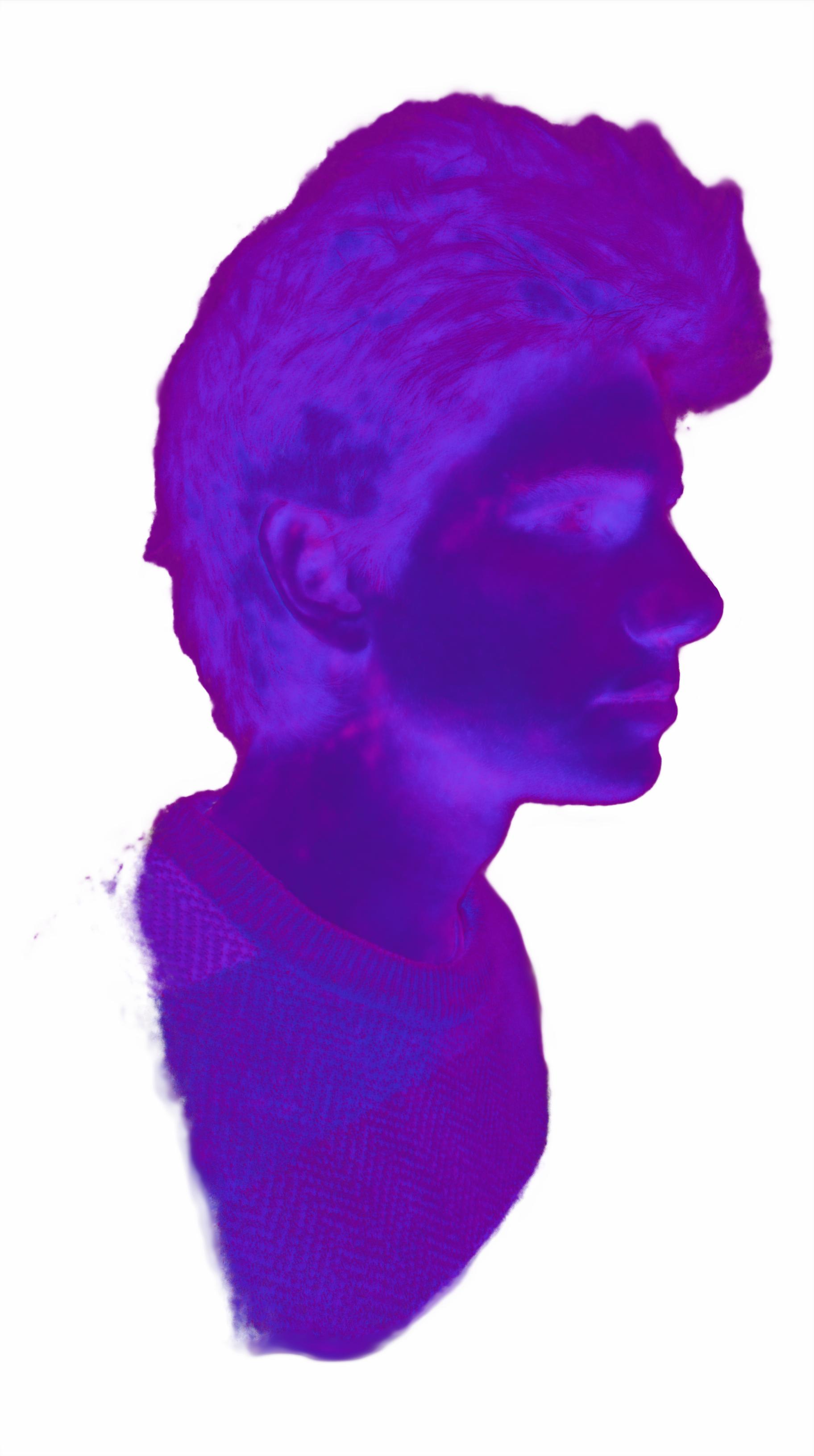}} &
        \raisebox{-0.5\height}{\adjincludegraphics[clip,trim={0 {.4\height} 0 0},width=.14\textwidth]{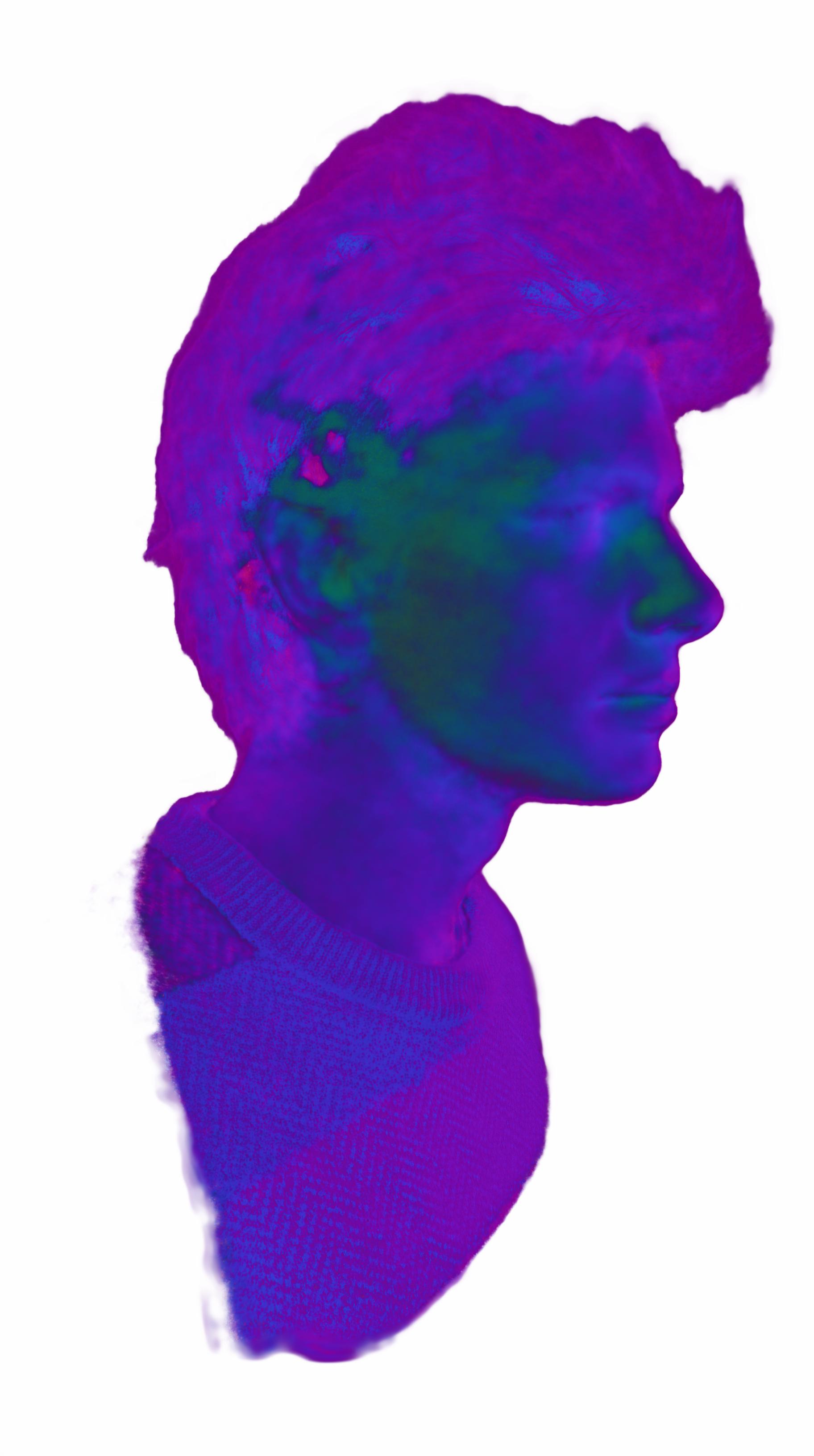}} &
        \raisebox{-0.5\height}{\adjincludegraphics[clip,trim={0 {.4\height} 0 0},width=.14\textwidth]{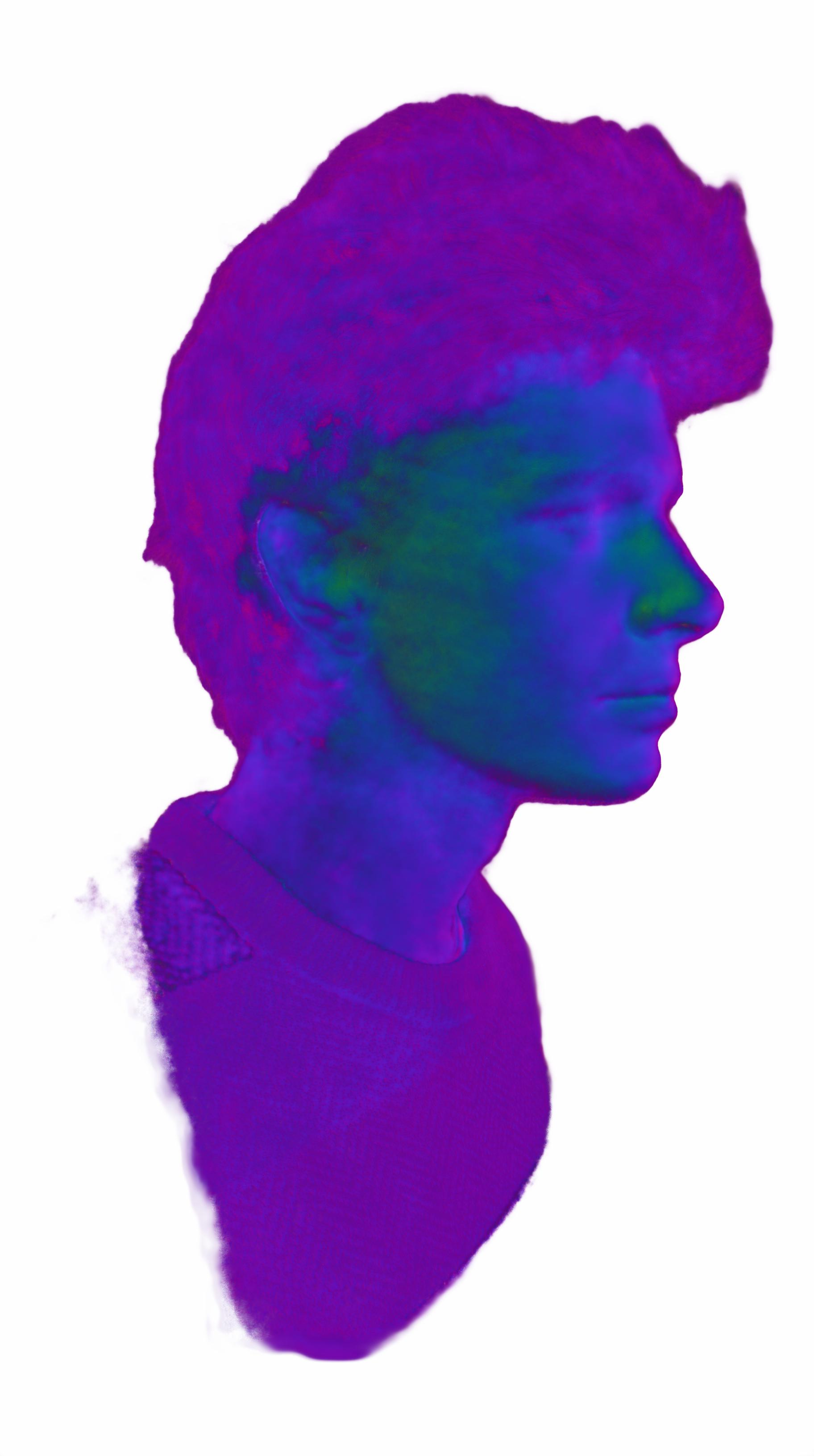}} &
        \raisebox{-0.5\height}{\adjincludegraphics[clip,trim={0 {.4\height} 0 0},width=.14\textwidth]{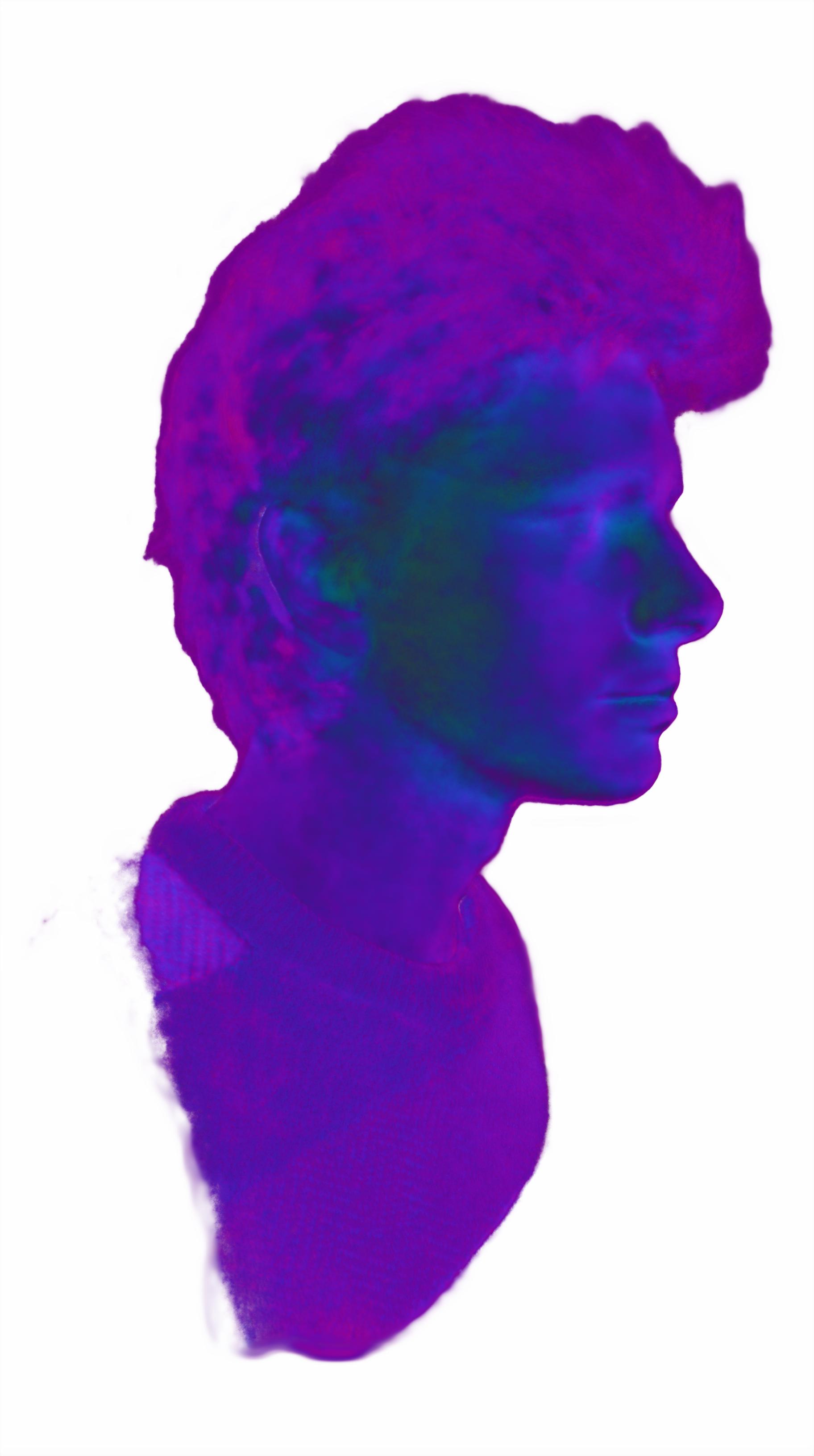}} &
        \raisebox{-0.5\height}{\adjincludegraphics[clip,trim={0 {.4\height} 0 0},width=.14\textwidth]{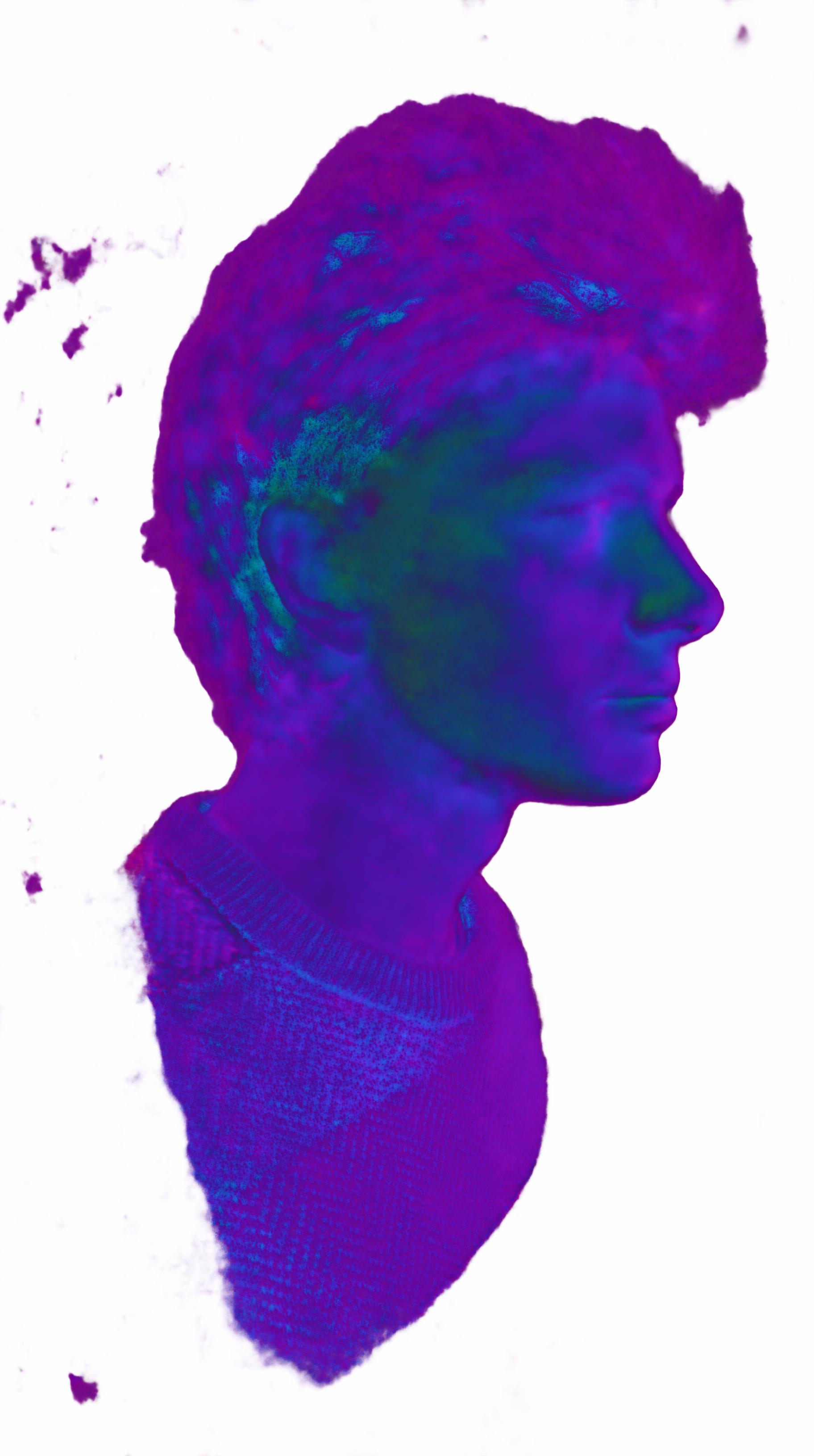}} &
        \raisebox{-0.5\height}{\adjincludegraphics[clip,trim={0 {.4\height} 0 0},width=.14\textwidth]{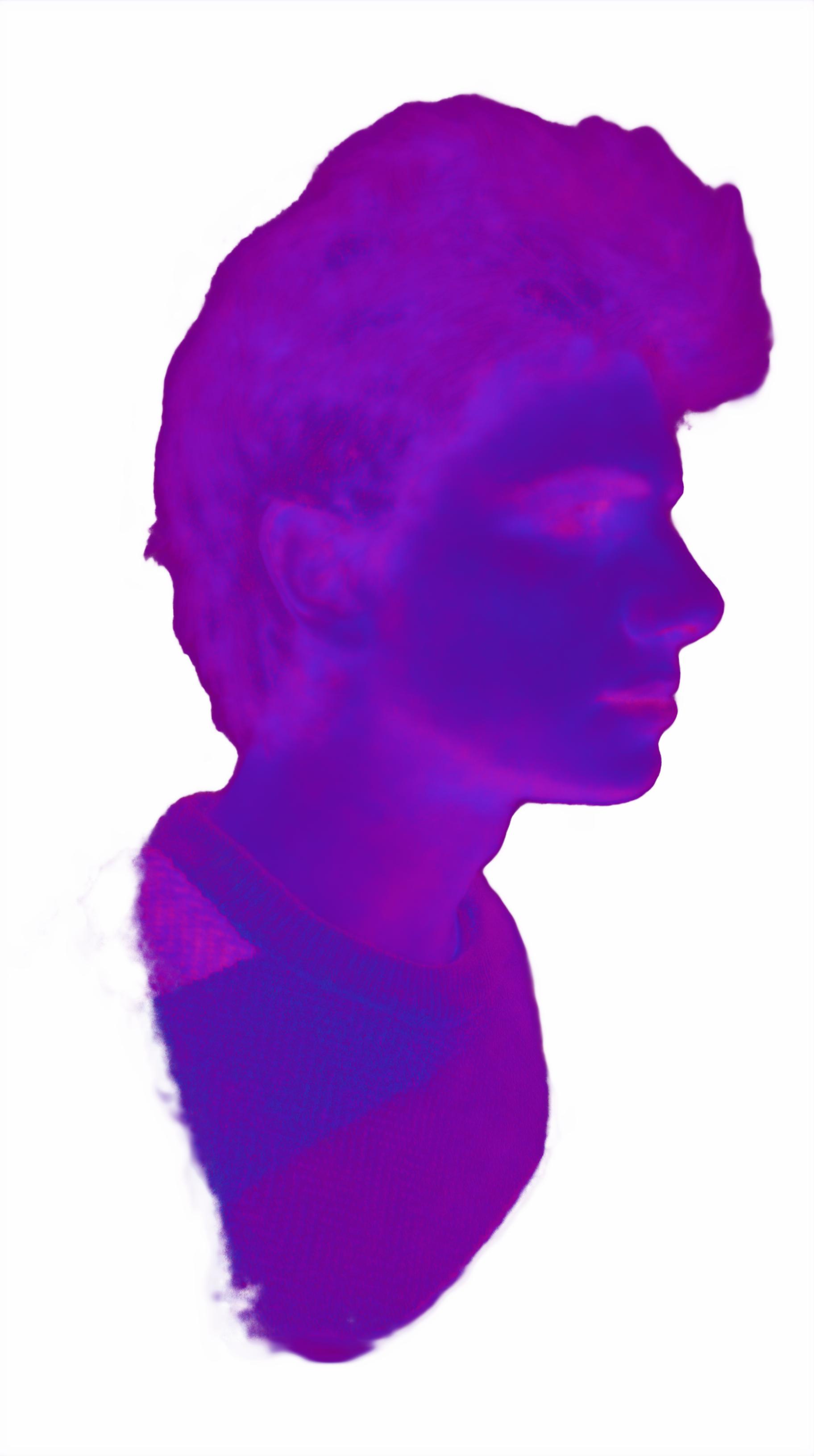}} \\
        
        \multirow{1}{*}{\rotatebox{90}{Shadows $S$}} &
        \raisebox{-0.5\height}{\adjincludegraphics[clip,trim={0 {.4\height} 0 0},width=.14\textwidth]{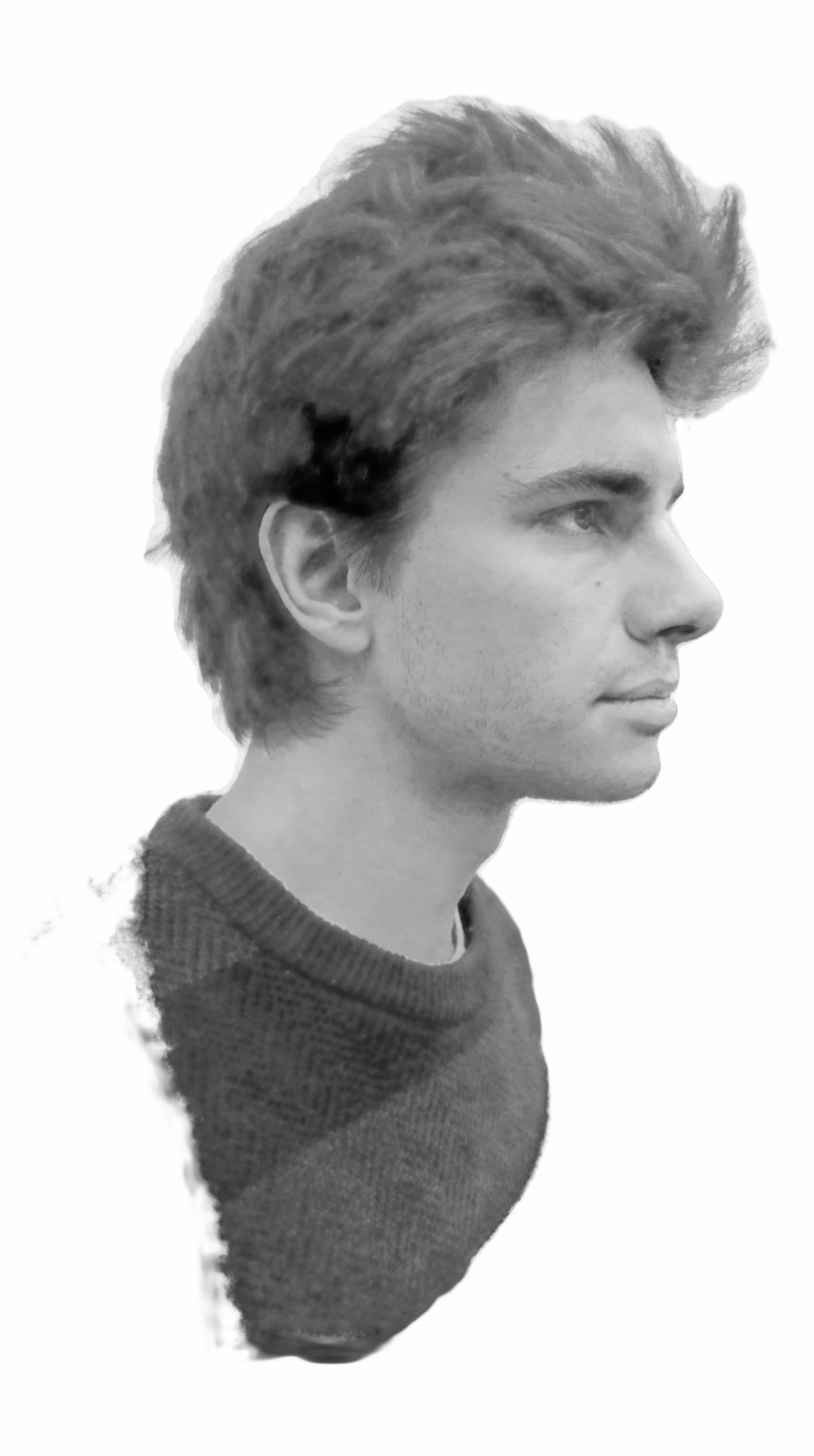}} & 
        \raisebox{-0.5\height}{\adjincludegraphics[clip,trim={0 {.4\height} 0 0},width=.14\textwidth]{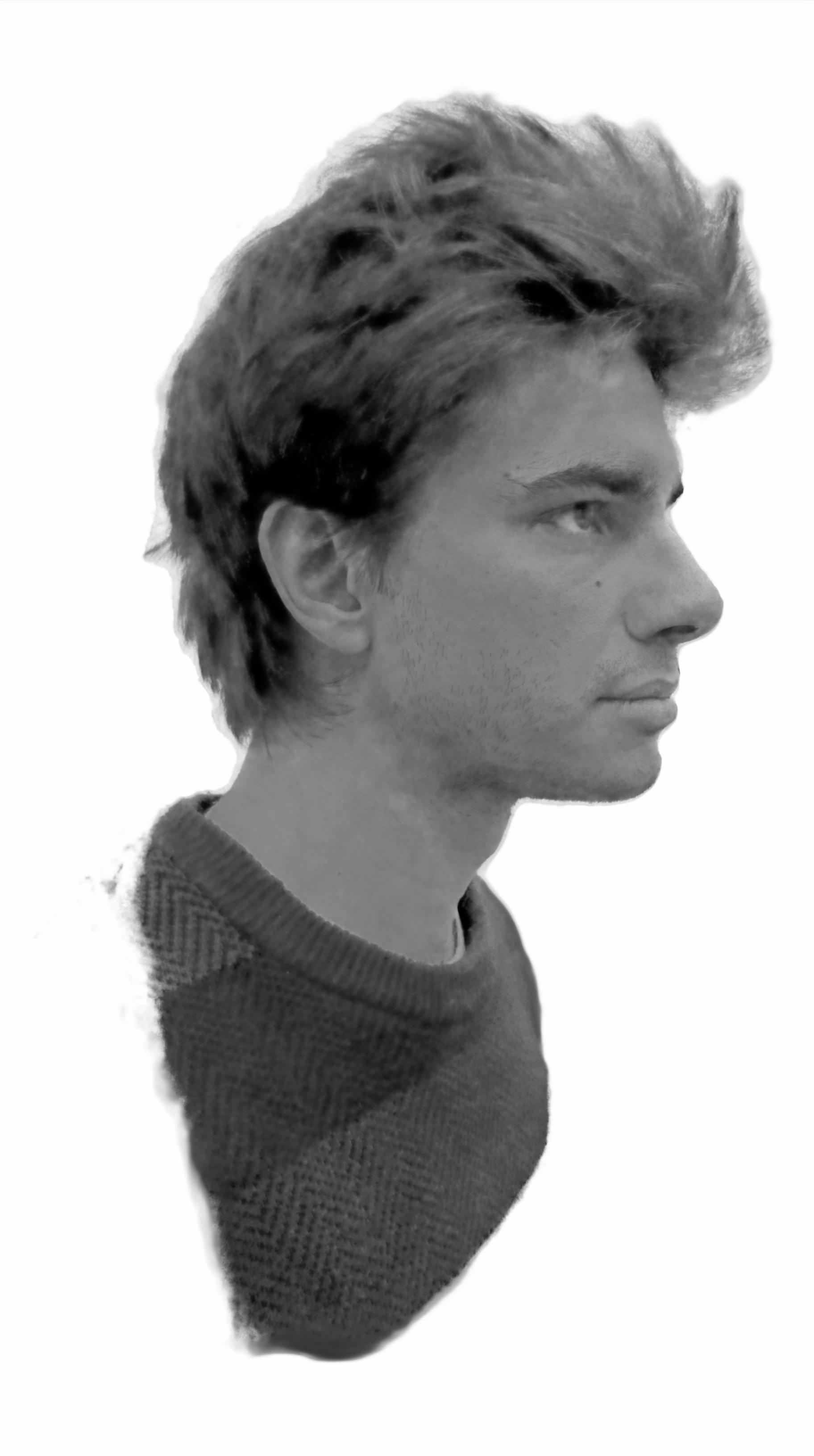}} &
        \raisebox{-0.5\height}{\adjincludegraphics[clip,trim={0 {.4\height} 0 0},width=.14\textwidth]{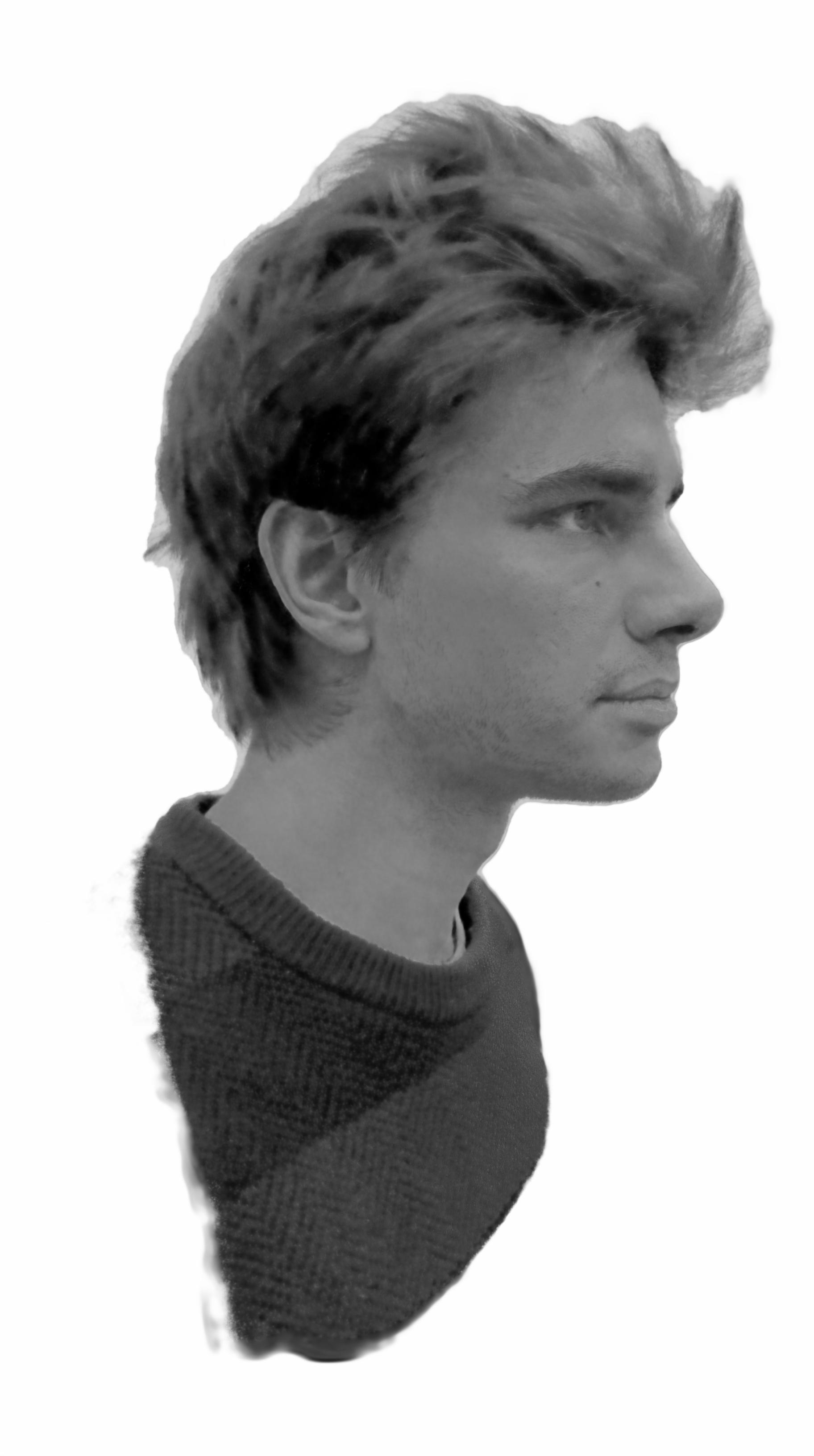}} &
        \raisebox{-0.5\height}{\adjincludegraphics[clip,trim={0 {.4\height} 0 0},width=.14\textwidth]{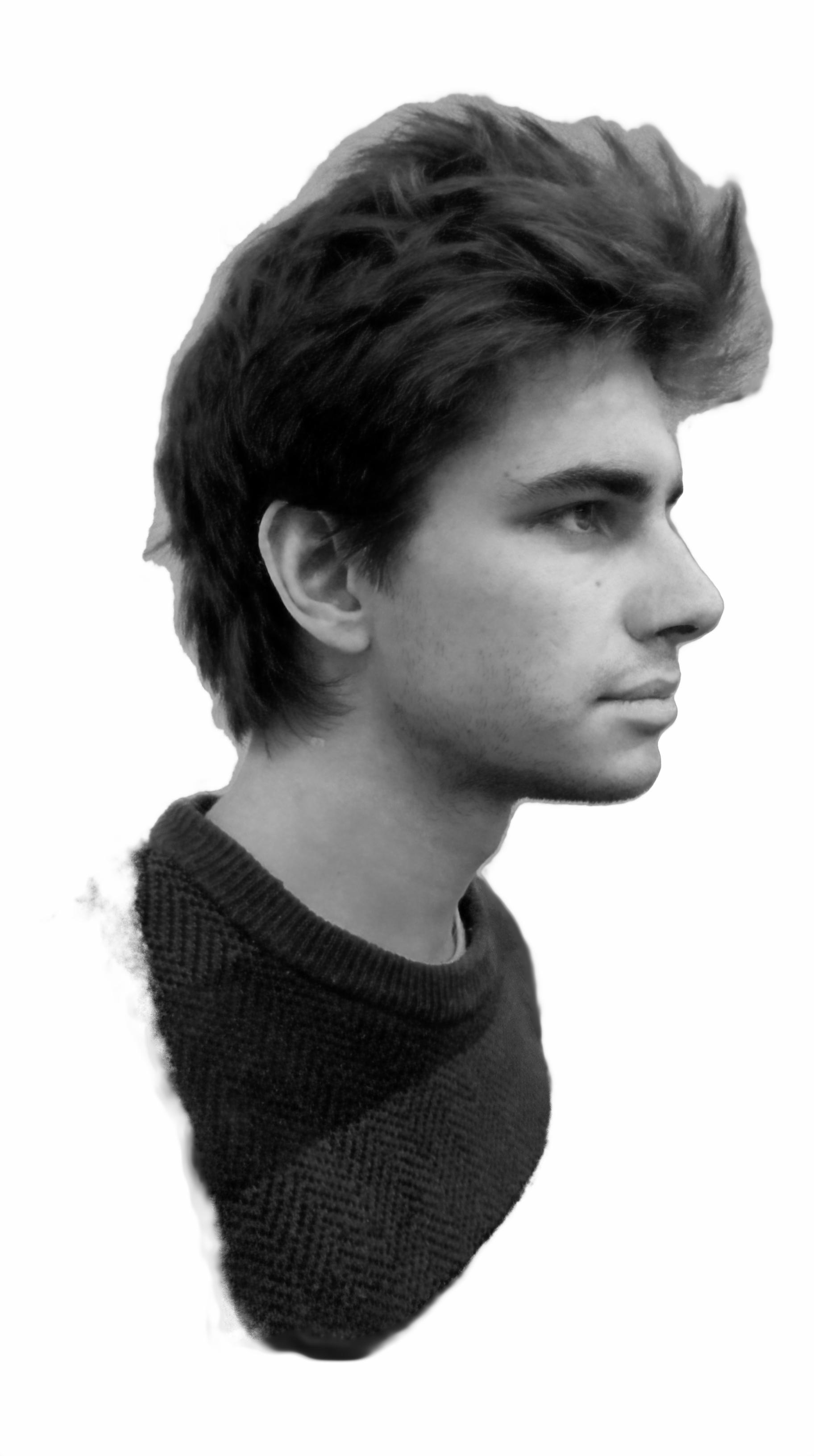}} &
        \raisebox{-0.5\height}{\adjincludegraphics[clip,trim={0 {.4\height} 0 0},width=.14\textwidth]{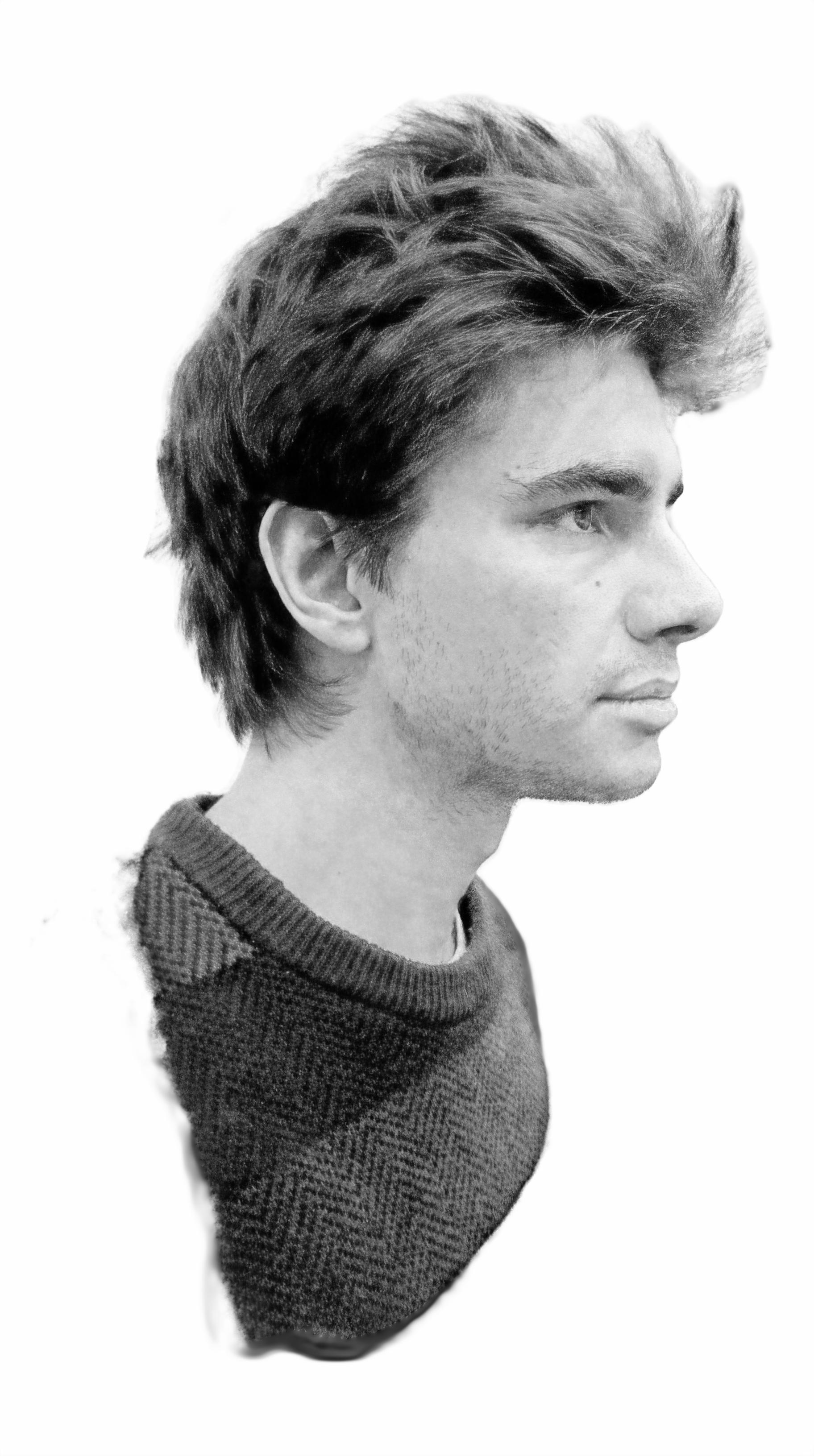}} &
        \raisebox{-0.5\height}{\adjincludegraphics[clip,trim={0 {.4\height} 0 0},width=.14\textwidth]{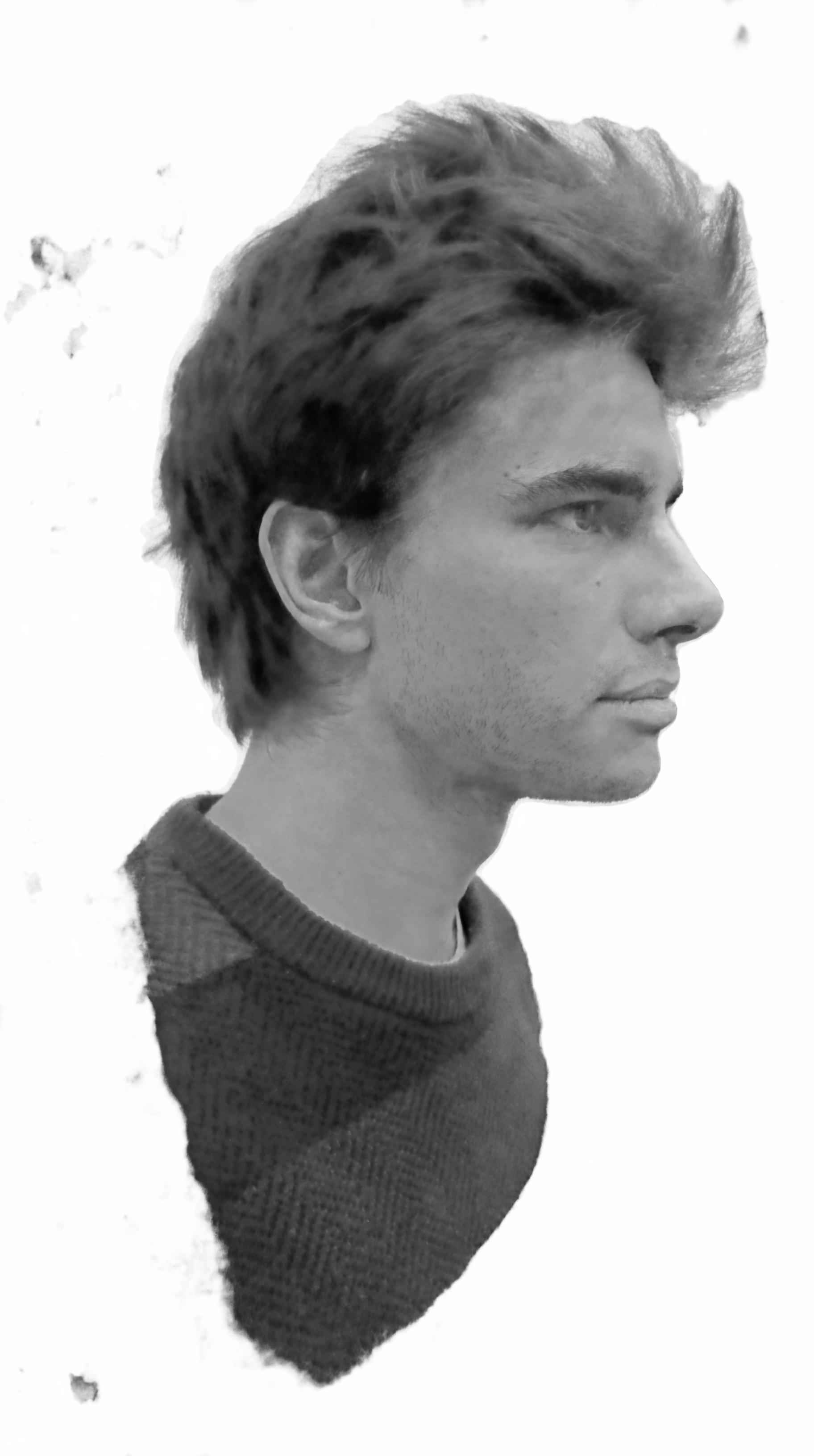}} &
        \raisebox{-0.5\height}{\adjincludegraphics[clip,trim={0 {.4\height} 0 0},width=.14\textwidth]{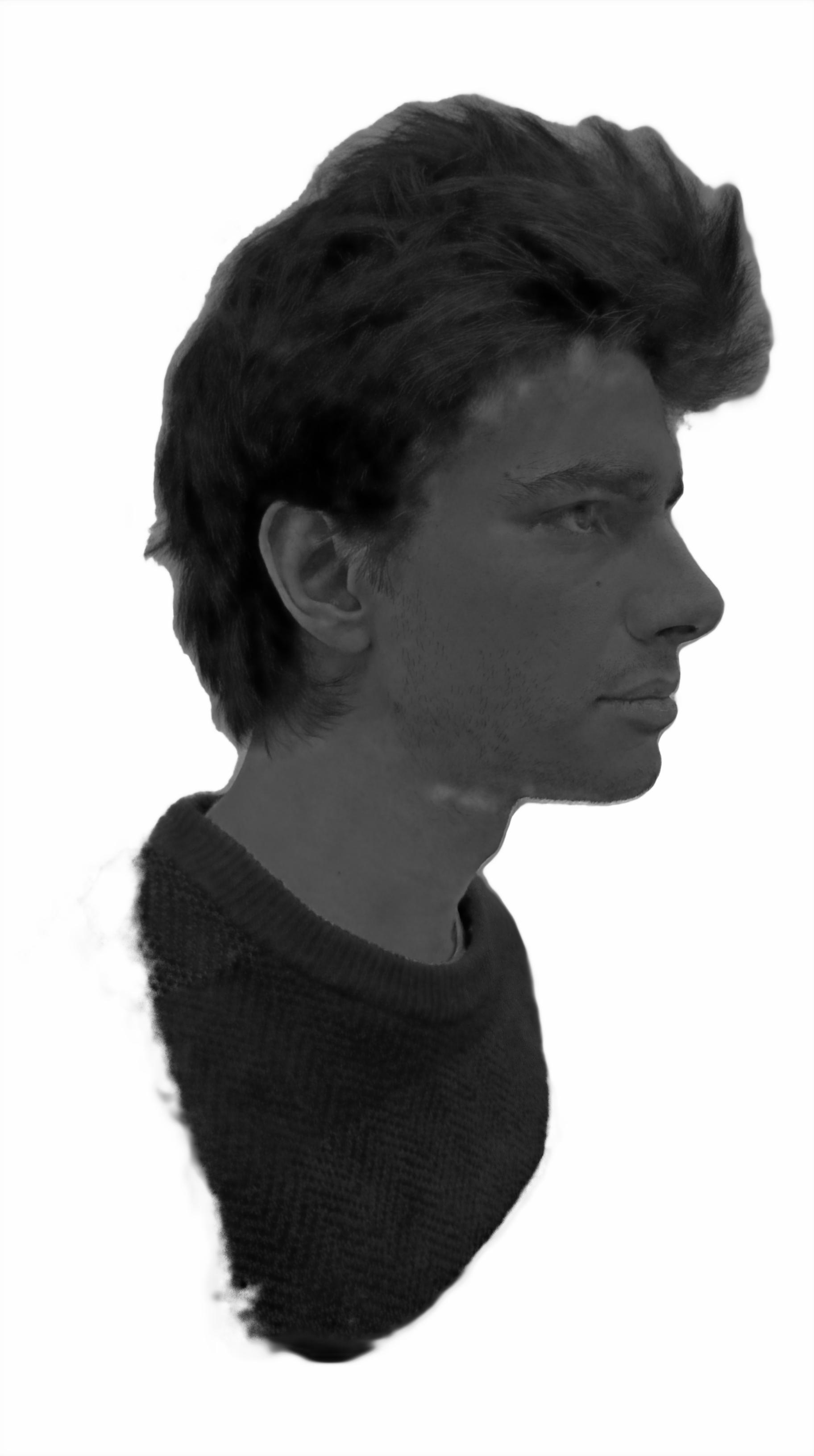}} \\
    \end{tabular}
    \captionof{figure}{Visual comparison of ablations, created by either removing one of the loss terms, excluding the preprocessing stage of filtering the point cloud by learned segmentation masks, or leaving only VGG term in $\Delta$ mismatch, which is used to evaluate $L_\mathrm{final},\, L_\mathrm{normal},\, L_\mathrm{symm}$. The viewpoint was not used for model fitting. \textit{Electronic zoom-in recommended.}}
    \label{fig:real_ablation}
\end{table*}

% \begin{figure*}[h]
    
%     \caption{Visual comparison of ablations.}
%     \label{fig:real_ablation}
% \end{figure*}

% \begin{table}[h]
%     \centering
%     \begin{tabular}{l|c|c|c|}
%                                               & VGG & FID & IS \\ \hline
%         Ours                                   &     &     &    \\ \hline \hline
%         w/o normals loss                       &     &     &    \\ \hline
%         w/o symmetry loss                      &     &     &    \\ \hline
%         w/o albedo color matching loss         &     &     &    \\ \hline
%         w/o room shading loss                  &     &     &    \\ \hline
%         w/o point cloud filtering, m.v. masks  &     &     &    \\ \hline
%         only VGG term in mismatch              &     &     &   
%     \end{tabular}\newline
    
%     \caption{Ablation study for real head portraits. The values of perceptual metrics are averaged over flashlighted validation views of $Person 1$ (both flashlighted and non-flashlighted). }
%     \label{table:ablation}
% \end{table}

\section{Discussion}

We have presented a pipeline for creating virtual portraits of humans based on smartphone-captured videos. Starting from the 3D rendering approach from Neural Point-Based Graphics~\cite{Aliev20}, we modify it by introducing the decomposition of a rendered image into albedo, normals, and room shadows. After training, the mappings can be rendered for an arbitrary viewpoint and with varying lighting. We demonstrate the relighting ability for simple models, such as directional light, and for more complex ones, such as spherical harmonics. Compared to most of the similar pipelines, ours does not require complex equipment for the data acquisition.

The method has certain limitations. First of all, the view interpolation ability is softly constrained by the trajectory of capturing, and the rendering quality can decrease when a viewpoint far from train is selected. Also, the approach models occlusions caused by other head parts (e.g. nose casting shadow on a cheek) only at train time, but not at test time. Finally, the method performs the best in the \textit{additional lighting} setting, i.e. the case when new lighting is added, but the room lighting is left in place.

%\section*{Acknowledgements}

\FloatBarrier
% \ifnum\value{page}>8 \errmessage{Number of pages exceeded!!!!}\fi

\balance
{\small
\bibliographystyle{ieee}
\bibliography{refs}

\begin{thebibliography}{10}\itemsep=-1pt

\bibitem{Agisoft}
{Agisoft}.
\newblock {\em Metashape software}.

\bibitem{Agrawal20}
S.~Agrawal, A.~Pahuja, and S.~Lucey.
\newblock High accuracy face geometry capture using a smartphone video.
\newblock In {\em The IEEE Winter Conference on Applications of Computer
  Vision}, pages 81--90, 2020.

\bibitem{Aliev20}
K.-A. Aliev, A.~Sevastopolsky, M.~Kolos, D.~Ulyanov, and V.~Lempitsky.
\newblock Neural point-based graphics.
\newblock {\em European Conference on Computer Vision (ECCV)}, 2020.

\bibitem{Boss20}
M.~Boss, V.~Jampani, K.~Kim, H.~Lensch, and J.~Kautz.
\newblock Two-shot spatially-varying brdf and shape estimation.
\newblock In {\em Proceedings of the IEEE/CVF Conference on Computer Vision and
  Pattern Recognition}, pages 3982--3991, 2020.

\bibitem{Cao20}
X.~Cao, M.~Waechter, B.~Shi, Y.~Gao, B.~Zheng, and Y.~Matsushita.
\newblock Stereoscopic flash and no-flash photography for shape and albedo
  recovery.
\newblock In {\em Proceedings of the IEEE/CVF Conference on Computer Vision and
  Pattern Recognition}, pages 3430--3439, 2020.

\bibitem{Chen20}
Z.~Chen, A.~Chen, G.~Zhang, C.~Wang, Y.~Ji, K.~N. Kutulakos, and J.~Yu.
\newblock A neural rendering framework for free-viewpoint relighting.
\newblock In {\em Proceedings of the IEEE/CVF Conference on Computer Vision and
  Pattern Recognition}, pages 5599--5610, 2020.

\bibitem{Dong14}
Y.~Dong, G.~Chen, P.~Peers, J.~Zhang, and X.~Tong.
\newblock Appearance-from-motion: Recovering spatially varying surface
  reflectance under unknown lighting.
\newblock {\em ACM Transactions on Graphics (TOG)}, 33(6):1--12, 2014.

\bibitem{Feng18}
Y.~Feng, F.~Wu, X.~Shao, Y.~Wang, and X.~Zhou.
\newblock Joint 3d face reconstruction and dense alignment with position map
  regression network.
\newblock In {\em Proceedings of the European Conference on Computer Vision
  (ECCV)}, pages 534--551, 2018.

\bibitem{Furukawa09}
Y.~Furukawa and J.~Ponce.
\newblock Accurate, dense, and robust multiview stereopsis.
\newblock {\em {T-PAMI}}, 32(8):1362--1376, 2009.

\bibitem{Gkitsas20}
V.~Gkitsas, N.~Zioulis, F.~Alvarez, D.~Zarpalas, and P.~Daras.
\newblock Deep lighting environment map estimation from spherical panoramas.
\newblock In {\em Proceedings of the IEEE/CVF Conference on Computer Vision and
  Pattern Recognition Workshops}, pages 640--641, 2020.

\bibitem{Guo19}
K.~Guo, P.~Lincoln, P.~Davidson, J.~Busch, X.~Yu, M.~Whalen, G.~Harvey,
  S.~Orts-Escolano, R.~Pandey, J.~Dourgarian, et~al.
\newblock The relightables: Volumetric performance capture of humans with
  realistic relighting.
\newblock {\em ACM Transactions on Graphics (TOG)}, 38(6):1--19, 2019.

\bibitem{OpenCamera20}
M.~Harman and et~al.
\newblock Open {C}amera application.
\newblock \url{https://sourceforge.net/p/opencamera/code/ci/master/tree/},
  2013--2020.

\bibitem{Heusel17}
M.~Heusel, H.~Ramsauer, T.~Unterthiner, B.~Nessler, and S.~Hochreiter.
\newblock Gans trained by a two time-scale update rule converge to a local nash
  equilibrium.
\newblock In {\em Advances in neural information processing systems}, pages
  6626--6637, 2017.

\bibitem{Johnson16}
J.~Johnson, A.~Alahi, and L.~Fei-Fei.
\newblock Perceptual losses for real-time style transfer and super-resolution.
\newblock In {\em European conference on computer vision}, pages 694--711.
  Springer, 2016.

\bibitem{Kajiya86}
J.~T. Kajiya.
\newblock The rendering equation.
\newblock In {\em Proceedings of the 13th annual conference on Computer
  graphics and interactive techniques}, pages 143--150, 1986.

\bibitem{Karras17}
T.~Karras, T.~Aila, S.~Laine, and J.~Lehtinen.
\newblock Progressive growing of gans for improved quality, stability, and
  variation.
\newblock {\em arXiv preprint arXiv:1710.10196}, 2017.

\bibitem{Kingma14}
D.~P. Kingma and J.~Ba.
\newblock Adam: A method for stochastic optimization.
\newblock {\em arXiv preprint arXiv:1412.6980}, 2014.

\bibitem{Kong14}
N.~Kong, P.~V. Gehler, and M.~J. Black.
\newblock Intrinsic video.
\newblock In {\em European Conference on Computer Vision}, pages 360--375.
  Springer, 2014.

\bibitem{Kutulakos00}
K.~N. Kutulakos and S.~M. Seitz.
\newblock A theory of shape by space carving.
\newblock {\em International journal of computer vision}, 38(3):199--218, 2000.

\bibitem{Lassner20}
C.~Lassner.
\newblock Fast differentiable raycasting for neural rendering using
  sphere-based representations.
\newblock {\em arXiv preprint arXiv:2004.07484}, 2020.

\bibitem{Lattas20}
A.~Lattas, S.~Moschoglou, B.~Gecer, S.~Ploumpis, V.~Triantafyllou, A.~Ghosh,
  and S.~Zafeiriou.
\newblock Avatarme: Realistically renderable 3d facial reconstruction.
\newblock In {\em Proceedings of the IEEE/CVF Conference on Computer Vision and
  Pattern Recognition}, pages 760--769, 2020.

\bibitem{Laurentini94}
A.~Laurentini.
\newblock The visual hull concept for silhouette-based image understanding.
\newblock {\em IEEE Transactions on pattern analysis and machine intelligence},
  16(2):150--162, 1994.

\bibitem{Li20}
Z.~Li, M.~Shafiei, R.~Ramamoorthi, K.~Sunkavalli, and M.~Chandraker.
\newblock Inverse rendering for complex indoor scenes: Shape, spatially-varying
  lighting and svbrdf from a single image.
\newblock In {\em Proceedings of the IEEE/CVF Conference on Computer Vision and
  Pattern Recognition}, pages 2475--2484, 2020.

\bibitem{Meka19}
A.~Meka, C.~Haene, R.~Pandey, M.~Zollh{\"o}fer, S.~Fanello, G.~Fyffe,
  A.~Kowdle, X.~Yu, J.~Busch, J.~Dourgarian, et~al.
\newblock Deep reflectance fields: high-quality facial reflectance field
  inference from color gradient illumination.
\newblock {\em ACM Transactions on Graphics (TOG)}, 38(4):1--12, 2019.

\bibitem{Meka20}
A.~Meka, R.~Pandey, C.~Haene, S.~Orts-Escolano, P.~Barnum, P.~Davidson,
  D.~Erickson, Y.~Zhang, J.~Taylor, S.~Bouaziz, C.~Legendre, W.-C. Ma,
  R.~Overbeck, T.~Beeler, P.~Debevec, S.~Izadi, C.~Theobalt, C.~Rhemann, and
  S.~Fanello.
\newblock Deep relightable textures - volumetric performance capture with
  neural rendering.
\newblock In {\em ACM Transactions on Graphics (Proceedings SIGGRAPH Asia)},
  volume~39, December 2020.

\bibitem{Mildenhall20}
B.~Mildenhall, P.~P. Srinivasan, M.~Tancik, J.~T. Barron, R.~Ramamoorthi, and
  R.~Ng.
\newblock Nerf: Representing scenes as neural radiance fields for view
  synthesis.
\newblock {\em arXiv preprint arXiv:2003.08934}, 2020.

\bibitem{Palma13}
G.~Palma, N.~Desogus, P.~Cignoni, and R.~Scopigno.
\newblock Surface light field from video acquired in uncontrolled settings.
\newblock In {\em 2013 Digital Heritage International Congress
  (DigitalHeritage)}, volume~1, pages 31--38. IEEE, 2013.

\bibitem{Qin20}
X.~Qin, Z.~Zhang, C.~Huang, M.~Dehghan, O.~R. Zaiane, and M.~Jagersand.
\newblock U2-net: Going deeper with nested u-structure for salient object
  detection.
\newblock {\em Pattern Recognition}, 106:107404, 2020.

\bibitem{Ramamoorthi01}
R.~Ramamoorthi and P.~Hanrahan.
\newblock An efficient representation for irradiance environment maps.
\newblock In {\em Proceedings of the 28th annual conference on Computer
  graphics and interactive techniques}, pages 497--500, 2001.

\bibitem{RenderPeople}
{RenderPeople}.
\newblock {\em RenderPeople: World's largest library of 3D people}.

\bibitem{Ronneberger15}
O.~Ronneberger, P.~Fischer, and T.~Brox.
\newblock U-net: Convolutional networks for biomedical image segmentation.
\newblock In {\em International Conference on Medical image computing and
  computer-assisted intervention}, pages 234--241. Springer, 2015.

\bibitem{Schoenberger2016sfm}
J.~L. Sch\"{o}nberger and J.-M. Frahm.
\newblock Structure-from-motion revisited.
\newblock In {\em Conference on Computer Vision and Pattern Recognition
  (CVPR)}, 2016.

\bibitem{Schoenberger2016mvs}
J.~L. Sch\"{o}nberger, E.~Zheng, M.~Pollefeys, and J.-M. Frahm.
\newblock Pixelwise view selection for unstructured multi-view stereo.
\newblock In {\em European Conference on Computer Vision (ECCV)}, 2016.

\bibitem{Simonyan14}
K.~Simonyan and A.~Zisserman.
\newblock Very deep convolutional networks for large-scale image recognition.
\newblock {\em CoRR}, abs/1409.1556, 2014.

\bibitem{Sorkine04}
O.~Sorkine, D.~Cohen-Or, Y.~Lipman, M.~Alexa, C.~R{\"o}ssl, and H.-P. Seidel.
\newblock Laplacian surface editing.
\newblock In {\em Proceedings of the 2004 Eurographics/ACM SIGGRAPH symposium
  on Geometry processing}, pages 175--184, 2004.

\bibitem{Sudre17}
C.~H. Sudre, W.~Li, T.~Vercauteren, S.~Ourselin, and M.~J. Cardoso.
\newblock Generalised dice overlap as a deep learning loss function for highly
  unbalanced segmentations.
\newblock In {\em Deep learning in medical image analysis and multimodal
  learning for clinical decision support}, pages 240--248. Springer, 2017.

\bibitem{Supervisely}
{Supervisely}.
\newblock {\em Supervisely Person Dataset}.

\bibitem{Taheri12}
S.~Taheri, A.~C. Sankaranarayanan, and R.~Chellappa.
\newblock Joint albedo estimation and pose tracking from video.
\newblock {\em IEEE transactions on pattern analysis and machine intelligence},
  35(7):1674--1689, 2012.

\bibitem{Thies19}
J.~Thies, M.~Zollh\"{o}fer, and M.~Nie\ss{}ner.
\newblock Deferred neural rendering: Image synthesis using neural textures.
\newblock {\em ACM Trans. Graph.}, 38(4), July 2019.

\bibitem{Vogiatzis07}
G.~Vogiatzis, C.~H. Esteban, P.~H. Torr, and R.~Cipolla.
\newblock Multiview stereo via volumetric graph-cuts and occlusion robust
  photo-consistency.
\newblock {\em {T-PAMI}}, 29(12):2241--2246, 2007.

\bibitem{Wiles20}
O.~Wiles, G.~Gkioxari, R.~Szeliski, and J.~Johnson.
\newblock Synsin: End-to-end view synthesis from a single image.
\newblock In {\em Proceedings of the IEEE/CVF Conference on Computer Vision and
  Pattern Recognition}, pages 7467--7477, 2020.

\bibitem{Wu20}
S.~Wu, C.~Rupprecht, and A.~Vedaldi.
\newblock Unsupervised learning of probably symmetric deformable 3d objects
  from images in the wild.
\newblock In {\em Proceedings of the IEEE/CVF Conference on Computer Vision and
  Pattern Recognition}, pages 1--10, 2020.

\bibitem{Xu20}
S.~Xu, J.~Yang, D.~Chen, F.~Wen, Y.~Deng, Y.~Jia, and X.~Tong.
\newblock Deep 3d portrait from a single image.
\newblock In {\em IEEE/CVF Conference on Computer Vision and Pattern
  Recognition (CVPR)}, June 2020.

\bibitem{Yu19}
J.~Yu, Z.~Lin, J.~Yang, X.~Shen, X.~Lu, and T.~S. Huang.
\newblock Free-form image inpainting with gated convolution.
\newblock In {\em Proceedings of the IEEE International Conference on Computer
  Vision}, pages 4471--4480, 2019.

\bibitem{Zhang16}
K.~Zhang, Z.~Zhang, Z.~Li, and Y.~Qiao.
\newblock Joint face detection and alignment using multitask cascaded
  convolutional networks.
\newblock {\em IEEE Signal Processing Letters}, 23(10):1499--1503, 2016.

\bibitem{Zhang18}
R.~Zhang, P.~Isola, A.~A. Efros, E.~Shechtman, and O.~Wang.
\newblock The unreasonable effectiveness of deep features as a perceptual
  metric.
\newblock In {\em Proceedings of the IEEE conference on computer vision and
  pattern recognition}, pages 586--595, 2018.

\bibitem{Zhang20b}
X.~Zhang, S.~Fanello, Y.-T. Tsai, T.~Sun, T.~Xue, R.~Pandey, S.~Orts-Escolano,
  P.~Davidson, C.~Rhemann, P.~Debevec, et~al.
\newblock Neural light transport for relighting and view synthesis.
\newblock {\em arXiv preprint arXiv:2008.03806}, 2020.

\bibitem{Zhang20a}
X.~C. Zhang, Y.-T. Tsai, R.~Pandey, X.~Zhang, R.~Ng, D.~E. Jacobs, et~al.
\newblock Portrait shadow manipulation.
\newblock {\em arXiv preprint arXiv:2005.08925}, 2020.

\bibitem{Zhi20}
T.~Zhi, C.~Lassner, T.~Tung, C.~Stoll, S.~G. Narasimhan, and M.~Vo.
\newblock Texmesh: Reconstructing detailed human texture and geometry from
  rgb-d video.
\newblock In {\em European Conference on Computer Vision}, pages 492--509.
  Springer, 2020.

\bibitem{Zhou19}
H.~Zhou, S.~Hadap, K.~Sunkavalli, and D.~W. Jacobs.
\newblock Deep single-image portrait relighting.
\newblock In {\em Proceedings of the IEEE International Conference on Computer
  Vision}, pages 7194--7202, 2019.

\end{thebibliography}
}
\end{document}